\documentclass{article}

\usepackage[final]{neurips_2023}
\usepackage{hyperref}       %

\usepackage{amsfonts,amsmath,bm}

\def\eqref#1{equation~\ref{#1}}

\def\1{\bm{1}}

\DeclareMathAlphabet{\mathsfit}{\encodingdefault}{\sfdefault}{m}{sl}
\SetMathAlphabet{\mathsfit}{bold}{\encodingdefault}{\sfdefault}{bx}{n}

\usepackage[utf8]{inputenc} %
\usepackage{lipsum} %

\usepackage[T1]{fontenc}

\usepackage{etoolbox}
\newbool{includeappendix}
\setbool{includeappendix}{true}

\ifdefined\isoverfull
\else
\fi

\usepackage[usenames, dvipsnames]{color} %
\usepackage{xcolor} %

\definecolor{my-full-blue}{HTML}{1F77B4}
\definecolor{my-dark-blue}{HTML}{0047AB}

\definecolor{my-full-orange}{HTML}{FF7F0E}

\definecolor{my-full-green}{HTML}{2CA02C}

\definecolor{my-full-red}{HTML}{d62728}
\definecolor{my-dark-red}{HTML}{C41E3A}

\definecolor{my-full-purple}{HTML}{9467bd}

\definecolor{my-dark-green}{HTML}{006e00}

\colorlet{my-blue}{my-full-blue!30}
\colorlet{my-orange}{my-full-orange!30}
\colorlet{my-green}{my-full-green!30}
\colorlet{my-red}{my-full-red!30}
\colorlet{my-purple}{my-full-purple!30}

\usepackage{listings}

\usepackage{textcomp}

\usepackage{xcolor}

\usepackage[scaled=0.8]{beramono}

\definecolor{ckeyword}{HTML}{7F0055}
\definecolor{ccomment}{HTML}{3F7F5F}
\definecolor{cstring}{HTML}{2A0099}

\lstdefinestyle{numbers}{
	numbers=left,
	framexleftmargin=20pt,
	numberstyle=\tiny,
	firstnumber=auto,
	numbersep=1em,
	xleftmargin=2em
}

\lstdefinestyle{layout}{
	frame=none,
	captionpos=b,
}

\lstdefinestyle{comment-style}{
	morecomment=[l]//,
	morecomment=[s]{/*}{*/},
	commentstyle={\color{ccomment}\itshape},
}

\lstdefinestyle{string-style}{
	morestring=[b]",%
	morestring=[b]',%
	stringstyle={\color{cstring}},
	showstringspaces=false,%
}

\lstdefinestyle{keyword-style}{
	keywordstyle={\ttfamily\bfseries},
	morekeywords={
		function,
		constructor,
		int,
		bool,
		return,
		returns,
		uint
	},
	morekeywords = [2]{},
	keywordstyle = [2]{\text},
	sensitive=true,
}

\lstdefinestyle{input-encoding}{
	inputencoding=utf8,
	extendedchars=true,
	literate=
	{ℝ}{$\reals$}1%
	{→}{$\rightarrow$}1%
	{α}{$\alpha$}1%
	{β}{$\beta$}1%
	{λ}{$\lambda$}1%
	{θ}{$\theta$}1%
	{ϕ}{$\phi$}1%
}

\lstdefinestyle{escaping}{
	moredelim={**[is][\color{blue}]{\%}{\%}},
	escapechar=|,
	mathescape=true
}

\lstdefinestyle{default-style}{
	basicstyle=\fontencoding{T1}\ttfamily\footnotesize,
	style=numbers,
	style=layout,
	style=comment-style,
	style=string-style,
	style=keyword-style,
	style=input-encoding,
	style=escaping,
	tabsize=2,
	upquote=true
}

\lstdefinelanguage{BASIC}{
	language=C++,
	style=default-style
}[keywords,comments,strings]%

\usepackage[capitalize]{cleveref}

\crefformat{section}{\S#2#1#3}

\crefrangeformat{section}{\S#3#1#4\crefrangeconjunction\S#5#2#6}

\crefmultiformat{section}{\S#2#1#3}{\crefpairconjunction\S#2#1#3}{\crefmiddleconjunction\S#2#1#3}{\creflastconjunction\S#2#1#3}

\newcommand{\crefrangeconjunction}{--}

\crefname{listing}{Lst.}{listings}
\crefname{line}{Lin.}{Lin.}
\crefname{appendix}{App.}{App.}

\newcommand{\app}[1]{%
	\ifbool{includeappendix}{\cref{#1}}{the appendix}%
}
\newcommand{\App}[1]{%
	\ifbool{includeappendix}{\cref{#1}}{The appendix}%
}

\usepackage{tikz}

\usetikzlibrary{calc,decorations,decorations.pathmorphing}
\usetikzlibrary{positioning,fit,arrows}
\usetikzlibrary{decorations.markings}
\usetikzlibrary{shapes,shapes.geometric}
\usetikzlibrary{shadows,patterns,snakes}
\usetikzlibrary{backgrounds,decorations.pathreplacing,calligraphy,automata}
\usetikzlibrary{intersections}
\usetikzlibrary{angles,quotes}
\usetikzlibrary{plotmarks}
\usetikzlibrary{patterns}
\usetikzlibrary{arrows.meta}
\usetikzlibrary{shapes.misc}
\usetikzlibrary{chains}

\usepackage[utf8]{inputenc} %
\usepackage[T1]{fontenc}    %
\usepackage{url}            %
\usepackage{booktabs}       %
\usepackage{nicefrac}       %
\usepackage{microtype}      %
\usepackage{xcolor}         %

\usepackage{graphicx}
\usepackage{adjustbox}
\usepackage{threeparttable}
\usepackage{thmtools, nameref}
\usepackage{multirow}
\usepackage{amssymb}
\usepackage{mathtools}
\usepackage{amsthm}
\usepackage[abbreviations]{foreign}
\usepackage{enumitem}
\usepackage{makecell}
\usepackage{algorithm}
\usepackage[noend]{algpseudocode} %
\usepackage{wrapfig}
\usepackage{pifont}
\usepackage{longtable}

\usepackage{subcaption}
\usepackage{caption}
\crefname{lemma}{Lemma}{lemmas}
\Crefname{lemma}{Lemma}{Lemmas}
\crefname{corollary}{corollary}{corollaries}
\Crefname{corollary}{Corollary}{Corollaries}

\newcommand{\IN}{\textsc{ImageNet}\xspace}
\newcommand{\SIN}{\textsc{Salient ImageNet}\xspace}
\newcommand{\HIN}{\textsc{Hard ImageNet}\xspace}
\newcommand{\INH}{\textsc{ImageNet-Hard}\xspace}
\newcommand{\INA}{\textsc{ImageNet-A}\xspace}
\newcommand{\INM}{\textsc{ImageNet-Major}\xspace}
\newcommand{\INX}{\textsc{ImageNet-X}\xspace}
\newcommand{\INV}{\textsc{ImageNetV2}\xspace}

\newcommand{\INST}{\textsc{Instagram}\xspace}
\newcommand{\WN}{\textsc{WordNet}\xspace}
\newcommand{\LAION}{\textsc{LAION-2B}\xspace}

\newcommand{\vit}{\texttt{ViT-3B}\xspace}
\newcommand{\rnf}{\texttt{ResNet50}\xspace}
\newcommand{\greedysoups}{\texttt{Greedy Soups}\xspace}

\newcommand{\clip}{\textsc{CLIP}\xspace}

\newcommand{\nscls}{161\xspace}
\newcommand{\nmodels}{962\xspace}

\newcommand{\toa}{top-1\xspace}
\newcommand{\tfa}{top-5\xspace}
\newcommand{\mla}{MLA\xspace}

\newcommand{\inc}[1]{\texttt{#1\xspace}}

\colorlet{c_sc}{my-full-blue!30}
\colorlet{c_wns}{my-full-orange!30}
\colorlet{c_wna}{my-full-green!30}

\newcommand{\newprotectedcommand}[2]{\newcommand{#1}{\protecting{#2}}}

\newprotectedcommand{\msc}{\tikz[]{\node[fill=c_sc, aspect=1, inner sep=0pt, minimum size=2.1mm, rounded corners=1.5pt]{};}\xspace}
\newprotectedcommand{\mwns}{\tikz[]{\node[fill=c_wns, aspect=1, inner sep=0pt, minimum size=2.1mm, rounded corners=1.5pt]{};}\xspace}
\newprotectedcommand{\mwna}{\tikz[]{\node[fill=c_wna, aspect=1, inner sep=0pt, minimum size=2.1mm, rounded corners=1.5pt]{};}\xspace}

\newprotectedcommand{\mplab}{\tikz[]{\node[fill=c_wna, aspect=1, inner sep=0pt, minimum size=2.1mm, rounded corners=1.5pt] (b1) {};
\node[fill=c_wns, aspect=1, inner sep=0pt, minimum size=2.1mm, rounded corners=1.5pt] (b2) at ($(b1)+(0.1,0)$){};
\node[fill=c_sc, aspect=1, inner sep=0pt, minimum size=2.1mm, rounded corners=1.5pt] (b3) at ($(b2)+(0.1,0)$){};
}\xspace}

\newprotectedcommand{\mpred}{\tikz[]{\node[aspect=1, draw=black, inner sep=0pt, minimum size=2.1mm, dash pattern=on 2.5pt off 1pt, rounded corners=1.5pt]{};}\xspace}
\newprotectedcommand{\mlab}{\tikz[]{\node[aspect=1, draw=black, inner sep=0pt, minimum size=2.1mm, rounded corners=1.5pt]{};}\xspace}

\newprotectedcommand{\markermultipgd}{\tikz[]{\draw[pgdarrow, dotted, dash pattern=on 1pt off 0.75pt] (0,0.2) -- (0.2,0.2) -- (0.2,0);}\xspace}

\title{Automated Classification of Model Errors on ImageNet}

\author{%
  Momchil Peychev\thanks{Equal contribution}, Mark Niklas Müller\footnotemark[1], Marc Fischer, Martin Vechev\\
  Department of Computer Science\\
  ETH Zurich, Switzerland\\
  \texttt{\{momchil.peychev, mark.mueller, marc.fischer, martin.vechev\}@inf.ethz.ch} \\
}
\begin{document}

\maketitle

\begin{abstract}
  While the ImageNet dataset has been driving computer vision research over the past decade, significant label noise and ambiguity have made top-1 accuracy an insufficient measure of further progress. To address this, new label-sets and evaluation protocols have been proposed for ImageNet showing that state-of-the-art models already achieve over $95\%$ accuracy and shifting the focus on investigating why the remaining errors persist.
Recent work in this direction employed a panel of experts to manually categorize all remaining classification errors for two selected models. However, this process is time-consuming, prone to inconsistencies, and requires trained experts, making it unsuitable for regular model evaluation thus limiting its utility. To overcome these limitations, we propose the first automated error classification framework, a valuable tool to study how modeling choices affect error distributions. We use our framework to comprehensively evaluate the error distribution of over 900 models. Perhaps surprisingly, we find that across model architectures, scales, and pre-training corpora, top-1 accuracy is a strong predictor for the \emph{portion} of all error types. In particular, we observe that the portion of severe errors drops significantly with top-1 accuracy indicating that, while it underreports a model's true performance, it remains a valuable performance metric.
We release all our code at \texttt{\href{https://github.com/eth-sri/automated-error-analysis}{https://github.com/eth-sri/automated-error-analysis}}.
  
\end{abstract}

\section{Introduction}

\IN \citep{DengDSLL009,RussakovskyDSKS15} has established itself as one of the most influential and widely used datasets in computer vision, driving progress in object recognition \citep{KrizhevskySH12}, object detection \citep{TanPL20}, and image segmentation \citep{MinaeeBPPKT22}. As state-of-the-art models have come close to, and by some metrics exceeded, human performance, the significance of further progress in \toa and \tfa accuracy has, been questioned in the face of label errors and systematic biases \citep{BeyerHKZO20,TsiprasSEIM20}. 

The most severe such bias is the lack of multi-label annotations in the original \IN dataset, with recent studies finding that roughly a fifth of images show multiple entities. While state-of-the-art models have learned to exploit labeling biasses on these images \citep{TsiprasSEIM20}, best practices have shifted to reporting multi-label accuracy (MLA) computed using new multi-label annotations \citep{BeyerHKZO20,ShankarRMFRS20}.
Further, many \IN and especially organism classes, are hard to distinguish even by trained humans \citep{HornBFHBIPB15,NorthcuttAM21,LeeAK17}, leading to persistent labeling errors. 

In the face of these challenges and with state-of-the-art models exceeding $95\%$ MLA, the focus has increasingly shifted towards analyzing and understanding the remaining model errors instead of blindly pursuing improvements in headline accuracy numbers. 
To this end, \citet{VasudevanCLFR22} review all remaining errors of two state-of-the-art models using a panel of experts and classify them with regards to both error category and severity. While they find that many of the remaining errors are minor or can be attributed to fine-grained class distinctions, they also find major classification errors that are not so easily explainable. 
We believe that tracking and analyzing the distribution of these error types over a large number of models can not only help us to better understand the impact of novel training techniques and architectures but also identify where the biggest challenges lie and thus how to address them. However, the manual review process employed by \citet{VasudevanCLFR22}   has several issues, preventing it from being employed for a large scale or repeated study of model errors: (i) it is time-consuming even for precise models, making it infeasible to repeat for a large number of potentially less precise models, (ii) it requires (a panel of) expert reviewers which need to be trained on the fine-grained class distinctions, and (iii) it is inconsistent as different reviewers or even the same reviewers at different times might classify the same error differently.

\paragraph{This Work} To overcome these challenges, we propose an automated error classification pipeline that we use to study the distribution of different types of errors across \nmodels models of different scales, architectures, training methods, and pre-training datasets. Our pipeline allows us to automatically detect all four error categories identified by \citet{VasudevanCLFR22}: (i) fine-grained classification errors are detected using a set of \nscls manually defined superclasses, (ii) fine-grained out-of-vocabulary errors are detected using a visual similarity based criterion for prediction quality and confirmation of their out-of-vocabulary nature using an open-world classifier, (iii) non-prototypical examples are identified using the exhaustive annotations by \citet{VasudevanCLFR22}, and (iv) spurious correlations are detected using a co-occurrence-frequency based criterion.

\paragraph{Main Findings} Our automated error classification pipeline allows us to, for the first time, study the \emph{distribution of different error types} across a large number of models, leading to the following insights:
(i) even MLA is a pessimistic measure of model progress with the \emph{portion} of severe model failures quickly decreasing with MLA, (ii) this reduction of model failure rate with MLA is more pronounced for larger (pre-)training corpora, \ie, models trained on more data make less severe errors even at the same \toa or multilabel accuracy, and (iii) organism and artifact classes exhibit very different trends and prevalences of error types, \eg, fine-grained classification errors are much more frequent for organisms than for artifacts, while artifacts suffer much more from spurious correlations and out-of-vocabulary errors.
We believe that these insights can help guide future research in computer vision and that studying the effects of new methods on the resulting error distribution can become an important part of the evaluation pipeline.

\section{Related Work} \label{sec:related}

\paragraph{Multi-Label Annotations}
While the \IN dataset is annotated with a single label per image, many images have been found to contain multiple entities \citep{BeyerHKZO20,ShankarRMFRS20,TsiprasSEIM20,YunOHHCC21}. Thus, multi-label accuracy (MLA) with respect to new multi-label annotations has been established as a more meaningful metric.
\citet{YunOHHCC21} generate pixel-wise labels for \IN by directly applying a classification layer to image embeddings before spatial pooling.
\citet{BeyerHKZO20} collect Reassessed Labels (ReaL) for the whole \IN validation set, by first identifying 6 models with high prediction coverage and accuracy, before manually annotating all images where these 6 models disagree using non-expert labelers. They discard $3\,163$ images where these labelers disagreed.
\citet{TsiprasSEIM20} collect multi-label annotations for $10\,000$ validation set images, and find that \toa accuracy is 10\% lower for multi- compared to single-label images while \mla accuracy is identical.%
~\citet{ShankarRMFRS20} collect multi-label annotation for $40\,000$ \IN and \INV validation images by manually reviewing all predictions made by a diverse set of 72 models, using human expert labelers.

\paragraph{Label Errors}
\citet{NorthcuttAM21} study label errors across 10 commonly used datasets, including \IN, using a confident learning framework to identify potential errors before validating them with Mechanical Turk (MTurk). Studying the whole validation set, they report an error rate of over $5.8\%$.
\citet{LeeAK17} manually review 400 randomly selected classification errors of an ensemble model and find the \IN label for a substantial portion to either be incorrect or not describe the main entity in the image.
\citet{VasudevanCLFR22} reviewed all remaining errors for two state-of-the-art models with a panel of experts and found that of 676 reviewed model mistakes, 298 were either correct, ambiguous, or the original ground truth incorrect or problematic.

\paragraph{Error Analysis}
Recent work has focused on understanding the types of errors models still make on \IN. To this end, one strand of work analyses the differences in errors between models. \citet{ManiaMSHR19} find that independently trained models make errors that correlate well beyond what can be expected from their accuracy alone. \citet{GeirhosMW20} analyze the consistency between errors made by humans and CNNs on a $16$ class version of \IN. They observe that while CNNs make remarkably similar errors, the consistency between humans and CNNs barely goes beyond chance. 
\citet{NguyenRK21} analyze wide and deep ResNets and find that these exhibit different error patterns.
\citet{ManiaS20} find that more precise models typically dominate less precise ones, \ie, their error set is a subset of that of less precise models. \citet{LopesDC22}, in contrast, find that different pretraining corpora and training objectives can significantly increase error diversity with \citet{AndreassenBNR21} showing that this diversity reduces significantly during finetuning.

\citet{VasudevanCLFR22} focus on the errors current state-of-the-art models make, identifying four categories: (i) \emph{fine-grained} errors describe a model's failure to distinguish between two very similar classes, (ii) \emph{fine-grained out-of-vocablary} errors occur when an image shows an entity not contained in the \IN vocabulary and the model instead predicts a similar class, (iii) \emph{spurious correlations} cause the model to predict a class that is not shown in the image in the presence of strongly correlated features, and (iv) \emph{non-prototypical} instantiations of a class are not recognized by the model.

\paragraph{Datasets for Error Analysis}
In addition to work analyzing model errors on fixed datasets, there has been a growing interest in datasets specifically designed to highlight and thus analyze specific error types.
\citet{SinglaF22} and \citet{Moayeri0F22} find that, for some classes, models rely heavily on correlated or spurious features suffering severely reduced accuracy if these are absent or removed. To study this effect, they introduce \SIN \citep{SinglaF22} and \HIN \citep{Moayeri0F22}, providing soft masks for causal and correlated features and segmentation masks, respectively.
\citet{HendrycksZBSS21} propose \INA, a dataset of $7\,500$ single-entity, natural adversarial examples, which induce high-confidence misclassifications in a set of \rnf and belong to $200$ \IN classes that were chosen to minimize class overlap and the potential for fine-grained errors (see \cref{sec:ina} for an analysis).
\citet{VasudevanCLFR22} collect images where state-of-the-art models fail in an unexplained manner in \INM.
\citet{TaesiriNHBN23} investigate the effect of zoom and crop on classifier performance and introduce \INH as the set of images (from a range of \IN-like datasets) that none of their considered models classified correctly for any crop.
\citet{IdrissiBBEHBVDL23} annotate \IN images with respect to how they differ from what is prototypical for their class, introducing \INX. This allows them to study the effect of these image-specific variations such as atypical pose, background, or lighting situation, on accuracy over a large number of models. In contrast, our (and \citet{VasudevanCLFR22}'s) work focuses more on systematic errors caused (partially) by labeling (set) choices rather than on what makes an individual image hard to classify.

\section{Categorizing ImageNet Errors} \label{sec:method}

In this section, we introduce our automated error classification pipeline, which aims to explain model errors by assigning it one of six error types. We consider errors that can not be explained in this way to be particularly severe \emph{model failures}.
While the definitions of our error types are heavily inspired by \citet{VasudevanCLFR22}, we address three main issues of their manual approach with our automated pipeline: (i) it is time-consuming even for precise models and intractable for imprecise ones, (ii) it requires a panel of expert reviewers, typically not available, and (iii) it introduces inconsistencies due to human expert error, disagreement, or ambiguity.

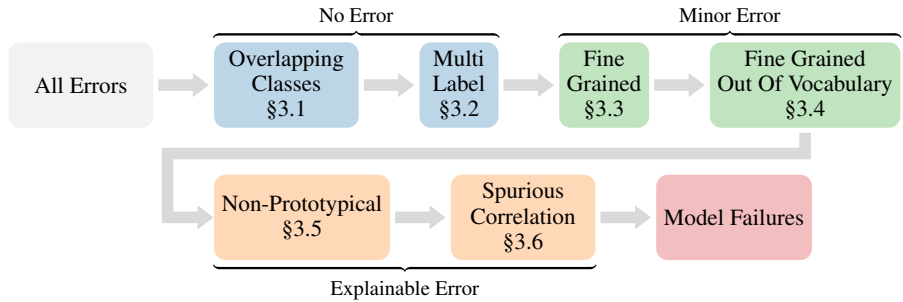
\begin{figure}[t]
    \vspace{-4mm}
    \centering
    \resizebox{0.86\textwidth}{!}{    
    \begin{tikzpicture}[scale=0.70]
    \tikzset{>=latex}
        \def \dx{1.3}
        \def \dxs{0.*\dx}
        \def \dy{-1.0}
        \def \dys{0*\dy}

        \def \aw{5pt}
        \def \tw{10pt}
        \def \tl{7pt}

        \def \dd{0.1}

        \colorlet{c_all}{black!05}
        \colorlet{c_no}{my-full-blue!30}
        \colorlet{c_minor}{my-full-green!30}
        \colorlet{c_exp}{my-full-orange!30}
        \colorlet{c_fail}{my-full-red!30}
        \colorlet{c_arrow}{black!15}

        \node[fill=c_all, rectangle, rounded corners=5pt, minimum width=2.2cm, minimum height=1.3cm, align=center] (all_errors) at (0, 0) {All Errors};

        \node[fill=c_no, rectangle, rounded corners=5pt, minimum width=2.2cm, minimum height=1.3cm, align=center, anchor=west] (overlap_errors) at ($(all_errors.east)+(\dx, \dys)$) {Overlapping \\ Classes\\ \cref{sec:class_overlap}};
       
        \node[fill=c_no, rectangle, rounded corners=5pt, minimum width=1.2cm, minimum height=1.3cm, align=center, anchor=west] (multi_label) at ($(overlap_errors.east)+(\dx, 0)$) {Multi \\ Label\\ \cref{sec:multi_label}};

        \node[fill=c_minor, rectangle, rounded corners=5pt, minimum width=1.2cm, minimum height=1.3cm, align=center, anchor=west] (fine_grained) at ($(multi_label.east)+(\dx, \dys)$) {Fine \\ Grained\\ \cref{sec:fine_grained}};

        \node[fill=c_minor, rectangle, rounded corners=5pt, minimum width=2.2cm, minimum height=1.3cm, align=center, anchor=west] (oov) at ($(fine_grained.east)+(\dx, 0)$) {Fine Grained \\ Out Of Vocabulary\\ \cref{sec:oov}};

        \node[fill=c_exp, rectangle, rounded corners=5pt, minimum width=2.2cm, minimum height=1.3cm, align=center, anchor=north west] (non_prot) at ($(overlap_errors.south west)+(0*\dx, \dy)$) {Non-Prototypical\\ \cref{sec:non_prototypical}};

        \node[fill=c_exp, rectangle, rounded corners=5pt, minimum width=2.2cm, minimum height=1.3cm, align=center, anchor=west] (spurious) at ($(non_prot.east)+(\dx, 0)$) {Spurious \\ Correlation\\ \cref{sec:spurious}};

        \node[fill=c_fail, rectangle, rounded corners=5pt, minimum width=2.2cm, minimum height=1.3cm, align=center, anchor=west] (failures) at ($(spurious.east)+(\dx, \dys)$) {Model Failures};

        \draw[-{Triangle[width=\tw,length=\tl]}, line width=\aw,c_arrow] ($(all_errors.east)+(\dd, 0 )$) -- ($(overlap_errors.west)+(-\dd, 0)$);
        \draw[-{Triangle[width=\tw,length=\tl]}, line width=\aw,c_arrow] ($(overlap_errors.east)+(\dd,0)$) -- ($(multi_label.west)+(-\dd,0)$);
        \draw[-{Triangle[width=\tw,length=\tl]}, line width=\aw,c_arrow] ($(multi_label.east)+(\dd,0)$) -- ($(fine_grained.west)+(-\dd,0)$);
        \draw[-{Triangle[width=\tw,length=\tl]}, line width=\aw,c_arrow] ($(fine_grained.east)+(\dd,0)$) -- ($(oov.west)+(-\dd,0)$);

        \draw[-{Triangle[width=\tw,length=\tl]}, line width=\aw,c_arrow] ($(oov.south)+(0,-\dd)$) -- ($(oov.south)+(0,0.55*\dy)$) -- ($(non_prot.north west)+(-1.0,-0.45*\dy)$) -- ($(non_prot.west)+(-1.0,-0.*\dy)$) -- ($(non_prot.west)+(0,0)$);
        \draw[-{Triangle[width=\tw,length=\tl]}, line width=\aw,c_arrow] ($(non_prot.east)+(\dd,0)$) -- ($(spurious.west)+(-\dd,0)$);
        \draw[-{Triangle[width=\tw,length=\tl]}, line width=\aw,c_arrow] ($(spurious.east)+(\dd,0)$) -- ($(failures.west)+(-\dd,0)$);

        \draw [decorate, decoration = {calligraphic brace}, line width=1pt] ($(overlap_errors.north west)+(0, \dd)$) -- node[pos=0.50,black,above=0.1] {\small No Error} ($(multi_label.north east)+(0, \dd)$);
        \draw [decorate, decoration = {calligraphic brace}, line width=1pt] ($(fine_grained.north west)+(0, \dd)$) -- node[pos=0.50,black,above=0.1] {\small Minor Error} ($(oov.north east)+(0, \dd)$);
        \draw [decorate, decoration = {calligraphic brace, mirror}, line width=1pt] ($(non_prot.south west)+(0, -\dd)$) -- node[pos=0.50,black,below=0.1] {\small Explainable Error} ($(spurious.south east)+(0, -\dd)$);

    \end{tikzpicture}
    }
    \vspace{-2.5mm}
    \caption{We first remove errors w.r.t. the original \IN labels caused by overlapping class definitions or missing multi-label annotations, yielding multi-label accuracy (\mla). We then, in this order, identify fine-grained misclassifications, fine-grained misclassifications where the true label of the main entity is not included in the \IN labelset, non-prototypical examples of a given class, and spurious correlations. This leaves us with severe model failures that are unexplained by our categorization.}%
    \label{fig:overview}
    \vspace{-4.5mm}
\end{figure}

Below, we first provide a brief overview of our pipeline (illustrated in \cref{fig:overview}), before discussing the different error types and how we identify them in more detail.

We consider errors due to overlapping class definitions (discussed in \cref{sec:class_overlap}) and missing multi-label annotations (\cref{sec:multi_label}) to be the least severe, as they can be interpreted as labeling errors rather than classification errors.
When a model predicts a class that is closely related to one of the labels, we call this a fine-grained classification error (\cref{sec:fine_grained}), as the model succeeds in recognizing the type of entity but fails during the fine-grained distinction.
When an image contains an entity that does not belong to any of the \IN classes and a model predicts a label that is closely related to this (out-of-vocabulary) class, we call this a fine-grained out-of-vocabulary error (\cref{sec:oov}).
We consider both types of fine-grained errors to be minor.
If a model fails on an image that shows a very non-prototypical instance of a given class, we call this a non-prototypical error (\cref{sec:non_prototypical}).
If a model predicts a class that commonly co-occurs with a ground-truth class but is not shown in the image, we attribute this error to spurious correlation and discuss it in \cref{sec:spurious}.
We denote both of these errors as explainable errors.
If an error can be attributed to multiple causes, we assign the least severe category and thus design our pipeline to consider error types in order of increasing severity (see \cref{fig:overview}).

Throughout this section, we include examples and trends for the kinds of errors we discuss. We generally describe these separately for images with a ground-truth class describing an organism (410 of 1000 classes) and artifacts (522 classes) (in line with previous work \citep{ShankarRMFRS20}), as we observe that these two groups of classes exhibit different error patterns. These two groups account for almost all \IN classes, with the remaining 68 classes being assigned the group ``other''.
Where we find interesting trends, we further distinguish models by their (pre-)training dataset ranging in size from one million (\IN) to multiple billion images (\INST, \LAION), their architecture, including a broad range of MLPs, CNNs, transformer-based, and hybrid models, and their (pre-) training method.
For the full details on all \nmodels models we consider, please refer to \cref{sec:setup}. We provide more examples of every error type in \cref{app:error-examples} and more detailed trends in \cref{sec:additional_results}.

\subsection{Class Overlap}\label{sec:class_overlap}
\begin{wrapfigure}[10]{r}{0.32\textwidth}
    \centering
    \vspace{-5mm}
    \scalebox{0.8}{
    \begin{tikzpicture}[scale=1.0]
    \tikzset{>=latex}
        \def \dx{1.3}
        \def \dxs{0.*\dx}
        \def \dy{-1.0}
        \def \dys{0*\dy}

        \def \aw{5pt}
        \def \tw{10pt}
        \def \tl{7pt}

        \def \dd{0.1}

        \colorlet{c_tusk}{black!15}
        \colorlet{c_af}{my-full-blue!40}
        \colorlet{c_ind}{my-full-green!40}

        \coordinate (tusker) at (0,0);
        \coordinate (african) at (-1.5,0);
        \coordinate (indian) at (1.60, 0.35);

        \draw[draw=none, fill=c_tusk] (tusker) ellipse (2.5cm and 1.5cm);
        \node at (tusker) {\inc{tusker}};

        \draw[draw=none, fill=c_af, opacity=0.7] (african) ellipse (0.9cm and 0.9cm);
        \node[align=center] at (african) {\inc{african}\\\inc{elephant}};

        \draw[draw=none, fill=c_ind, opacity=0.7] (indian) ellipse (0.9cm and 0.9cm);
        \node[align=center] at (indian) {\inc{indian}\\\inc{elephant}};

    \end{tikzpicture}

    
    }
    \caption{Venn-Diagram of the \inc{tusker}, \inc{indian elephant}, and \inc{african elephant} classes.}
    \label{fig:tusker}
\end{wrapfigure}
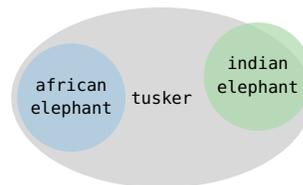

Prior work has established that a small number of \IN classes exhibit extreme overlap \citep{NorthcuttJC21,VasudevanCLFR22}. We illustrate one such example in \cref{fig:tusker}: \inc{tusker} is defined as ``an animal with tusks''\footnote{According to the \href{https://www.merriam-webster.com/dictionary/tusker}{Merriam-Webster Dictionary}.},
which describes a strict superset of \inc{african elephant} and has a significant overlap with \inc{indian elephant} (females have short and sometimes no tusks).
To avoid penalizing a model for correct predictions that do not match the ground truth, we consider all predictions describing a \emph{superset or equivalent} of the ground-truth class to be correct. For example, we accept \inc{tusker} for an image labeled \inc{african elephant}, but not vice-versa, as the latter might actually show a boar. We follow the mappings of equivalence and containment from \citet{VasudevanCLFR22} and refer to their App. C for full details.

\begin{wrapfigure}[11]{r}{0.565\textwidth}
    \centering
    \vspace{-4mm}
    \begin{minipage}[t]{0.48 \linewidth}
        \centering
        \includegraphics[width=0.85\linewidth]{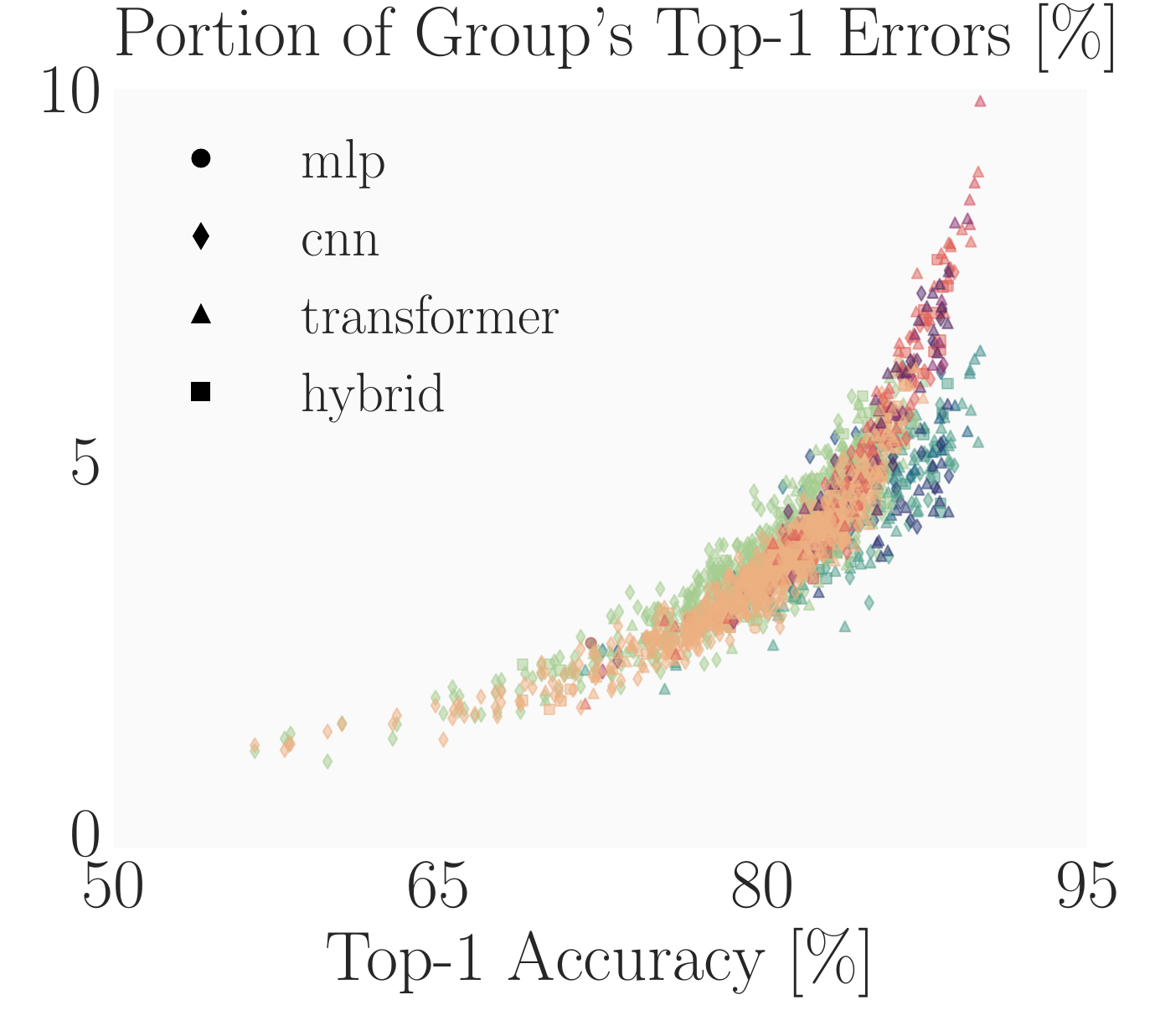}
    \end{minipage}
    \hfil
    \begin{minipage}[t]{0.48 \linewidth}
        \centering
        \includegraphics[width=0.92\linewidth]{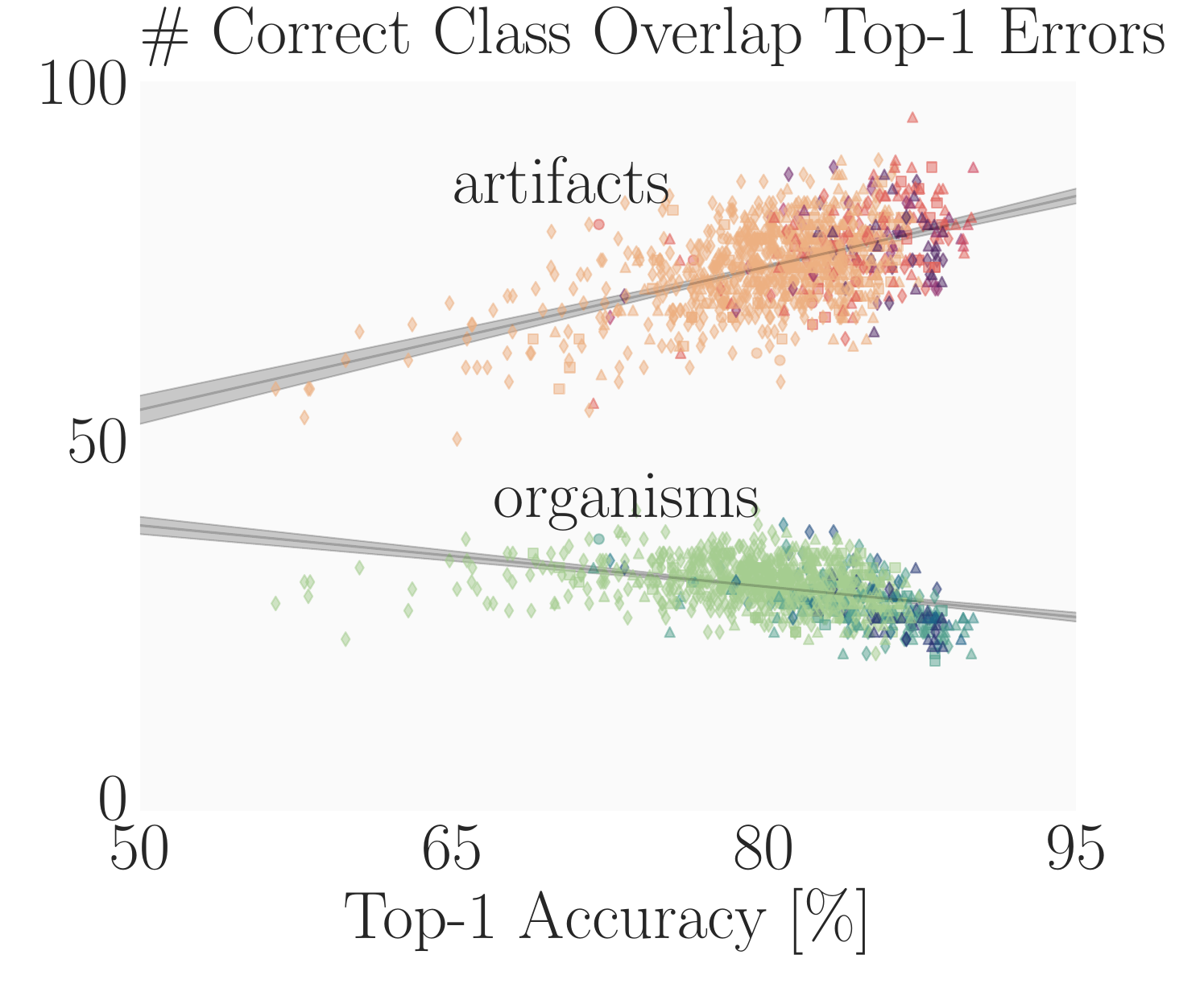}
    \end{minipage}
    \vspace{-2mm}
    \caption{Portion (left) and number (right) of \toa errors caused by \emph{class overlap} by group -- organisms (green) and artifacts (red). A 95\% confidence interval linear fit is shown on the right.}
    \label{fig:class_overlap}
\end{wrapfigure}
In \cref{fig:class_overlap}, we visualize the portion and number of \toa errors identified to be correct, separating images with a ground-truth class describing an organism (green hues) and artifacts (red hues) and encoding the pre-training dataset size by assigning darker colors to larger datasets.
While trends for artifacts and organisms seem very similar for portions of \toa errors, we observe a clear difference when looking at their absolute numbers. There, two competing effects are at play. With increasing accuracy, more samples get classified correctly, including as overlapping classes, thus increasing the number of such errors. On the other hand, more accurate models leverage labeling biases to pick the ``correct'' among equivalent classes \citep{TsiprasSEIM20}, thus reducing the number of such errors. While the former effect seems to dominate for artifacts, the latter dominates for organisms.

\subsection{Missing Multi-Label Annotations}\label{sec:multi_label}
\begin{wrapfigure}[10]{r}{0.26\textwidth}
    \vspace{-4mm}
    \centering
    \includegraphics[width=0.8\linewidth]{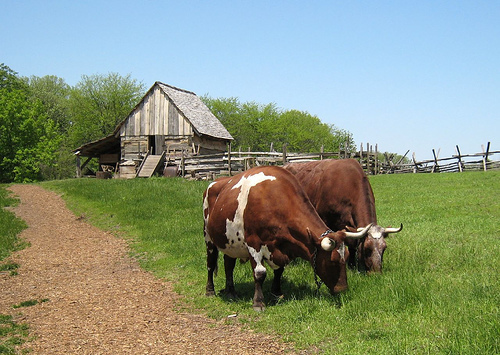}
    \caption{Image with label \inc{ox}, but also showing the classes \inc{barn} and \inc{fence}. Example from \citet{YunOHHCC21}}
    \label{fig:multi_label_example}
\end{wrapfigure}
While the \IN dataset is annotated with a single label per image, many images contain multiple entities \citep{BeyerHKZO20,ShankarRMFRS20,TsiprasSEIM20}. Consequently, a model predicting the class of any such entity, different from the original \IN label, is considered incorrect when computing top-1 accuracy.
For example, the image shown in \cref{fig:multi_label_example} contains \inc{ox}, \inc{barn}, and \inc{fence} while only being labeled \inc{ox}.
To remedy this issue, multi-label accuracy (MLA) considers all shown classes, according to some multi-label annotation, to be correct. In this work, we use the annotations collected by \citet{ShankarRMFRS20} and improved by \citet{VasudevanCLFR22} combined with the mean class-wise accuracy definition of MLA \citep{ShankarRMFRS20}.

\begin{wrapfigure}[11]{r}{0.50\textwidth}
    \centering
    \vspace{-4mm}
    \begin{minipage}[t]{0.49 \linewidth}
        \centering
        \includegraphics[width=0.93\linewidth]{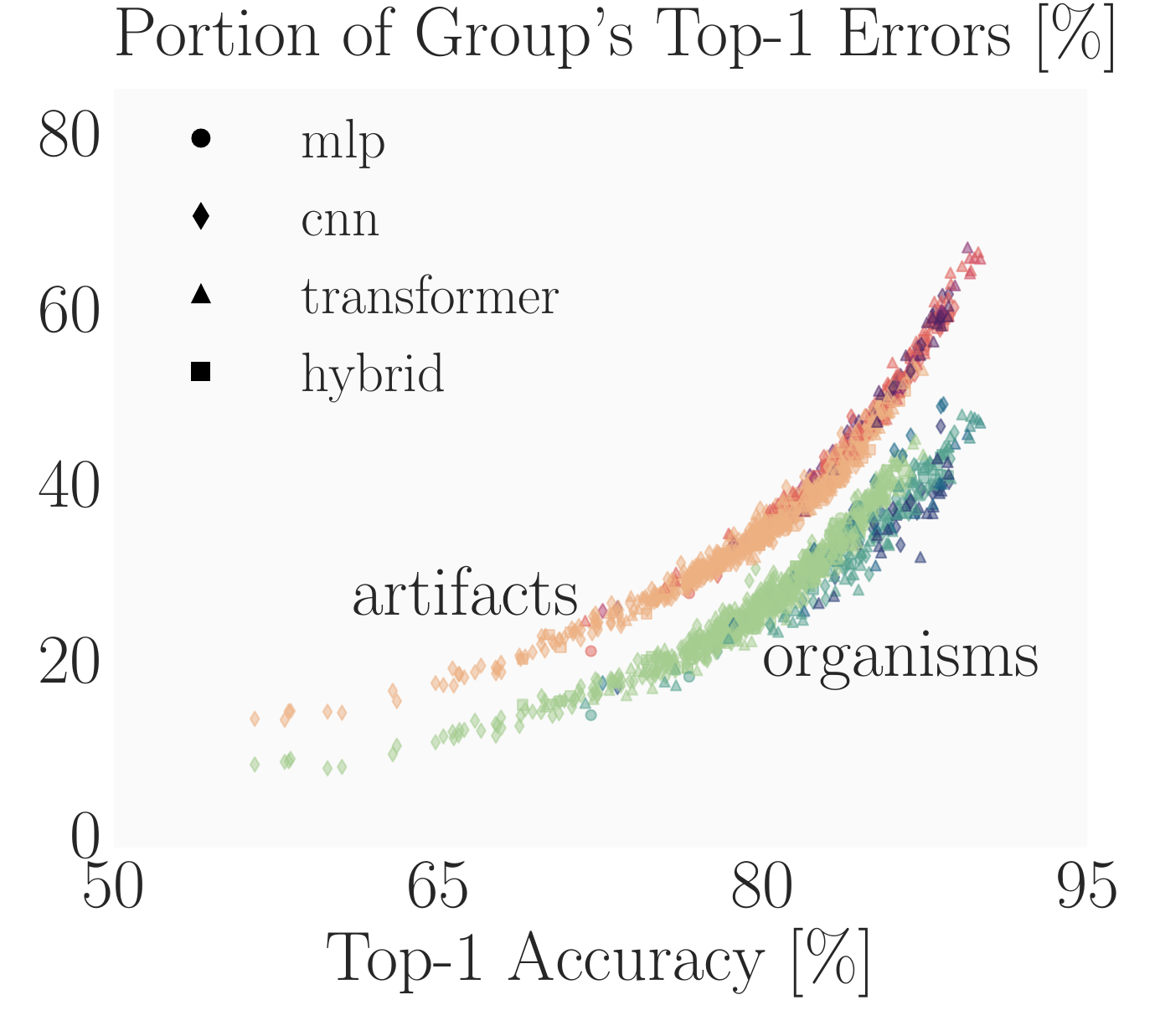}
    \end{minipage}
    \hfil
    \begin{minipage}[t]{0.49 \linewidth}
        \centering
        \includegraphics[width=0.98\linewidth]{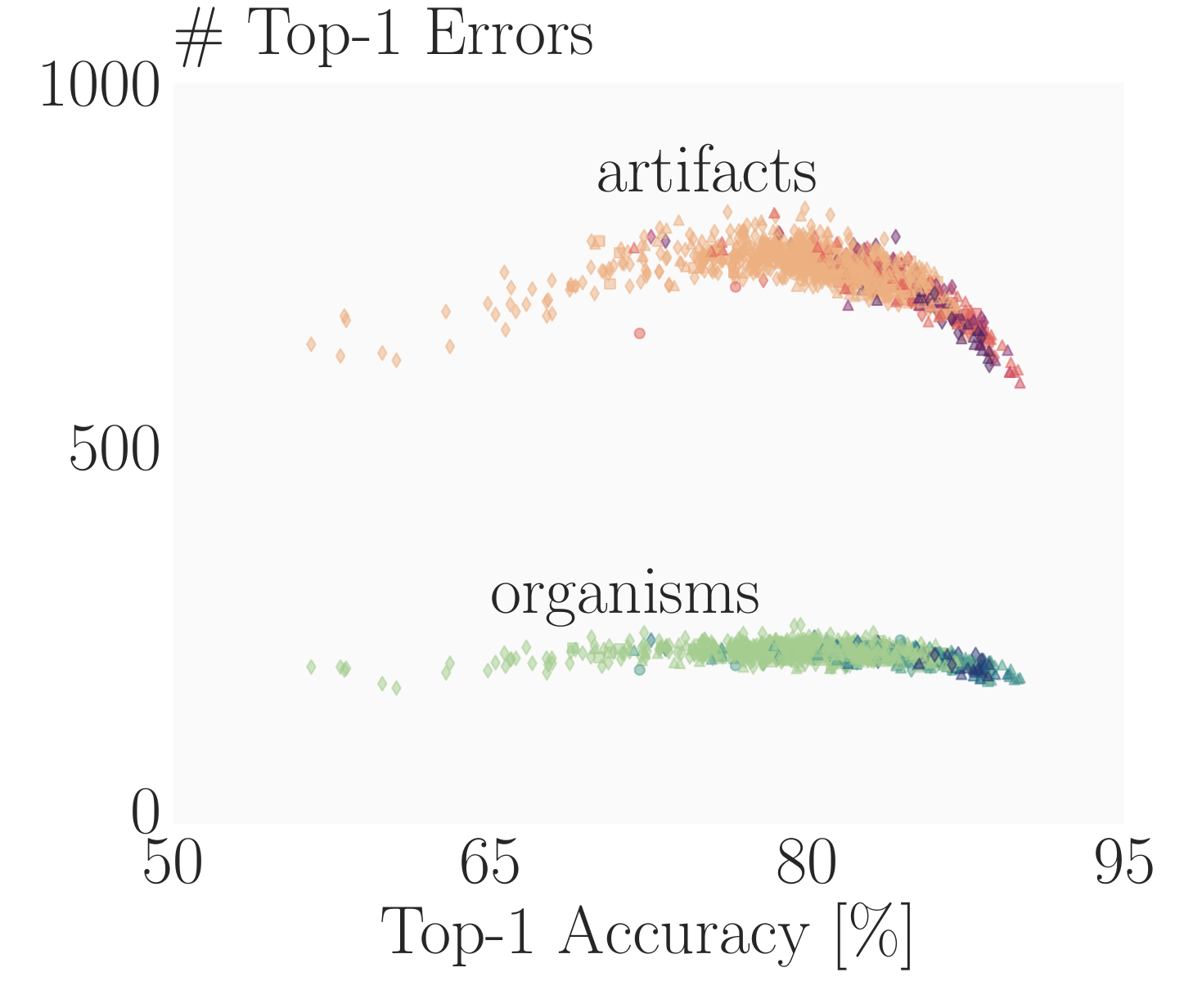}
    \end{minipage}
    \vspace{-5mm}
    \caption{Portion (left) and number (right) of \toa errors caused by \emph{missing multi-label annotations} by group -- organisms (green) and artifacts (red).}
    \label{fig:multi_label_annotations}
\end{wrapfigure}

We visualize the portion and number of top-1 errors that turn out to be correct under multi-label evaluation in \cref{fig:multi_label_annotations}. When looking at the portion of errors (\cref{fig:multi_label_annotations} left), we observe a similar trend for organisms and artifacts with missing multi-label annotations accounting for an increasing portion of \toa errors as \mla increases and artifacts consistently more affected than organisms. When looking at the absolute numbers, we again observe the number of errors explained by multi-label annotations to first increase with accuracy as models become more precise, before decreasing again as models start to leverage labeling biases \citep{TsiprasSEIM20}.
As missing multi-label annotations explain up to $60\%$ of model errors, we henceforth use MLA instead of top-1 accuracy as the reference for model comparison.

\subsection{Fine-Grained Classification Errors}\label{sec:fine_grained}
\begin{wrapfigure}[10]{r}{0.27\textwidth}
    \centering
    \vspace{-4mm}
    \includegraphics[width=0.8\linewidth]{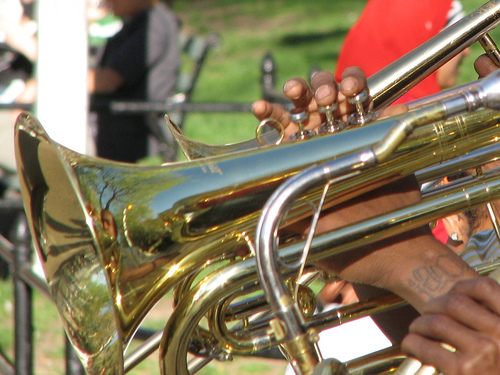}
    \vspace{-1mm}
    \caption{Image labeled \inc{french horn}, but predicted to show a \inc{cornet}.}
    \label{fig:fine_grained_example}
\end{wrapfigure}

Many of the \IN classes and especially the organism classes are very similar, making their fine-grained distinction challenging even for trained humans. For example, \cref{fig:fine_grained_example} is classified to \inc{cornet} while actually showing a \inc{french horn}, both brass wind instruments.
While these errors can be corrected for the relatively small validation set using expert human judgment \citep{HornBFHBIPB15,VasudevanCLFR22}, the much larger training set remains uncorrected. In this light, we believe a comprehensive model evaluation and comparison, should consider the failure to distinguish between very similar classes to be less severe than the failure to distinguish very different classes.

\begin{wrapfigure}[11]{r}{0.50\textwidth}
    \centering
    \vspace{-4mm}
    \begin{minipage}[t]{0.49 \linewidth}
        \centering
        \includegraphics[width=1.03\linewidth]{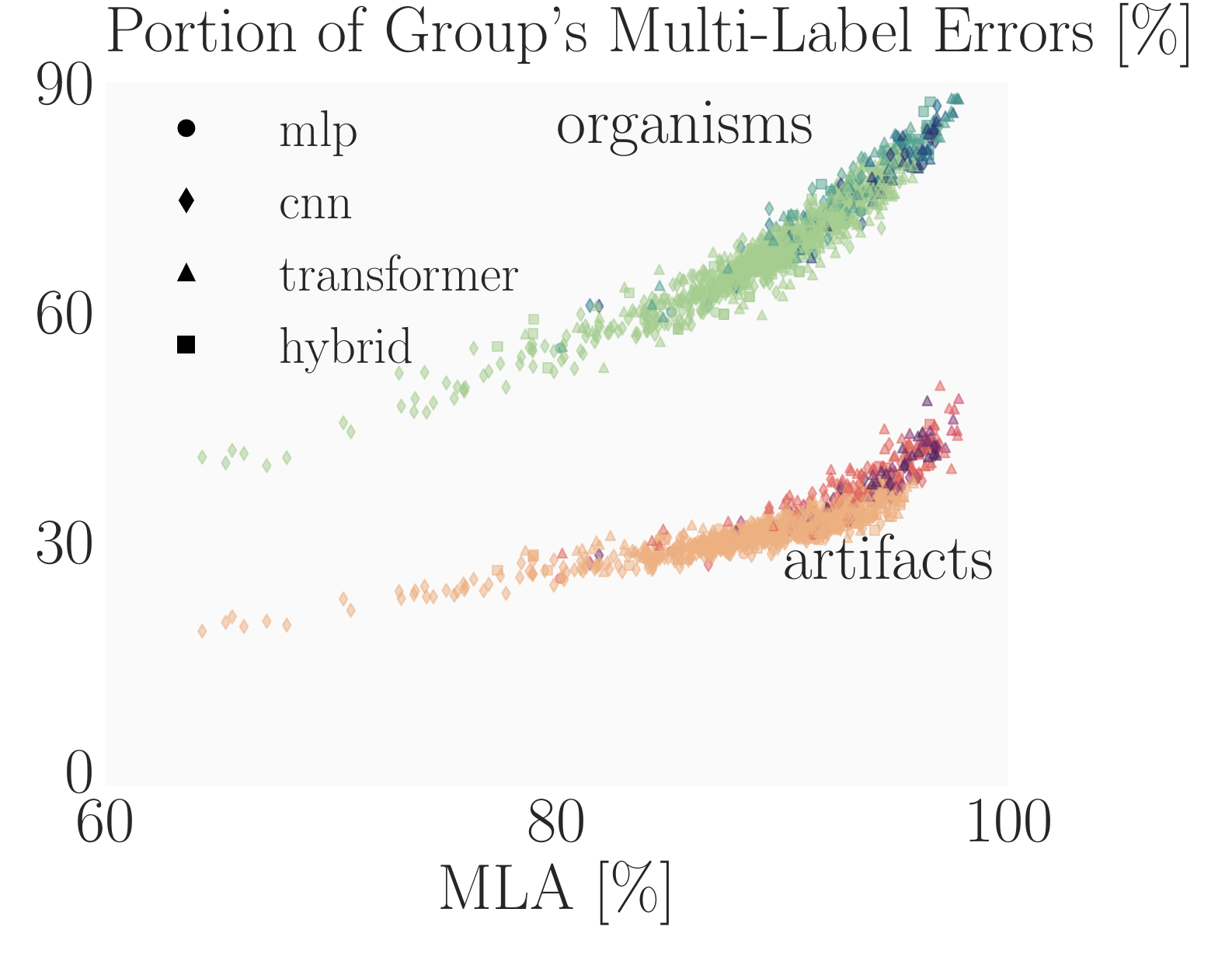}
    \end{minipage}
    \hfil
    \begin{minipage}[t]{0.49 \linewidth}
        \centering
        \includegraphics[width=0.95\linewidth]{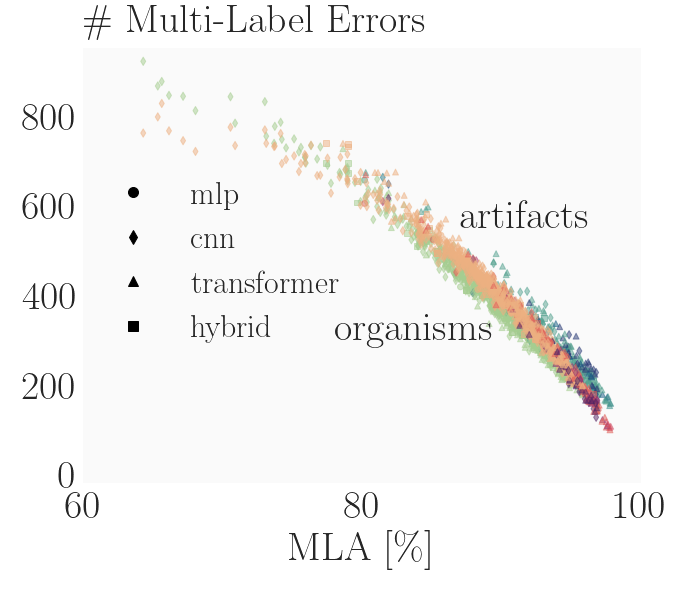}
    \end{minipage}
    \vspace{-5mm}
    \caption{Portion (left) and number (right) of \mla errors caused by \emph{fine-grained misclassifications} by group -- organisms (green) and artifacts (red).}
    \label{fig:fine_grained}
\end{wrapfigure}
To automatically detect such fine-grained classification errors, we manually review all 1000 \IN classes and define semantically similar superclasses guided by whether an untrained human could reasonably confuse them. We obtain $\nscls$ superclasses, containing between $2$ and $31$ classes with an average of $6.7$ and a median of $4$. An additional $74$ classes are not part of any superclass. $50$ superclasses contain organisms ($9.8$ on average) and $101$ contain artifacts ($5.3$ on average). %
We visualize the portion and number of fine-grained errors in \cref{fig:fine_grained}. As expected, we observe significantly more fine-grained errors for organisms than for artifacts, explaining up to $88\%$ and $50\%$ of multi-label errors, respectively. %

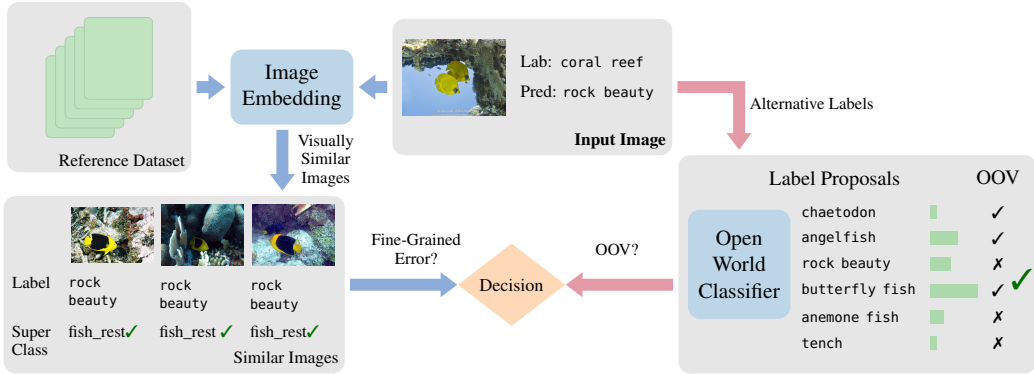
\begin{figure}[t]
    \vspace{-4mm}
    \centering
    \resizebox{\textwidth}{!}{
    \begin{tikzpicture}[scale=0.70]
    \tikzset{>=latex}
        \def \dx{1.8}
        \def \dxs{0.6*\dx}
        \def \dy{3.3}
        \def \dys{0.16*\dy}

        \def \ddx{0.2}

        \def \ddy{0.55}

        \def \aw{5pt}
        \def \tw{10pt}
        \def \tl{7pt}

        \def \dd{0.1}

        \colorlet{c_model}{my-full-blue!30}
        \colorlet{c_input}{black!05}
        \colorlet{c_ds}{my-full-green!30}
        \colorlet{c_dsb}{c_ds!90!black}
        \colorlet{c_bg}{black!10}
        \colorlet{c_prob}{my-full-green!40}

        \colorlet{c_decision}{my-full-orange!30}
        \colorlet{c_oov_arrow}{my-dark-red!45}
        \colorlet{c_fg_arrow}{my-dark-blue!45}

        \node[fill=c_model, rectangle, rounded corners=5pt, minimum width=1.8cm, minimum height=1.1cm, align=center] (clip) at (0, 0) {\small Image \\\small Embedding};

        \node[fill=c_bg, rectangle, rounded corners=5pt, minimum width=2.7cm, minimum height=2.5cm, anchor=east] (ds) at  ($(clip.west)+(-\dx+1.0, 0)$) {};
        \node[anchor=south east, scale=0.8] at  ($(ds.south east)+(0, -0.00)$) {\small Reference Dataset};
        \node[fill=c_ds, draw=c_dsb, rectangle, rounded corners=2pt, minimum width=1.0cm, minimum height=1.2cm, anchor=center] (input_1) at ($(ds)+(-2*\ddx, -2*\ddx+0.2)$) {};
        \node[fill=c_ds, draw=c_dsb, rectangle, rounded corners=2pt, minimum width=1.0cm, minimum height=1.2cm, anchor=center] (input_2) at ($(input_1)+(\ddx, \ddx)$) {};
        \node[fill=c_ds, draw=c_dsb, rectangle, rounded corners=2pt, minimum width=1.0cm, minimum height=1.2cm, anchor=center] (input_3) at ($(input_2)+(\ddx, \ddx)$) {};
        \node[fill=c_ds, draw=c_dsb, rectangle, rounded corners=2pt, minimum width=1.0cm, minimum height=1.2cm, anchor=center] (input_4) at ($(input_3)+(\ddx, \ddx)$) {};
        \node[fill=c_ds, draw=c_dsb, rectangle, rounded corners=2pt, minimum width=1.0cm, minimum height=1.2cm, anchor=center] (input_5) at ($(input_4)+(\ddx, \ddx)$) {};

        \node[fill=c_bg, rectangle, rounded corners=5pt, minimum width=4.1cm, minimum height=2.cm, anchor=west] (image_bg) at  ($(clip.east)+(\dx-1, 0)$) {};
        \node[anchor=south east, scale=0.8] at  ($(image_bg.south east)+(0, -0.00)$) {\small \bf Input Image};
        \node (image) at ($(image_bg.west)+(1.3, 0.2)$) {\includegraphics[width=1.5cm]{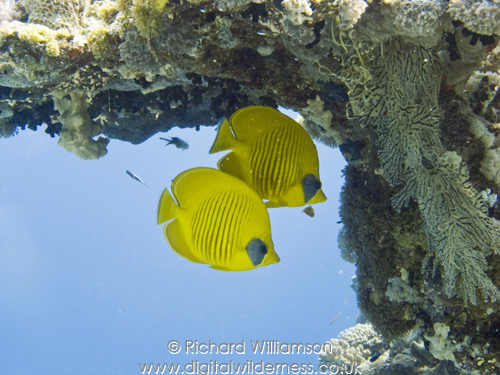}};
        \node[anchor=south east, scale=0.8, anchor=west, align=left] at ($(image.east)+(0, -0.00)$) {\small Lab: $\,$\inc{coral reef} \\[1ex] \small Pred: \inc{rock beauty}};

        \node[fill=c_bg, rectangle, rounded corners=5pt, minimum width=5.0cm, minimum height=2.6cm, anchor=north east] (retrieved_bg) at  ($(clip.south east)+(-0.1*\dx, -2.8*\dys)$) {};
        \node[anchor=south east, scale=0.8] at  ($(retrieved_bg.south east)+(0, -0.00)$) {\small Similar Images};
        \node (retrieved_2) at ($(retrieved_bg.north)+(0.6, -0.825)$) {\includegraphics[width=1.2cm]{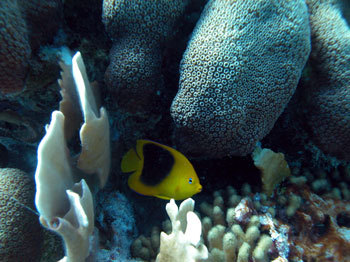}};
        \node (retrieved_1) at ($(retrieved_2.west)+(-0.8*\dxs, 0*\dys)$) {\includegraphics[width=1.2cm]{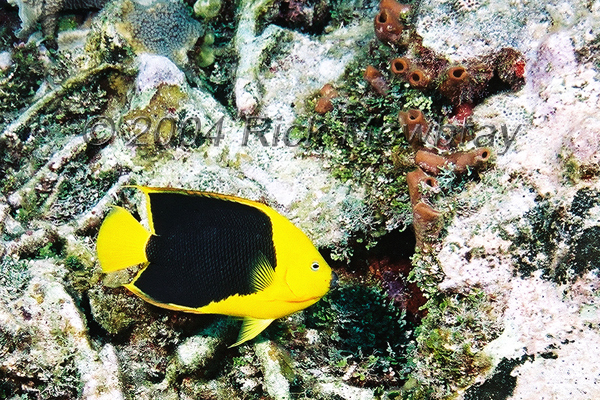}};
        \node (retrieved_3) at ($(retrieved_2.east)+(0.8*\dxs, -0*\dys)$) {\includegraphics[width=1.2cm]{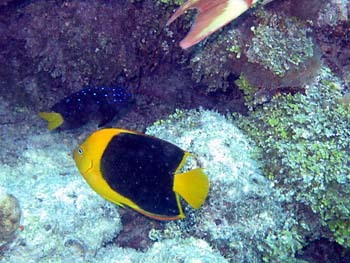}};

        \node[minimum width=1.1cm,anchor=north west, align=left, scale=0.7] (r_labels_2) at ($(retrieved_2.south west)+(0*\ddx, -0.0)$) {\inc{rock}\\ \inc{beauty}};
        \node[minimum width=1.1cm,anchor=north west, align=left, scale=0.7] (r_labels_1) at ($(retrieved_1.south west)+(0*\ddx, -0.0)$) {\inc{rock}\\ \inc{beauty}};
        \node[minimum width=1.1cm,anchor=north west, align=left, scale=0.7] (r_labels_3) at ($(retrieved_3.south west)+(0*\ddx, -0.0)$) {\inc{rock}\\ \inc{beauty}};
        \node[minimum width=1.1cm,anchor=north west, align=left, scale=0.7] (r_labels_2_a) at ($(r_labels_2.south west)+(0*\ddx, -0.05)$) {fish\_rest \textcolor{my-dark-green}{\ding{51}}};
        \node[minimum width=1.1cm,anchor=north west, align=left, scale=0.7] (r_labels_1_a) at ($(r_labels_1.west |- r_labels_2_a.north)$) {fish\_rest\textcolor{my-dark-green}{\ding{51}}};
        \node[minimum width=1.1cm,anchor=north west, align=left, scale=0.7] (r_labels_3_a) at ($(r_labels_3.west |- r_labels_2_a.north)$) {fish\_rest\textcolor{my-dark-green}{\ding{51}}};

        \node[minimum width=1.1cm,anchor=north east, align=left, scale=0.7] (r_labels_2) at ($(r_labels_1.north west)+(-0.1, -0.00)$) {Label};
        \node[minimum width=1.1cm,anchor=north east, align=left, scale=0.7] (r_labels_2) at ($(r_labels_1_a.north west)+(-0.1, -0.0)$) {Super\\Class};

        \node[fill=c_bg, rectangle, rounded corners=5pt, minimum width=5.25cm, minimum height=3.2cm, anchor=south west] (class_bg) at  ($(image_bg.east |- retrieved_bg.south)+(0.1*\dxs, 0)$) {};
        \node[fill=c_model, rectangle, rounded corners=5pt, minimum width=1.5cm, minimum height=1.6cm, align=center, anchor=west] (class) at ($(class_bg.west)+(0.2, 0)$) {\small Open \\ \small World\\\small Classifier};

        \node[anchor=west, align=left, scale=0.85] (p_label_1) at ($(class.east)+(-0.6, 3.2*\ddy)$) {Label Proposals};
        \node[anchor=south west, align=left, scale=0.85] (p_label_1_oov) at ($(p_label_1.south east)+(1.3,0.1)$) {OOV};
        \node[minimum height=0.8cm, anchor=west, scale=0.8, align=left, text width=2.1cm] (p_label_2) at ($(class.east)+(0.4 * \ddx, 2*\ddy)$) {\small\inc{chaetodon}};
        \node[minimum height=0.8cm, anchor=west, scale=0.8, align=left, text width=2.1cm] (p_label_3) at ($(class.east)+(0.4 * \ddx, 1*\ddy)$) {\small\inc{angelfish}};
        \node[minimum height=0.8cm, anchor=west, align=left, scale=0.8, text width=2.1cm] (p_label_4) at ($(class.east)+(0.4 * \ddx, 0*\ddy)$) {\small\inc{rock beauty}};
        \node[minimum height=0.8cm, anchor=west, align=left, scale=0.8, text width=2.1cm] (p_label_5) at ($(class.east)+(0.4 * \ddx, -1*\ddy)$) {\small\inc{butterfly fish}};
        \node[minimum height=0.8cm, anchor=west, align=left, scale=0.8, text width=2.1cm] (p_label_6) at ($(class.east)+(0.4 * \ddx, -2*\ddy)$) {\small\inc{anemone fish}};
        \node[minimum height=0.8cm, anchor=west, align=left, scale=0.8, text width=2.1cm] (p_label_7) at ($(class.east)+(0.4 * \ddx, -3*\ddy)$) {\small\inc{tench}};

        \node[fill=c_prob, rectangle, minimum width=0.1cm, minimum height=0.2cm, inner sep=0.1pt, anchor=west] (p_bar_2) at ($(p_label_2.east)+(0.75 * \ddx, 0)$) {};
        \node[fill=c_prob, rectangle, minimum width=0.4cm, minimum height=0.2cm, inner sep=0.1pt, anchor=west] (p_bar_3) at ($(p_label_3.east)+(0.75 * \ddx, 0)$) {};
        \node[fill=c_prob, rectangle, minimum width=0.3cm, minimum height=0.2cm, inner sep=0.1pt, anchor=west] (p_bar_4) at ($(p_label_4.east)+(0.75 * \ddx, 0)$) {};
        \node[fill=c_prob, rectangle, minimum width=0.7cm, minimum height=0.2cm, inner sep=0.1pt, anchor=west] (p_bar_5) at ($(p_label_5.east)+(0.75 * \ddx, 0)$) {};
        \node[fill=c_prob, rectangle, minimum width=0.2cm, minimum height=0.2cm, inner sep=0.1pt, anchor=west] (p_bar_7) at ($(p_label_6.east)+(0.75 * \ddx, 0)$) {};
        \node[fill=c_prob, rectangle, minimum width=0.1cm, minimum height=0.2cm, inner sep=0.1pt, anchor=west] (p_bar_6) at ($(p_label_7.east)+(0.75 * \ddx, 0)$) {};

        \node[anchor=center, align=left, scale=0.7] (oov_2) at ($(p_bar_2 -| p_label_1_oov)+(0, -0.00)$) {\ding{51}};
        \node[anchor=center, align=left, scale=0.7] (oov_3) at ($(p_bar_3 -| p_label_1_oov)+(0, -0.00)$) {\ding{51}};
        \node[anchor=center, align=left, scale=0.7] (oov_4) at ($(p_bar_4 -| p_label_1_oov)+(0, -0.00)$) {\ding{55}};
        \node[anchor=center, align=left, scale=0.7] (oov_5) at ($(p_bar_5 -| p_label_1_oov)+(0, -0.00)$) {\ding{51}};
        \node[anchor=center, align=left, scale=0.7] (oov_6) at ($(p_bar_6 -| p_label_1_oov)+(0, -0.00)$) {\ding{55}};
        \node[anchor=center, align=left, scale=0.7] (oov_7) at ($(p_bar_7 -| p_label_1_oov)+(0, -0.00)$) {\ding{55}};

        \node[anchor=west, align=left, scale=1.2] (checkmark) at ($(oov_5.east)+(-0.2, 0.25)$) {\textcolor{my-dark-green}{\ding{51}}};

        \node[fill=c_decision, diamond, minimum width=1.2cm, minimum height=1.2cm, align=center, anchor=center, scale=0.75, aspect=1.3] (decision) at ($(retrieved_bg.east)!0.5!(class_bg.west |- retrieved_bg.east)+(0, 0)$) {Decision};

        \draw[-{Triangle[width=\tw,length=\tl]}, line width=\aw,c_fg_arrow] ($(ds.east)+(\dd, 0 )$) -- ($(clip.west)+(-\dd, 0)$);
        \draw[-{Triangle[width=\tw,length=\tl]}, line width=\aw,c_fg_arrow] ($(image_bg.west)+(-\dd, 0 )$) -- ($(clip.east)+(\dd, 0)$);

        \draw[-{Triangle[width=\tw,length=\tl]}, line width=\aw,c_oov_arrow] ($(image_bg.east)+(\dd, 0 )$) -- (image_bg.east -| class.north) -- ($(class_bg.north -| class)+(0, \dd)$);
        \node [anchor=north west, scale=0.7, align=center] () at ($(image_bg.east -| class.north)+(0.1, - 0.1)$) {Alternative Labels};

        \draw[-{Triangle[width=\tw,length=\tl]}, line width=\aw,c_fg_arrow] ($(clip.south)+(-0.2, -\dd )$) -- ($(retrieved_bg.north -| clip)+(-0.2, \dd)$);
        \node [anchor=west, scale=0.7, align=center] () at ($(clip.south)!0.5!(retrieved_bg.north -| clip)+(-0.0, 0)$) {Visually  \\ Similar \\Images};

        \draw[-{Triangle[width=\tw,length=\tl]}, line width=\aw,c_oov_arrow] ($(class_bg.west |- decision)+(-\dd, 0 )$) -- ($(decision.east)+(0, 0)$);
        \draw[-{Triangle[width=\tw,length=\tl]}, line width=\aw,c_fg_arrow] ($(retrieved_bg.east)+(\dd, 0 )$) -- ($(decision.west)+(0, 0)$);
        
        \node[anchor=east, align=center, scale=0.7] at ($(decision.west)+(0.2, 0.75)$) {Fine-Grained \\Error?};
        \node[anchor=west, align=center, scale=0.7] at ($(decision.east)+(0.5, 0.75)$) {OOV?};

    \end{tikzpicture}
    }
    \caption{To confirm a fine-grained out-of-vocabulary error we proceed as follows. Given a misclassified input image, we first retrieve the 10 most visually similar images from the \IN train set. If the superclass of any of these images matches the model prediction, we conclude that an entity with a class similar to our prediction is shown in the image. To confirm that this entity is indeed out-of-vocabulary, we first collect a set of proposal labels from WordNet \citep{miller1995wordnet} which are partially in and out of vocabulary. Then, we use an Open World Classifier to score each of the proposed labels. If the highest scored among these is not included in the \IN labelset, we consider the analyzed error to be OOV.}
    \label{fig:oov_method}
    \vspace{-4mm}
\end{figure}

\subsection{Fine-Grained Out-of-Vocabulary Errors}\label{sec:oov}
\begin{wrapfigure}[10]{r}{0.27\textwidth}
    \centering
    \vspace{-4mm}
    \includegraphics[width=0.85\linewidth]{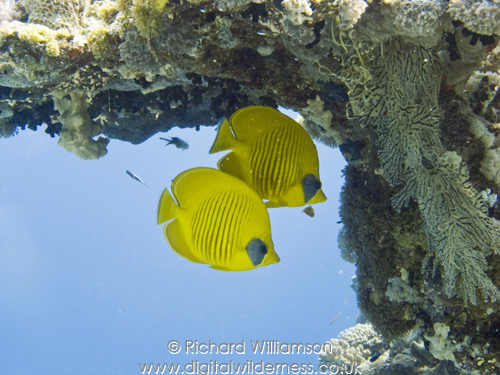}
    \vspace{-1mm}
    \caption{Image labeled \inc{coral reef}, but prediction \inc{rock beauty} (fish).}
    \label{fig:fine_grained_oov_example}
\end{wrapfigure}

Often images contain entities that do not belong to any \IN class, we call these out-of-vocabulary. For example, \cref{fig:fine_grained_oov_example} shows an image of two blue-cheeked butterflyfish, which is a class not part of the \IN labelset. Instead, the target label is \inc{coral reef}, describing the background of the image. The classifier predicts \inc{rock beauty}, which is a \WN child of \inc{butterflyfish} (despite being part of the angelfish and not the butterflyfish family) and the closest class in the \IN labelset to the blue-cheeked butterflyfish.
While an optimal classifier could be expected to always predict a class from the intersection of the contained entities and the \IN labelset, the following is often observed in practice \citep{VasudevanCLFR22}. The classifier tries to classify a prominent entity in the image, but as it is not part of the \IN labelset, it predicts a closely related class instead. We argue, that these cases should be treated at least as leniently as a fine-grained classification error, but they are harder to detect.
As the entity the model tried to classify can not be assigned any \IN label, it will not be included in any multi-label annotation and can thus not be captured by defining groups of closely related classes.

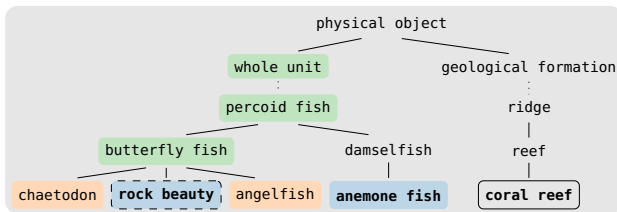
\begin{wrapfigure}[12]{r}{0.60\textwidth}
    \vspace{-4mm}
    \centering
    \resizebox{\linewidth}{!}{
    \begin{tikzpicture}[scale=0.70]
    \tikzset{>=latex}
        \def \dx{1.8}
        \def \dy{3.3}
        \def \dd{0.4}
        \def \ddd{0.05}

        \def \ddx{0.4}
        \def \ddy{0.3}

        \colorlet{c_sc}{my-full-blue!30}
        \colorlet{c_wns}{my-full-orange!30}
        \colorlet{c_wna}{my-full-green!30}
        \colorlet{c_bg}{black!10}

        \node[fill=c_bg, rectangle, rounded corners=5pt, minimum width=8.6cm, minimum height=2.9cm, anchor=east] (bg) at (0, 0) {};

        \node[anchor=north, scale=0.8, align=center] (pobject) at ($(bg.north)+(1.35, -0.10)$) {\small \inc{physical object}};
        
        \node[anchor=north east, fill=c_wna, rounded corners=2pt, scale=0.8, align=center] (wunit) at ($(pobject.south west)+(\ddx, -\ddy)$) {\small \inc{whole unit}};
        \draw[-] ($(pobject.south)+(-\dd, 0 )$) -- ($(wunit.north)+(\dd, \ddd)$);

        \node[anchor=north, fill=c_wna, rounded corners=2pt, scale=0.8, align=center] (pfish) at ($(wunit.south)+(0, -\ddy)$) {\small \inc{percoid fish}};
        \draw[dotted] ($(wunit.south)+(0, -\ddd)$) -- ($(pfish.north)+(0, \ddd)$);

        \node[anchor=north east, fill=c_wna, rounded corners=2pt, scale=0.8, align=center] (bfish) at ($(pfish.south west)+(0.8*\ddx, -\ddy)$) {\small \inc{butterfly fish}};
        \draw[-] ($(pfish.south)+(-\dd, -\ddd)$) -- ($(bfish.north)+(\dd, \ddd)$);

        \node[anchor=north east, fill=c_wns, rounded corners=2pt, scale=0.8, align=center, minimum height=4.9mm] (chaetodon) at ($(bfish.south west)+(0.3*\ddx, -\ddy)$) {\small \inc{chaetodon}};
        \draw[-] ($(bfish.south)+(-\dd, -\ddd)$) -- ($(chaetodon.north)+(\dd, \ddd)$);

        \node[anchor=north, draw=black, style=dashed, fill=c_sc, rounded corners=2pt, scale=0.8, align=center, minimum height=4.9mm] (rbeauty) at ($(bfish.south)+(0, -\ddy)$) {\small \bf \inc{rock beauty}};
        \draw[-] ($(bfish.south)+(0, -\ddd)$) -- ($(rbeauty.north)+(0, \ddd)$);

        \node[anchor=north west, fill=c_wns, rounded corners=2pt, scale=0.8, align=center, minimum height=4.9mm] (angelfish) at ($(bfish.south east)+(-0.3*\ddx, -\ddy)$) {\small \inc{angelfish}};
        \draw[-] ($(bfish.south)+(\dd, -\ddd)$) -- ($(angelfish.north)+(-\dd, \ddd)$);

        \node[anchor=north west, scale=0.8, align=center] (damselfish) at ($(pfish.south east)+(-0.0*\ddx, -\ddy)$) {\small \inc{damselfish}};
        \draw[-] ($(pfish.south)+(\dd, -\ddd)$) -- ($(damselfish.north)+(-\dd, \ddd)$);

        \node[anchor=north, fill=c_sc, rounded corners=2pt, scale=0.8, align=center, minimum height=4.9mm] (afish) at ($(damselfish.south)+(0, -\ddy-0.06)$) {\small \bf \inc{anemone fish}};
        \draw[-] ($(damselfish.south)+(0, 0)$) -- ($(afish.north)+(0, \ddd)$);

        \node[anchor=north west, scale=0.8, align=center] (gformation) at ($(pobject.south east)+(-\ddx, -\ddy)$) {\small \inc{geological formation}};
        \draw[-] ($(pobject.south)+(\dd, 0 )$) -- ($(gformation.north)+(-\dd, 0)$);

        \node[anchor=north, scale=0.8, align=center] (ridge) at ($(gformation.south)+(0, -\ddy+0.05)$) {\small \inc{ridge}};
        \draw[dotted] ($(gformation.south)+(0, 0 )$) -- ($(ridge.north)+(0, 0)$);

        \node[anchor=north, scale=0.8, align=center] (reef) at ($(ridge.south)+(0, -\ddy)$) {\small \inc{reef}};
        \draw[-] ($(ridge.south)+(0, 0 )$) -- ($(reef.north)+(0, 0)$);

        \node[draw=black, rounded corners=2pt, anchor=north, scale=0.8, align=center, minimum height=4.9mm] (creef) at ($(reef.south)+(0, -\ddy-0.05)$) {\small \bf \inc{coral reef}};
        \draw[-] ($(reef.south)+(0, 0 )$) -- ($(creef.north)+(0, \ddd)$);

    \end{tikzpicture}
    }
    \caption{Illustration of proposal labels \mplab. In-vocabulary {\bf \inc{labels}} are shown in bold. Classes in the same superclass as the prediction \mpred are shown in blue boxes \msc, direct \WN siblings in orange \mwns, and ancestors not shared with the label \mlab in green \mwna.}
    \label{fig:proposal_labels}
    \vspace{-2mm}
\end{wrapfigure}

To still detect fine-grained out-of-vocabulary (OOV) errors, we propose the following approach, illustrated in \cref{fig:oov_method} for the above example.
We first retrieve the 10 most visually similar images from the \IN training set using cosine similarity in the \clip embedding space \citep{RadfordKHRGASAM21}, in this case all labeled \inc{rock beauty}. If a label of any of these images belongs to the same superclass as the model's prediction (as defined in \cref{sec:fine_grained}), we conclude that an entity similar to our prediction is shown in the image and proceed with our analysis. Otherwise, we conclude that the analyzed error is not fine-grained. To confirm that the shown entity is indeed out-of-vocabulary, we first collect a set of proposal labels from the following (illustrated in \cref{fig:proposal_labels}):
all \IN labels in the same superclass as the model's prediction (IV), all direct \WN siblings of the model's prediction (IV and OOV), and all \WN ancestors of the model's prediction, up to but excluding the first common ancestor with the \IN label (IV and OOV).
Finally, we use \clip as an open-world classifier to score each of the proposed labels. If the highest scoring class is not included in the \IN label set, we consider the shown entity to be OOV and conclude that this error was indeed fine-grained OOV.

\begin{wrapfigure}[12]{r}{0.55\textwidth}
    \centering
    \begin{minipage}[t]{0.49 \linewidth}
        \centering
        \includegraphics[width=0.97\linewidth]{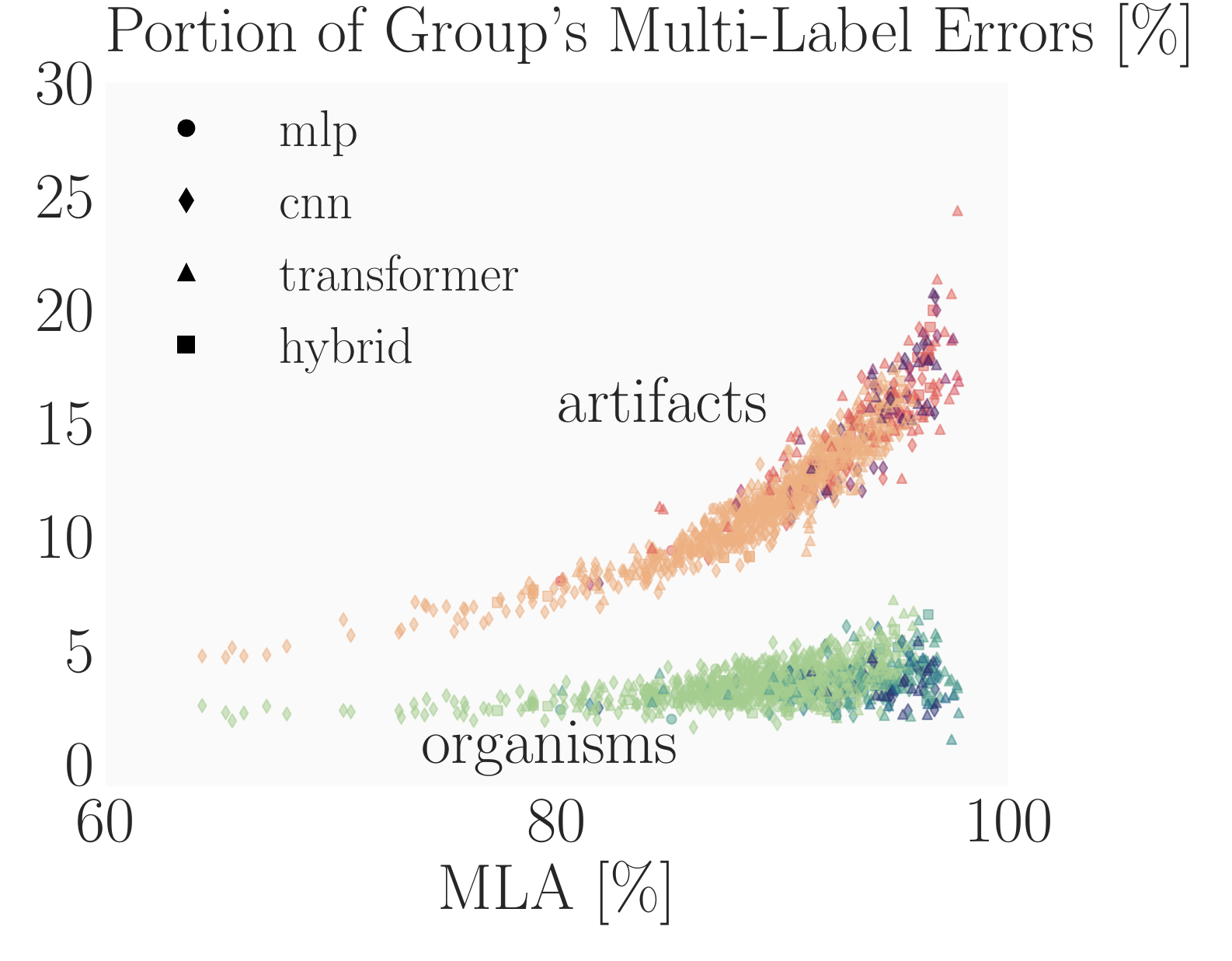}
    \end{minipage}
    \hfil
    \begin{minipage}[t]{0.49 \linewidth}
        \centering
        \includegraphics[width=0.88\linewidth]{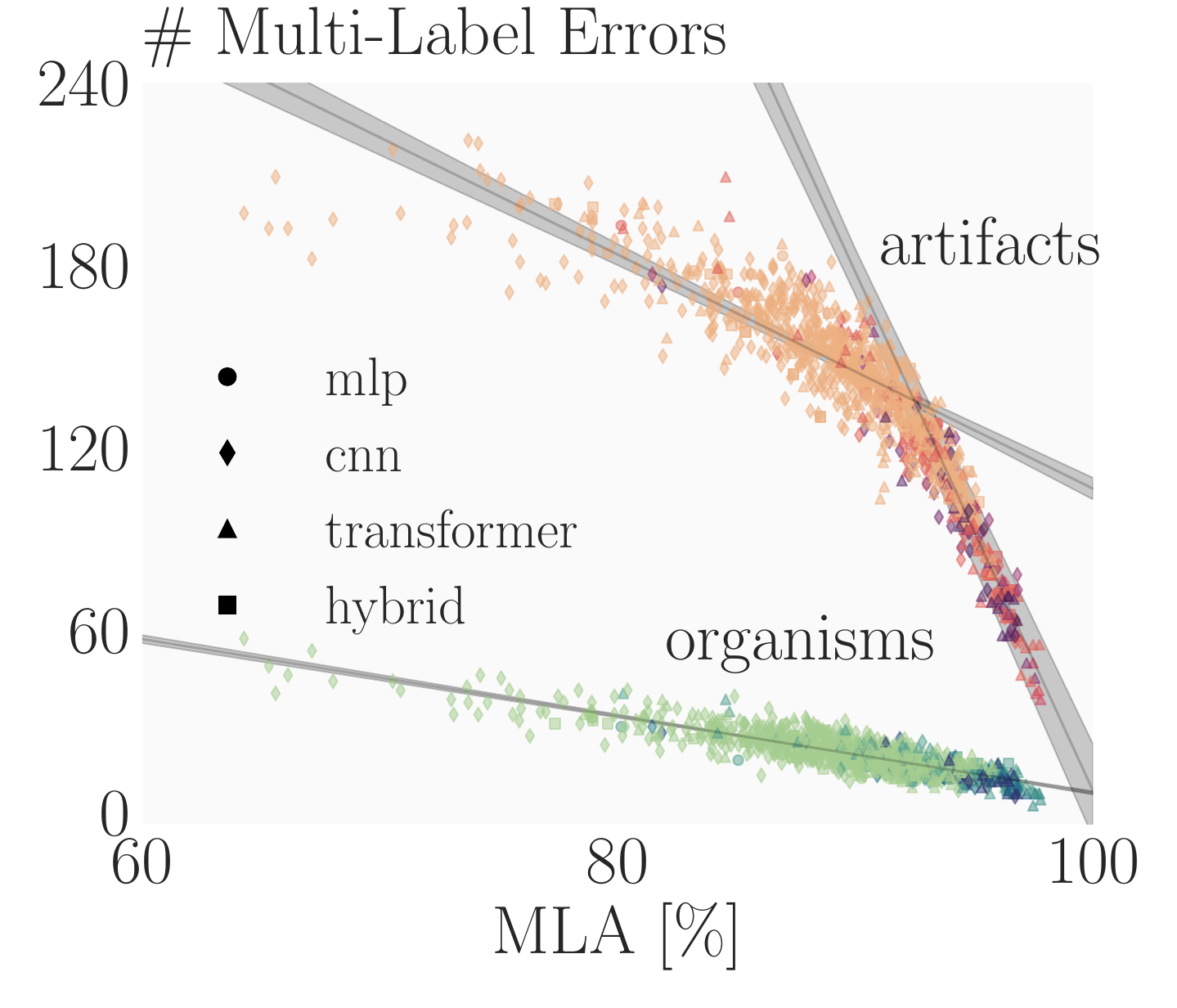}
    \end{minipage}
    \vspace{-5mm}
    \caption{Portion (left) and number (right) of \mla errors identified as \emph{fine-grained OOV} by group -- organisms (green) and artifacts (red). 95\% confidence interval linear fit is shown on the right. For artifacts, models are divided at $93\%$ \mla.}
    \label{fig:fine_grained_oov}
\end{wrapfigure}
We visualize the portion and number of multi-label errors categorized as fine-grained OOV in \cref{fig:fine_grained_oov}.
Interestingly, we observe that fine-grained OOV errors are not only much more prevalent in artifacts but also that their portion increases much more quickly with MLA for artifacts.
We hypothesize that this is due to the \IN labels covering organisms occurring in the validation set much more comprehensively than artifacts, and many images of artifacts being cluttered and full of other (OOV) objects. We note that this might be a reason why trained humans still outperform even state-of-the-art models on artifacts but not organisms \citep{ShankarRMFRS20}. Interestingly, we observe that, across pretraining datasets and model sizes, the number of fine-grained OOV errors for artifact drops much quicker with \mla above around $93\%$ \mla (see confidence intervals in \cref{fig:fine_grained_oov} right).

\vspace{-0.5mm}
\subsection{Non-Prototypical Instances}\label{sec:non_prototypical}
\vspace{-0.5mm}
\begin{wrapfigure}[8]{r}{0.27\textwidth}
    \centering
    \vspace{-5mm}
    \includegraphics[width=0.4\linewidth]{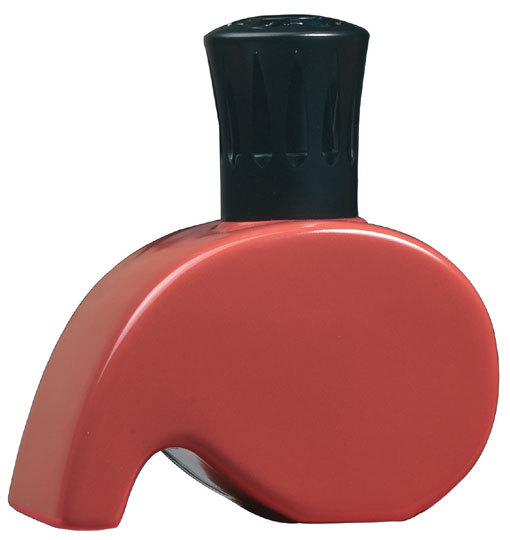}
    \vspace{-1mm}
    \caption{Image with label \inc{whistle}, but predcition \inc{perfume}.}
    \label{fig:non_prot_example}
\end{wrapfigure}
Many concepts described by any one \IN class label are broad, depend on social and geographic factors, and change over time. Combined with the search-based data collection process, this has led to a biased dataset with skewed representations of the concepts described by the contained classes.
Therefore it is unsurprising that models perform worse on non-prototypical instances or pictures of any given object, \ie, instances that, while clearly belonging to a given class, can be considered outliers in the \IN distribution of that class.

\begin{wrapfigure}[10]{r}{0.55\textwidth}
    \centering
    \vspace{-4mm}
    \begin{minipage}[t]{0.49 \linewidth}
        \centering
        \includegraphics[width=0.97\linewidth]{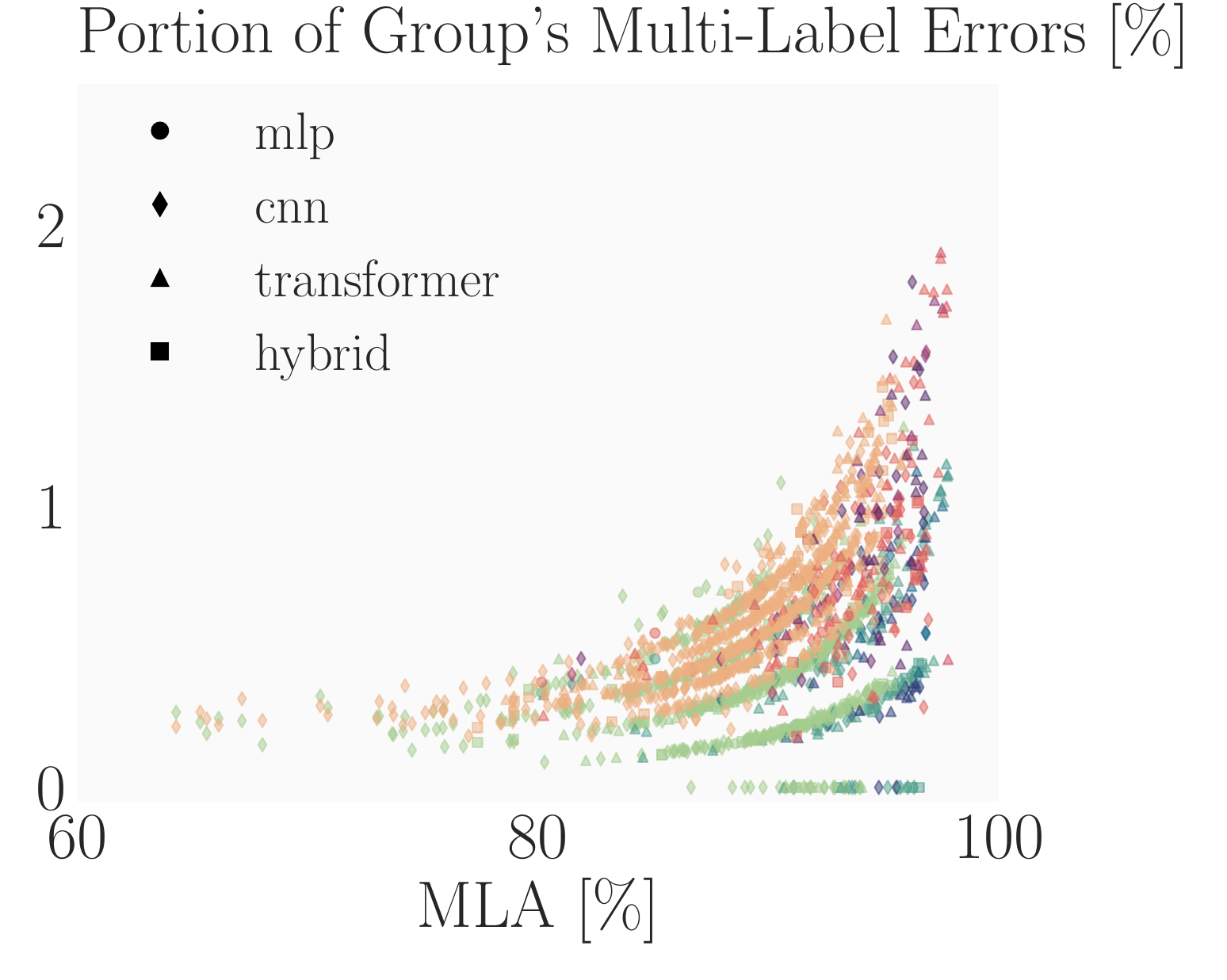}
    \end{minipage}
    \hfil
    \begin{minipage}[t]{0.49 \linewidth}
        \centering
        \includegraphics[width=0.88\linewidth]{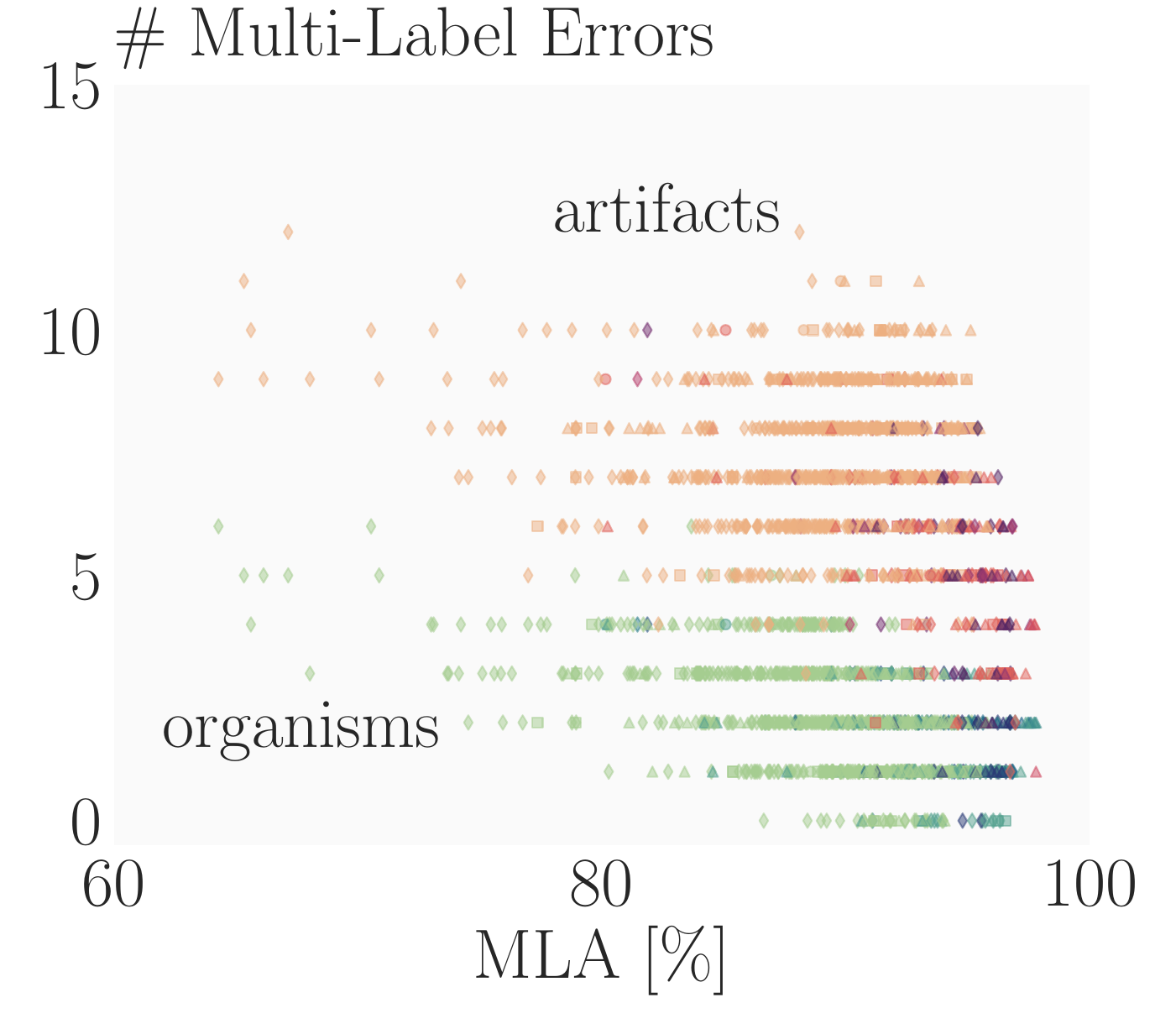}
    \end{minipage}
    \vspace{-6mm}
    \caption{Portion (left) and number (right) of \mla errors identified as \emph{non-prototypical sample}.}
    \label{fig:non_prot}
\end{wrapfigure}

For example, in \cref{fig:non_prot_example} we see a non-prototypical \inc{whistle}.
We believe it is interesting to track progress on these hard, non-prototypical instances as they are likely to be a good indicator to what extent a model has learned to recognize the underlying concepts of a class.
However, defining what constitutes a non-prototypical instance of any class is hard and to a large extent subjective. Fortunately, this error type is independent of the (incorrect) prediction made by the model, allowing us to directly leverage the manual categorization of non-prototypical images by \citet{VasudevanCLFR22}. We thus implicitly decide that all images that are classified correctly by the state-of-the-art models (\vit and \greedysoups) they considered are not sufficiently non-prototypical to explain an error by another model.

We visualize the number of errors caused by non-prototypical instances in \cref{fig:non_prot}. Interestingly and in contrast to all other error types, there is no strong correlation between performance on non-prototypical examples and overall \mla. This suggests that these non-prototypical examples are not inherently hard to classify for all models. Further surprisingly, non-prototypical examples account for a very similar portion of errors for artifacts and organisms, despite the appearance of artifacts of the same class varying significantly more, which we expected would lead to a larger portion of artifact errors being explained by their non-prototypical appearance.

\vspace{-0.5mm}

\subsection{Spurious Correlations}\label{sec:spurious}
\vspace{-0.5mm}
\begin{wrapfigure}[8]{r}{0.28\textwidth}
    \centering
    \vspace{-11.5mm}
    \includegraphics[trim={4mm 6mm 4mm 20mm},clip,width=0.90\linewidth]{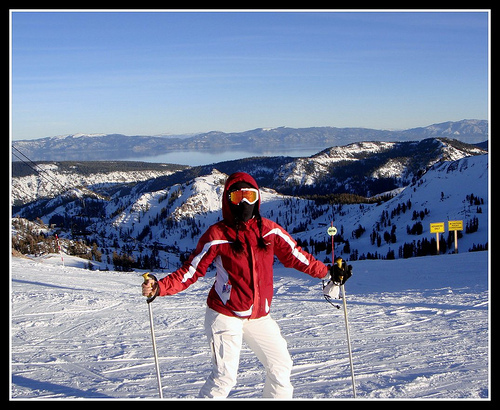}
    \caption{An image with label \inc{ski mask} (and also multi-class label \inc{alp}), but prediction \inc{ski}.}
    \label{fig:spurious_example}
\end{wrapfigure}
Entities of many classes frequently co-occur with features that have no causal relationship to that class. For example, oxen are frequently observed on green pastures (such as in \cref{fig:multi_label_example}). We call these correlated features \citep{NeuhausABH22}. Inherently this is not a problem, however, it has been observed that models frequently rely on these correlated features to make predictions \citep{ChoiTW12,BeeryHP18,NeuhausABH22,Moayeri0F22}, for example, by predicting \inc{ox} when shown only a green pasture or \inc{camel} when shown a cow in the desert. When correlated features lead to prediction errors, we call these spurious correlations. Here, we focus on the case where the presence of a correlated spurious feature causes a misclassification to the correlated class, despite no corresponding entity being shown. In \cref{fig:spurious_example}, we show an example of such a spurious correlation where the presence of a \inc{ski mask} and \inc{alp}s causes a model to predict \inc{ski}, despite the image containing no skis. We do not consider the error mode, where the absence of a correlated feature causes the model to \emph{not} predict the correlated class, despite a corresponding entity being shown, investigated by \citet{SinglaF22,Moayeri0F22}.

\begin{wrapfigure}[12]{r}{0.55\textwidth}
    \centering
    \vspace{-4mm}
    \begin{minipage}[t]{0.49 \linewidth}
        \centering
        \includegraphics[width=0.97\linewidth]{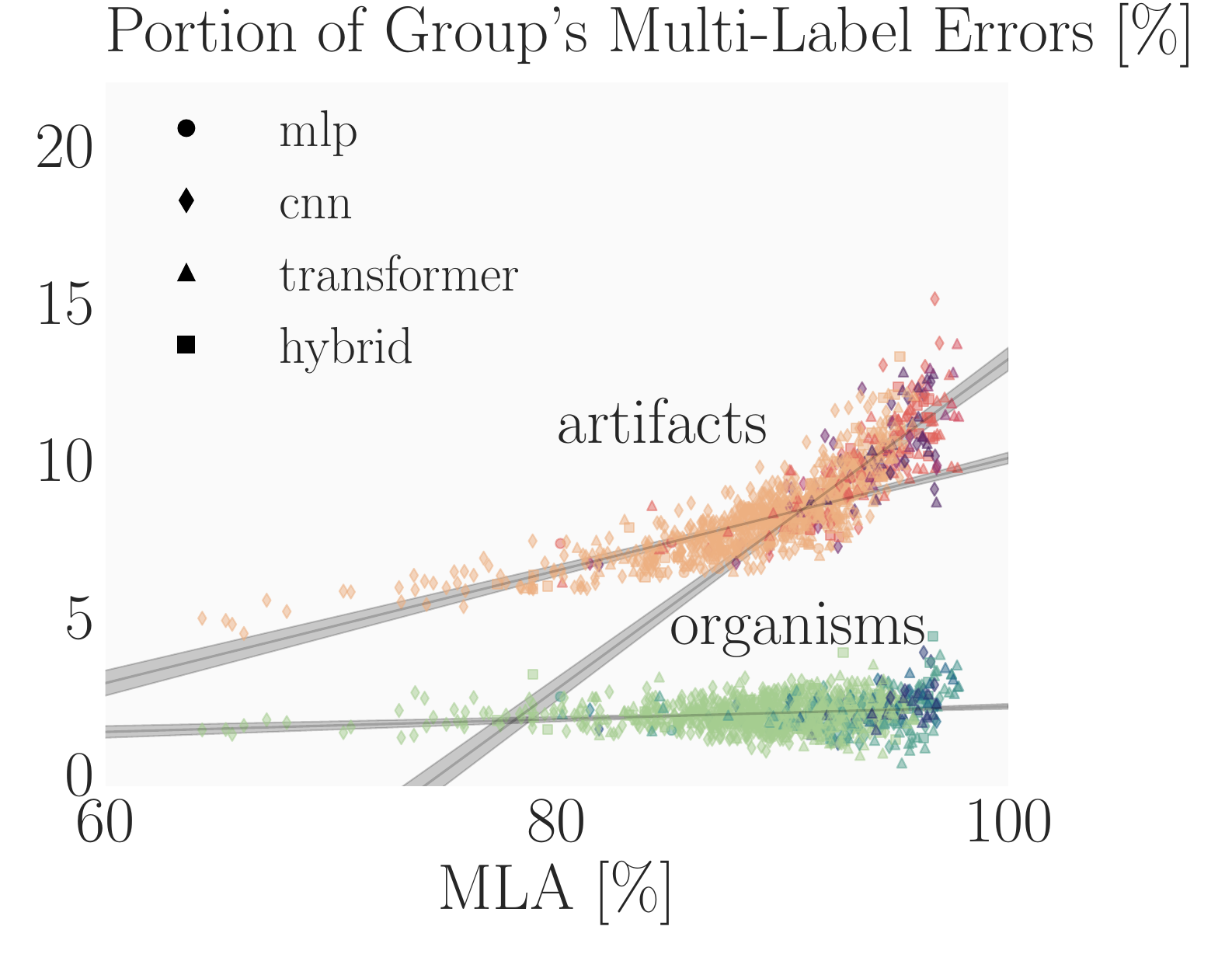}
    \end{minipage}
    \hfil
    \begin{minipage}[t]{0.49 \linewidth}
        \centering
        \includegraphics[width=0.88\linewidth]{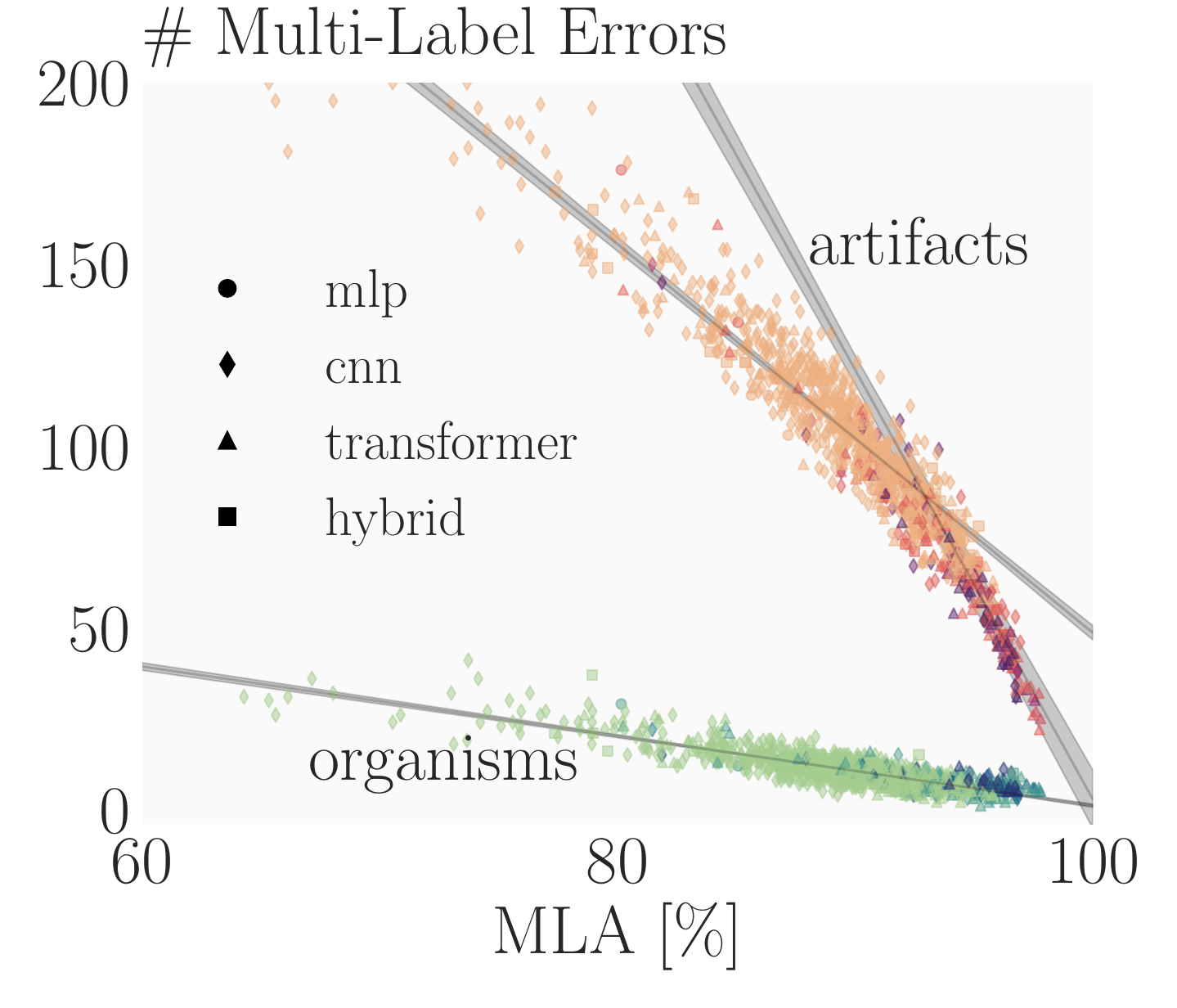}
    \end{minipage}
    \vspace{-5mm}
    \caption{Portion (left) and number (right) of \mla errors caused by \emph{spurious correlations} by group. A 95\% confidence interval linear fit is shown on the right. For artifacts, models are divided into those with more and less than $93\%$ \mla.}
    \label{fig:spurious}
\end{wrapfigure}

To identify errors caused by spurious correlations, we identify pairs of commonly co-occurring classes and then categorize errors as spurious correlations if an incorrect model prediction and a multi-label form such a co-occurrence pair.
More concretely, we first extract all pairs of co-occurring labels from the ReaL multi-label annotations \citep{BeyerHKZO20}, excluding samples we evaluate on. We then filter out pairs that either belong to the same superclass as defined in \cref{sec:fine_grained} or only co-occur once. Using this process, we extract $13\,090$ label pairs from $6\,622$ images with more than one label, yielding $1019$ unique co-occurrence pairs after filtering, which indicate a spurious correlation if they occur.

We visualize the portion and number of errors caused by spurious correlation in \cref{fig:spurious}. We observe that artifacts and organisms follow notably different trends. Not only is the portion of errors caused by spurious correlations much larger for artifacts than organisms, but it also increases with \mla for artifacts (at an increased rate for higher \mla), while it stays roughly constant for organisms. For state-of-the-art models, spurious correlations explain up to $15\%$ of errors on artifacts making them the second largest error source we identify.

\section{Analysis of Model Errors on ImageNet} \label{sec:analysis}
In this section, we discuss global trends, analyze their interaction with architecture and training choices, and validate our automatic pipeline against the manual analysis of \citet{VasudevanCLFR22}.

\paragraph{Model Failures}

\begin{wrapfigure}[13]{r}{0.55\textwidth}
    \centering
    \vspace{-4mm}
    \begin{minipage}[t]{0.49 \linewidth}
        \centering
        \includegraphics[width=0.97\linewidth]{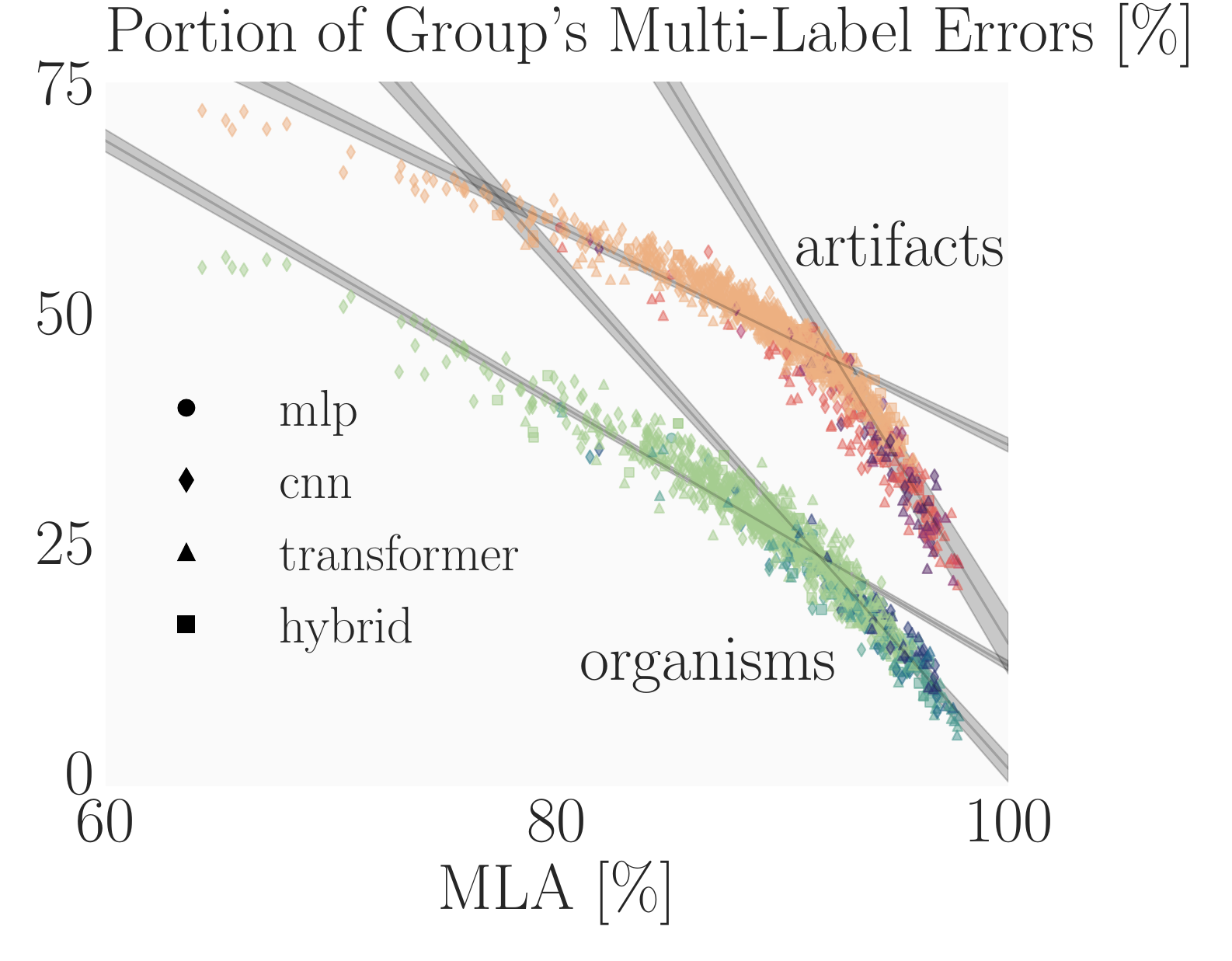}
    \end{minipage}
    \hfil
    \begin{minipage}[t]{0.49 \linewidth}
        \centering
        \includegraphics[width=0.88\linewidth]{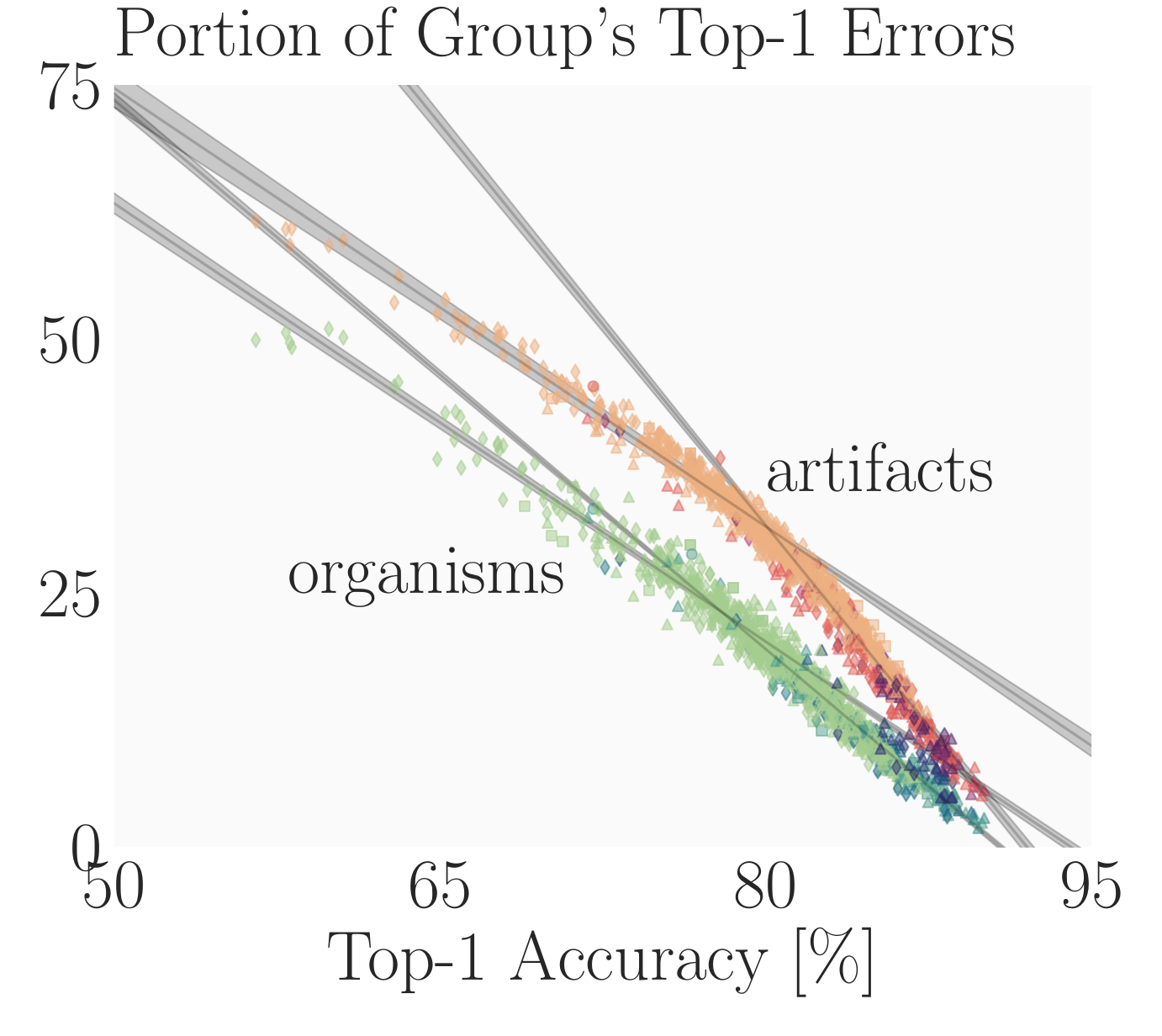}
    \end{minipage}
    \vspace{-5mm}
    \caption{Portion of \mla (left) and \toa (right) errors that can not be explained and are thus considered \emph{model failures} -- organisms (green) and artifacts (red). 95\% confidence interval linear fit is shown shaded. Models are divided by \mla (at $93\%$) and \toa accuracy (at $80\%$) for linear fits.}
    \label{fig:mle}
\end{wrapfigure}

After removing all minor (see \cref{sec:fine_grained,sec:oov}) and explainable (see \cref{sec:non_prototypical,sec:spurious}) errors, from the multi-label errors (MLE), we are left with a set of particularly severe, unexplained model failures (MLF).
In \cref{fig:mle} we visualize the portion of these unexplained model failures over multi-label accuracy (MLA) and standard top-1 accuracy, again split by artifact (red) and organism (green). A ratio of $1$ corresponds to none of the MLEs being explainable by our pipeline, while a ratio of $0$ corresponds to all MLEs being explainable, thus consisting only of less severe error types.
Surprisingly, both top-1 accuracy and MLA are pessimistic when it comes to reporting model progress, with unexplained model failures decreasing at a much quicker rate than both top-1 errors and MLE.

Further, we observe different error distributions for organisms and artifacts.
Firstly, the portion of model failures is much higher for artifacts than for organisms.
Secondly, while the portion of unexplained errors decreases for both organisms and artifacts with \mla and \toa accuracy, indicating that as models get more accurate they make not only less but also less severe errors, the rate of this change differs. For artifacts, this decrease is initially slower but then at around 93\% MLA or 80\% \toa error, the portion of severe model failures starts to drop rapidly (roughly three times as fast as before), while the decrease is roughly linear for organisms. This phase change becomes particularly apparent when viewing the trend over MLA, where it is even visible for organisms.

\subsection{(Pre-)training Dataset}
\begin{wrapfigure}[16]{r}{0.335\textwidth}
    \centering
    \vspace{-5mm}
    \includegraphics[width=0.95\linewidth]{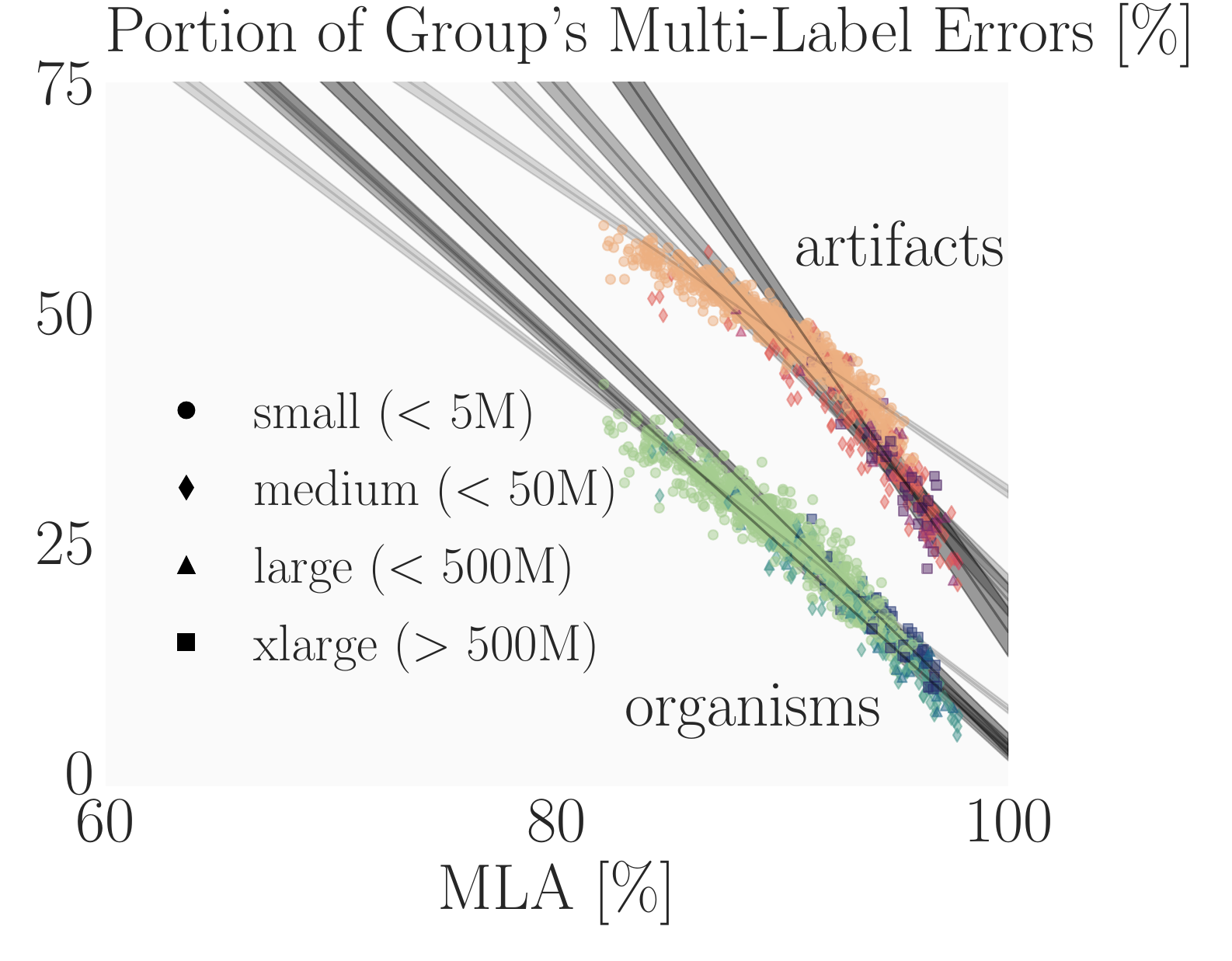}
    \vspace{-2mm}
    \caption{Portion of model failures (MLF) over \mla depending on pretraining dataset size with 95$\%$ confidence intervals for models with at least $82\%$ \mla. Confidence intervals are shown darker for larger pretraining dataset sizes.}
    \label{fig:dataset_size}
\end{wrapfigure}
Throughout \cref{sec:method}, we have illustrated the pretraining dataset size with the marker hue. Generally, we observe that conditioned on identical \mla, the effect of pre-training dataset size is rather modest. Here, we divide the 12 datasets we consider into 4 categories from ``small'' ($<$5M) to ``xlarge'' ($>$ 500M) depending on the number of included images, illustrating separate fits in \cref{fig:dataset_size}, with a darker shade corresponding to a larger dataset. We observe that across a wide range of accuracies, larger pretraining datasets lead to a faster reduction of model failures with \mla for both artifacts and organisms. While this effect is partially explained by larger datasets more frequently leading to higher \mla, it is also observed for narrow \mla ranges. Considering individual error types, we observe that in particular fine-grained errors are more frequent for larger pretraining datasets. We hypothesize that this is due to these larger datasets leading to better underlying representations, allowing the model to succeed in the coarse classification for harder samples, while still failing on the fine-grained distinctions.

\subsection{Model Architecture}
\begin{wrapfigure}[15]{r}{0.335\textwidth}
    \centering
    \vspace{-5mm}
    \includegraphics[width=0.95\linewidth]{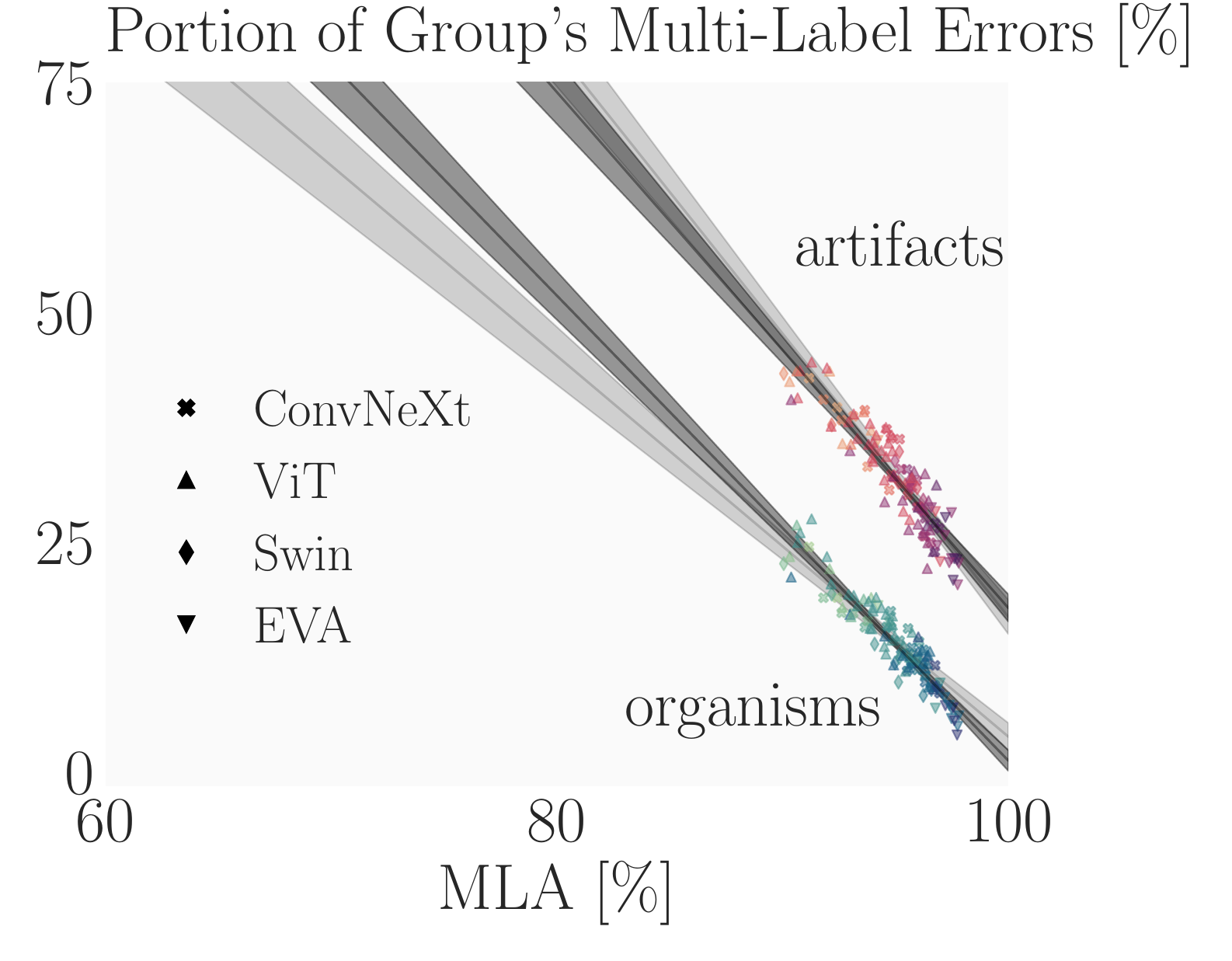}
    \vspace{-2mm}
    \caption{Portion of model failures (MLF) over MLA depending on model architecture. 95$\%$ confidence interval for transformer-based models is shown darker and for CNN-based models lighter.}
    \label{fig:model_params}
\end{wrapfigure}
In \cref{fig:model_params}, we again show the portion of unexplained model failures over MLA, this time broken down by architecture type and colored by the number of parameters (larger models have darker hues). To investigate whether modern convolutional architectures (ConvNeXts \citep{0003MWFDX22}) and vision transformers (ViT \citep{DosovitskiyB0WZ21}, Swin \citep{LiuL00W0LG21}, EVA \citep{FangWXSWW0WC23}) exhibit different trends in terms of model failures, we focus on state-of-the-art models with at least $92\%$ \mla that were trained using a dataset with at least 50M images, leaving 28 CNN-based and 79 transformer-based models.
While we observe clear trends of larger models performing better, the rate of model failures seems to be largely independent of model size when conditioned on MLA. We do observe a slight trend where the model failure rate of transformers decreases faster with MLA for organisms but slower for artifacts when compared to ConvNeXts. We speculate that this might be due to the ConvNeXts leveraging textures more heavily \citep{GeirhosRMBWB19}, which could be particularly helpful for fine-grained class distinctions on organisms leading to a larger portion of multi-label errors being model failures at the same \mla.

\subsection{Comparison to \citet{VasudevanCLFR22}}\label{sec:dough-bagel-eval}
To evaluate the alignment of our automated error analysis with a panel of human
experts, we compare its results to those of \citet{VasudevanCLFR22} on both
networks they consider, here for \vit and in \cref{app:ext_comp} for \greedysoups.

Comparing the results in \cref*{tbl:dough-bagel-cmp-vit}, we confirm that our pipeline
is conservative, classifying $16.4\%$ ($62$ / $378$) of the errors assigned an explanation by \citet{VasudevanCLFR22} as model failures.
On the remaining errors, our pipeline agrees with the panel of human
experts in $73.1\%$ of the cases. By manually inspecting the $85$ differently
classified errors, we determine that for only $32$ of these ($8.4\%$ of all errors), the categorization by \citet{VasudevanCLFR22} is clearly preferable to ours.
Furthermore, our pipeline encodes a minimal-severity bias, absent in \citet{VasudevanCLFR22}'s categorization. That is, when multiple error explanations would be valid, our pipeline consistently chooses the least severe one.
This highlights a further advantage of our approach. While any human judgment is inherently subjective and thus prone to differ between expert panels or even for the same panel over time, as acknowledged by \citet{VasudevanCLFR22}, our approach is consistent and repeatable.

Observing similar trends for \greedysoups in \cref{app:ext_comp}, we confirm that
on all models for which human expert error categorizations are available, our automated pipeline is well aligned with their judgment while providing consistent, repeatable, and conservative error categorizations.

\begin{table*}[tb]
	\caption{
    \footnotesize
		Comparison of our automated analysis of the errors of a \vit model with the manual annotations from~\citep{VasudevanCLFR22}. Rows correspond to the error categories assigned by a human expert while columns correspond to the error types assigned by our automated pipeline.
	}
    \vspace{-2mm}
	\begin{center}
		\resizebox{0.95\linewidth}{!}{\begin{tabular}{@{}lcccccc@{}}
    \toprule
    Error categories & FG &  FG OOV & Non-prot. & Spur. Corr. & Model failures & Total (row) \\
    \midrule
    Fine-grained & $192$ & $15$ & $0$  & $10$ &  $25$  & $242$ \\
    Fine-grained OOV & $9$ & $20$ & $0$ & $11$  &  $14$ & $54$ \\
    Non-prototypical & $13$ & $2$ & $12$ & $3$ &  $0$  & $30$ \\
    Spurious Correlation & $10$ & $12$ & $0$ & $7$  &  $23$  &  $52$ \\
    \midrule
    Total (col) & $224$ &  $49$ & $12$ & $31$ & $62$ & $378$ \\
    \bottomrule
\end{tabular}
}
	\end{center}
    \vspace{-2mm}
	\label{tbl:dough-bagel-cmp-vit}
\end{table*}

\section{Limitations}
In this section, we briefly reflect on the limitations of our work.

\paragraph{Personal Bias}
Our choices in the implementation of the error analysis pipeline reflect our personal biases and decisions, such as the superclasses we picked in \cref{sec:fine_grained}. This is not only limited to our personal biases, but, as our pipeline relies on prior work in several places, it also encodes the biases of their authors.
For example, we rely on the class overlap mappings from \citet{VasudevanCLFR22} and multi-class labels by \citet{ShankarRMFRS20} and \citet{VasudevanCLFR22}.

\paragraph{Extension to New Datasets}
To be applied to a new dataset, our pipeline requires multi-label and non-prototypicality annotations and superclass definitions.
While less precise than using a human-annotated gold standard, multi-label annotations could be sourced using a Re-Label \citep{YunOHHCC21} approach, allowing (given sufficient scale) spurious correlation pairs to be extracted from co-occurrence frequencies. After defining superclasses manually as required by their very definition, this would allow all error types except for the rare non-prototypical instances to be categorized. With the rest of the pipeline in place, non-prototypical instances could be annotated by reviewing the uncategorized errors shared by multiple well-performing models. However, while feasible these steps still require a significant amount of manual work.
We showcase a (partial) adaption to the similar \INA dataset in \cref{sec:ina}.

\section{Conclusion}
As state-of-the-art models come close to and exceed human performance on \IN, focus is increasingly shifting towards understanding the last remaining errors. Towards this goal, we propose an automated error categorization pipeline that we use to study the distribution of different error types across \nmodels models, of different scales, architectures, training methods, and pre-training datasets. We distinguish between minor errors, constituting failures on fine-grained class distinctions both when the ground truth was in and out-of-vocabulary, explainable errors, constituting failures on non-prototypical examples and due to spurious correlations, and unexplained model failures, constituting all remaining errors. 
We find that even \mla is a pessimistic measure of model progress with the portion of severe errors quickly decreasing with multi-label-accuracy. Further, we find that organism and artifact classes exhibit very different trends and prevalences of error types. For example, we observe that fine-grained class distinctions are a much bigger challenge for organisms than for artifacts, while artifacts suffer much more from spurious correlations and out-of-vocabulary errors, with these trends becoming increasingly pronounced as models become more accurate.
We believe that such an analysis of a new method's effects on model error distributions can become an important part of the evaluation pipeline and lead to insights guiding future research in computer vision.

\section*{Acknowledgements}
We thank our anonymous reviewers for their constructive comments and insightful feedback.

This work has been done as part of the EU grant ELSA (European Lighthouse on Secure and Safe AI, grant agreement no. 101070617) and the SERI grant SAFEAI (Certified Safe, Fair and Robust Artificial Intelligence, contract no. MB22.00088). Views and opinions expressed are however those of the authors only and do not necessarily reflect those of the European Union or European Commission. Neither the European Union nor the European Commission can be held responsible for them. 

The work has received funding from the Swiss State Secretariat for Education, Research and Innovation (SERI).

\message{^^JLASTBODYPAGE \thepage^^J}

\bibliography{references}
\bibliographystyle{IEEEtranN}

\message{^^JLASTREFERENCESPAGE \thepage^^J}

\ifbool{includeappendix}{%
	\clearpage
	\appendix
	\section{Extended Comparison to \citet{VasudevanCLFR22}} \label{app:ext_comp}

Similarly to the analysis in Sec.~\ref{sec:dough-bagel-eval}, we now evaluate our automated error classification pipeline on \greedysoups, the only other model for which manual error categorizations exist~\citep{VasudevanCLFR22}, showing the results in \cref{tbl:dough-bagel-cmp}. 
We do not assign an explanation to $11.6\%$ ($29$ / $249$) of the errors, classifying them as model failures. 
On the remaining errors, our pipeline agrees with the human expert annotation in $74.1\%$ ($163$ / $220$)
of the cases. After a manual review of the remaining 57 differently classified errors,
we determine that the categorization provided by \citet{VasudevanCLFR22} is clearly
preferable to ours for only 26 samples ($10.4\%$ of all errors). Overall, the results
for \greedysoups are very similar to those for \vit in \cref{sec:dough-bagel-eval}, thus further validating our modeling choices.
Therefore, we believe that our automatic categorization is indeed aligned with
the opinion of human experts and note that an efficient and automatic
classification pipeline is valuable even if it is imperfect, as it enables the
study of models and trends at scale.

\begin{table*}[h]
	\caption{
    \footnotesize
    Comparison of our automated analysis of the errors of a \greedysoups model with the manual annotations from~\citep{VasudevanCLFR22}. Rows correspond to the error categories assigned by a human expert while columns correspond to the error types assigned by our automated pipeline.
	}
    \vspace{-2mm}
	\begin{center}
		\resizebox{0.95\linewidth}{!}{
            \begin{tabular}{@{}lcccccc@{}}
                \toprule
                Error categories & FG &  FG OOV & Non-prot. & Spur. Corr. & Model failures & Total (row) \\
                \midrule
                Fine-grained            & $139$ & $14$ & $1$ & $7$ & $11$ & $172$ \\
                Fine-grained OOV        & $7$   & $8$  & $0$ & $5$ & $6$  & $26$ \\
                Non-prototypical        & $8$   & $1$  & $8$ & $2$ & $0$  & $19$ \\
                Spurious Correlation    & $7$   & $5$  & $0$ & $8$ & $12$ & $32$ \\
                \midrule
                Total (col)             & $161$ &  $28$ & $9$ & $22$ & $29$ & $249$ \\
                \bottomrule
            \end{tabular}
        }
	\end{center}
    \vspace{-2mm}
	\label{tbl:dough-bagel-cmp}
\end{table*}

\newpage
\section{Additional Results} \label{sec:additional_results}

In addition to the analysis presented in the main paper, which focused on the portion of errors due to the different error categories for artifacts and organisms, we provide more detailed results in this section, considering absolute error counts as well as portions for artifacts, organisms, the remaining classes, and all classes jointly.

Figures \ref{fig:class_overlap_app}-\ref{fig:spurious_app} contain these statistics grouped by error type:
\begin{itemize}[nosep]
	\item \cref{fig:class_overlap_app} -- top-1 errors caused by class overlap (\cref{sec:class_overlap});
	\item \cref{fig:multi_label_annotations_app} -- top-1 errors due to missing multi-label annotations (\cref{sec:multi_label});
	\item \cref{fig:fine_grained_app} -- fine-grained multi-label errors (\cref{sec:fine_grained});
	\item \cref{fig:fine_grained_oov_app} -- fine-grained out-of-vocabulary multi-label errors (\cref{sec:oov});
	\item \cref{fig:non_prot_app} -- multi-label errors due to non-prototypical instances (\cref{sec:non_prototypical});
	\item \cref{fig:spurious_app} -- multi-label errors due to various spurious correlations (\cref{sec:spurious}).
\end{itemize}
We show the number of errors in relative and absolute terms. In each figure, subfigures (a) and (b) show the results for each group of classes separated in ``artifacts'', ``organisms'' and ``others'' (the rest of the labels), where for (a) the relative portions of top-1 or multi-label errors are computed \emph{per group}, \ie by diving by the number of top-1 or multi-label errors in the respective group. Red hues again correspond to artifacts, green hues -- to organisms, and blue hues-- to the remaining classes (others). Subfigures (c) and (d) present the relative and absolute statistics for all classes jointly.

\begin{figure}[H]
	\centering
	\subfloat[Relative portion per group.]{
		\includegraphics[width=0.23\linewidth]{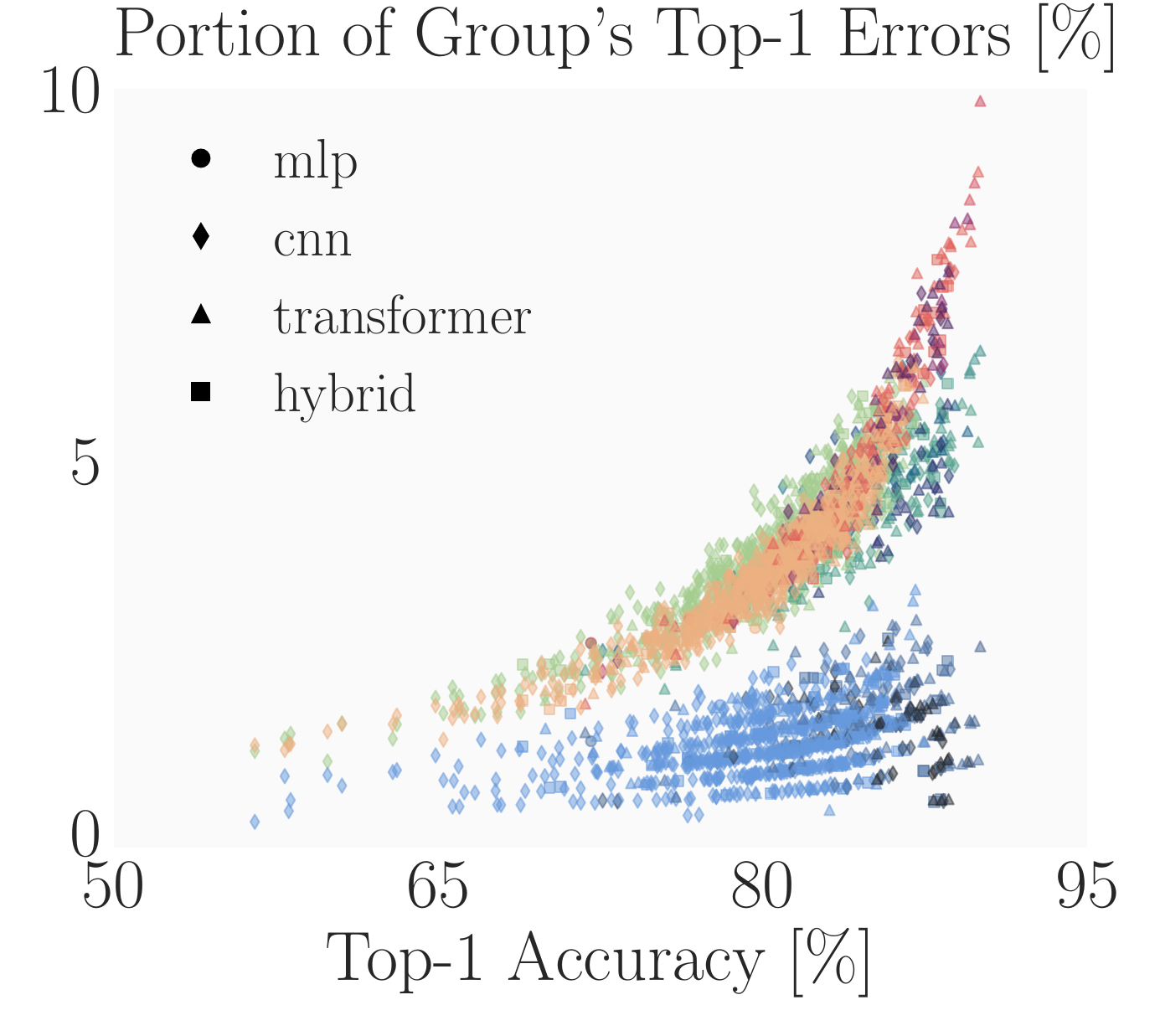}
	}\hfil
	\subfloat[Abs. number per group.]{
		\includegraphics[width=0.237\linewidth]{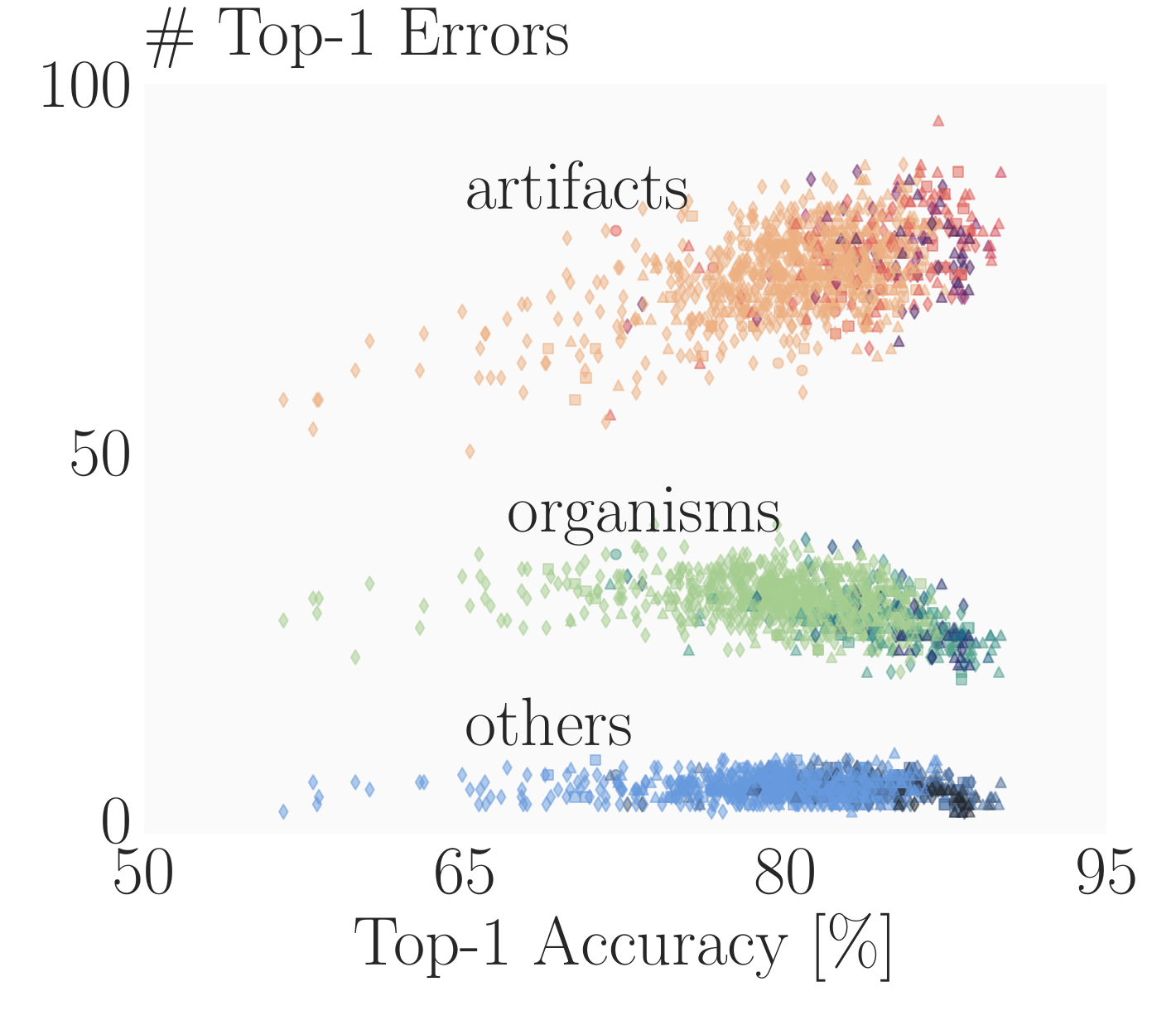}
	}\hfil
	\subfloat[Relative portion for all samples.]{
		\includegraphics[width=0.23\linewidth]{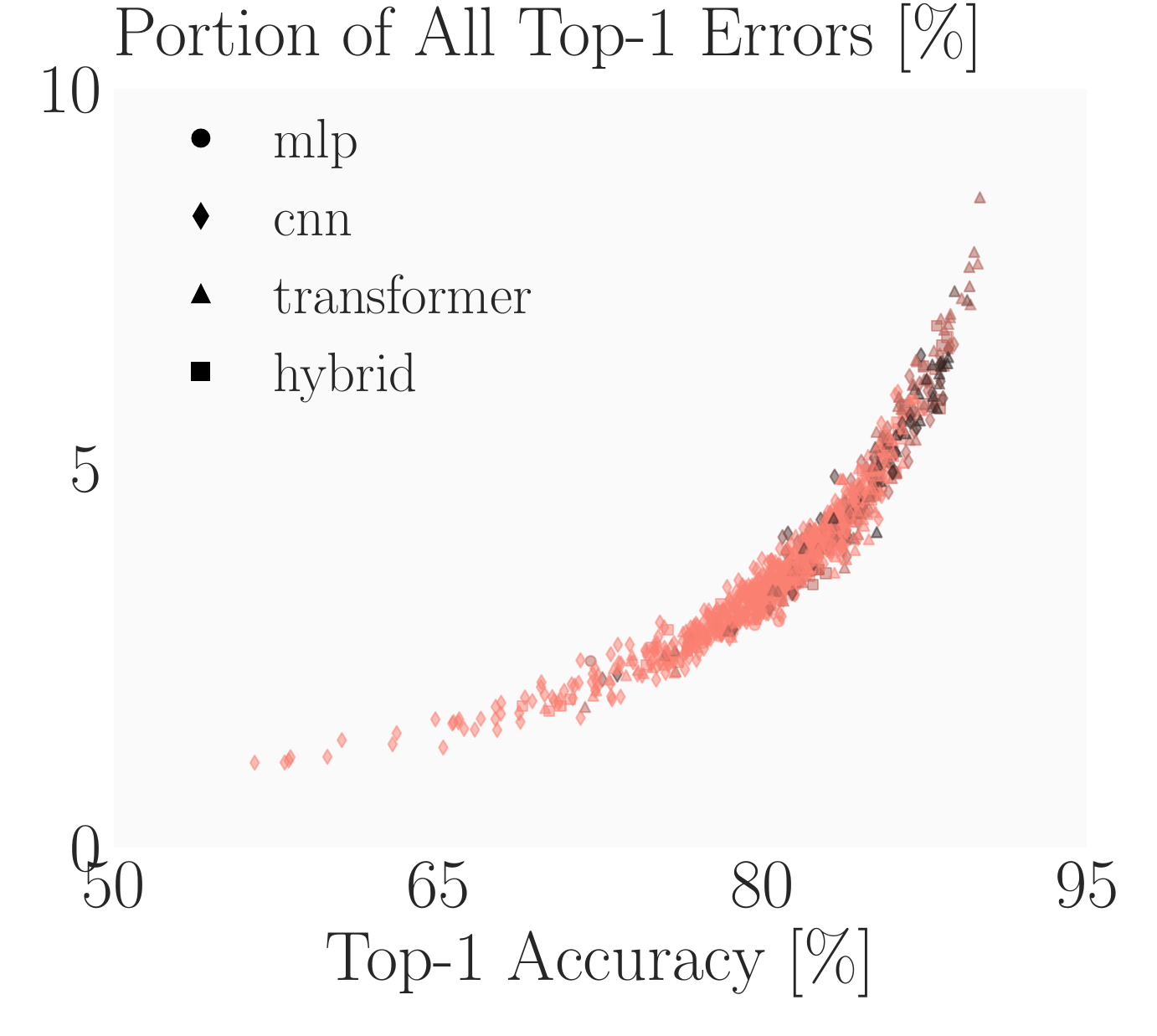}
	}\hfil
	\subfloat[Abs. number for all samples.]{
		\includegraphics[width=0.237\linewidth]{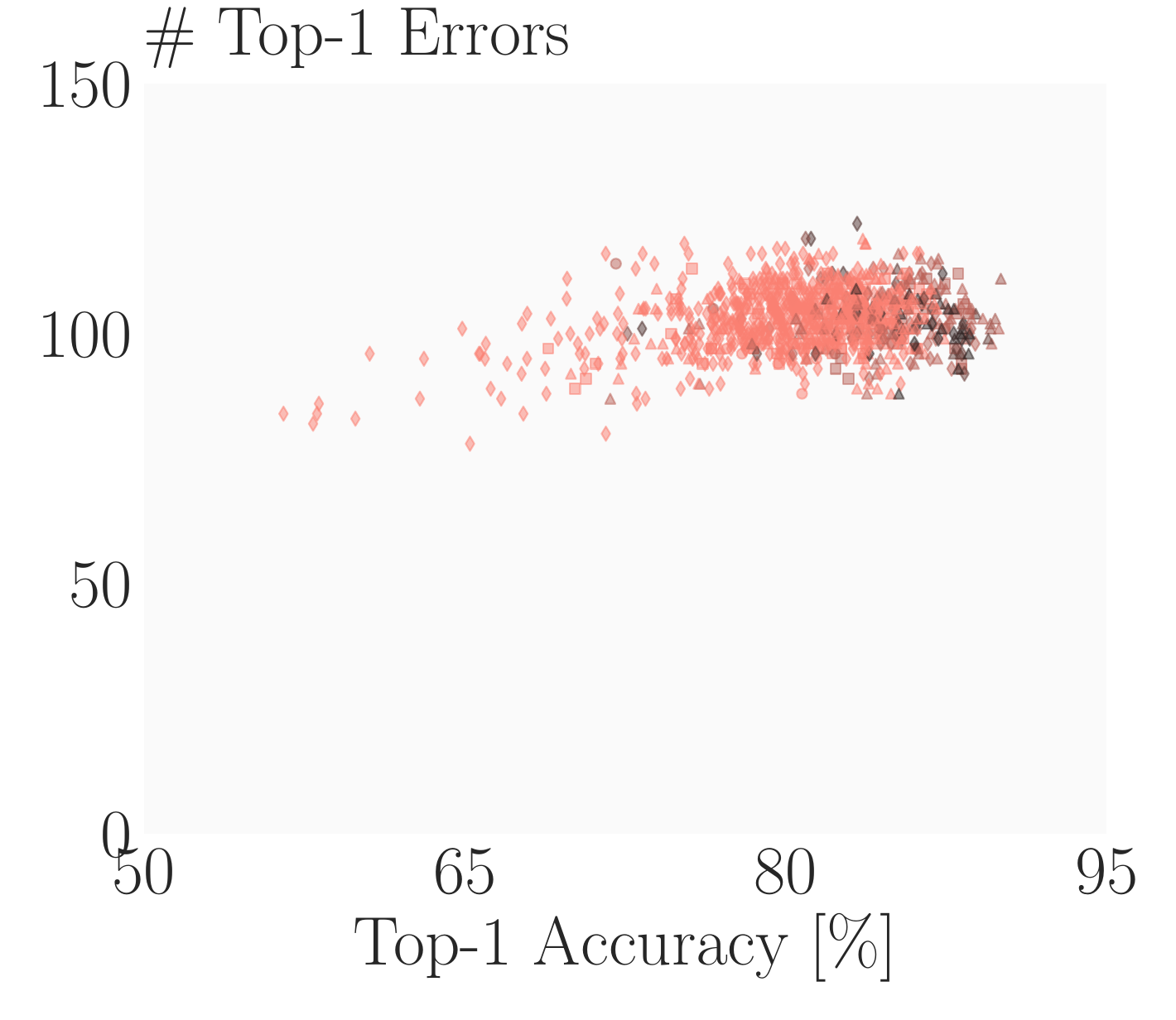}
	}

	\caption{
		Top-1 errors caused by class overlap. The relative portion of these errors on ``others'' is the smallest and the trend is noisier compared to ``artifacts'' and ``organisms''. This is possibly due to the low absolute number of classes in this group (68 out of 1000). In absolute terms, the number of this kind of error remains relatively constant.
	}
	\label{fig:class_overlap_app}
\end{figure}

\begin{figure}[H]
	\centering
	\subfloat[Relative portion per group.]{
		\includegraphics[width=0.23\linewidth]{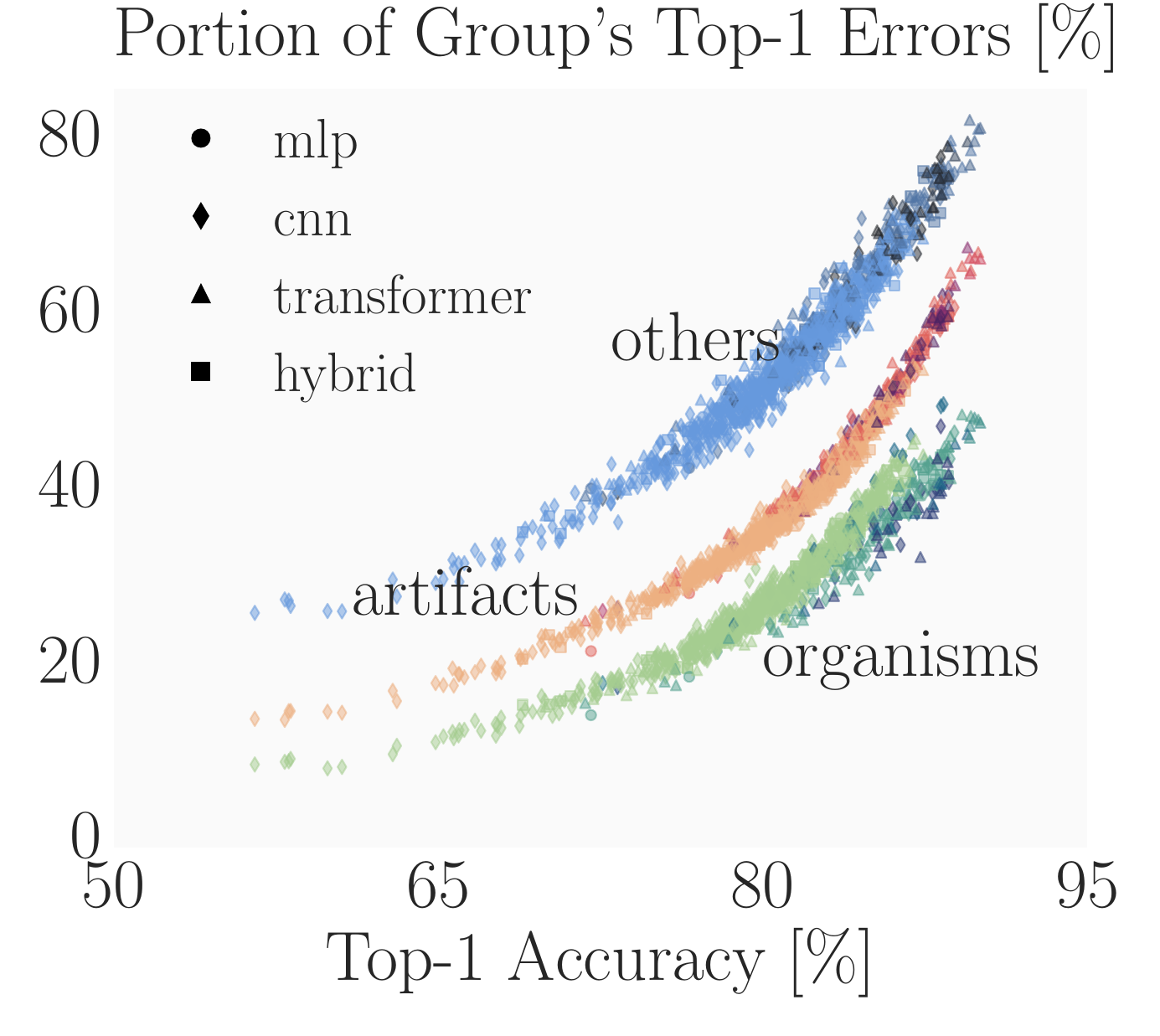}
	}\hfil
	\subfloat[Abs. number per group.]{
		\includegraphics[width=0.237\linewidth]{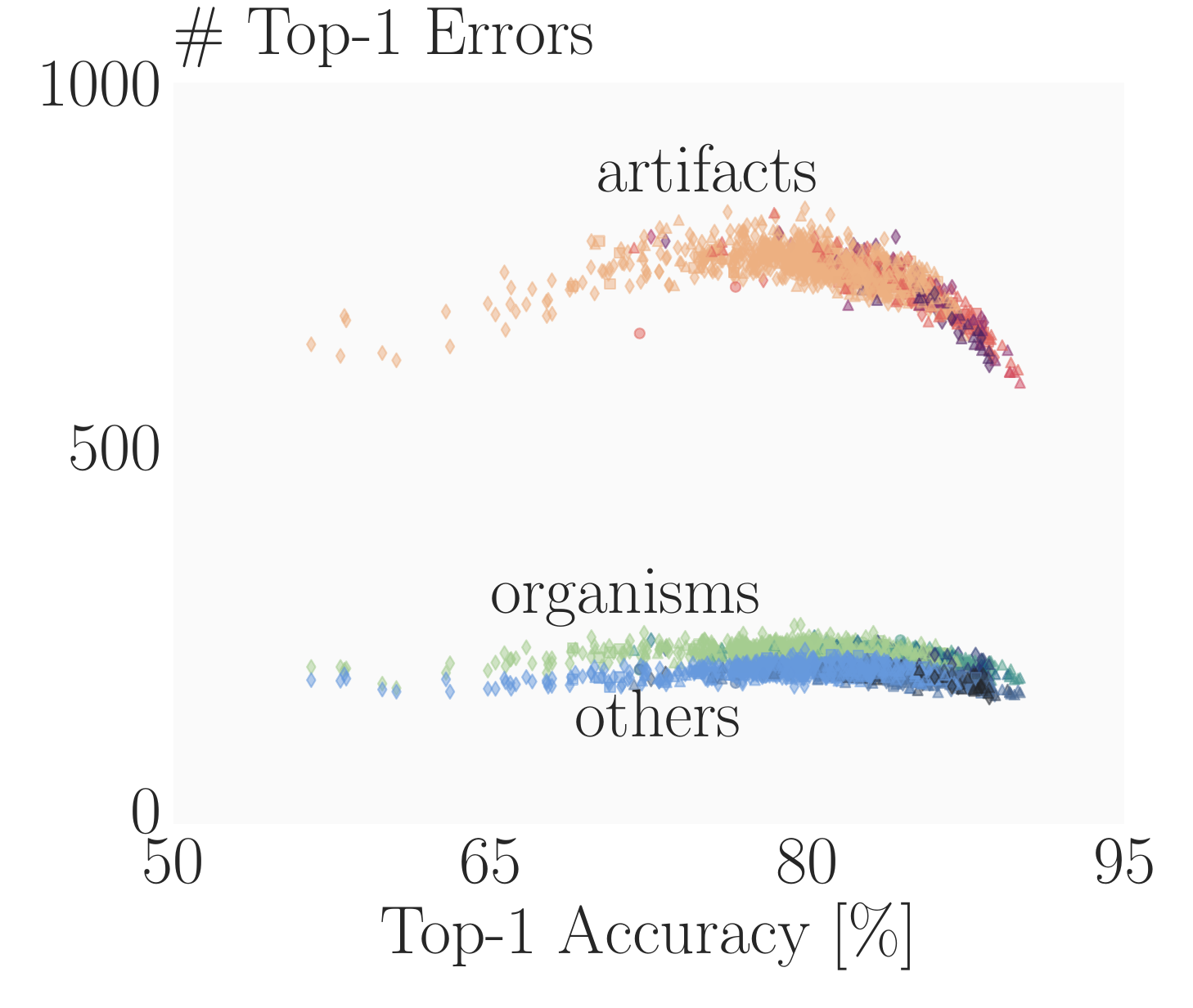}
	}\hfil
	\subfloat[Relative portion for all samples.]{
		\includegraphics[width=0.23\linewidth]{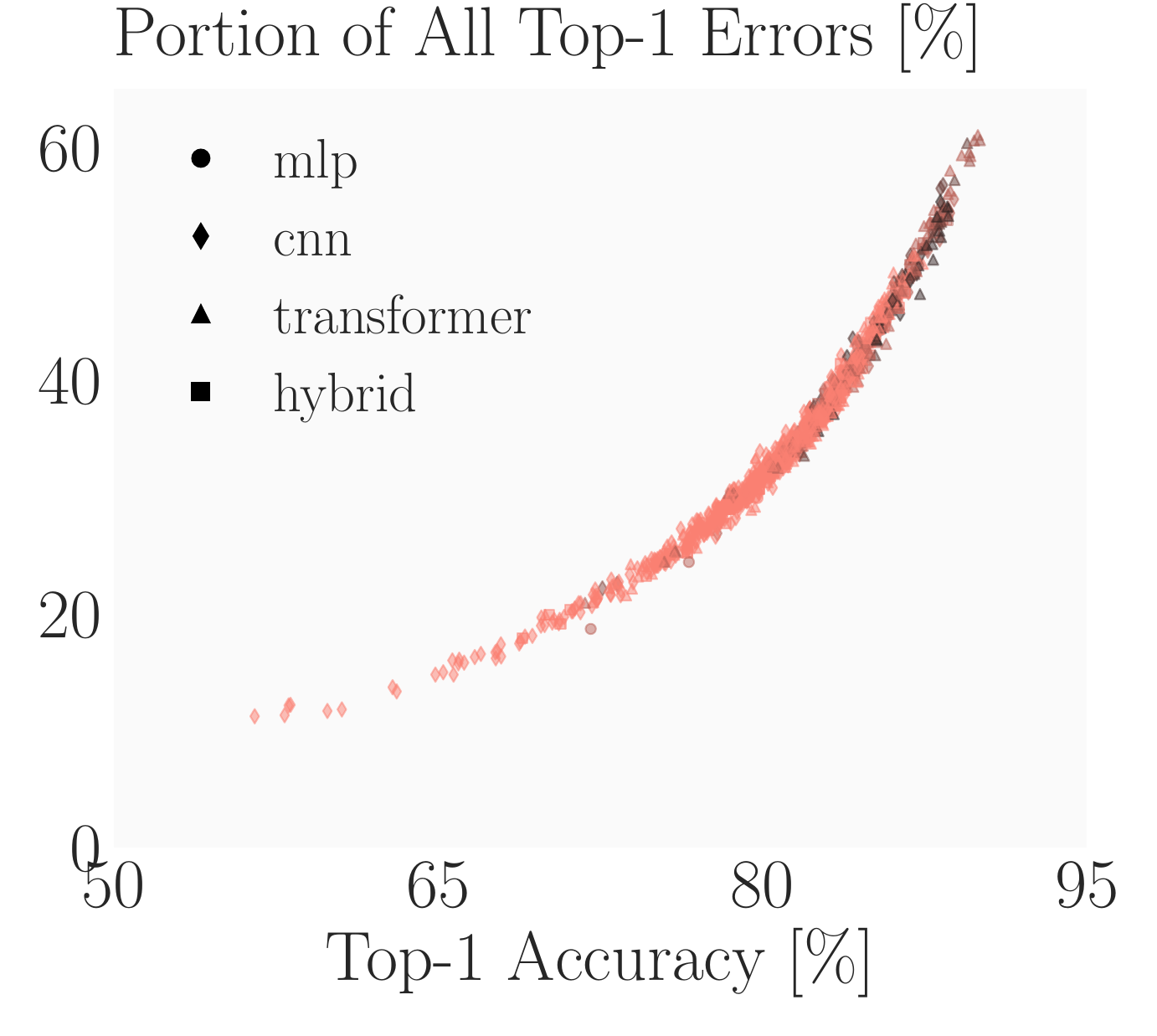}
	}\hfil
	\subfloat[Abs. number for all samples.]{
		\includegraphics[width=0.237\linewidth]{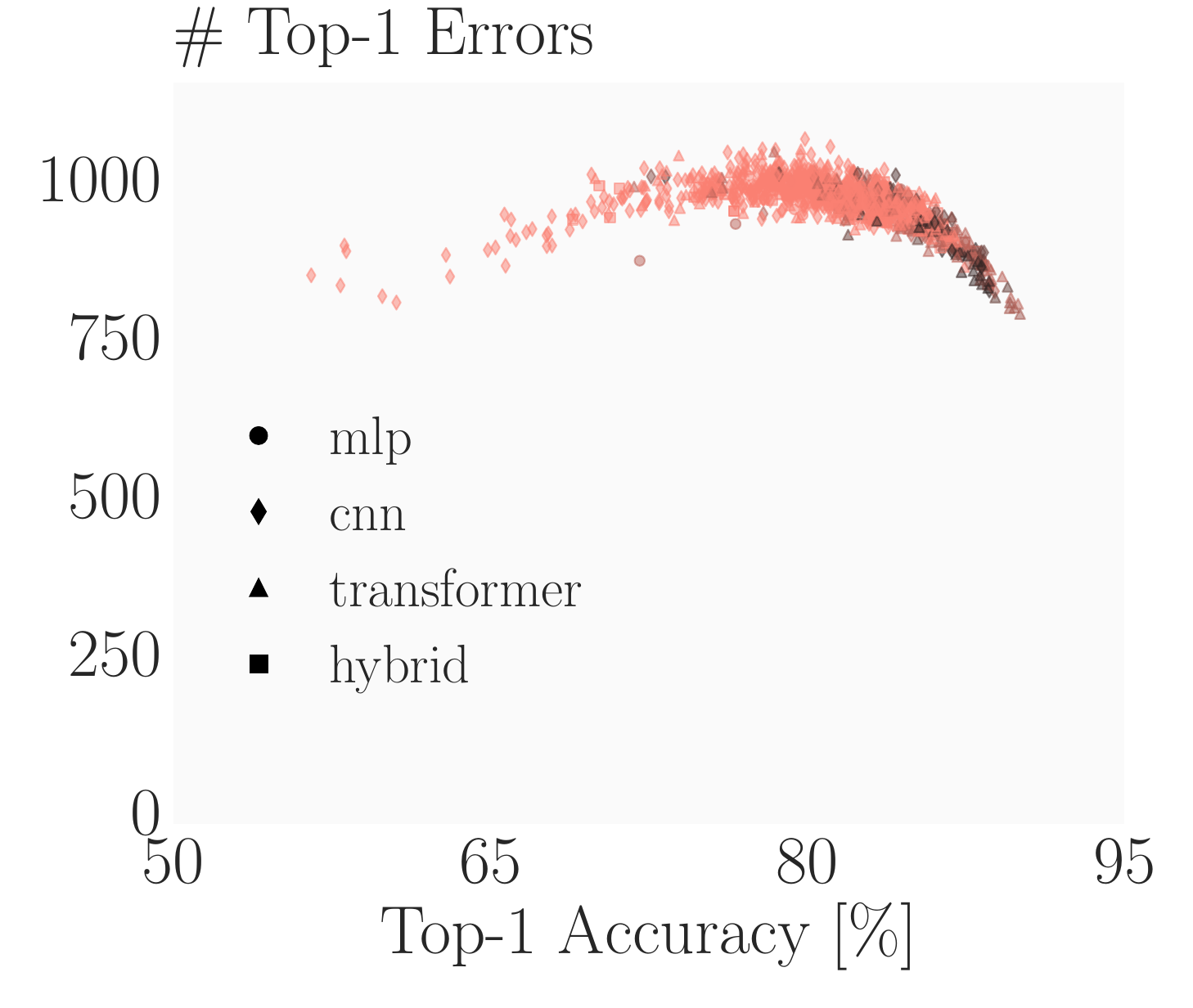}
	}

	\caption{
		Top-1 errors caused by missing multi-label annotations. Surprisingly, the portion of these errors is the highest on the ``others'' group. In absolute terms, ``organisms'' and ``others'' result in roughly the same constant amount of errors, despite the much larger number of organism classes (410 vs. 68). For ``artifacts'', it first increases and then starts dropping, possibly due to more accurate models learning to leverage \IN labeling biases.
	}
	\label{fig:multi_label_annotations_app}
\end{figure}

\begin{figure}[H]
	\centering
	\subfloat[Relative portion per group.]{
		\includegraphics[width=0.248\linewidth]{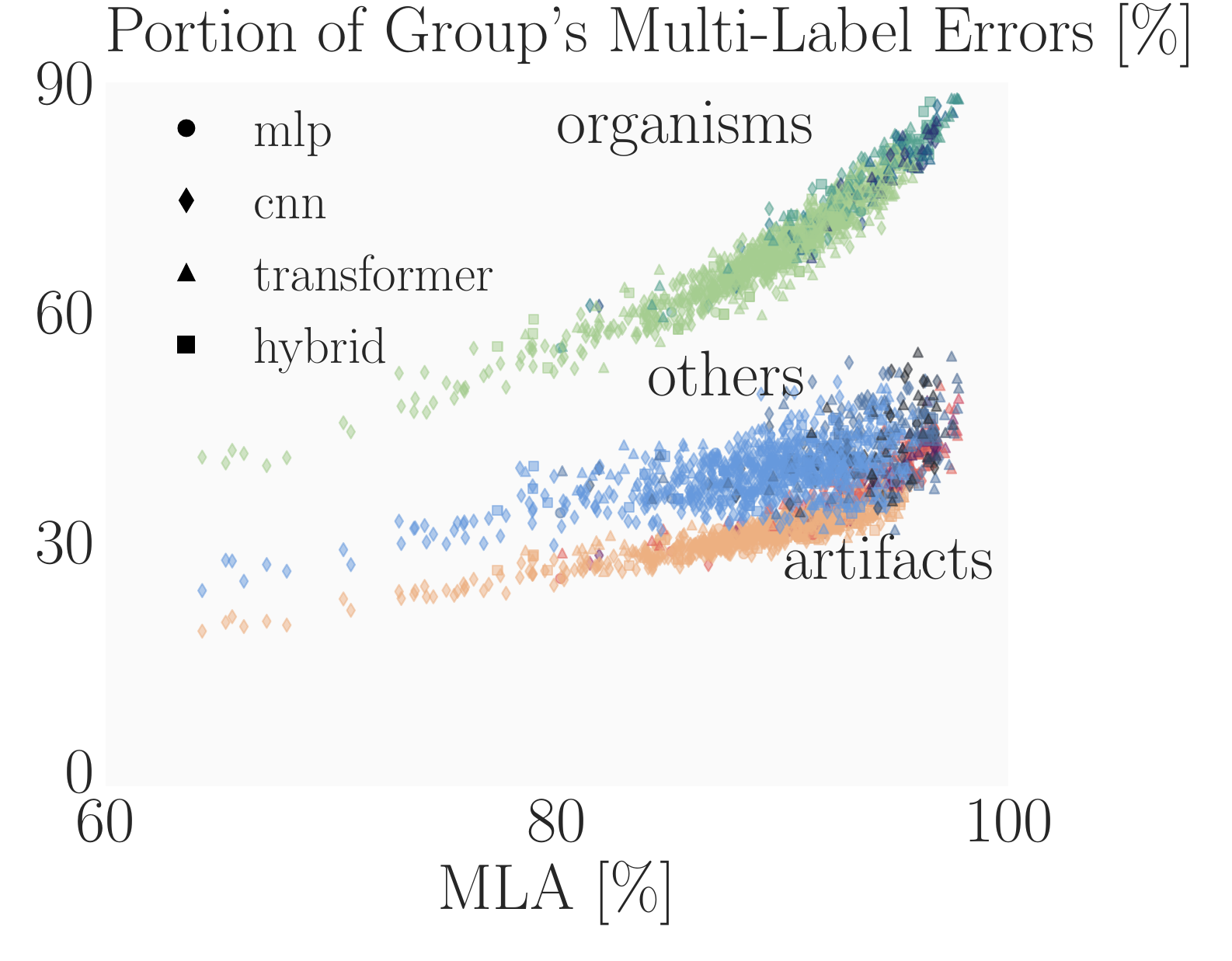}
	}\hfil
	\subfloat[Abs. number per group.]{
		\includegraphics[width=0.227\linewidth]{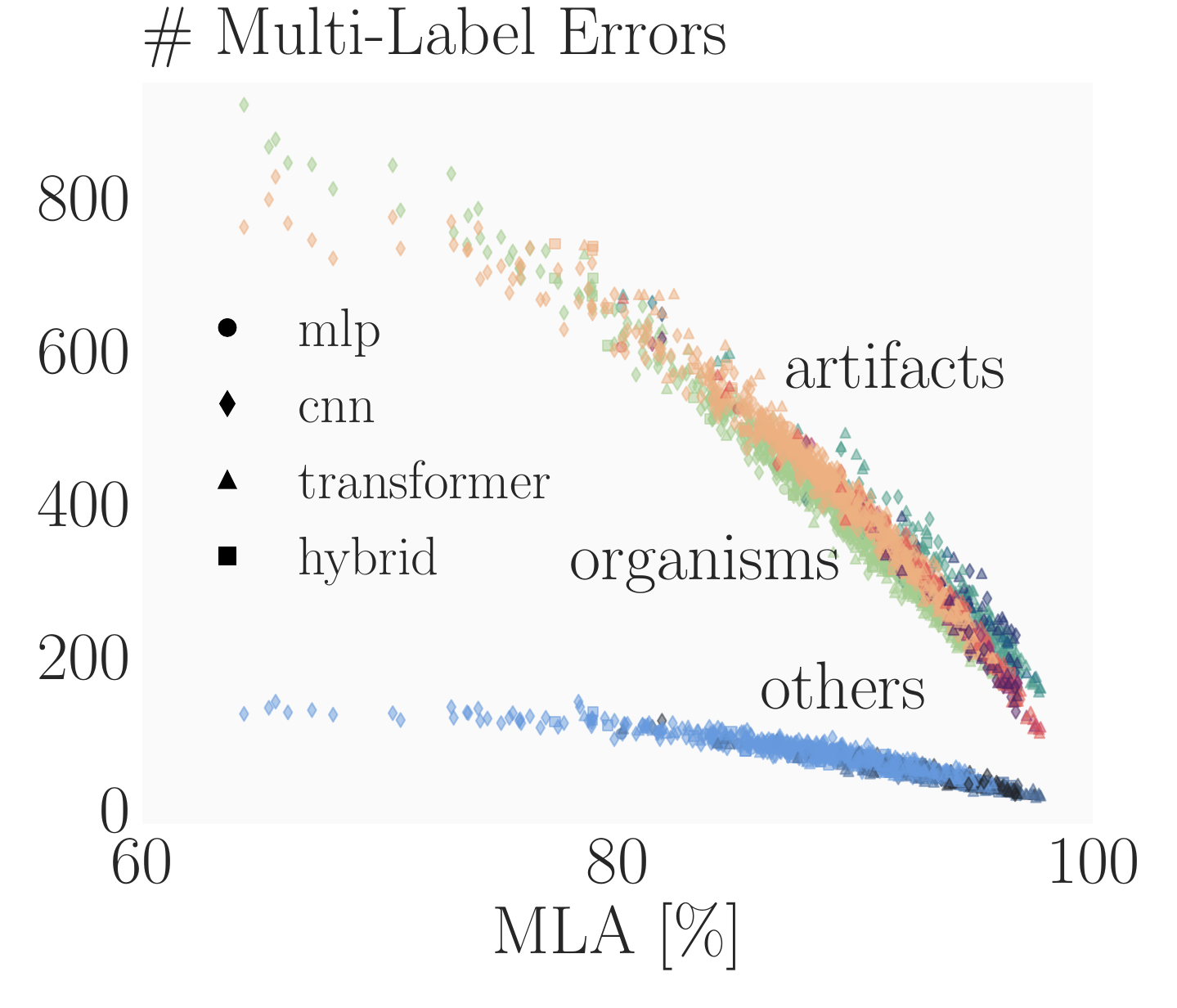}
	}\hfil
	\subfloat[Relative portion for all samples.]{
		\includegraphics[width=0.223\linewidth]{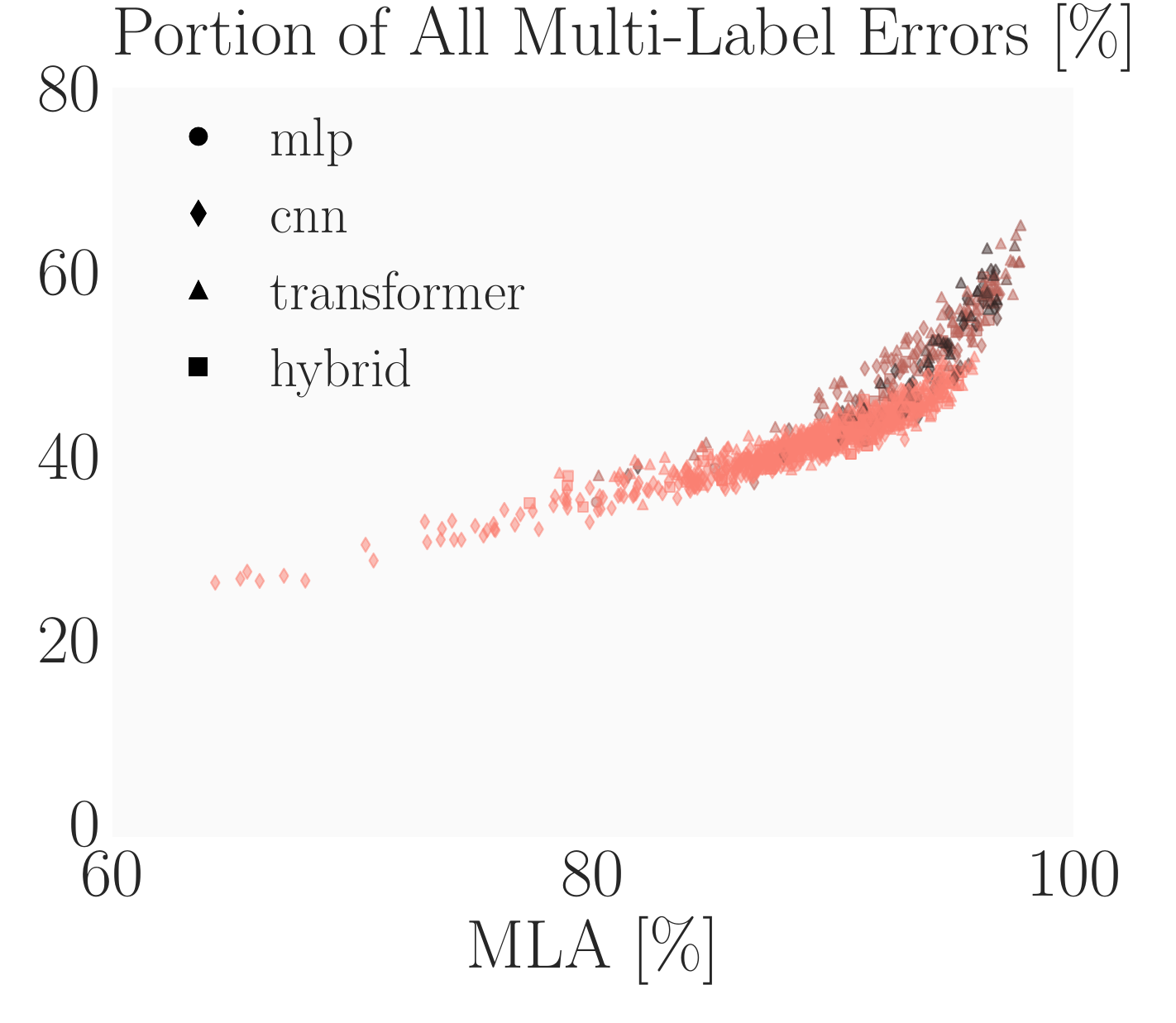}
	}\hfil
	\subfloat[Abs. number for all samples.]{
		\includegraphics[width=0.232\linewidth]{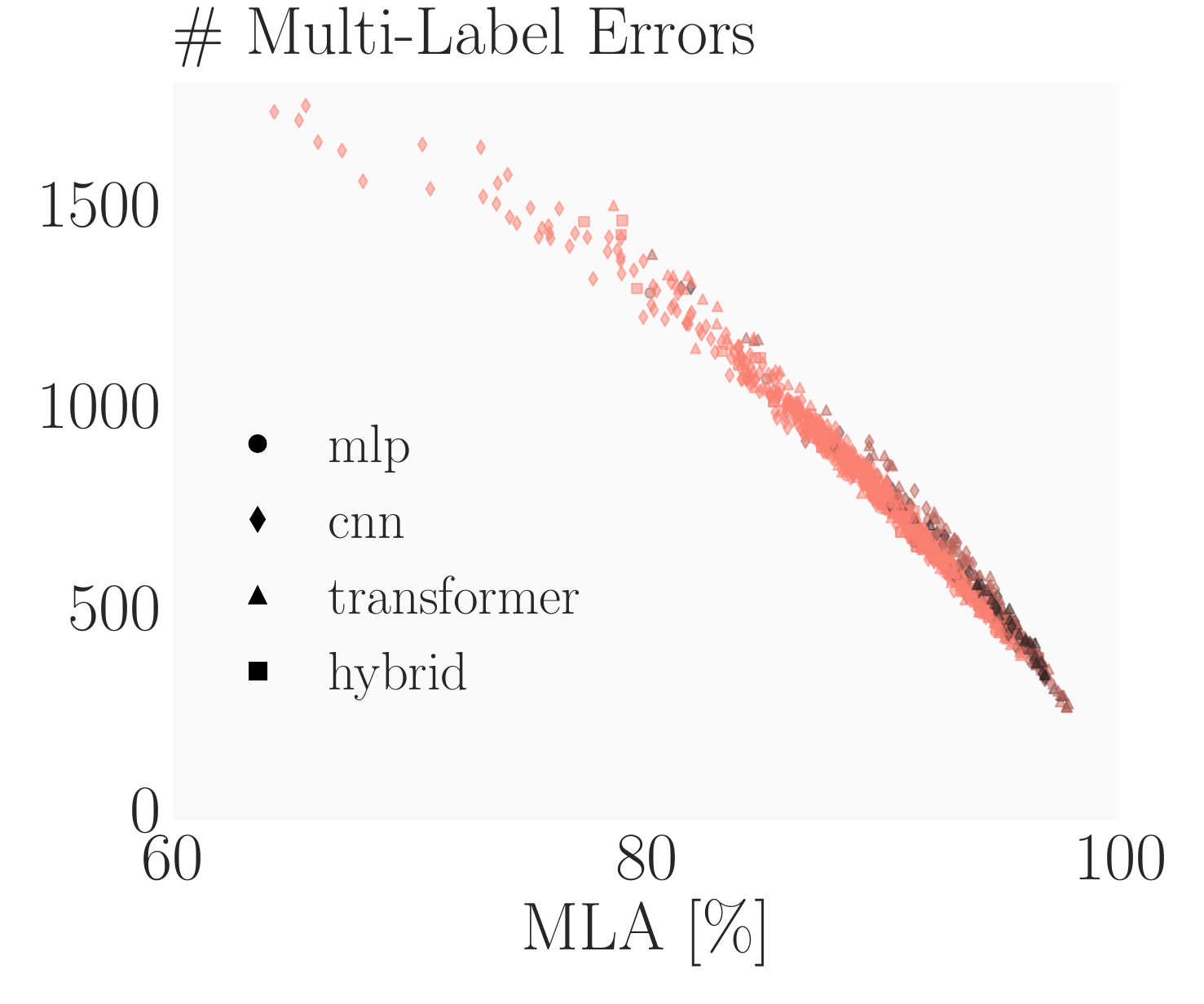}
	}

	\caption{
		Multi-label errors due to fine-grained misclassifications. The portion of fine-grained errors on ``others'' is larger than those on ``artifacts'' but smaller than ``organisms''.
	}
	\label{fig:fine_grained_app}
\end{figure}

\begin{figure}[H]
	\centering
	\subfloat[Relative portion per group.]{
		\includegraphics[width=0.248\linewidth]{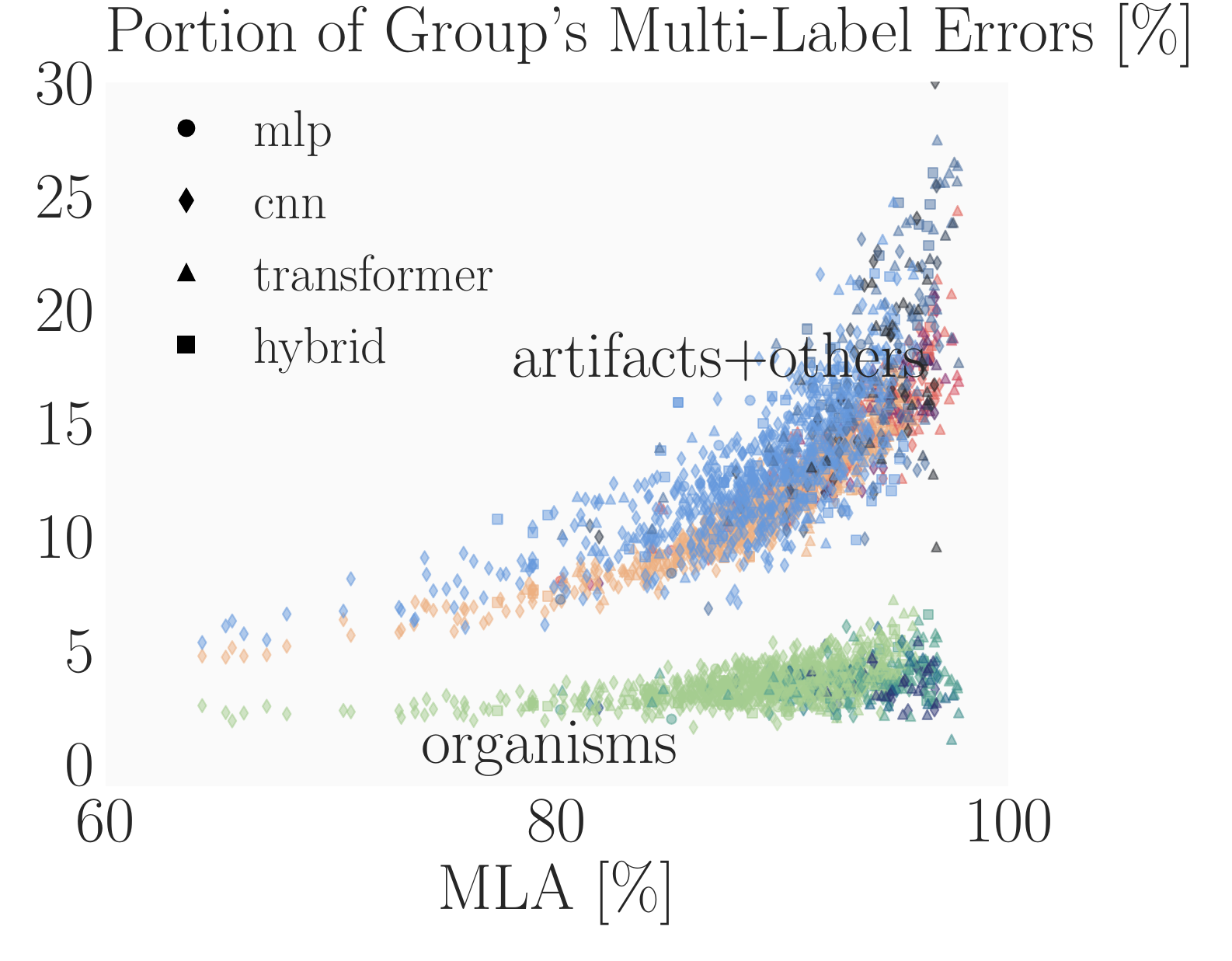}
	}\hfil
	\subfloat[Abs. number per group.]{
		\includegraphics[width=0.227\linewidth]{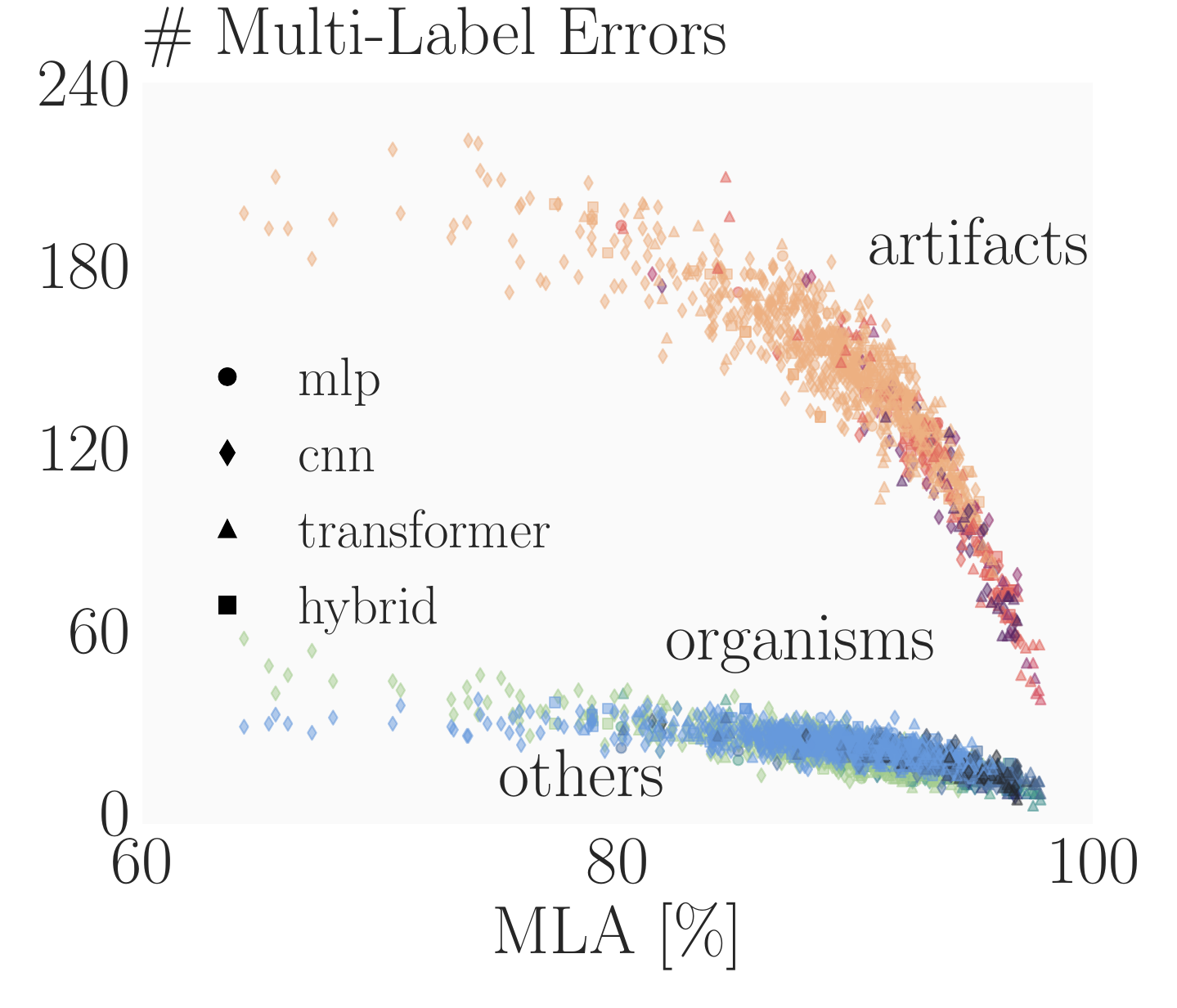}
	}\hfil
	\subfloat[Relative portion for all samples.]{
		\includegraphics[width=0.223\linewidth]{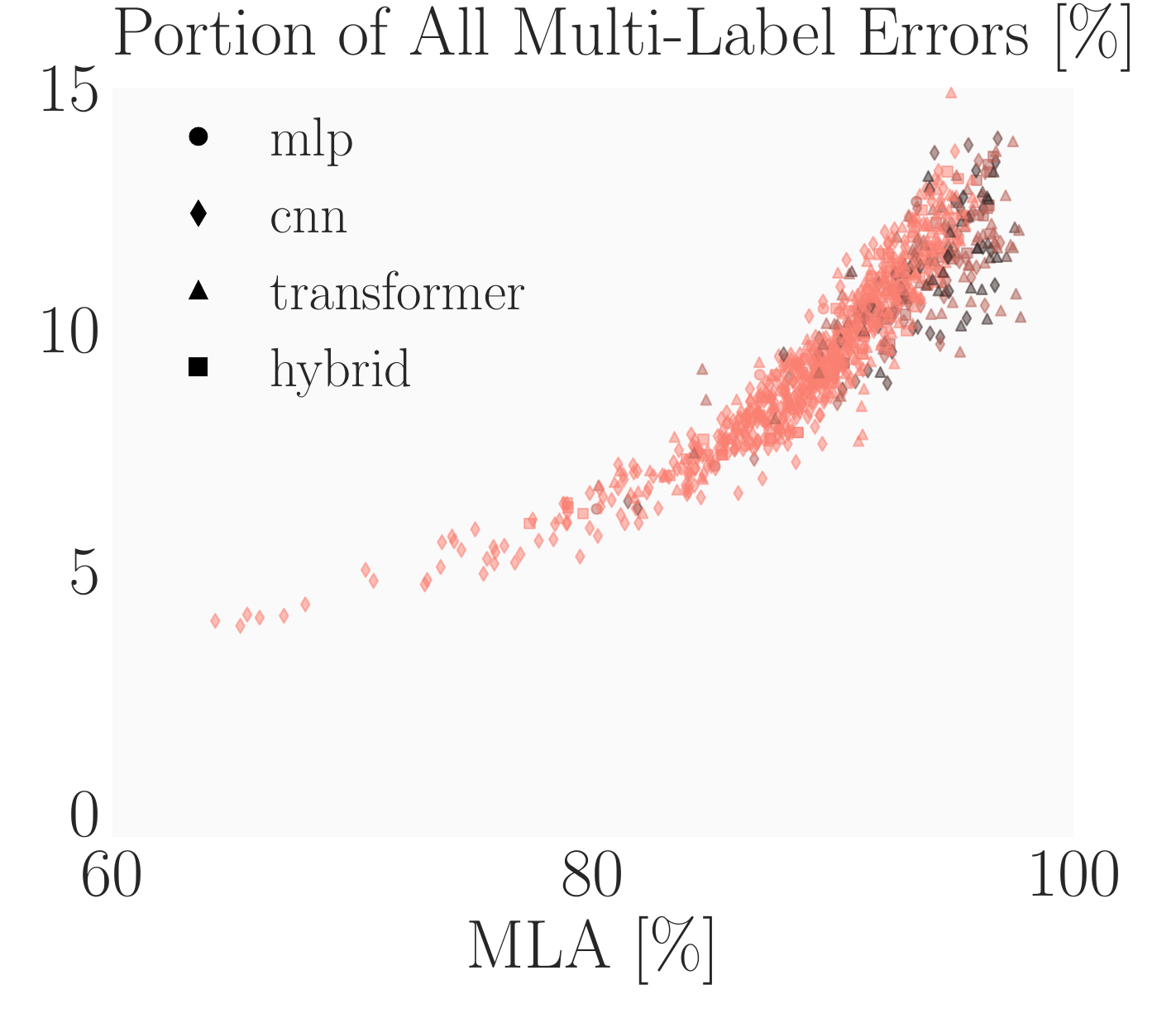}
	}\hfil
	\subfloat[Abs. number for all samples.]{
		\includegraphics[width=0.226\linewidth]{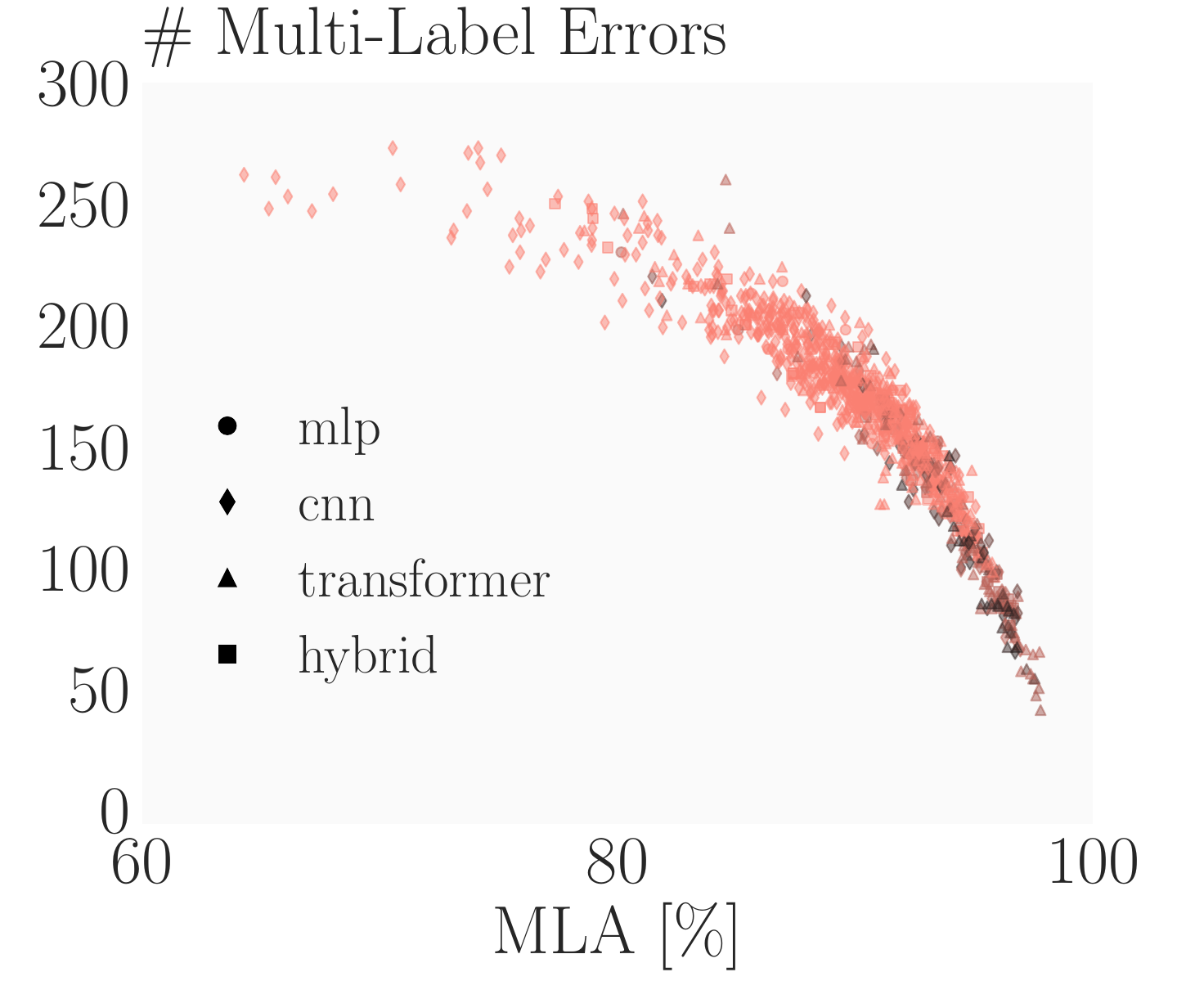}
	}

	\caption{
		Fine-grained out-of-vocabulary multi-label errors. The portion of errors follows a similar trend for ``artifacts'' and ``others'', while the sudden sharp drop in absolute numbers for ``artifacts'' is absent for ``others''.
	}
	\label{fig:fine_grained_oov_app}
\end{figure}

\begin{figure}[H]
	\centering
	\subfloat[Relative portion per group.]{
		\includegraphics[width=0.248\linewidth]{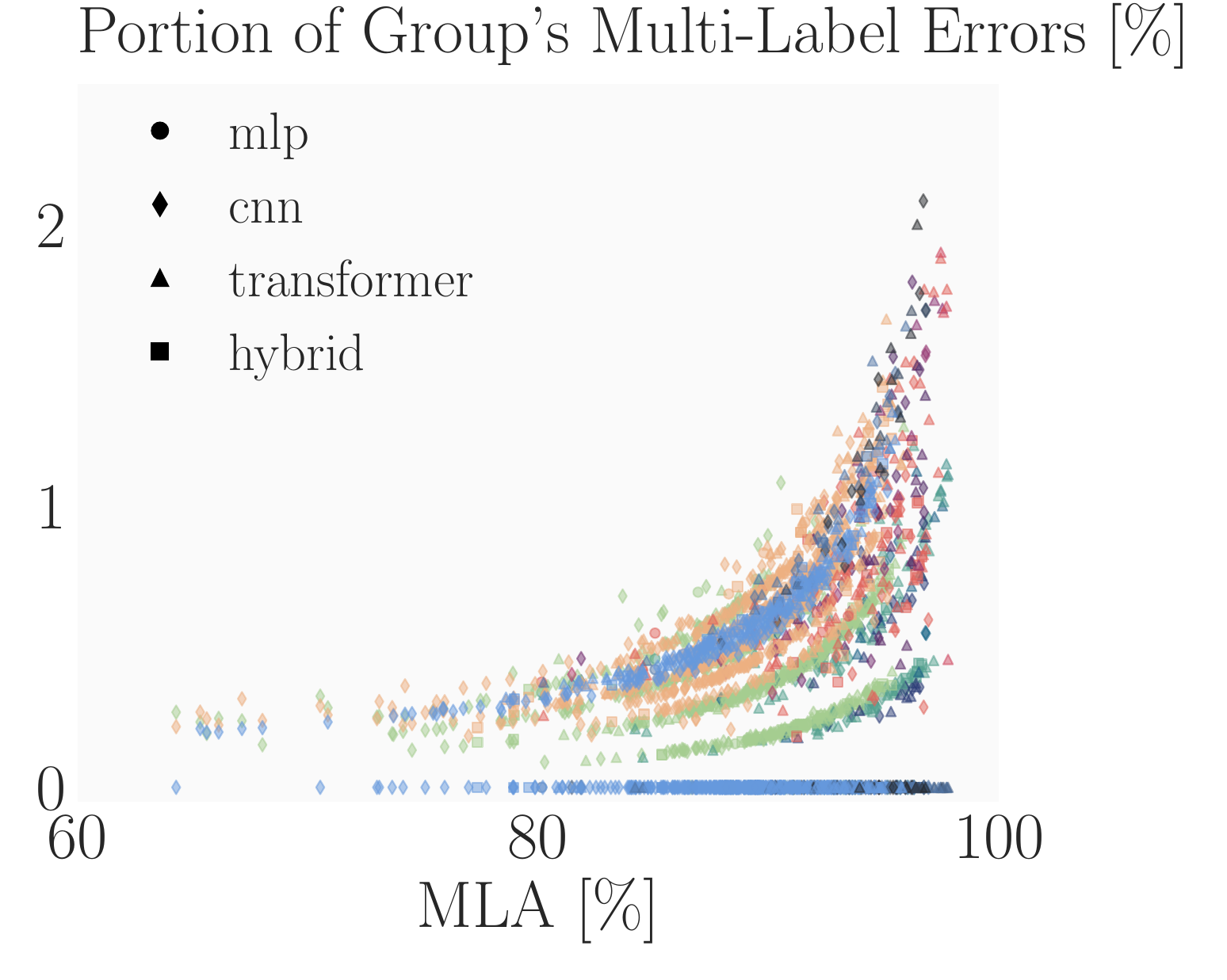}
	}\hfil
	\subfloat[Abs. number per group.]{
		\includegraphics[width=0.227\linewidth]{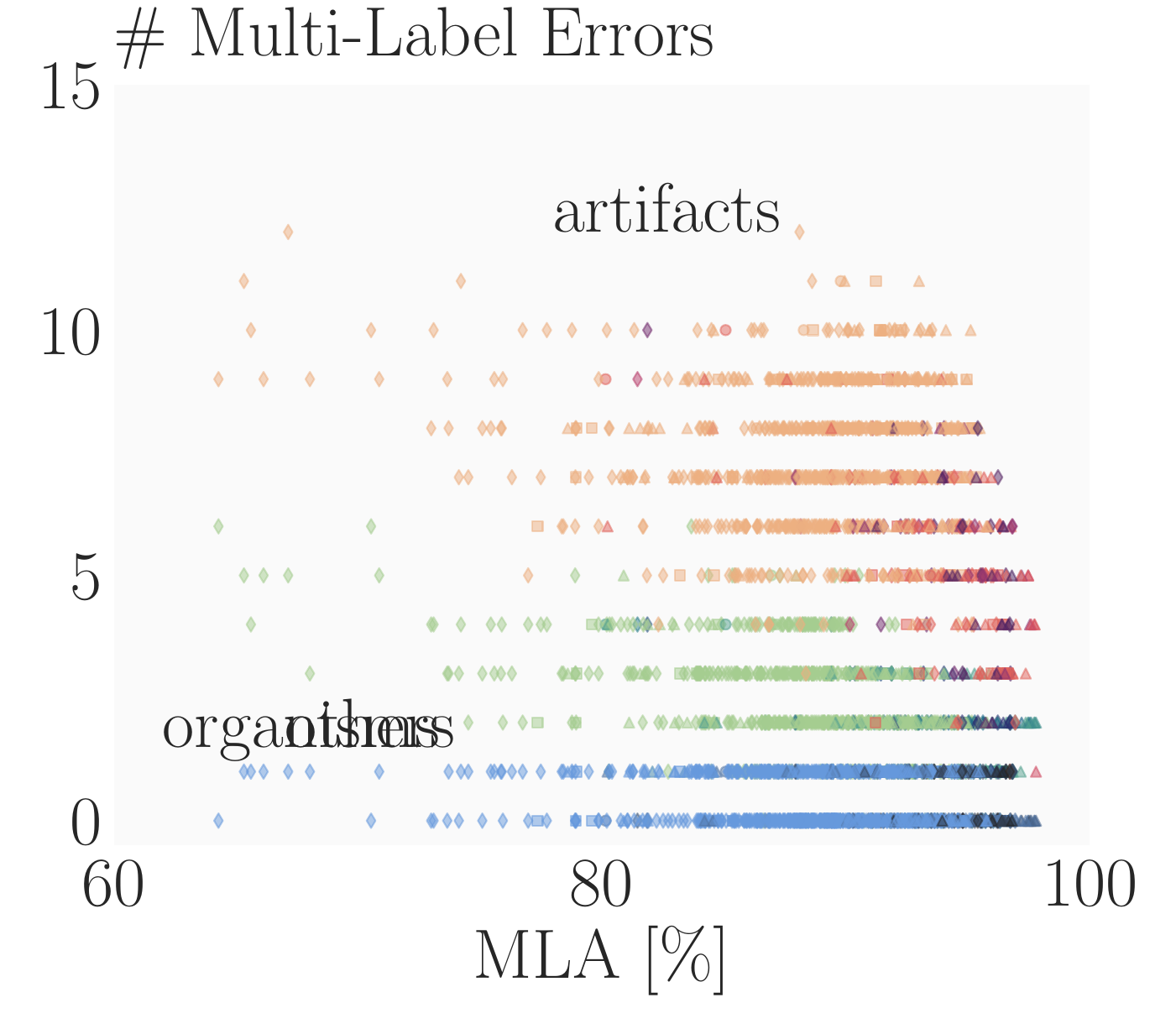}
	}\hfil
	\subfloat[Relative portion for all samples.]{
		\includegraphics[width=0.223\linewidth]{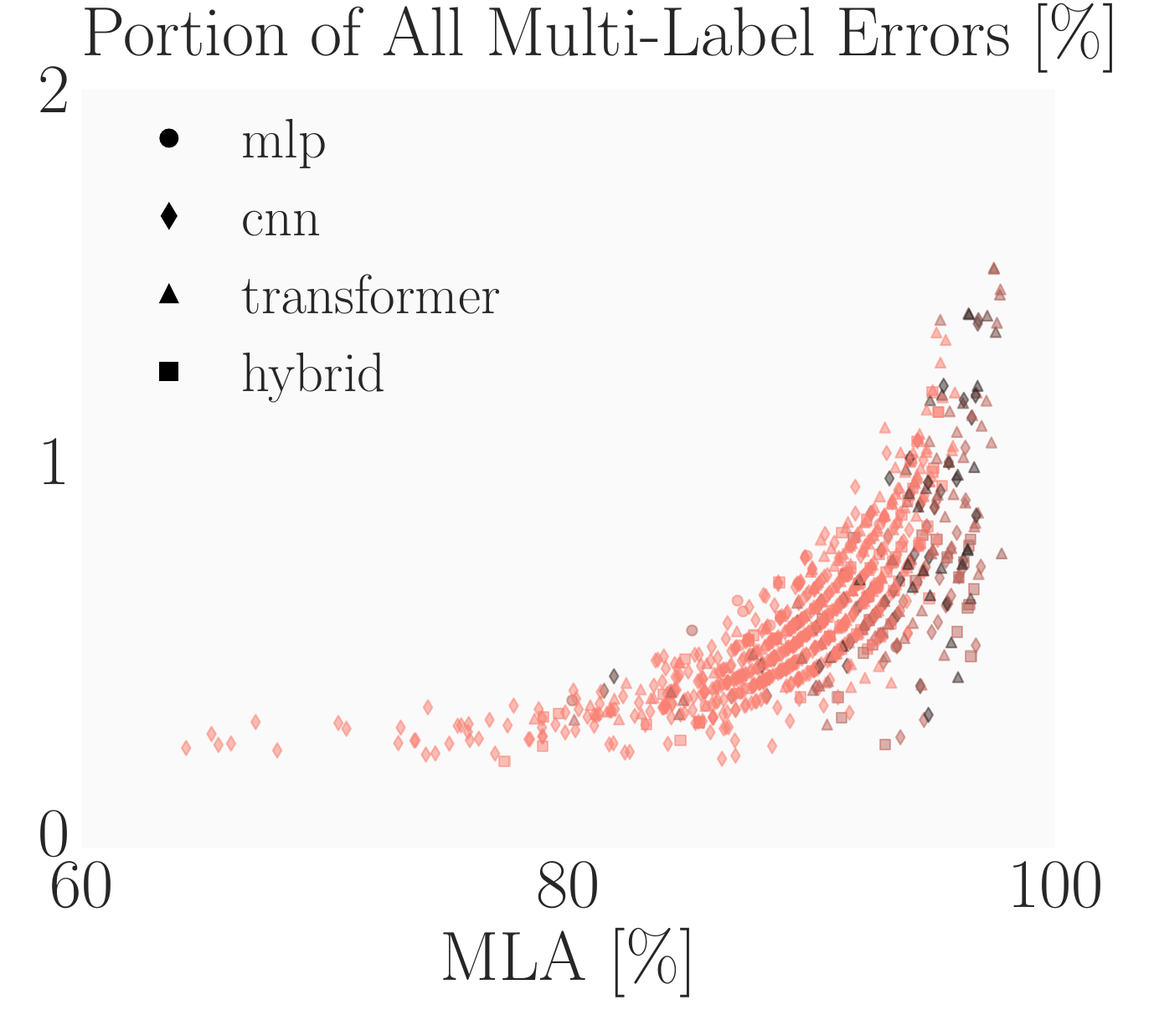}
	}\hfil
	\subfloat[Abs. number for all samples.]{
		\includegraphics[width=0.226\linewidth]{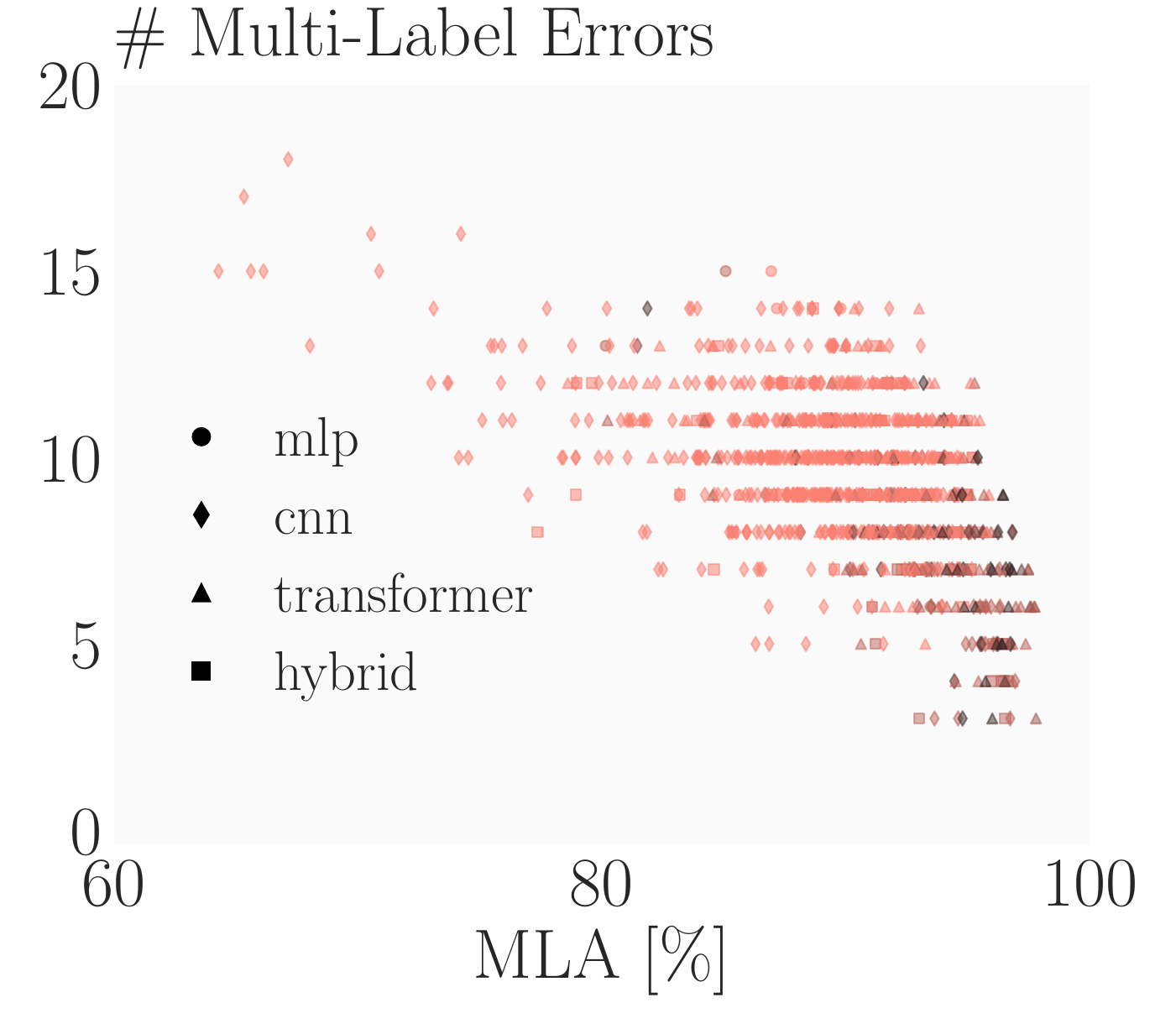}
	}

	\caption{
		Multi-label errors identified as non-prototypical instances. The number of samples identified as non-prototypical by \citet{VasudevanCLFR22} is inherently small (36) and the majority of them are artifacts, leading to noisy results. As the models get better at other types of errors, the relative portion of multi-label errors due to non-prototypical images increases.
	}
	\label{fig:non_prot_app}
\end{figure}

\begin{figure}[H]
	\centering
	\subfloat[Relative portion per group.]{
		\includegraphics[width=0.248\linewidth]{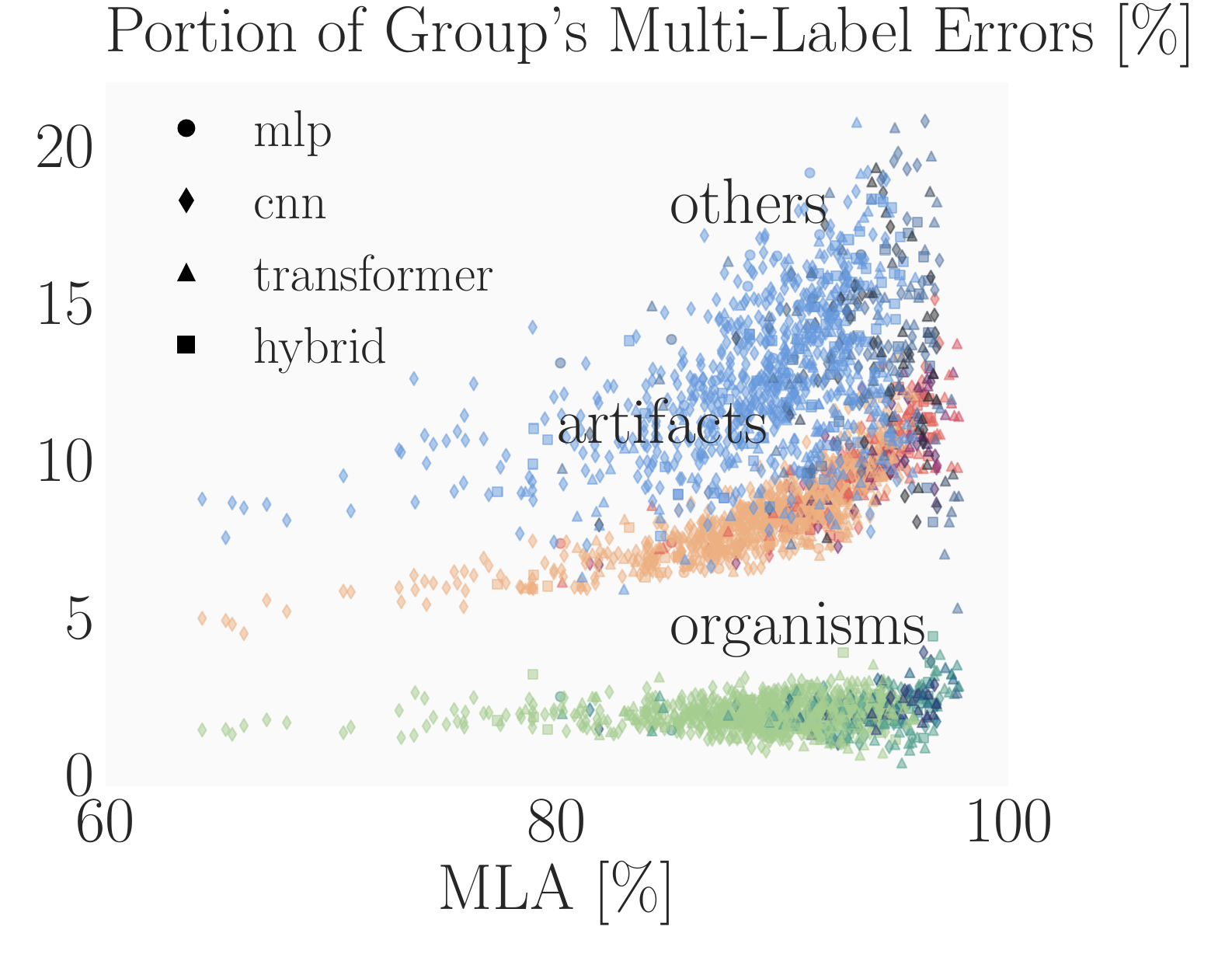}
	}\hfil
	\subfloat[Abs. number per group.]{
		\includegraphics[width=0.227\linewidth]{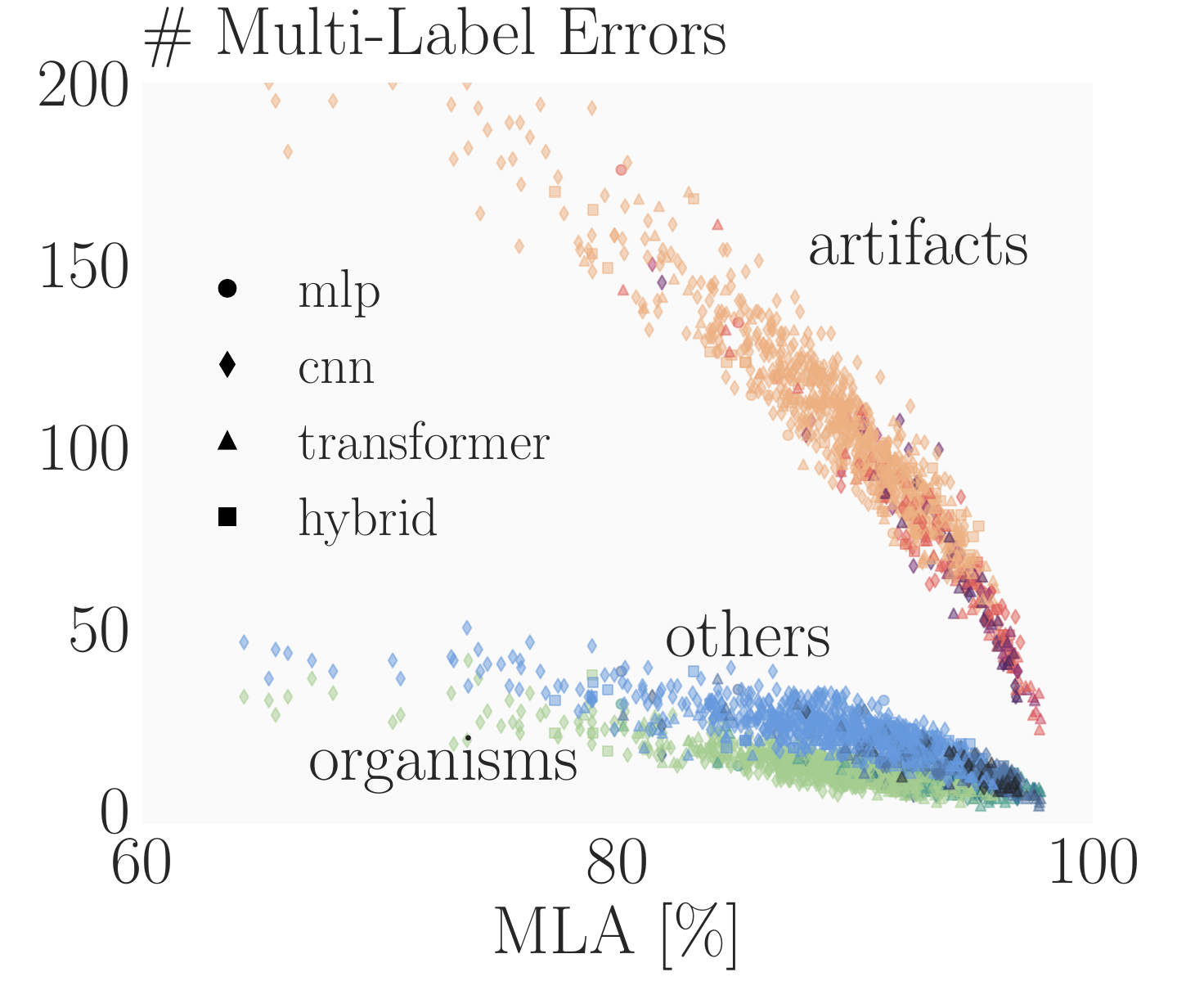}
	}\hfil
	\subfloat[Relative portion for all samples.]{
		\includegraphics[width=0.223\linewidth]{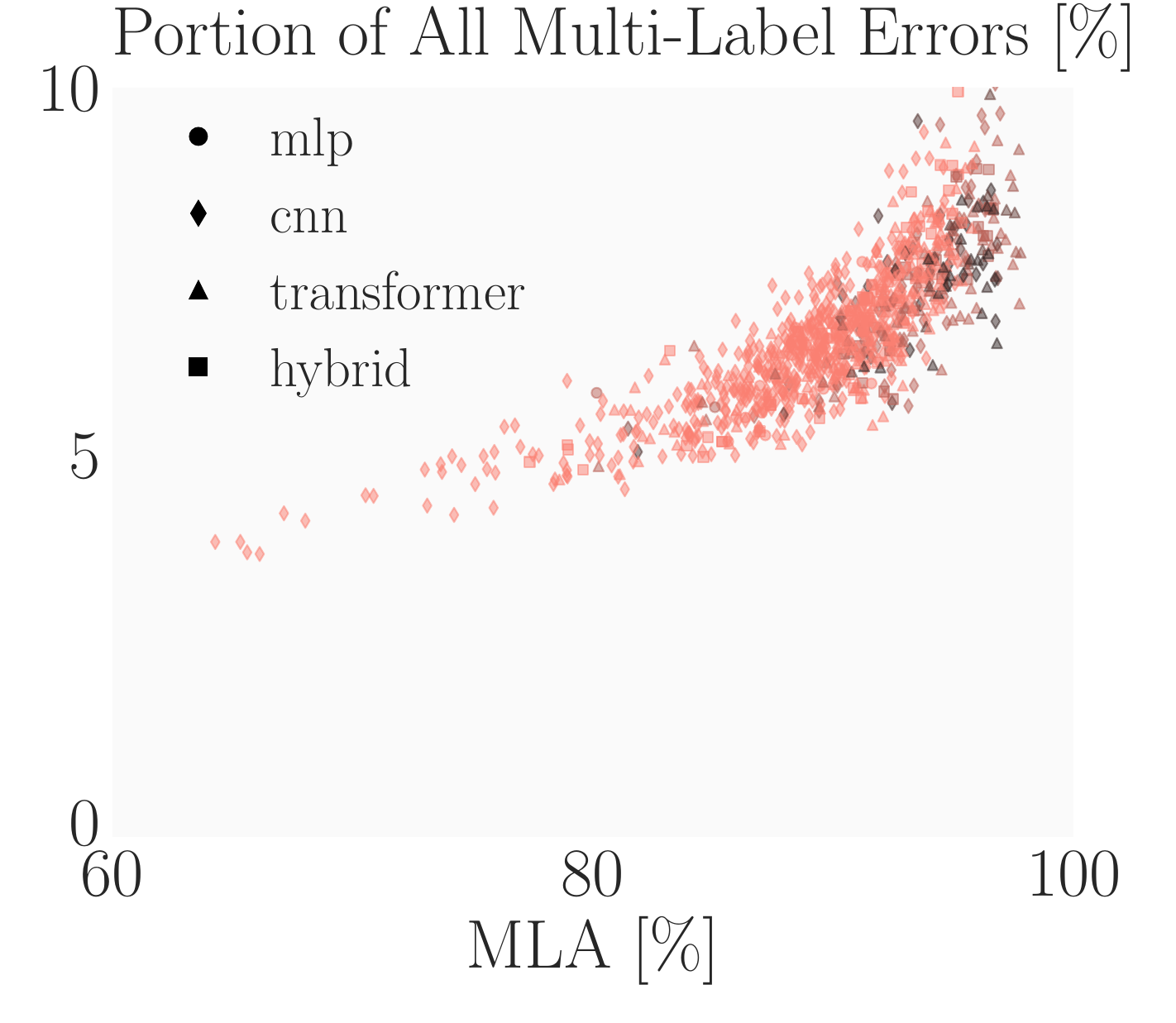}
	}\hfil
	\subfloat[Abs. number for all samples.]{
		\includegraphics[width=0.226\linewidth]{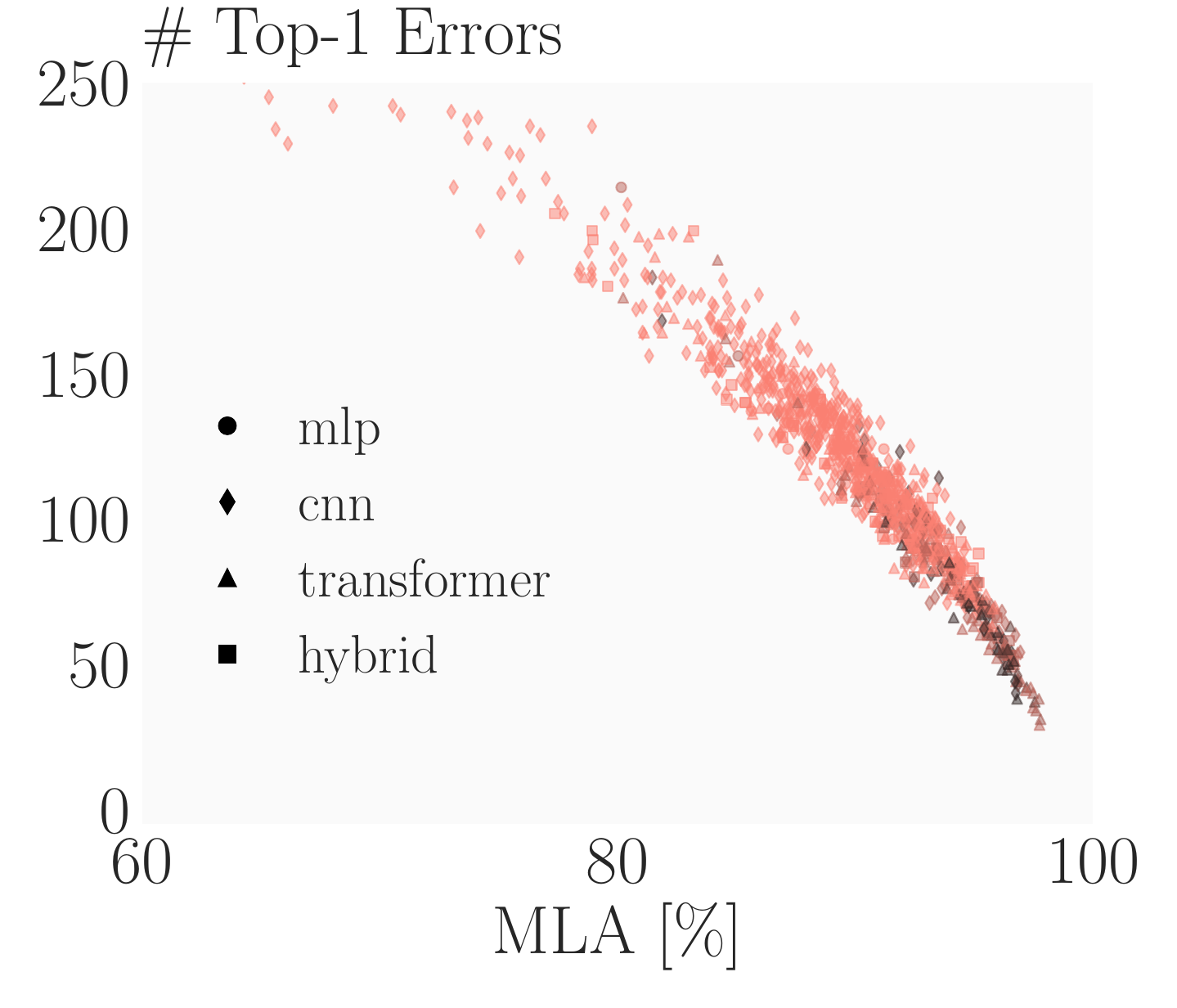}
	}

	\caption{
		Multi-label errors caused by spurious correlations. In relative terms, ``artifacts'' and ``others'' follow a similar trend which is noisier for the ``others'' group.
	}
	\label{fig:spurious_app}
\end{figure}

Finally, in \cref{fig:mle_app} we present the ratio of unexplained model failures (MLF) over multi-label accuracy (MLA) and standard top-1 accuracy. Again, subfigures (a) and (b) are organized per group, while  (c) and (d) show the results for all samples.

\begin{figure}[H]
	\centering
	\subfloat[Portion of unexplained model failures per group.]{
		\includegraphics[width=0.251\linewidth]{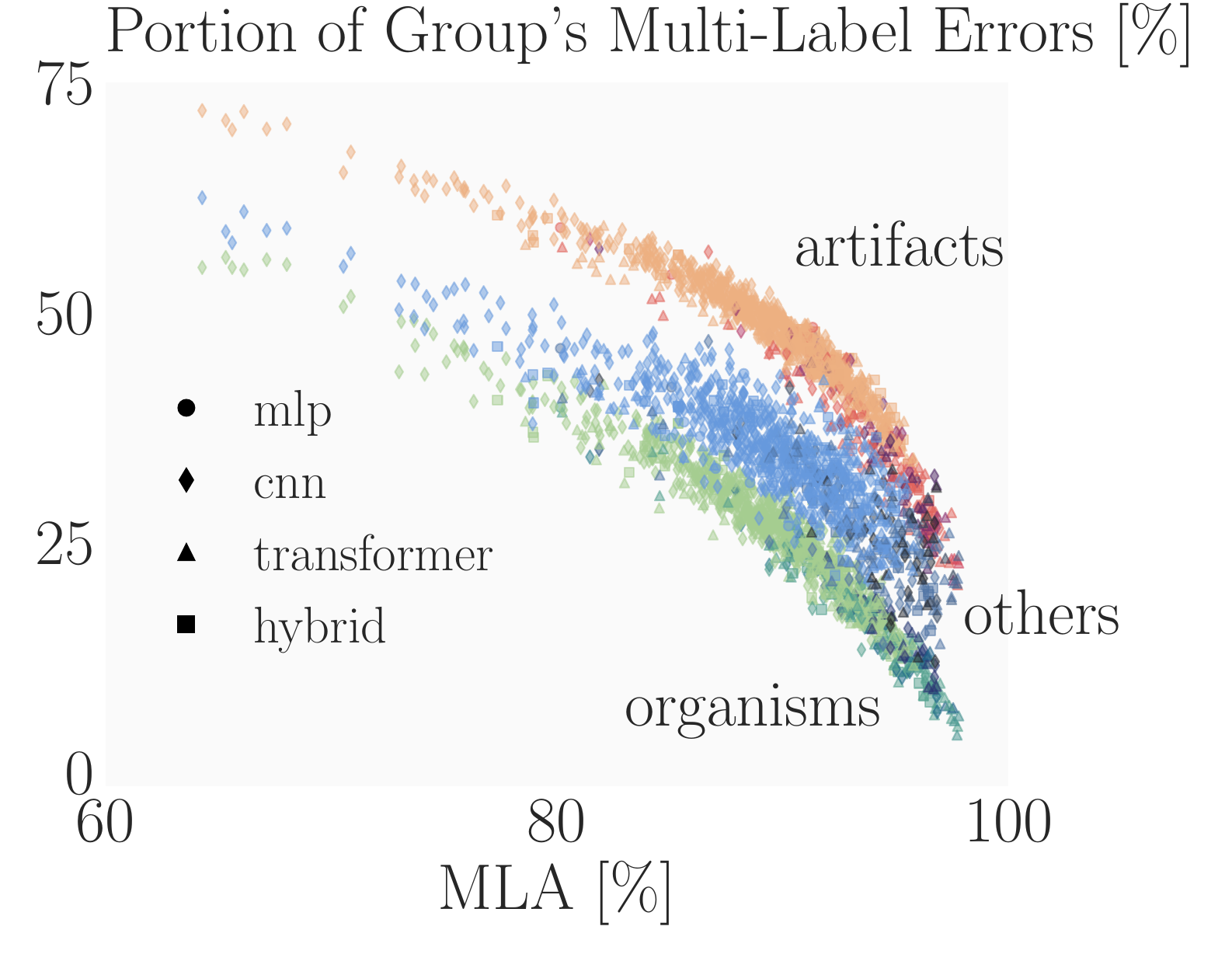}
	}\hfil
	\subfloat[Portion of unexplained model failures per group.]{
		\includegraphics[width=0.22\linewidth]{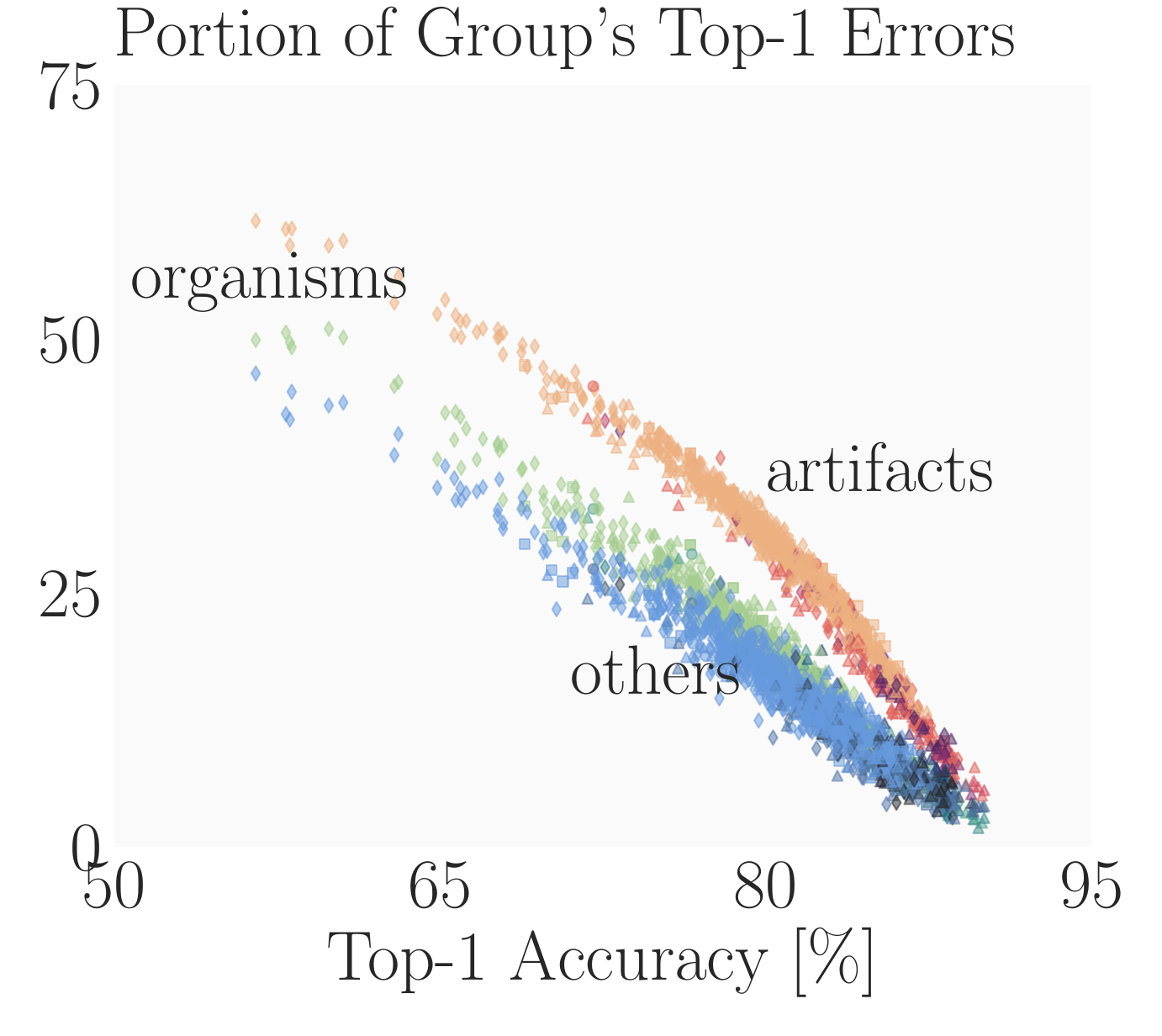}
	}\hfil
	\subfloat[Portion of unexplained model failures for all samples.]{
		\includegraphics[width=0.226\linewidth]{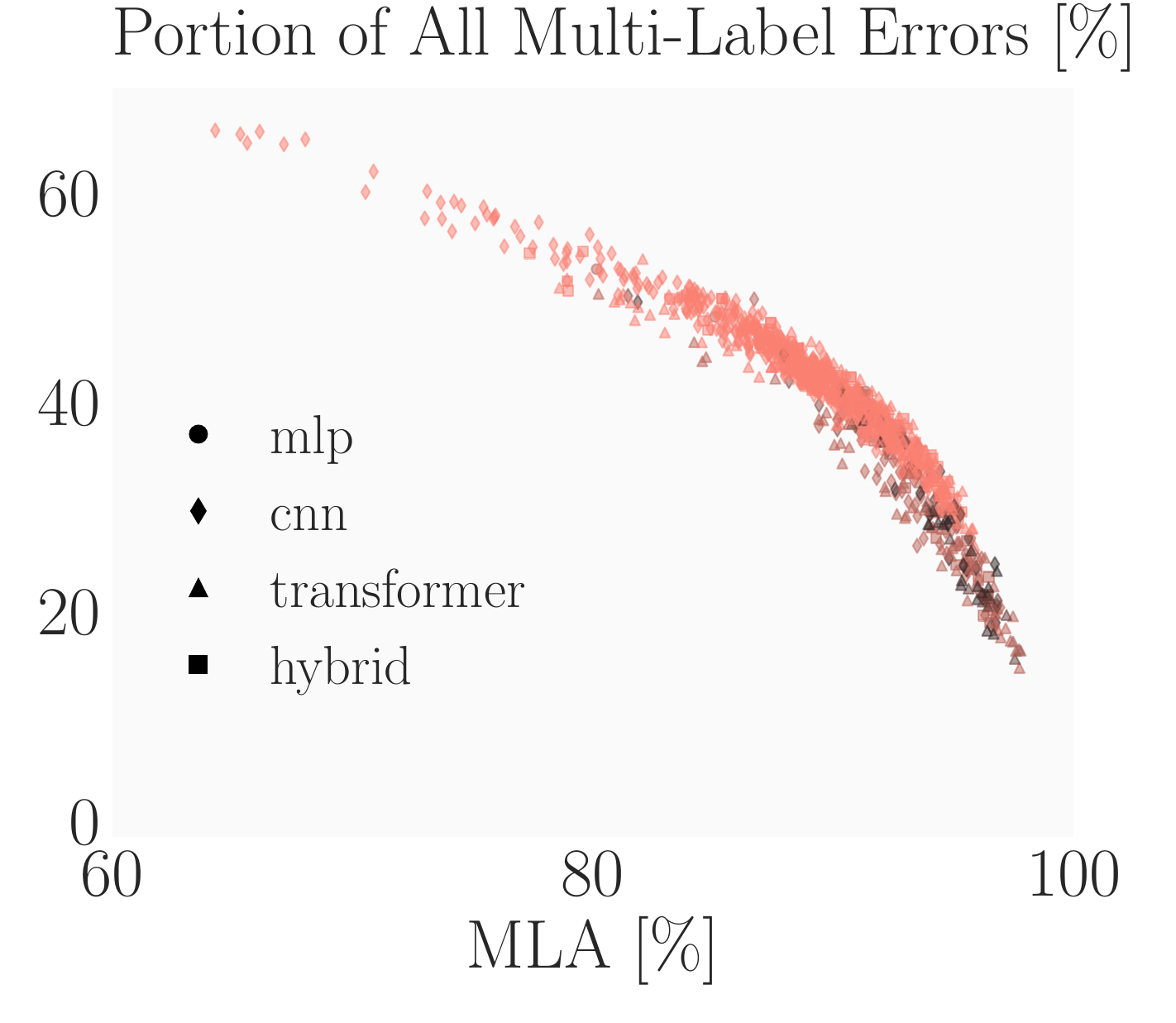}
	}\hfil
	\subfloat[Portion of unexplained model failures for all samples.]{
		\includegraphics[width=0.22\linewidth]{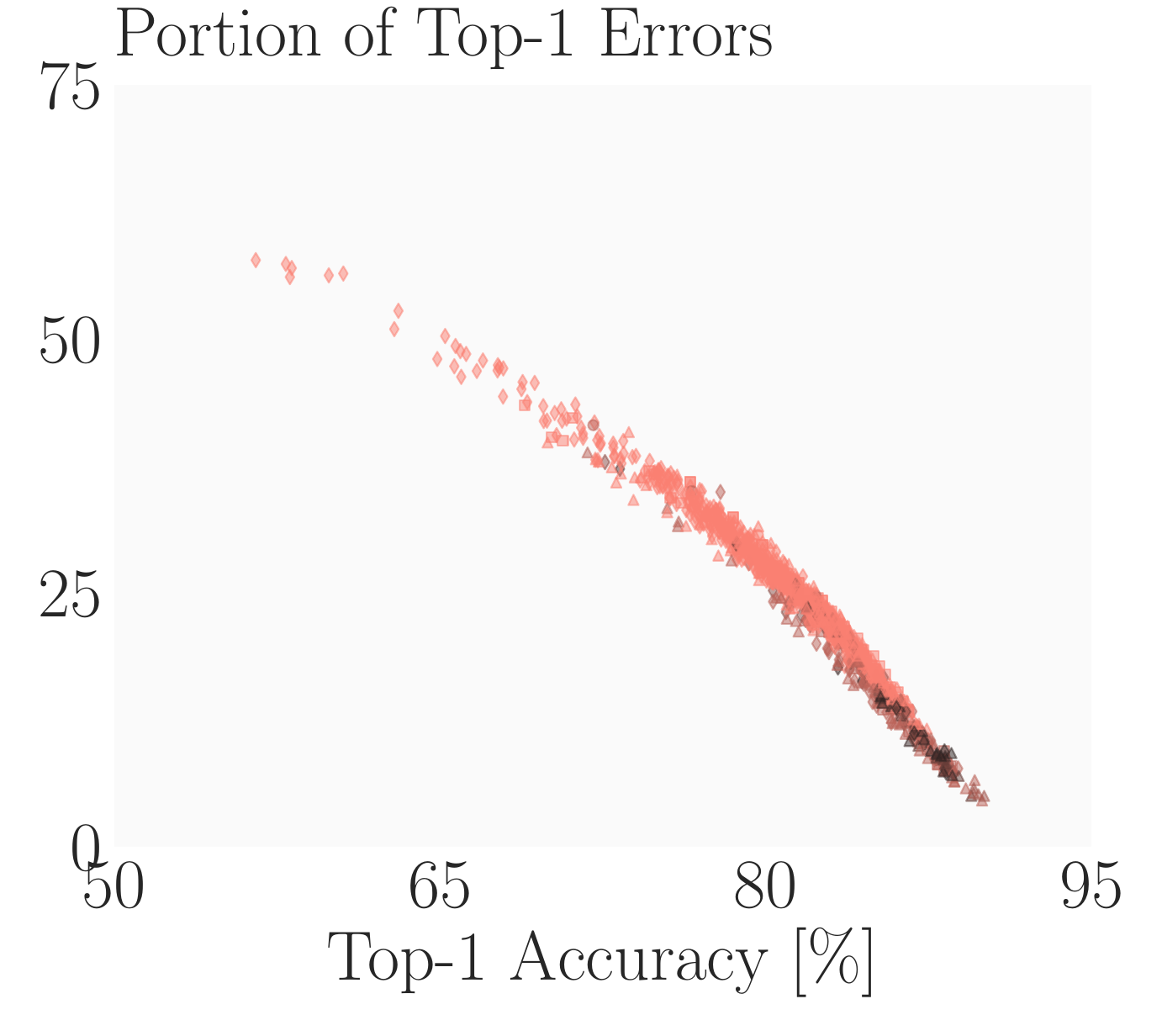}
	}

	\caption{
		Ratio of unexplained model failures (MLF) over multi-label accuracy (MLA -- (a) and (c)) and standard top-1 accuracy ((b) and (d)). In general, our pipeline explains the biggest portion of errors made on ``organisms'' and the lowest portion of errors on ``artifacts''. However, better models quickly close this gap. For the best models, our pipeline explains up to $85\%$ of all multi-label errors and $95\%$ of all top-1 errors.
	}
	\label{fig:mle_app}
\end{figure}

\newpage
\section{Analysis of \INA} \label{sec:ina}

\begin{wrapfigure}[12]{r}{0.6\textwidth}
    \centering
    \vspace{-4mm}
    \begin{minipage}[t]{0.49\linewidth}
        \centering
        \includegraphics[width=0.9\linewidth]{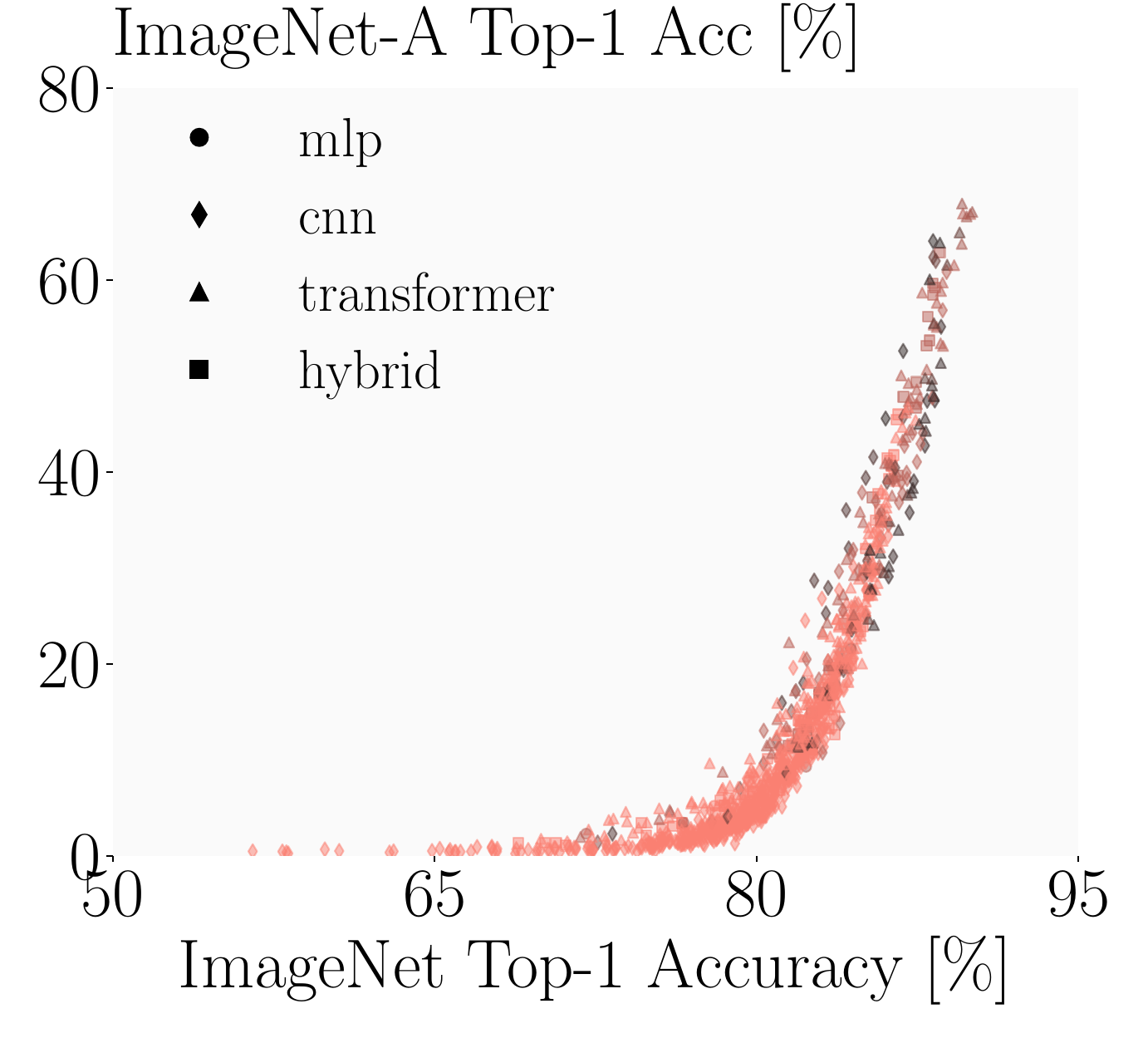}
    \end{minipage}
    \hfil
    \begin{minipage}[t]{0.49\linewidth}
        \centering
        \includegraphics[width=0.9\linewidth]{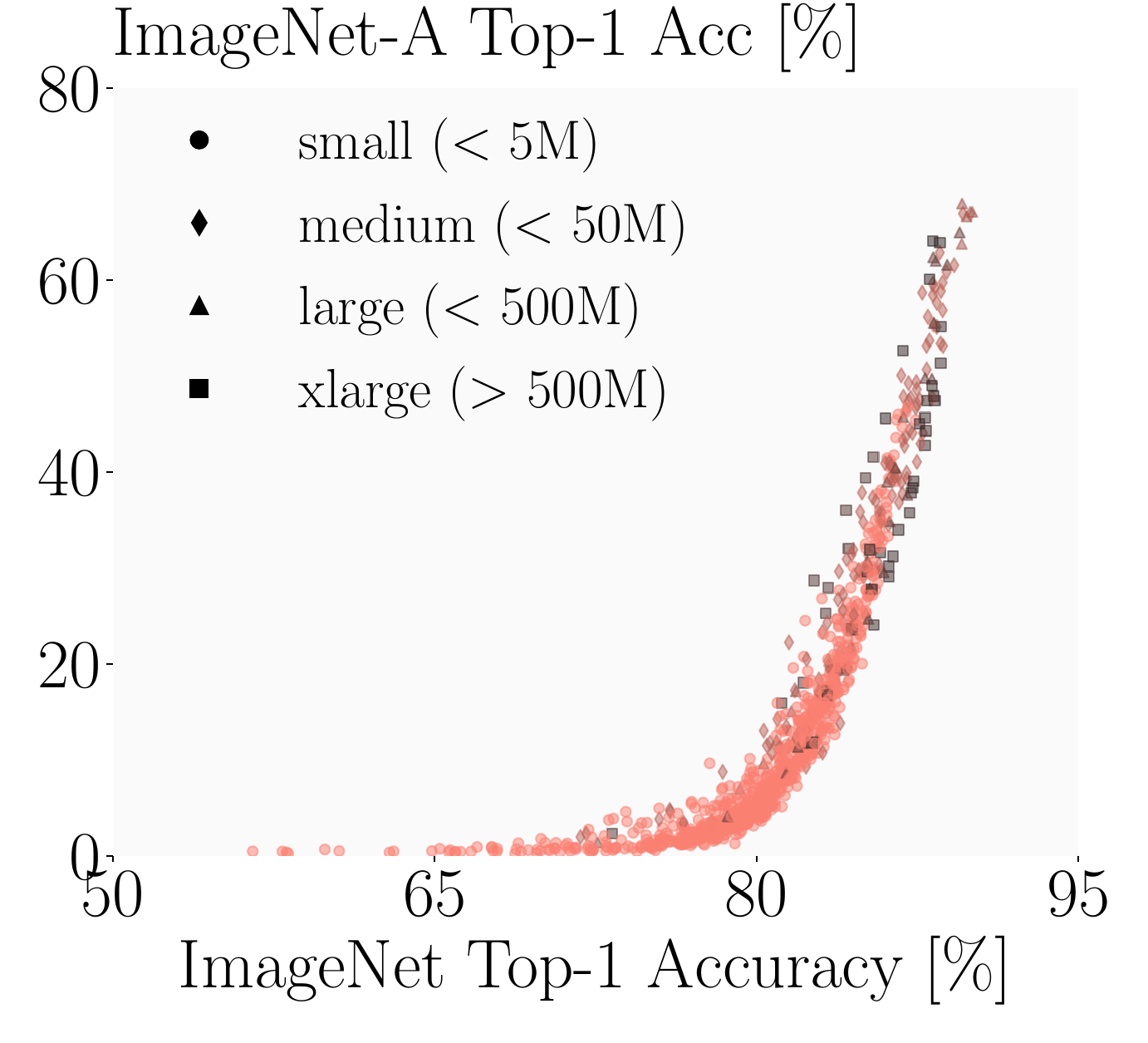}
    \end{minipage}
    \caption{\INA \toa Accuracy vs \IN \toa Accuracy grouped by architecture (left) and training data size (right).}
    \label{fig:imagenet_a_vs_imagenet}
\end{wrapfigure}
To demonstrate the applicability of our automated error analysis pipeline to other datasets, we now apply it to \INA.
\INA was introduced by \citet{HendrycksZBSS21} as an \IN-like test set of ``Natural Adversarial Examples''. It contains $7500$ images from $200$ \IN classes, picked to fool a set of ResNet-50 classifiers, while still constituting high-quality, single-entity images, clearly depicting an entity of the target class.
A comparison of \IN and \INA accuracy is shown in \cref{fig:imagenet_a_vs_imagenet}.
In this section, marker color encodes model parameter counts, i.e., larger models have darker markers.

As \INA images were selected to be challenging for \IN classifiers at the time, we see that many classifiers with \IN accuracies below $80\%$ have almost $0$ accuracy on \INA. In fact, 801 of the \nmodels models we consider have an accuracy of less than $30\%$ on \INA. 
As classifiers improved ($ > 80\%$ \toa \IN accuracy) \INA accuracy catches up rapidly. 
Among other improvements, the main driver seems to be the model parameter count.
Note that due to the much lower \INA accuracy (of some models), the x-axis of all further plots starts at $0\%$ rather than $60\%$, as before.

As \INA uses \IN classes, we keep the split into organisms (77 out of 200 classes, $38.5\%$), artifacts (104 / 200, $52\%$), and other classes (19 / 200, $9.5\%$).

\paragraph{Class Overlap}
\begin{wrapfigure}[12]{r}{0.6\textwidth}
    \centering
    \vspace{-4mm}
    \begin{minipage}[t]{0.49 \linewidth}
        \centering
        \includegraphics[width=0.9\linewidth]{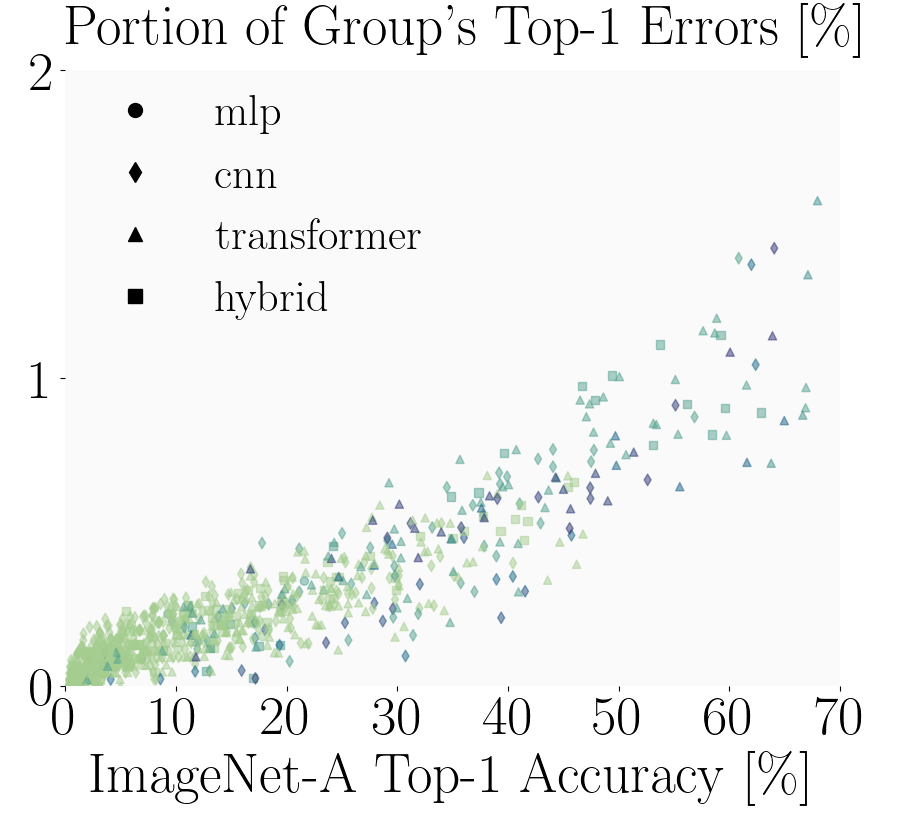}
    \end{minipage}
    \hfil
    \begin{minipage}[t]{0.49 \linewidth}
        \centering
        \includegraphics[width=0.9\linewidth]{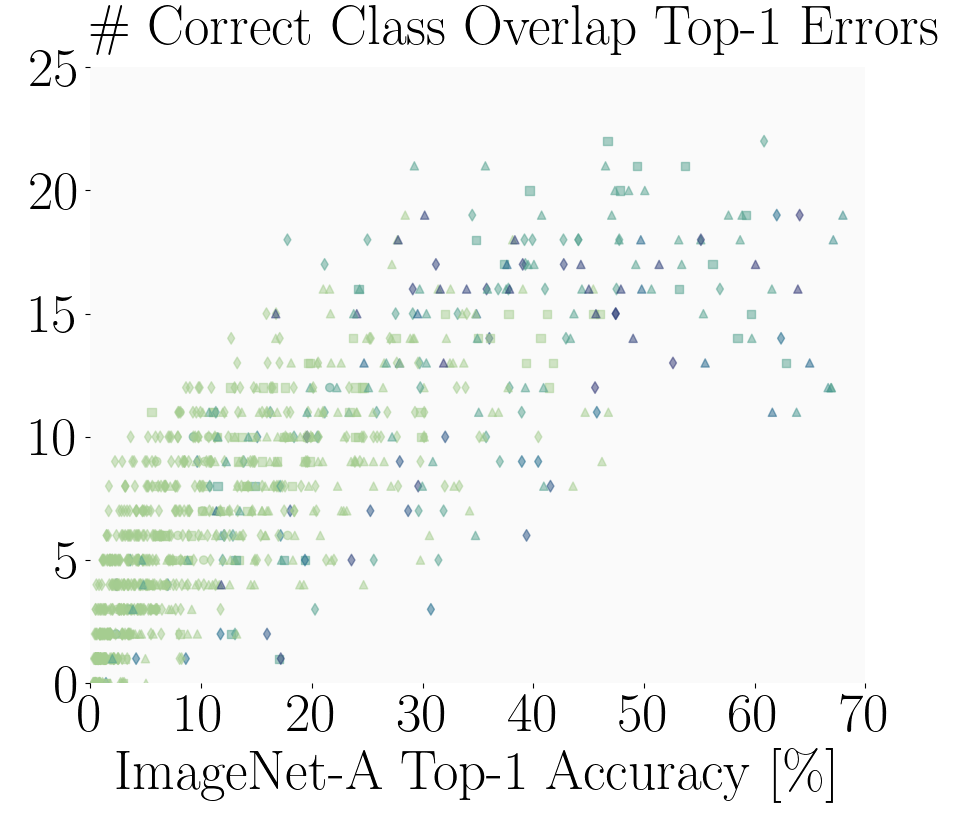}
    \end{minipage}
    \vspace{-2mm}
    \caption{Portion (left) and number (right) of \toa errors caused by \emph{class overlap} for organisms (green). For the artifacts and other groups we observe no errors.}
    \label{fig:imagenet_a:class_overlap}
\end{wrapfigure}

Due to the shared class structure, we can use the same analysis as for \IN and show the results in \cref{fig:imagenet_a:class_overlap}.
The 200 classes in \INA were selected from the 1000 \IN classes to avoid class overlap and label ambiguity. Thus, as expected, we see a much lower overall amount of such errors (below $0.8\%$ instead of up to $8.5\%$ for \IN). In particular, for artifacts and other classes, we see no class overlap errors at all.
Interestingly, for organisms, we still observe a small number ($\leq 22$) of errors caused by class overlap. We observe that errors become more frequent (albeit still rare) for models with higher accuracy. As with \IN this is due to more accurate models classifying more images correctly, including to overlapping classes.

\paragraph{Multi-Label Annotations}
In contrast to \IN, \INA was designed to only include single-entity images. Thus, we skip the multi-label analysis for \INA and use \toa accuracy/error rather than \mla accuracy/error for all further analyses.

\paragraph{Fine-Grained Classification Errors}
\begin{wrapfigure}[10]{r}{0.6\textwidth}
    \centering
    \vspace{-8mm}
    \begin{minipage}[t]{0.49 \linewidth}
        \centering
        \includegraphics[width=0.97\linewidth]{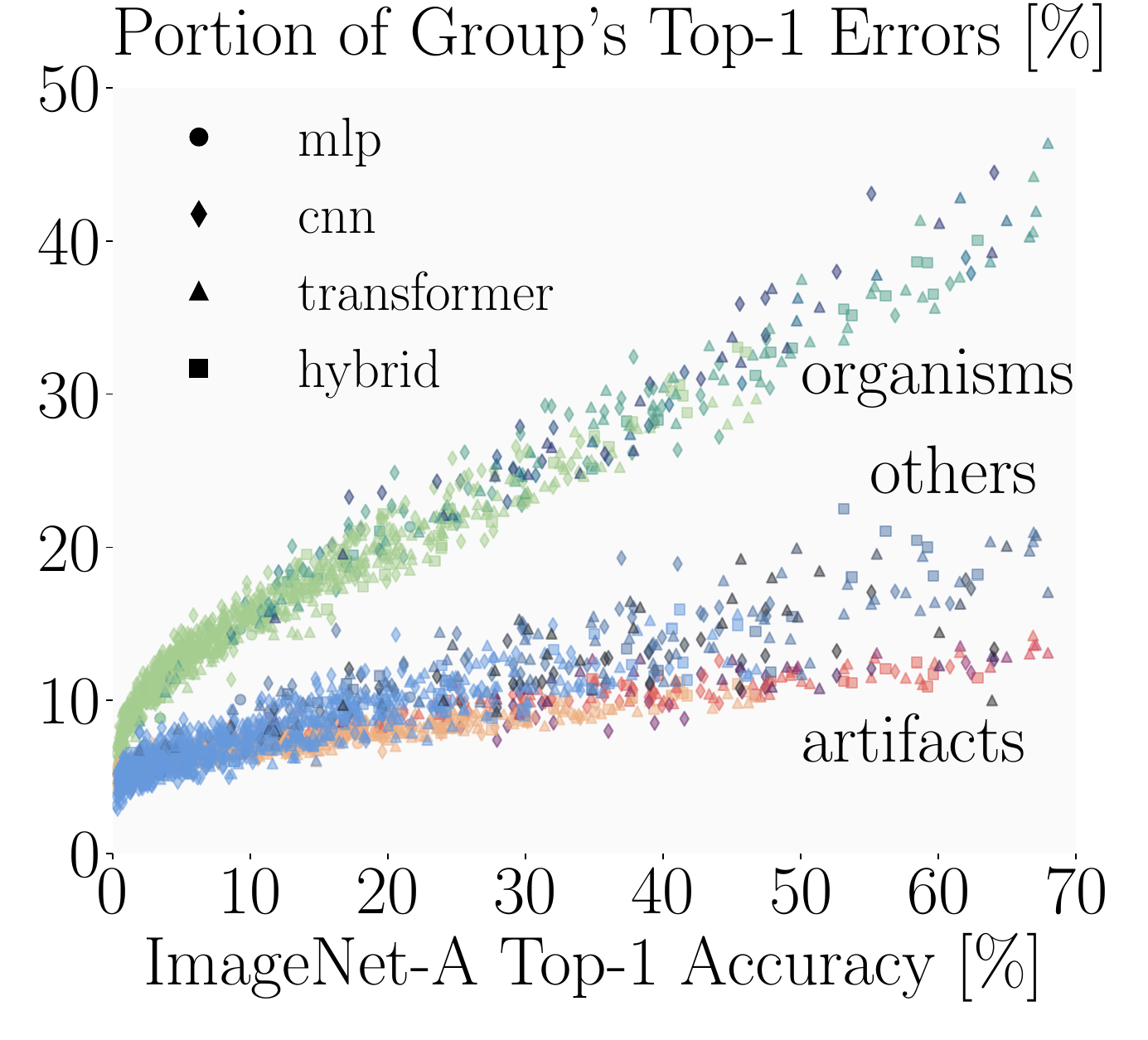}
    \end{minipage}
    \hfil
    \begin{minipage}[t]{0.49 \linewidth}
        \centering
        \includegraphics[width=0.97\linewidth]{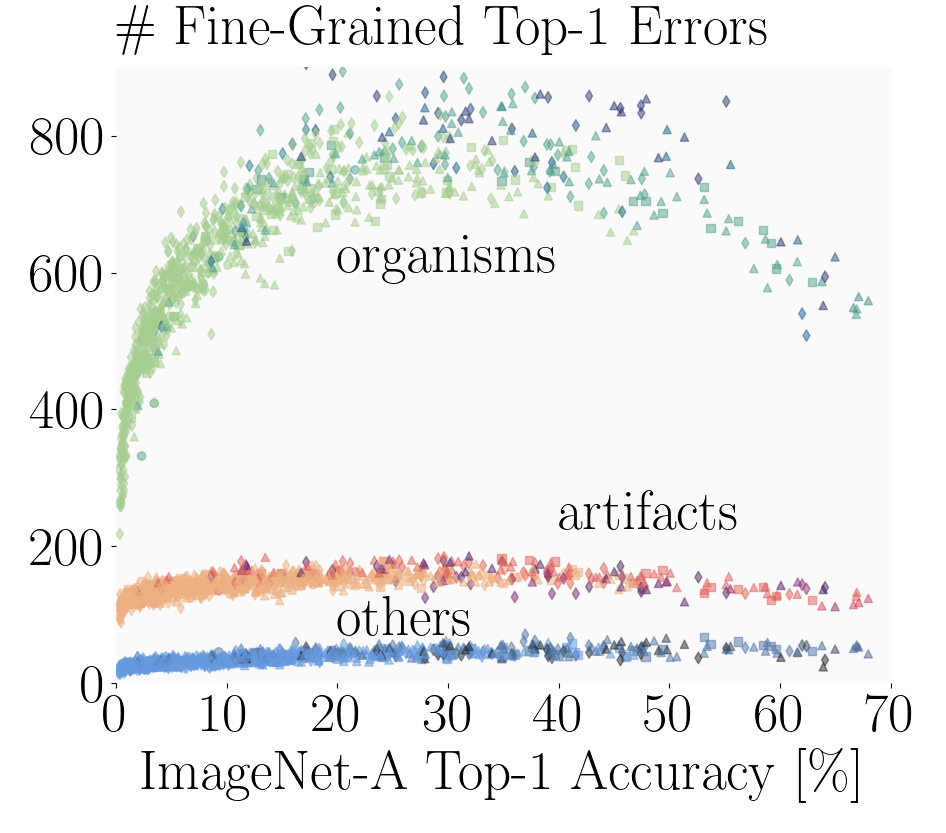}
    \end{minipage}
    \vspace{-2mm}
    \caption{Portion (left) and number (right) of \toa errors caused by \emph{fine-grained misclassifications} by group -- organisms (green), artifacts (red), and others (blue).}
    \label{fig:imagenet_a:fine_grained}
\end{wrapfigure}
Due to the shared class structure, we can apply the same analysis including superclasses for fine-grained classification errors as on \IN.
We show results in \cref{fig:imagenet_a:fine_grained}.

As with \IN, the group for which we observe most of these errors are organisms (explaining up to $47\%$ of the group's errors). However, the portion of explained errors is much lower than for \IN, where up to $88\%$ of \mla errors for organisms can be explained by fine-grained classification errors.
We see the percentage of errors increase linearly with accuracy. Looking at absolute counts, especially for organisms, we see that fine-grained errors initially become more prevalent as model accuracy increases before dropping off as models more frequently also succeed in the fine-grained class distinction.

\paragraph{Fine-Grained Out-of-Vocabulary Errors}
\begin{wrapfigure}[14]{r}{0.6\textwidth}
    \centering
    \begin{minipage}[t]{0.49 \linewidth}
        \centering
        \includegraphics[width=0.9\linewidth]{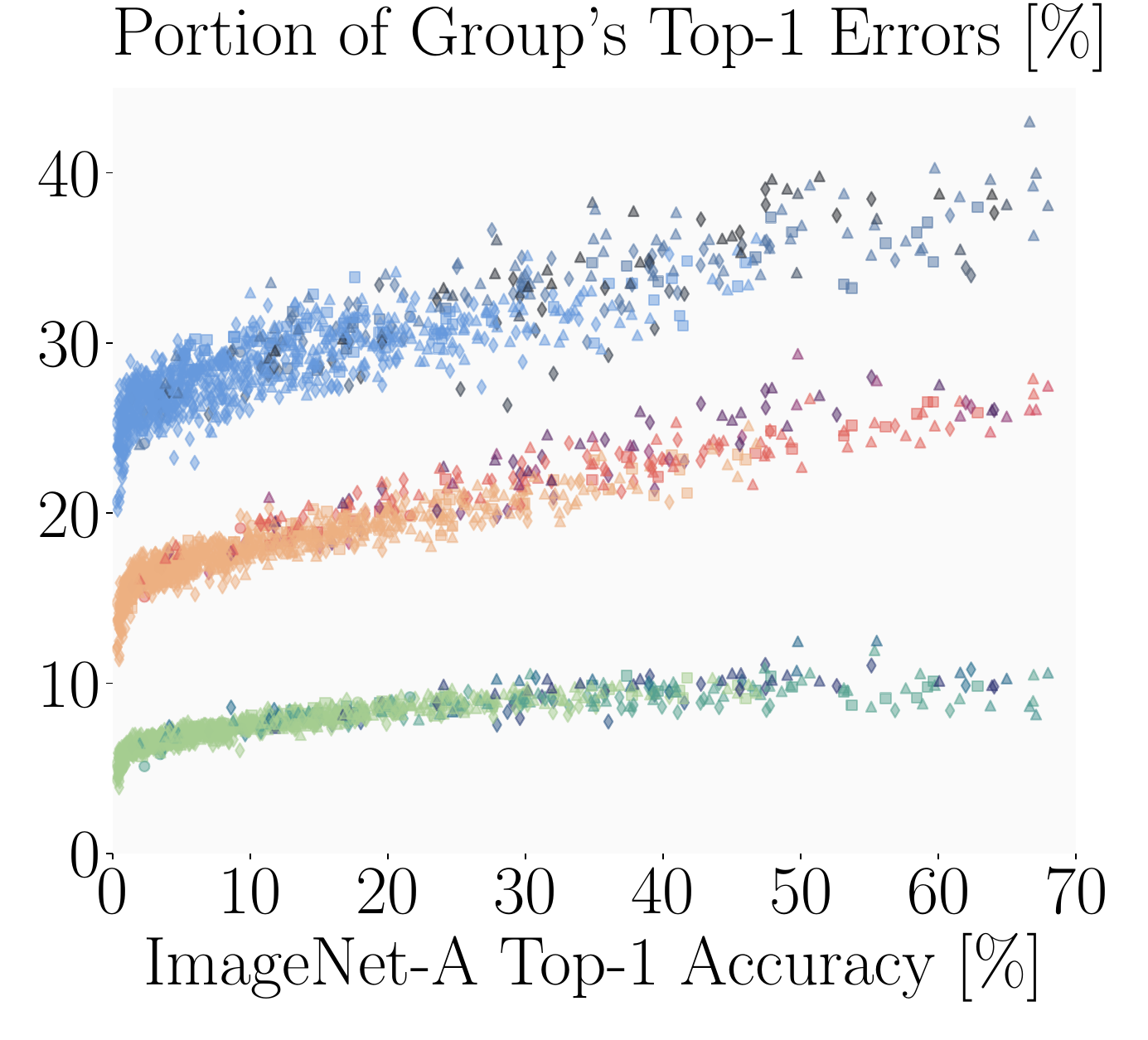}
    \end{minipage}
    \hfil
    \begin{minipage}[t]{0.49 \linewidth}
        \centering
        \includegraphics[width=0.9\linewidth]{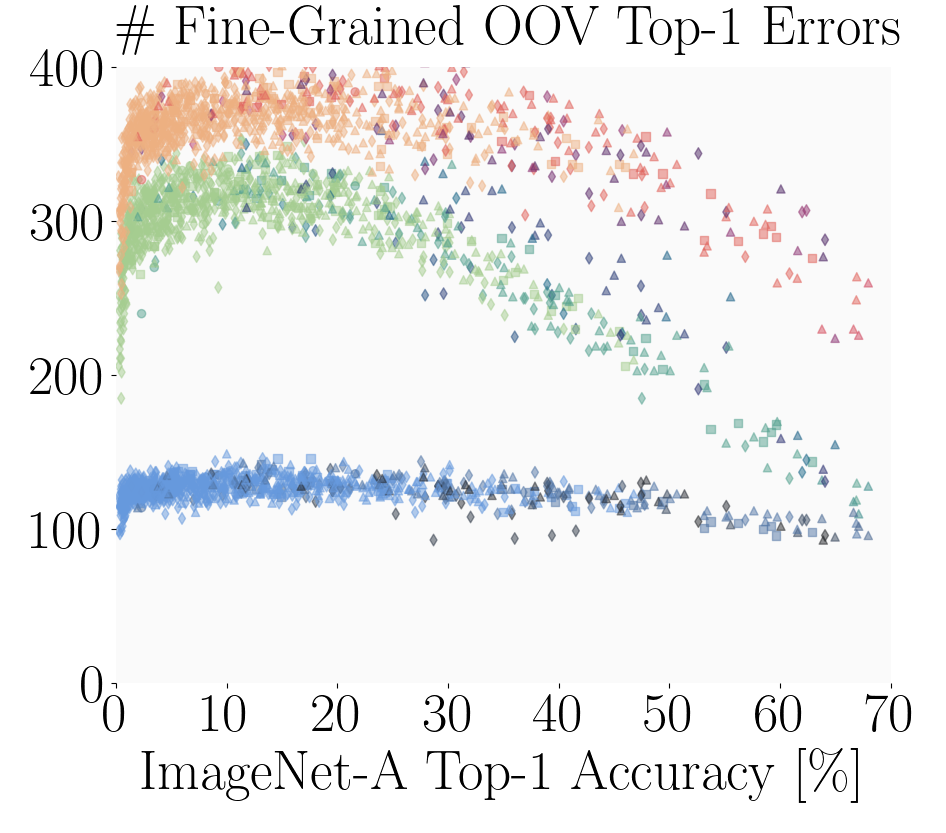}
    \end{minipage}
    \caption{Portion (left) and number (right) of \toa errors identified as \emph{fine-grained OOV} by group -- organisms (green), artifacts (red) and other (blue).}
    \label{fig:imagenet_a:fine_grained_oov}
\end{wrapfigure}
Again, we can use the same superclass definition, CLIP embedding and open world classifier as for \IN to identify fine-grained OOV errors.
We generally observe slightly larger portions of OOV errors than on \IN and a much clearer linear dependence on accuracy (see \cref{fig:imagenet_a:fine_grained_oov}). Notably, the superlinear increase observed on \IN (see \cref{fig:fine_grained_oov}) is absent here.
Looking at absolute counts, after an initial increase and stagnation, we observe a linear decrease with model accuracy. In contrast to \IN, we again do not observe a slope change.

\paragraph{Non-Prototypical Instances}
Our analysis pipeline for \IN uses the manual annotation of non-prototypical instances from \citep{VasudevanCLFR22}. As we do not have these available for \INA we skip this analysis step here.
However, for standard \IN we only observe a small number of such errors ($\leq 2\%$ of MLE) and note that the \INA images were manually selected to only include high-quality images \citep{HendrycksZBSS21}. Thus, we expect their portion to be even smaller here.

\paragraph{Spurious Correlations}
\begin{wrapfigure}[12]{r}{0.6\textwidth}
    \centering
    \vspace{-5mm}
    \begin{minipage}[t]{0.49 \linewidth}
        \centering
        \includegraphics[width=0.9\linewidth]{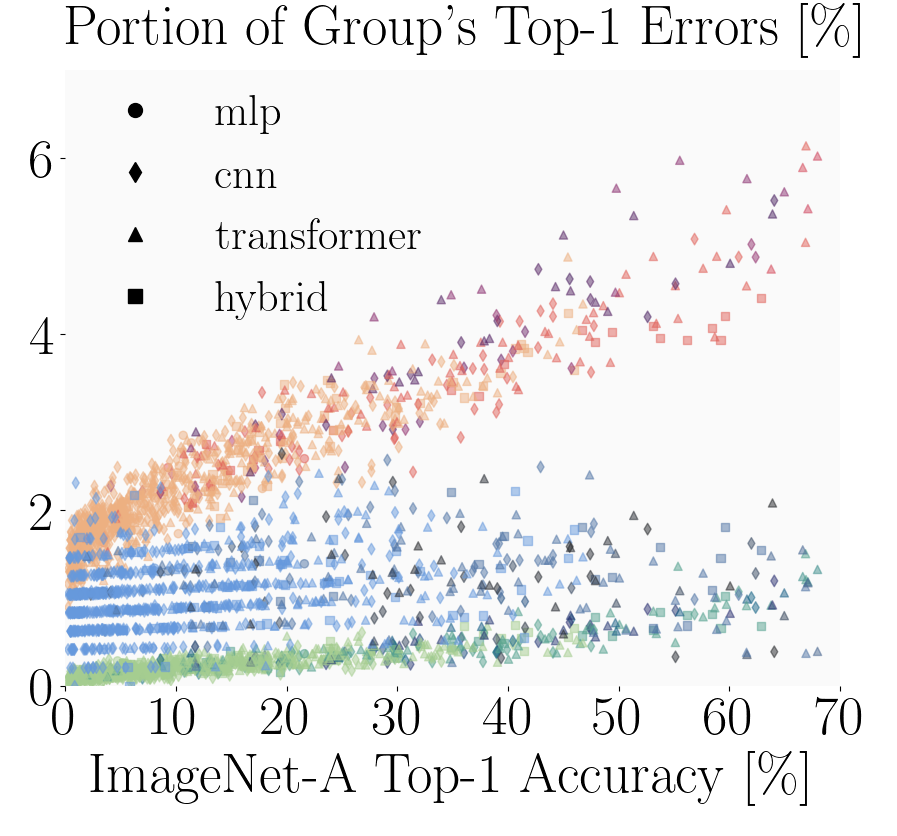}
    \end{minipage}
    \hfil
    \begin{minipage}[t]{0.49 \linewidth}
        \centering
        \includegraphics[width=0.9\linewidth]{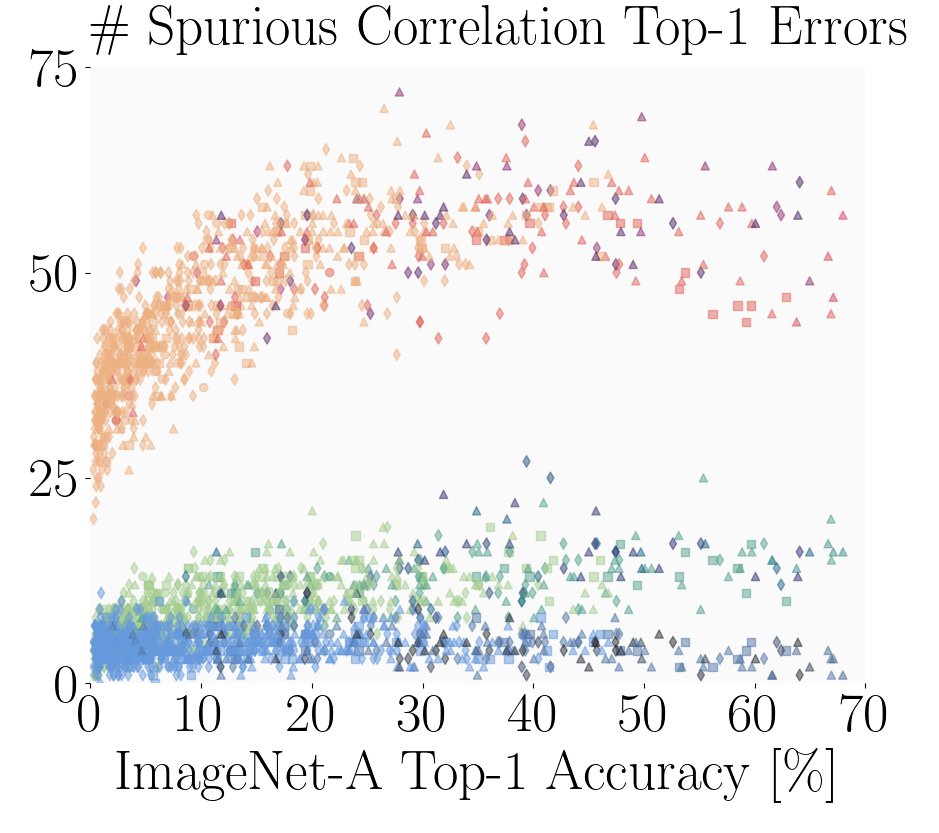}
    \end{minipage}
    \vspace{-3mm}
    \caption{Portion (left) and number (right) of \toa errors identified as \emph{spurious correlations} by group -- organisms (green), artifacts (red) and other (blue).}
    \label{fig:imagenet_a:spurious_correlations}
\end{wrapfigure}
Again following our \IN approach for spurious correlation errors we obtain the results in \cref{fig:imagenet_a:spurious_correlations}.
At first glance, we observe much less spurious correlations than for \IN ($6.2\%$ compared to $15\%$ for artifacts). However, we observe a more linear instead of super-linear increase in the portion of errors explained by spurious correlations, leading to this reduction being smaller for less accurate models.

\paragraph{Model Failures}
\begin{wrapfigure}[11]{r}{0.6\textwidth}
    \centering
    \vspace{-4mm}
    \begin{minipage}[t]{0.49 \linewidth}
        \centering
        \includegraphics[width=0.9\linewidth]{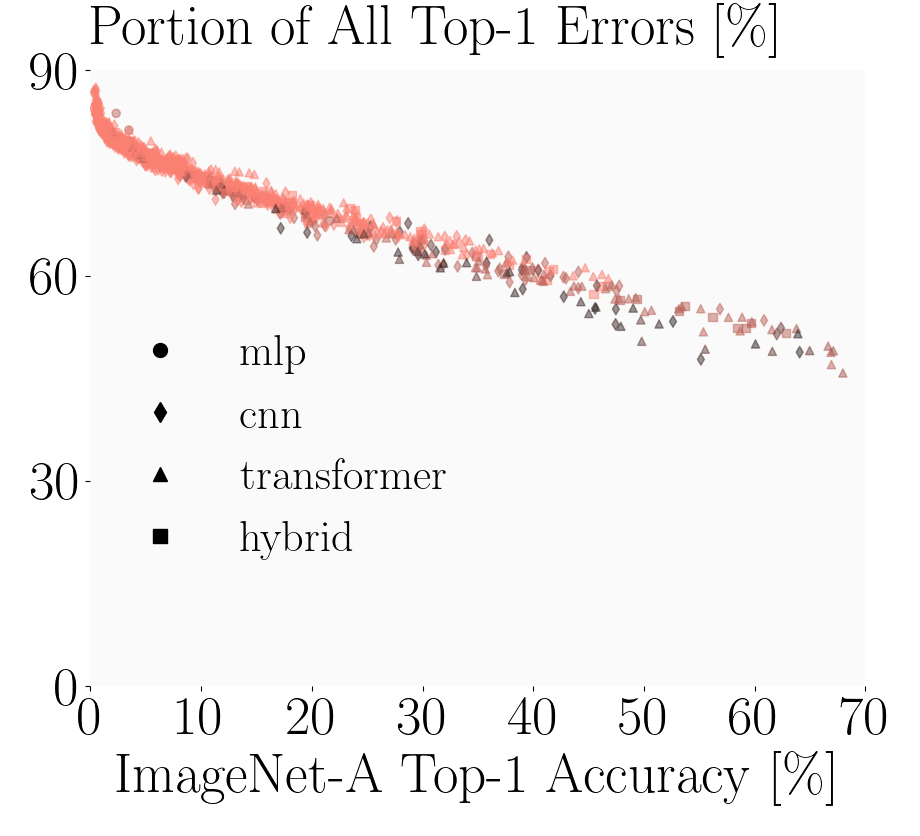}
    \end{minipage}
    \hfil
    \begin{minipage}[t]{0.49 \linewidth}
        \centering
        \includegraphics[width=0.9\linewidth]{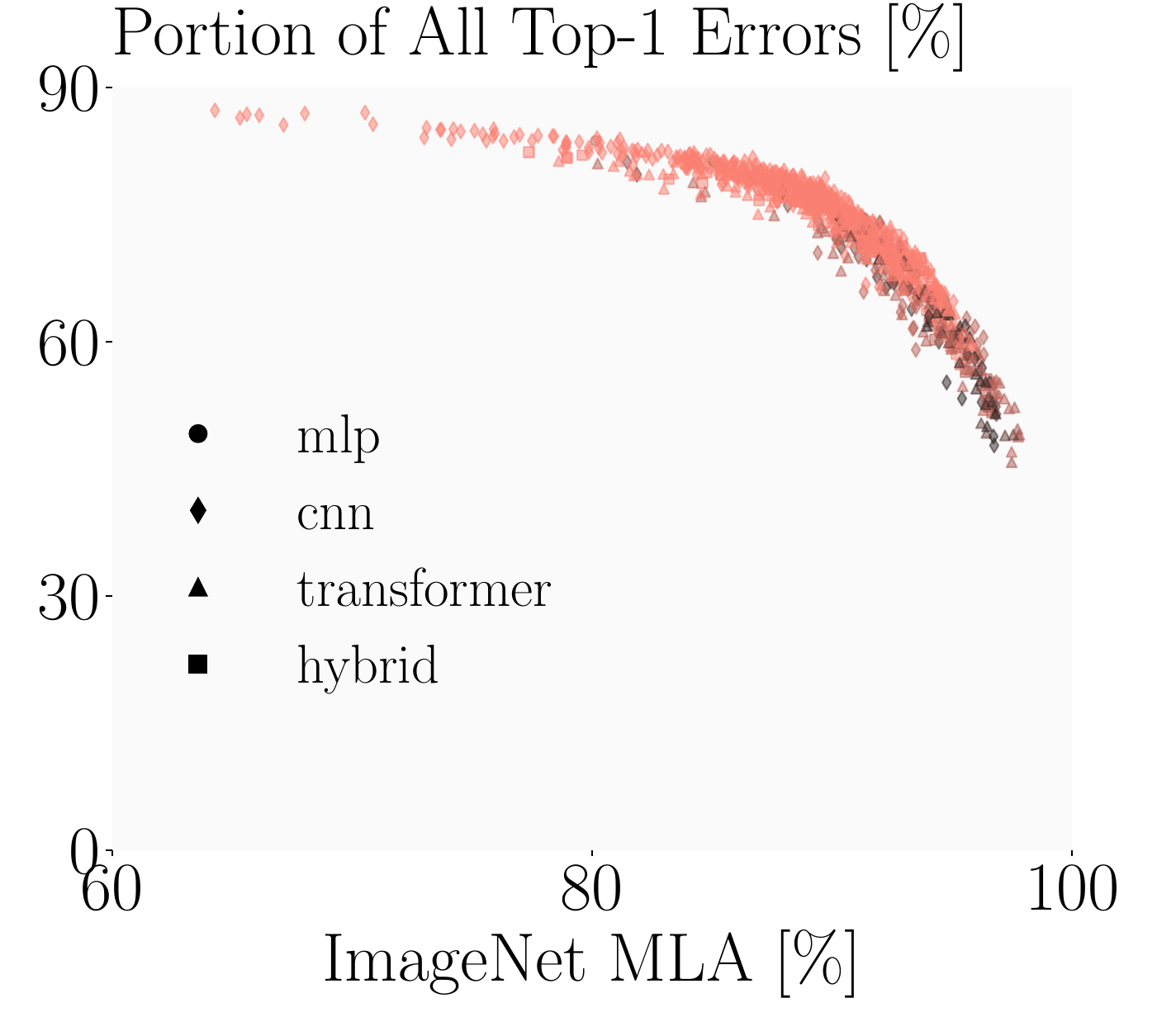}
    \end{minipage}
    \vspace{-3mm}
    \caption{Portion for remaining failures over \INA \toa (left) and \IN \mla (right).}
    \label{fig:imagenet_a:failures}
\end{wrapfigure}
Looking at the remaining failures, we see that we can only explain up to half of the errors for the best-performing models. Not only are \toa and \mla error rates much higher than for \IN but also a much larger portion of the remaining errors constitutes severe model failures, highlighting that \INA is a much harder dataset.

However, analyzing \cref{fig:imagenet_a:failures}, we observe that both the \INA \toa and \IN \mla accuracy are good predictors for the portion of model failures which decreases with increasing model accuracy. This highlights again and perhaps surprisingly that \IN \mla underreports progress in model performance, even on particularly challenging subdistributions like \INA.

\newpage
\section{Extended Examples on Fine-Grained OOV Error Classification}
\label{app:fg-OOV-pipeline}

In this section we provide further step by step examples of the procedure
that determines if a given model mistake is categorized as a fine-grained OOV error.
The sample in~\cref{app:fig-fg-OOV-pipeline1} is confirmed to be fine-grained OOV,
the one in~\cref{app:fig-fg-OOV-pipeline2} is not fine-grained
and the one in~\cref{app:fig-fg-OOV-pipeline3} is rejected as OOV.

\begin{figure}[H]
    \captionsetup[subfloat]{labelformat=empty}
	\centering
    \subfloat[GT: \inc{suit} \\Pred: \inc{bicycle-built-for-two} \\ Pred superclass: motor\_cycle]{
		\includegraphics[width=0.25\linewidth]{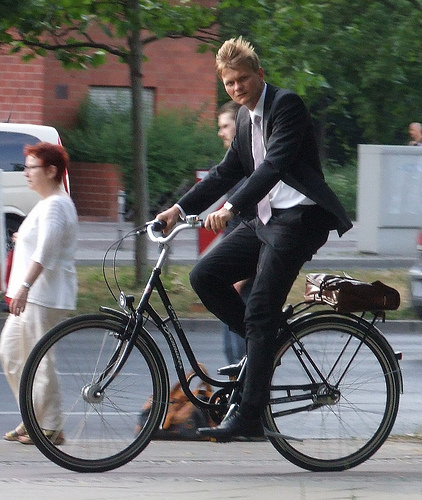}
	}\hfil

	\subfloat[(1) GT: \inc{suit} \\ suits]{
		\includegraphics[width=0.23\linewidth]{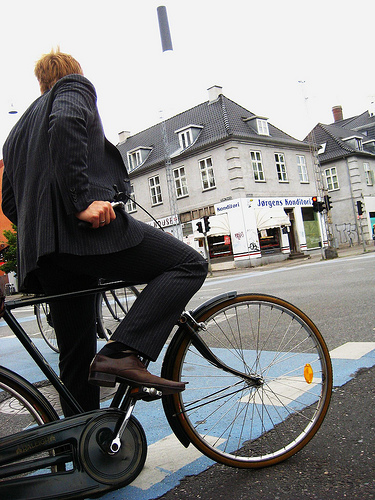}
	}\hfil
    \subfloat[(2) GT: \inc{suit} \\ suits]{
		\includegraphics[width=0.23\linewidth]{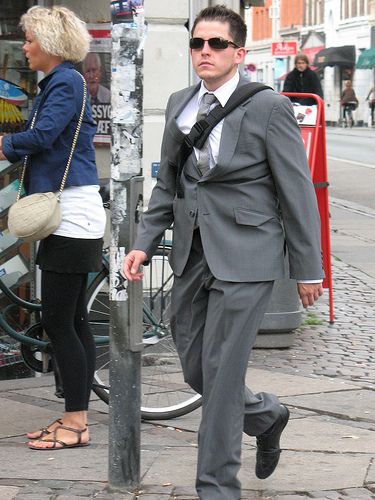}
	}\hfil
    \subfloat[(3) GT: \inc{unicycle} \\ motor\_cycle \textcolor{my-dark-green}{\ding{51}}]{
		\includegraphics[width=0.23\linewidth]{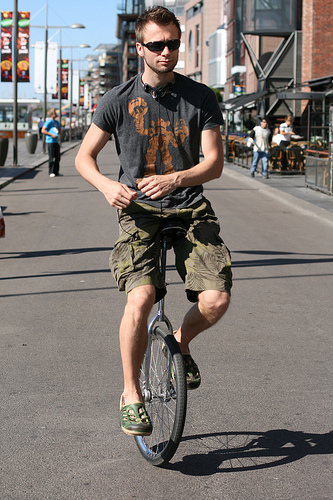}
	}\hfil
    \subfloat[(4) GT: \inc{backpack} \\ bag]{
		\includegraphics[width=0.23\linewidth]{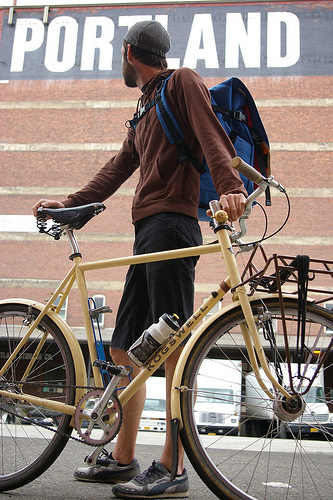}
	}\hfil
	\subfloat[(5) GT: \inc{suit} \\ suits]{
		\includegraphics[width=0.23\linewidth]{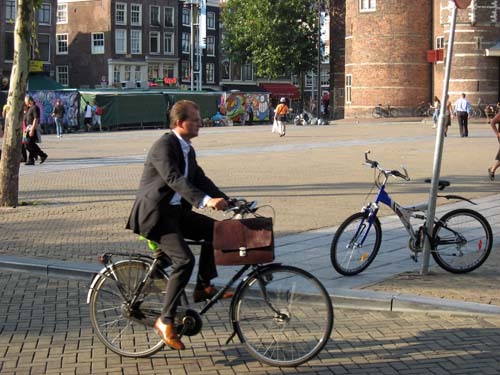}
	}\hfil
	\subfloat[(6) GT: \inc{moped} \\ motor\_cycle \textcolor{my-dark-green}{\ding{51}}]{
		\includegraphics[width=0.23\linewidth]{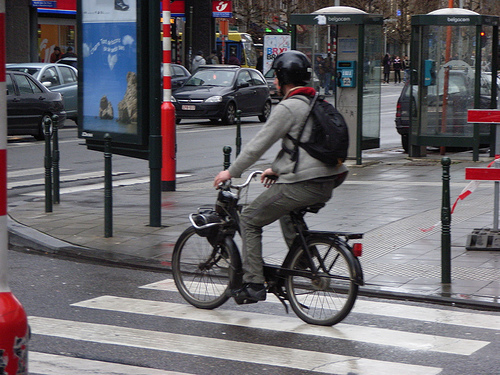}
	}\hfil
    \subfloat[(7) GT: \inc{traffic light} \\ No superclass]{
		\includegraphics[width=0.23\linewidth]{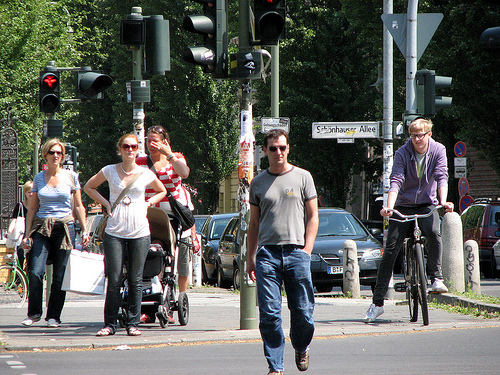}
	}\hfil
	\subfloat[(8) GT: \inc{plastic bag} \\ No superclass]{
		\includegraphics[width=0.23\linewidth]{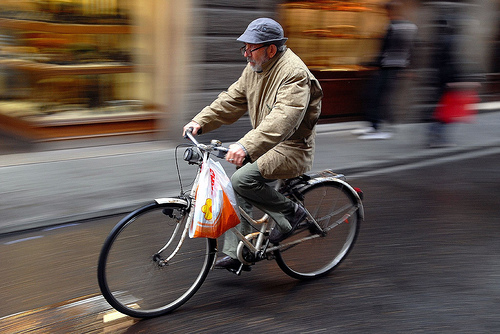}
	}\hfil
	\subfloat[(9) GT: \inc{bicycle-built-for-two} \\ motor\_cycle \textcolor{my-dark-green}{\ding{51}}]{
		\includegraphics[width=0.23\linewidth]{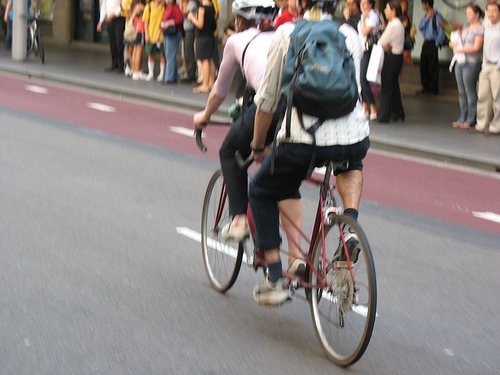}
	}\hfil
	\subfloat[(10) GT: \inc{bicycle-built-for-two} \\ motor\_cycle \textcolor{my-dark-green}{\ding{51}}]{
		\includegraphics[width=0.23\linewidth]{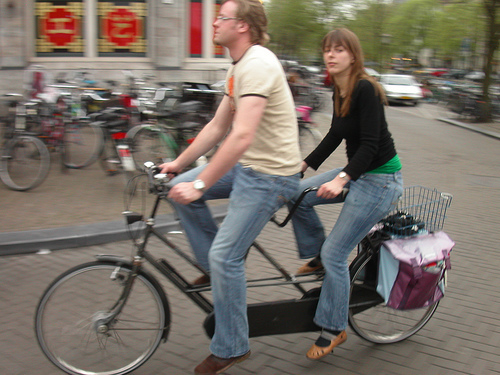}
	}\hfil
    \caption{
		\textbf{Fine-grained OOV confirmed}.
        The evaluated validation sample (top) has a ground-truth \IN label (GT)
        \inc{suit}, but the model predicts \inc{bicycle-built-for-two} which falls in
        the ``motor\_cycle'' superclass. Below, we show the top-10 training images that
        are the most visually similar to the validation sample (according to CLIP
        similarity) together with their \IN labels and the superclasses
        of these labels. The labels of images 3, 6, 9 and 10 are in the same
        superclass as the predicted label (motor\_cycle), hence we proceed with the
        procedure. After obtaining the in-vocabulary and OOV label proposals,
        as described in Sec.~\ref{sec:oov}, the proposals with
        the highest probability according to CLIP are: \inc{safety bicycle},
        \inc{ordinary bicycle}, \inc{velocipede} and \inc{bicycle}, all of which are
        OOV, confirming that the error is indeed fine-grained OOV.
	}
	\label{app:fig-fg-OOV-pipeline1}
\end{figure}

\newpage

\begin{figure}[H]
    \captionsetup[subfloat]{labelformat=empty}
	\centering
    \subfloat[GT: \inc{airship}, \inc{analog clock} \\Pred: \inc{jinrikisha} \\ Pred superclass: open\_cart]{
		\includegraphics[width=0.25\linewidth]{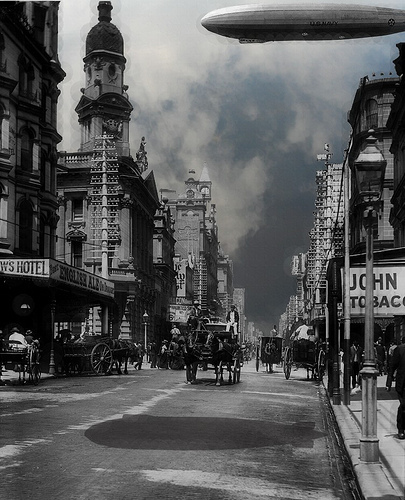}
	}\hfil

    \subfloat[GT: \inc{wall clock} \\ clock]{
		\includegraphics[width=0.23\linewidth]{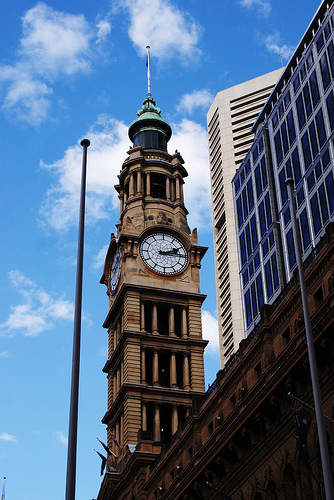}
	}\hfil
    \subfloat[GT: \inc{bell cote} \\ religious\_castle\_palace]{
		\includegraphics[width=0.23\linewidth]{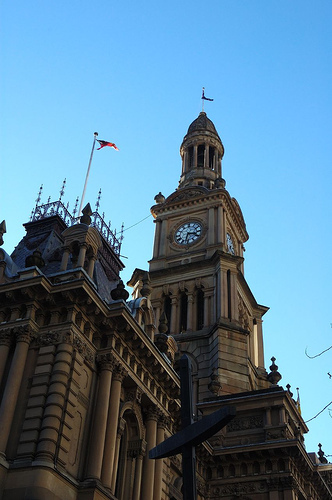}
	}\hfil
    \subfloat[GT: \inc{cab} \\ car\_bus\_truck]{
		\includegraphics[width=0.23\linewidth]{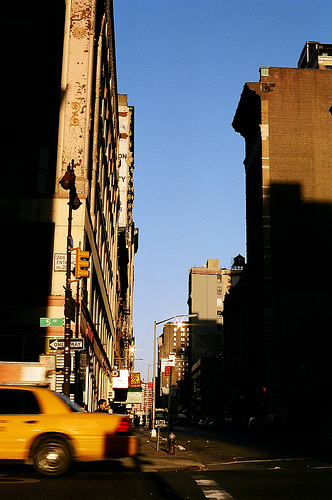}
	}\hfil

    \subfloat[GT: \inc{tram} \\ train]{
		\includegraphics[width=0.23\linewidth]{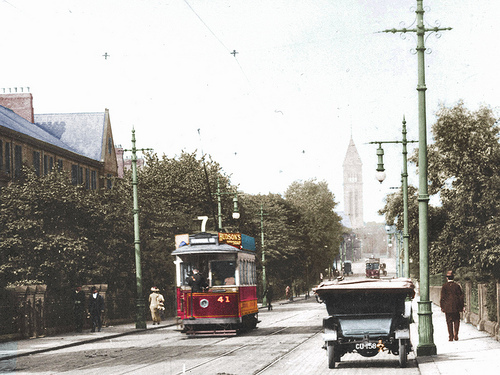}
	}\hfil
    \subfloat[GT: \inc{tram} \\ train]{
		\includegraphics[width=0.23\linewidth]{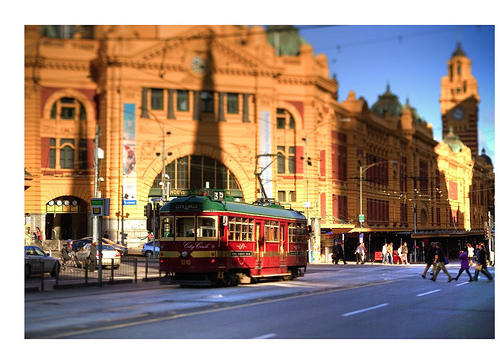}
	}\hfil
	\subfloat[GT: \inc{tram} \\ train]{
		\includegraphics[width=0.23\linewidth]{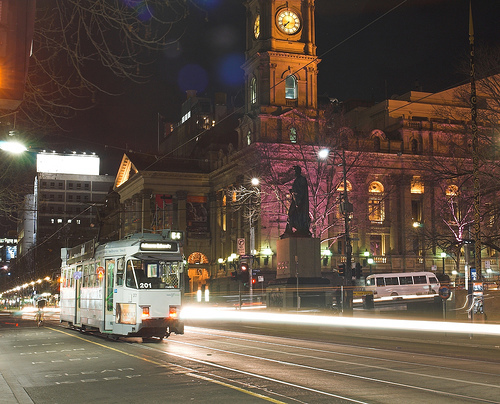}
	}\hfil
	\subfloat[GT: \inc{tram} \\ train]{
		\includegraphics[width=0.23\linewidth]{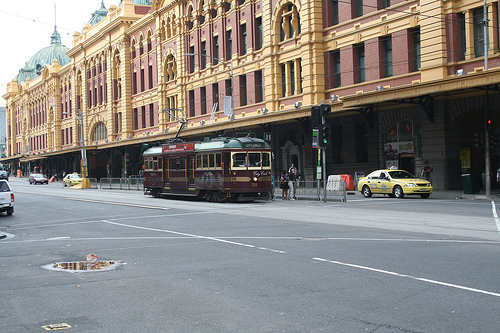}
	}\hfil
    \subfloat[GT: \inc{tram} \\ train]{
		\includegraphics[width=0.23\linewidth]{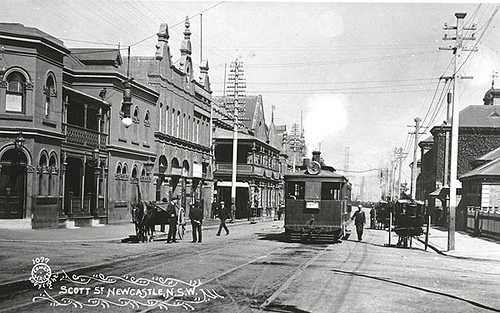}
	}\hfil
	\subfloat[GT: \inc{umbrella} \\ No superclass]{
		\includegraphics[width=0.23\linewidth]{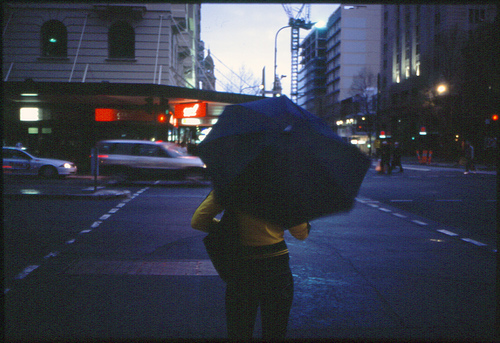}
	}\hfil
	\subfloat[GT: \inc{tram} \\ train]{
		\includegraphics[width=0.23\linewidth]{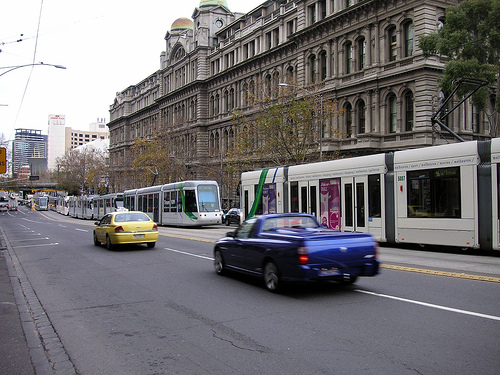}
	}\hfil
    \caption{
		\textbf{Fine-grained OOV rejected}.
        The evaluated validation sample (top) has ground-truth \IN multi-labels (GT)
        \inc{airship} and \inc{analog clock}, but the model predicts \inc{jinrikisha},
        arguably attempting to classify the carriages on the street. The predicted
        label is in the ``open\_cart'' superclass. Next, we show the top-10 training
        images that are the most visually similar to the validation sample
        (according to CLIP similarity) together with their \IN labels and the
        superclasses of these labels.
        None of the labels is in the ``open\_cart'' superclass, thus we terminate
        the procedure and reject the mistake as a possible fine-grained OOV error.
	}
	\label{app:fig-fg-OOV-pipeline2}
\end{figure}

\newpage

\begin{figure}[H]
    \captionsetup[subfloat]{labelformat=empty}
	\centering
    \subfloat[GT: \inc{barrow} \\Pred: \inc{shopping basket} \\ Pred superclass: box]{
		\includegraphics[width=0.25\linewidth]{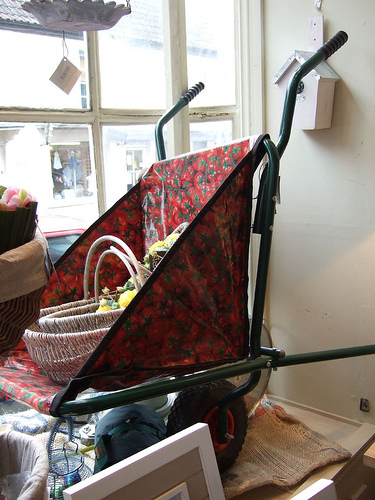}
	}\hfil

    \subfloat[(1) GT: \inc{barrow} \\ open\_cart]{
		\includegraphics[width=0.23\linewidth]{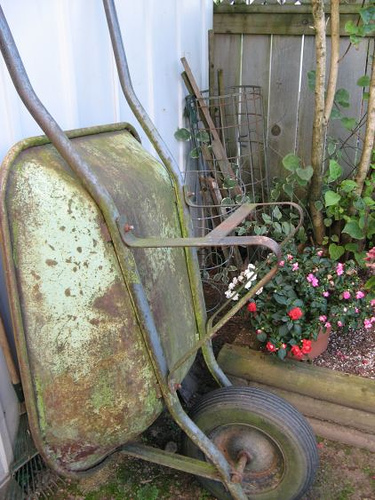}
	}\hfil
    \subfloat[(2) GT: \inc{lawn mower} \\ work\_cart]{
		\includegraphics[width=0.23\linewidth]{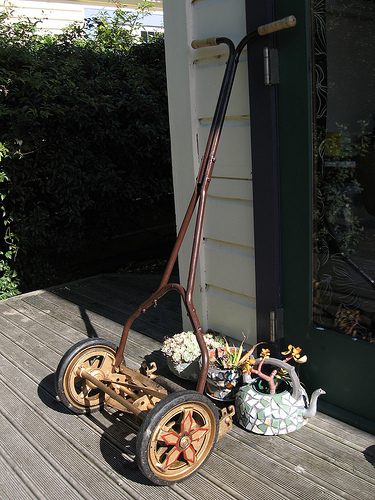}
	}\hfil
    \subfloat[(3) GT: \inc{shopping basket} \\ box \textcolor{my-dark-green}{\ding{51}}]{
		\includegraphics[width=0.23\linewidth]{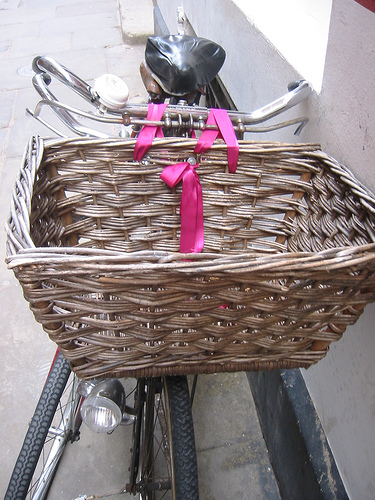}
	}\hfil

    \subfloat[(4) GT: \inc{shopping basket} \\ box \textcolor{my-dark-green}{\ding{51}}]{
		\includegraphics[width=0.23\linewidth]{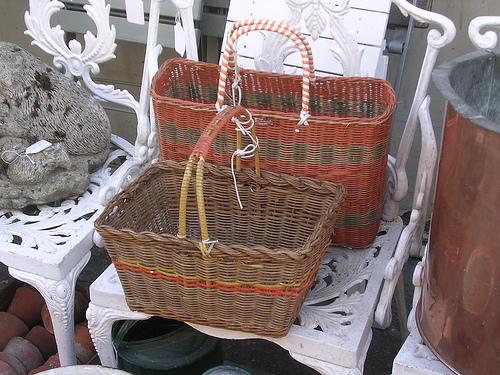}
	}\hfil
    \subfloat[(5) GT: \inc{barrow} \\ open\_cart]{
		\includegraphics[width=0.23\linewidth]{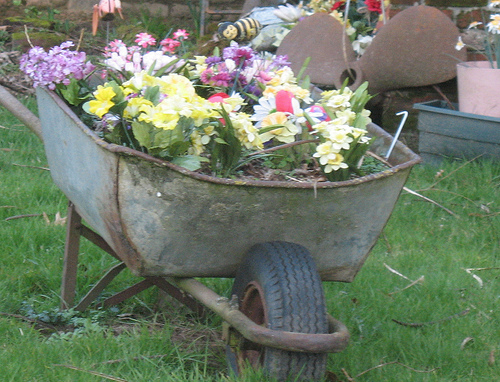}
	}\hfil
    \subfloat[(6) GT: \inc{tricycle} \\ motor\_cycle]{
		\includegraphics[width=0.23\linewidth]{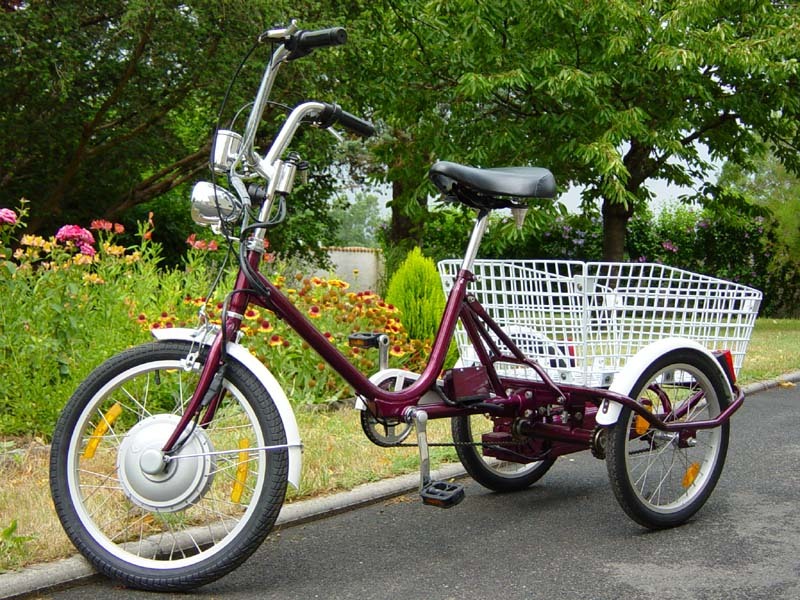}
	}\hfil
	\subfloat[(7) GT: \inc{barrow} \\ open\_cart]{
		\includegraphics[width=0.23\linewidth]{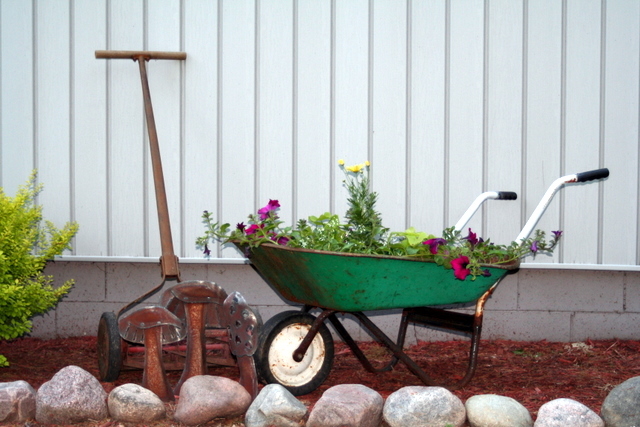}
	}\hfil
    \subfloat[(8) GT: \inc{bassinet} \\ baby\_bed]{
		\includegraphics[width=0.23\linewidth]{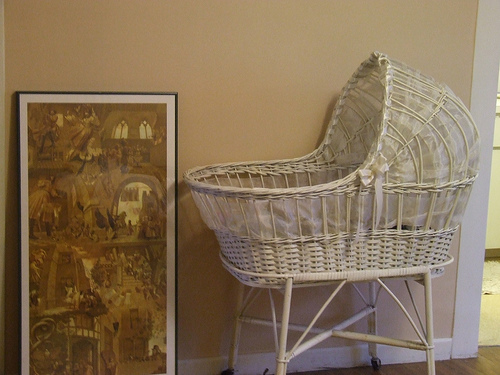}
	}\hfil
	\subfloat[(9) GT: \inc{tricycle} \\ motor\_cycle]{
		\includegraphics[width=0.23\linewidth]{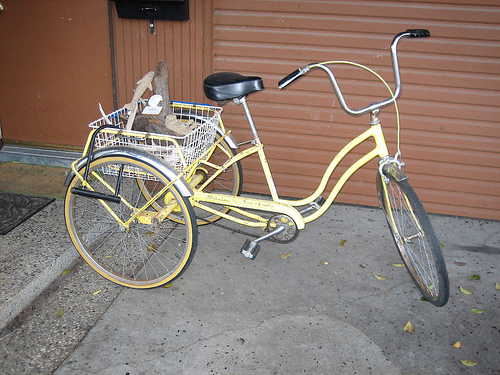}
	}\hfil
	\subfloat[(10) GT: \inc{cradle} \\ baby\_bed]{
		\includegraphics[width=0.23\linewidth]{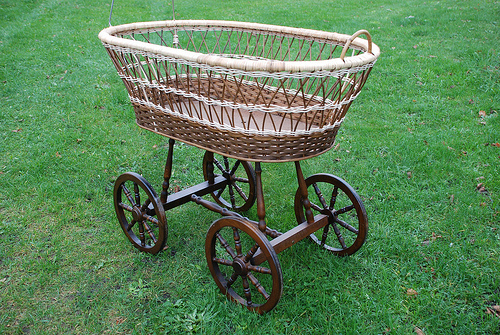}
	}\hfil
    \caption{
		\textbf{Fine-grained OOV rejected}.
        The evaluated validation sample (top) has a ground-truth \IN label (GT)
        \inc{barrow}, but the model predicts \inc{shopping basket},
        arguably attempting to classify the basket in the cart. The predicted
        label is in the ``box'' superclass.
        Next, we show the top-10 training images that
        are the most visually similar to the validation sample (according to CLIP
        similarity) together with their \IN labels and the superclasses
        of these labels. The labels of images 3 and 4 are in the same
        superclass as the predicted label (box), hence we proceed with the
        procedure. After obtaining the in-vocabulary (IV) and OOV label proposals,
        as described in Sec.~\ref{sec:oov}, the proposal with
        the highest probability according to CLIP is \inc{shopping cart}, which is
        in-vocabulary, therefore, the mistake is not categorized as a
        fine-grained OOV error.
        The other most likely proposals are: \inc{shopping basket} (IV),
        \inc{bushel basket} (OOV), \inc{basket} (OOV) and \inc{handbasket} (OOV).
	}
	\label{app:fig-fg-OOV-pipeline3}
\end{figure}

\newpage
\section{Mistake Examples by Error Type}
\label{app:error-examples}
\begin{figure}[H]
    \captionsetup[subfloat]{labelformat=empty}
	\centering
	\subfloat[GT: \inc{weasel}, \inc{polecat}, \inc{mink},\\ \inc{black-footed ferret}\\Pred: \inc{polecat}]{
		\includegraphics[width=0.23\linewidth]{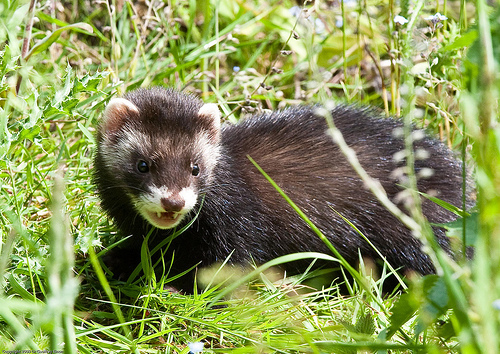}
	}\hfil
	\subfloat[GT: \inc{screen}, \inc{monitor}\\Pred: \inc{monitor}]{
		\includegraphics[width=0.23\linewidth]{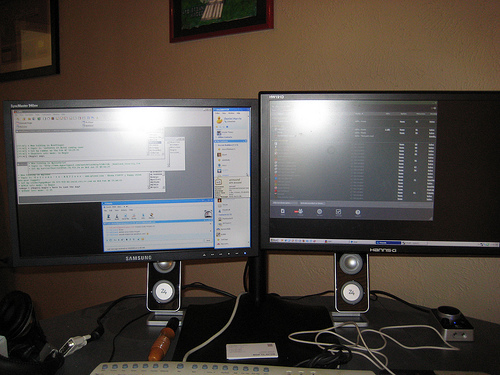}
	}\hfil
	\subfloat[GT: \inc{laptop}, \inc{notebook}\\Pred: \inc{notebook}]{
		\includegraphics[width=0.23\linewidth]{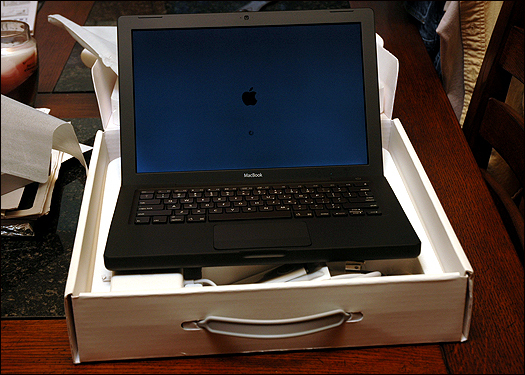}
	}\hfil
	\subfloat[GT: \inc{projectile}, \inc{missile}\\Pred: \inc{missile}]{
		\includegraphics[width=0.23\linewidth]{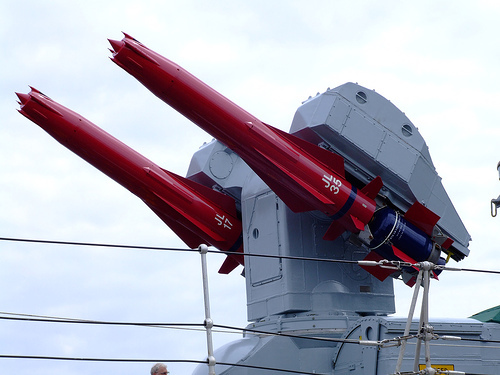}
	}\hfil
    \subfloat[GT: \inc{Indian elephant}, \inc{tusker}\\Pred: \inc{tusker}]{
		\includegraphics[width=0.23\linewidth]{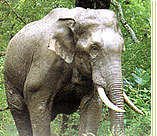}
	}\hfil
	\subfloat[GT: \inc{coffee mug}, \inc{cup}\hspace*{3em}\\Pred: \inc{cup}]{
		\includegraphics[width=0.23\linewidth]{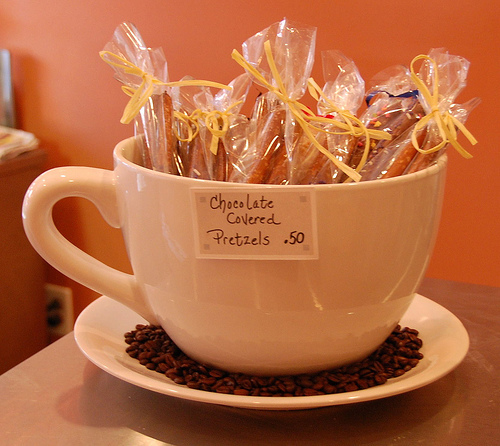}
	}\hfil
	\subfloat[GT: \inc{cassette player},\\ \inc{tape player}\\Pred: \inc{tape player}]{
		\includegraphics[width=0.23\linewidth]{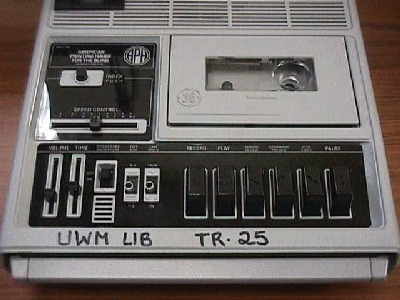}
	}\hfil
	\subfloat[GT: \inc{Siberian husky}, \inc{Eskimo dog}\\Pred: \inc{Eskimo dog}]{
		\includegraphics[width=0.23\linewidth]{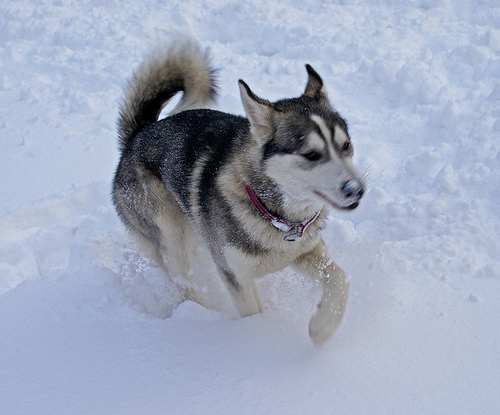}
	}\hfil
	\subfloat[GT: \inc{maillot}, \inc{tank suit}\\Pred: \inc{tank suit}]{
		\includegraphics[width=0.23\linewidth]{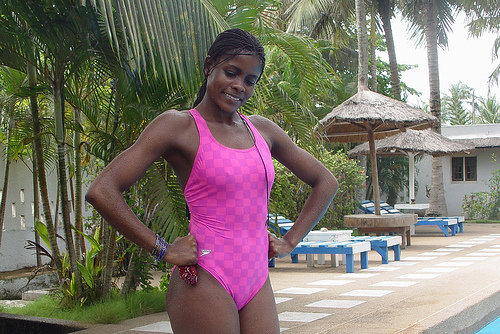}
	}\hfil
	\subfloat[GT: \inc{sunglass}, \inc{dark glasses}\\Pred: \inc{dark glasses}]{
		\includegraphics[width=0.23\linewidth]{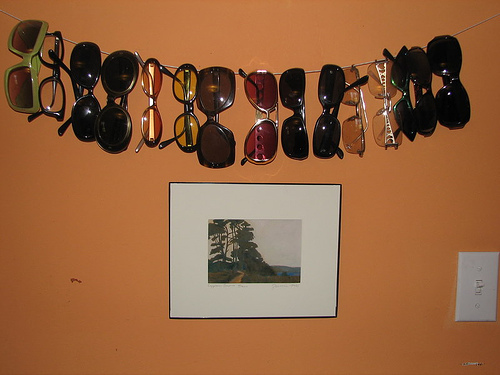}
	}\hfil
	\subfloat[GT: \inc{bathtub}, \inc{tub}, \inc{washbasin}\\Pred: \inc{tub}]{
		\includegraphics[width=0.23\linewidth]{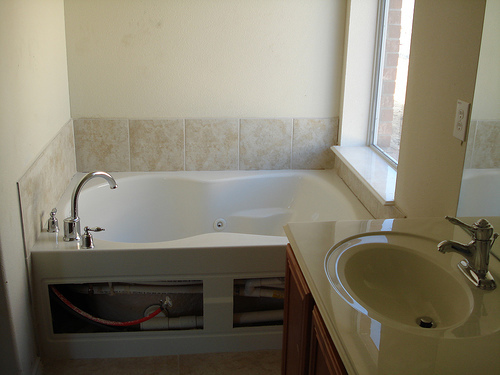}
	}
    \caption{
		\textbf{Class overlap}: examples of common prediction errors (Pred)
		due to overlapping classes.
		The correct (individual) multi-labels (GT) are separated by commas;
		the original \IN label is listed first.
	}
	\label{app:fig-class-overlap}
\end{figure}

\newpage

\begin{figure}[H]
    \captionsetup[subfloat]{labelformat=empty}
	\centering
	\subfloat[GT: \inc{monastery}, \inc{altar}\\Pred: \inc{altar}]{
		\includegraphics[width=0.23\linewidth]{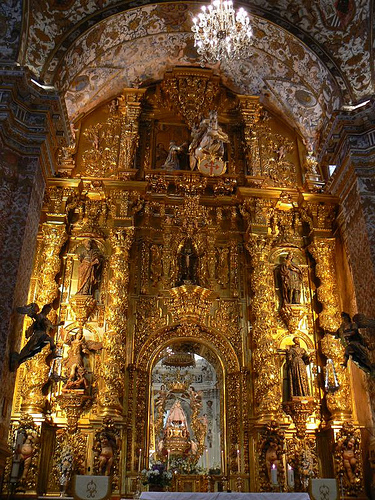}
	}\hfil
	\subfloat[GT: \inc{medicine chest}, \inc{washbasin}, \\ \inc{shower curtain} \\Pred: \inc{shower curtain}]{
		\includegraphics[width=0.23\linewidth]{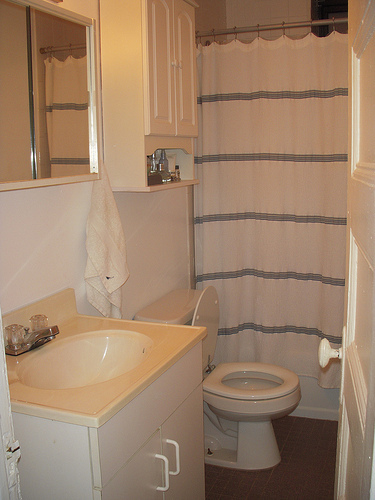}
	}\hfil
	\subfloat[GT: \inc{English foxhound}, \inc{beagle}\\Pred: \inc{beagle}]{
		\includegraphics[width=0.23\linewidth]{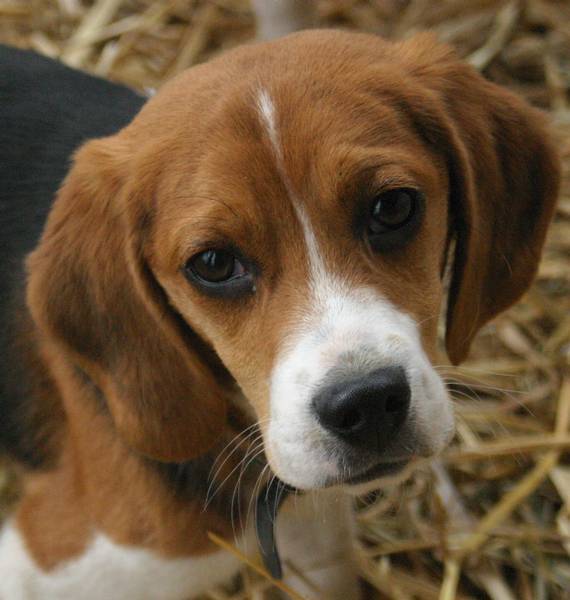}
	}\hfil
	\subfloat[GT: \inc{paper towel}, \inc{dock}\\Pred: \inc{dock}]{
		\includegraphics[width=0.23\linewidth]{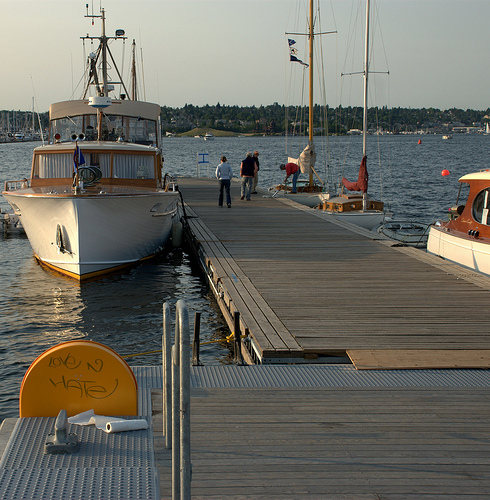}
	}\hfil
	\subfloat[GT: \inc{green snake}, \inc{green mamba}\\Pred: \inc{green mamba}]{
		\includegraphics[width=0.23\linewidth]{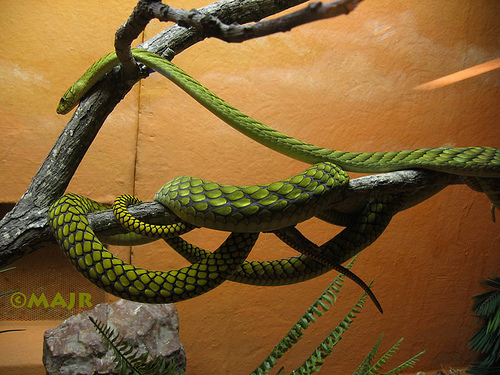}
	}\hfil
	\subfloat[GT: \inc{partridge}, \inc{ptarmigan}\\Pred: \inc{ptarmigan}]{
		\includegraphics[width=0.23\linewidth]{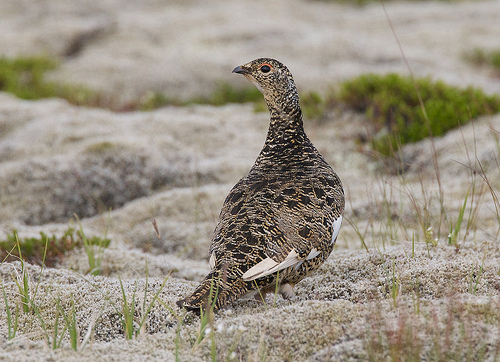}
	}\hfil
	\subfloat[GT: \inc{ground beetle},\\ \inc{tiger beetle}\\Pred: \inc{tiger beetle}]{
		\includegraphics[width=0.23\linewidth]{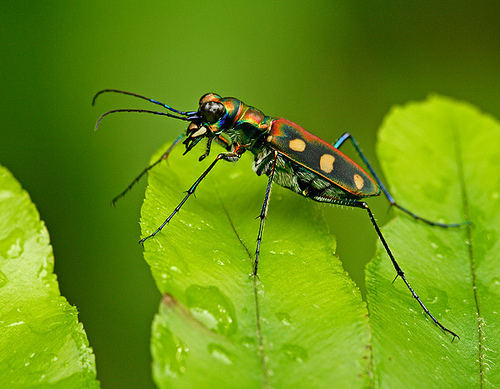}
	}\hfil
    \subfloat[GT: \inc{lumbermill},\\ \inc{chainlink fence}\\Pred: \inc{chainlink fence}]{
		\includegraphics[width=0.23\linewidth]{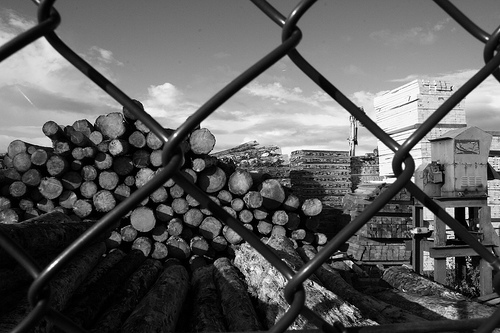}
	}\hfil
	\subfloat[GT: \inc{minibus}, \inc{school bus}\\Pred: \inc{school bus}]{
		\includegraphics[width=0.23\linewidth]{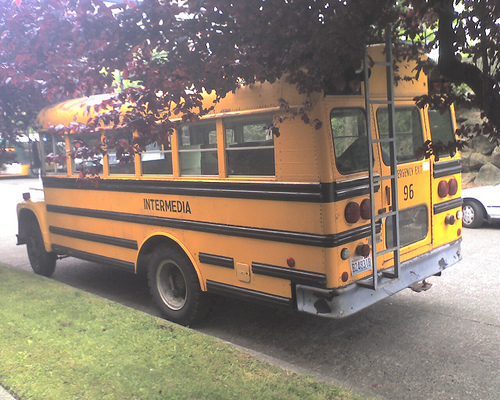}
	}\hfil
	\subfloat[GT: \inc{computer mouse}, \inc{mousetrap}\\Pred: \inc{mousetrap}]{
		\includegraphics[width=0.23\linewidth]{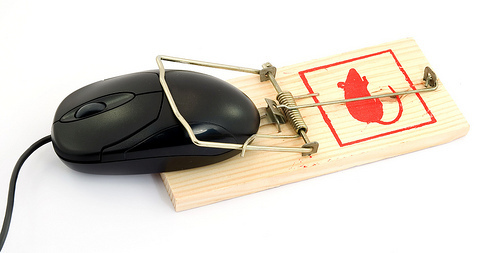}
	}\hfil
	\subfloat[GT: \inc{computer mouse},\\ \inc{desktop computer}\\Pred: \inc{desktop computer}]{
		\includegraphics[width=0.23\linewidth]{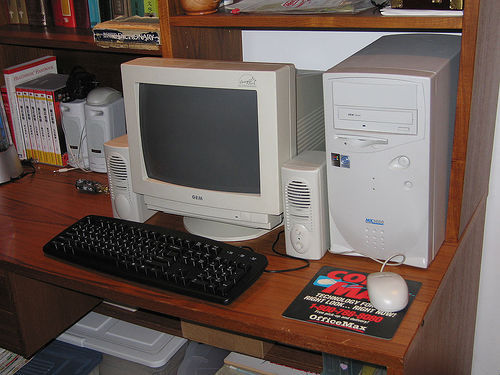}
	}\hfil
	\subfloat[GT: \inc{ice cream}, \inc{chocolate sauce}\\Pred: \inc{chocolate sauce}]{
		\includegraphics[width=0.23\linewidth]{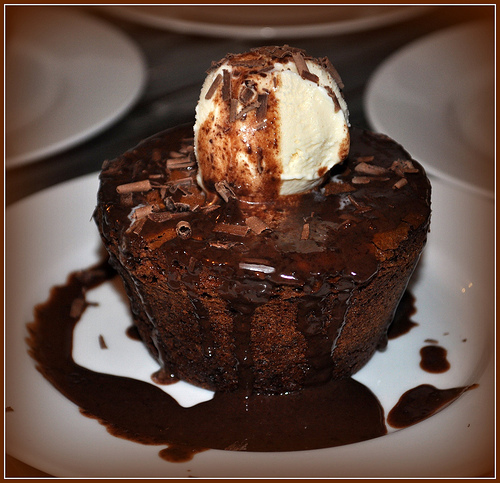}
	}\hfil
	\subfloat[GT: \inc{cucumber}, \inc{zucchini}\\Pred: \inc{zucchini}]{
		\includegraphics[width=0.23\linewidth]{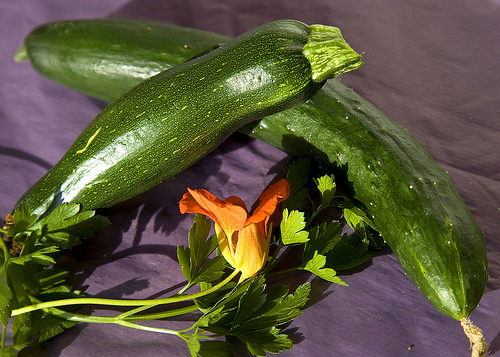}
	}\hfil
	\subfloat[GT: \inc{alp}, \inc{ibex} \hspace*{5em} \\Pred: \inc{ibex}]{
		\includegraphics[width=0.23\linewidth]{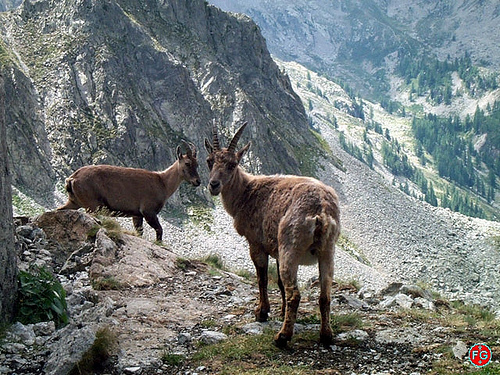}
	}\hfil
	\subfloat[GT: \inc{honeycomb}, \inc{pillow}, \inc{mailbag} \\Pred: \inc{pillow}]{
		\includegraphics[width=0.23\linewidth]{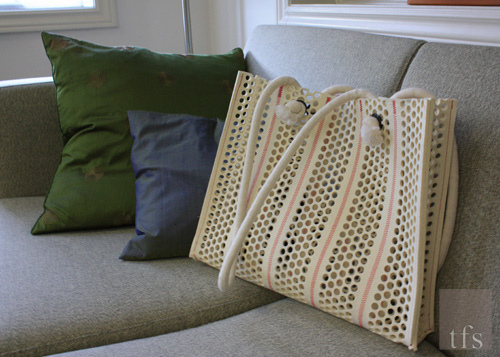}
	}\hfil
	\subfloat[GT: \inc{plate}, \inc{burrito} \\Pred: \inc{burrito}]{
		\includegraphics[width=0.23\linewidth]{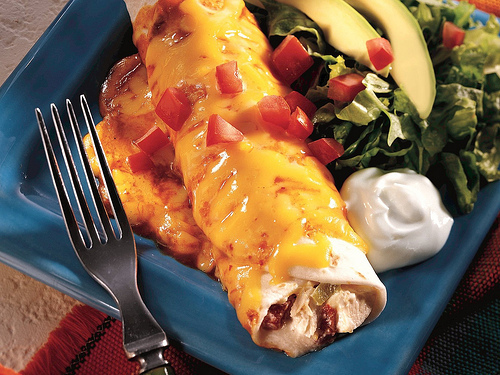}
	}\hfil
	\subfloat[GT: \inc{container ship}, \inc{dock}, \\ \inc{drilling platform} \\Pred: \inc{drilling platform}]{
		\includegraphics[width=0.23\linewidth]{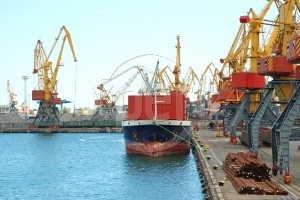}
	}\hfil
	\subfloat[GT: \inc{computer mouse}, \inc{monitor} \\ \inc{desk}, \inc{desktop computer}, \inc{screen} \\Pred: \inc{desktop computer}]{
		\includegraphics[width=0.23\linewidth]{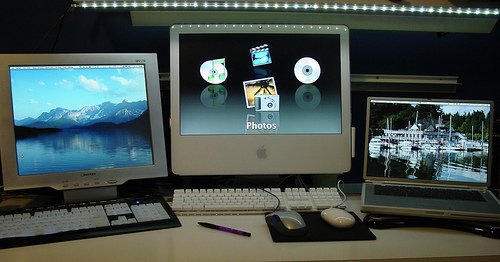}
	}\hfil
	\subfloat[GT: \inc{space bar}, \hspace*{4em} \\ \inc{computer keyboard} \\Pred: \inc{computer keyboard}]{
		\includegraphics[width=0.23\linewidth]{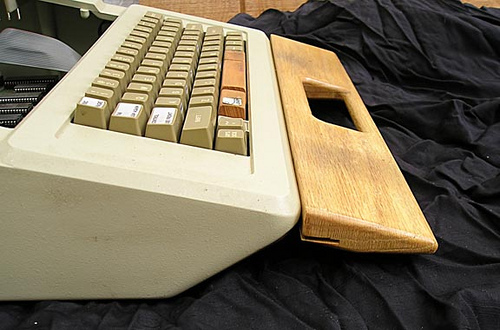}
	}\hfil
	\subfloat[GT: \inc{shoji}, \inc{patio}, \inc{tub} \\ \inc{sliding door} \\Pred: \inc{sliding door}]{
		\includegraphics[width=0.23\linewidth]{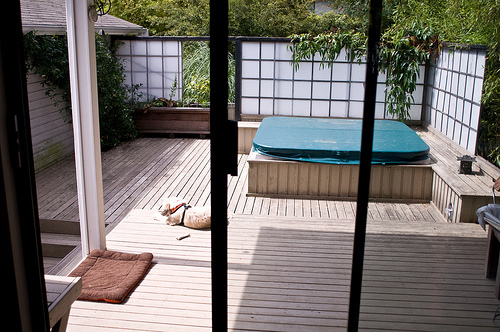}
	}
    \caption{
		\textbf{Multi-label}:  examples of common prediction errors (Pred)
		due to multi-label errors.
		The correct (individual) multi-labels (GT) are separated by commas;
		the original \IN label is listed first.
	}
	\label{app:fig-multi-label}
\end{figure}

\newpage

\begin{figure}[H]
    \captionsetup[subfloat]{labelformat=empty}
	\centering
	\subfloat[GT: \inc{china cabinet} \hspace*{3em}\\Pred: \inc{wardrobe}]{
		\includegraphics[width=0.23\linewidth]{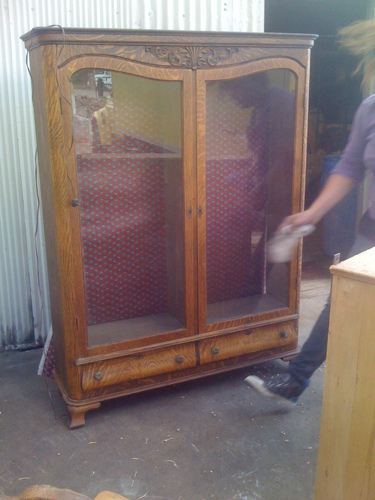}
	}\hfil
	\subfloat[GT: \inc{water jug} \hspace*{5em} \\Pred: \inc{pitcher}]{
		\includegraphics[width=0.23\linewidth]{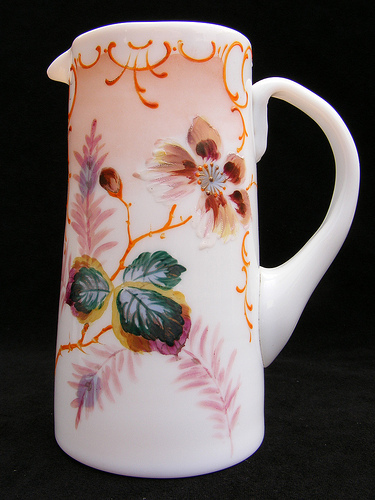}
	}\hfil
	\subfloat[GT: \inc{hare} \hspace*{5em} \\Pred: \inc{wood rabbit}]{
		\includegraphics[width=0.23\linewidth]{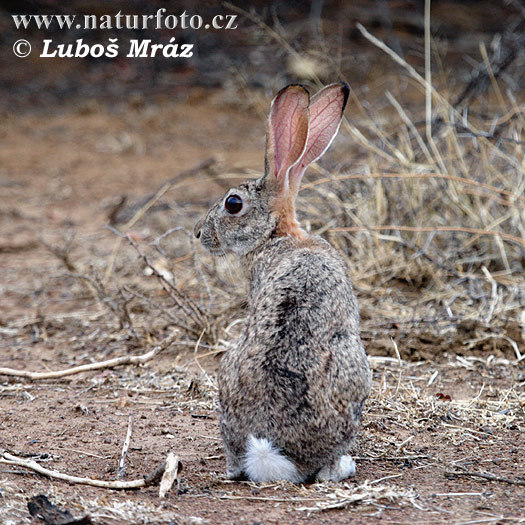}
	}\hfil
	\subfloat[GT: \inc{monastery}, \inc{stone wall}\\Pred: \inc{castle}]{
		\includegraphics[width=0.23\linewidth]{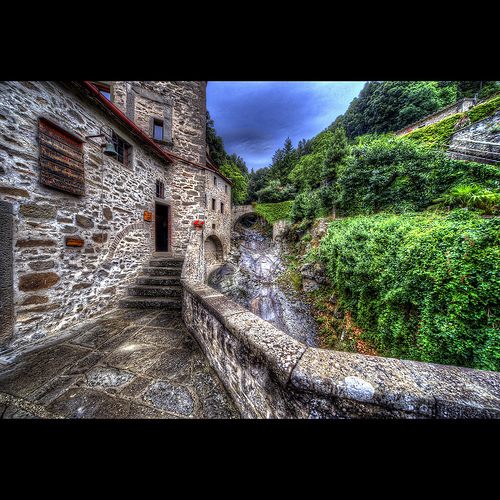}
	}\hfil
	\subfloat[GT: \inc{red wolf} \hspace*{4em}\\Pred: \inc{coyote}]{
		\includegraphics[width=0.23\linewidth]{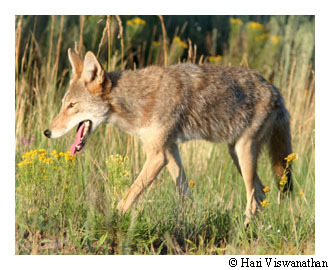}
	}\hfil
	\subfloat[GT: \inc{tabby cat} \hspace*{2em}\\Pred: \inc{Egyptian cat}]{
		\includegraphics[width=0.23\linewidth]{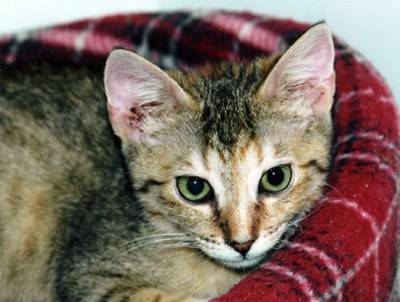}
	}\hfil
	\subfloat[GT: \inc{moving van} \hspace*{3em}  \\Pred: \inc{tow truck}]{
		\includegraphics[width=0.23\linewidth]{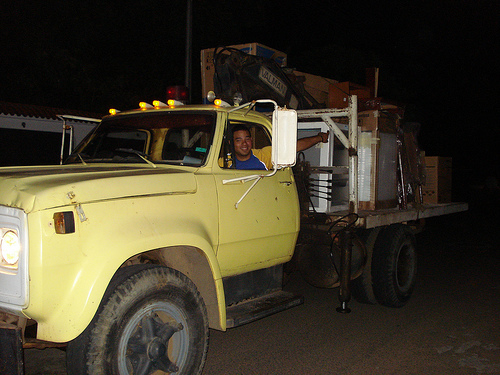}
	}\hfil
	\subfloat[GT: \inc{cauliflower} \hspace*{4em}\\Pred: \inc{broccoli}]{
		\includegraphics[width=0.23\linewidth]{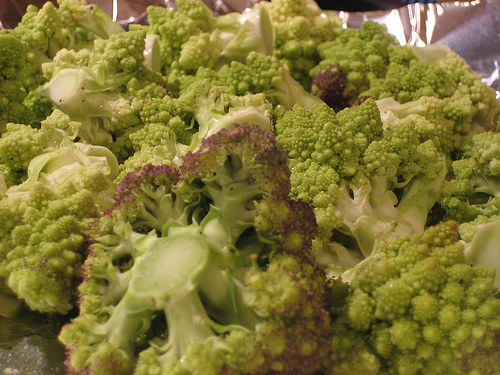}
	}\hfil
    \subfloat[GT: \inc{bloodhound} \hspace*{4em}\\Pred: \inc{redbone}]{
		\includegraphics[width=0.23\linewidth]{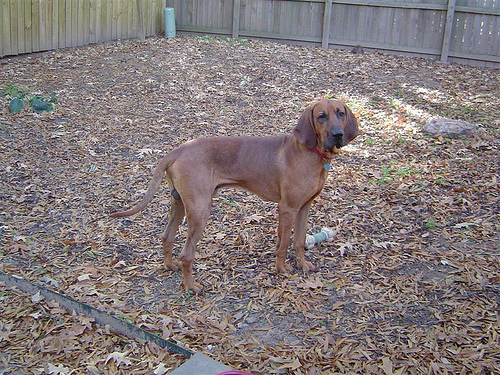}
	}\hfil
	\subfloat[GT: \inc{trimaran} \hspace*{4em}\\Pred: \inc{catamaran}]{
		\includegraphics[width=0.23\linewidth]{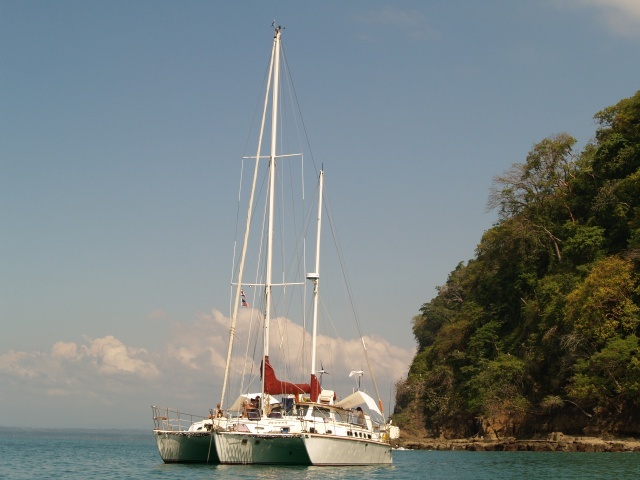}
	}\hfil
	\subfloat[GT: \inc{tobacco shop} \hspace{3em}\\Pred: \inc{barbershop}]{
		\includegraphics[width=0.23\linewidth]{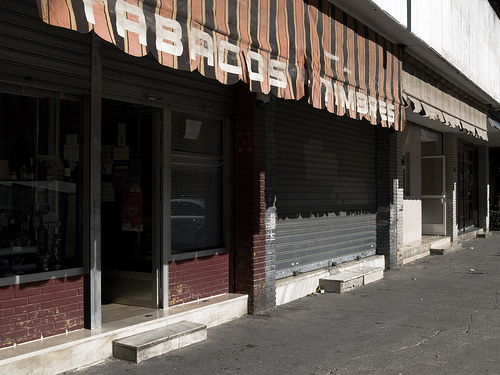}
	}\hfil
	\subfloat[GT: \inc{sloth bear} \hspace*{3em} \\Pred: \inc{American black bear}]{
		\includegraphics[width=0.23\linewidth]{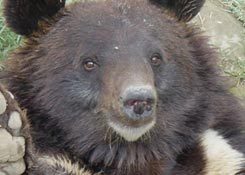}
	}\hfil
	\subfloat[GT: \inc{squirrel monkey} \hspace*{3em} \\Pred: \inc{titi monkey}]{
		\includegraphics[width=0.23\linewidth]{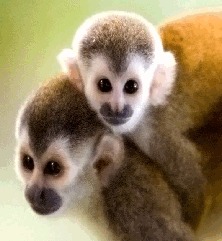}
	}\hfil
	\subfloat[GT: \inc{teapot} \hspace*{5em} \\Pred: \inc{coffeepot}]{
		\includegraphics[width=0.23\linewidth]{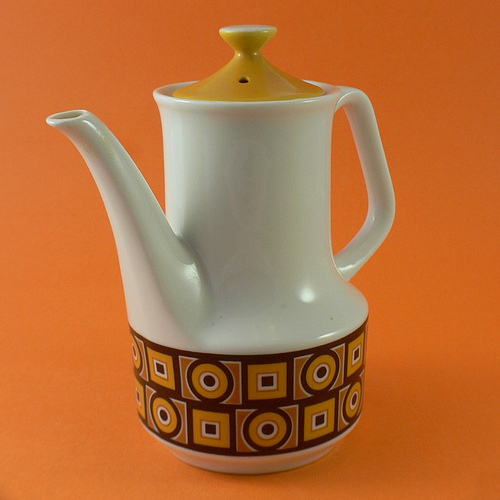}
	}\hfil
	\subfloat[GT: \inc{book jacket} \hspace*{3em} \\Pred: \inc{comic book}]{
		\includegraphics[width=0.23\linewidth]{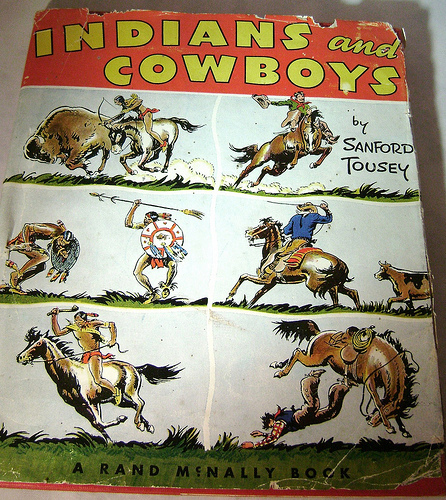}
	}\hfil
	\subfloat[GT: \inc{whippet} \hspace*{4em}\\Pred: \inc{Saluki}]{
		\includegraphics[width=0.23\linewidth]{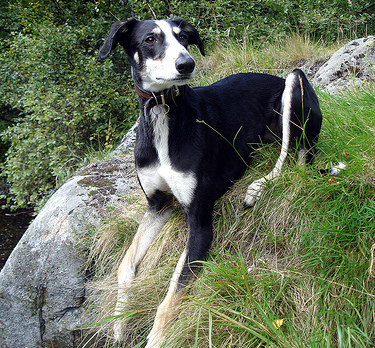}
	}\hfil
	\subfloat[GT: \inc{grey fox} \hspace*{4em} \\Pred: \inc{kit fox}]{
		\includegraphics[width=0.23\linewidth]{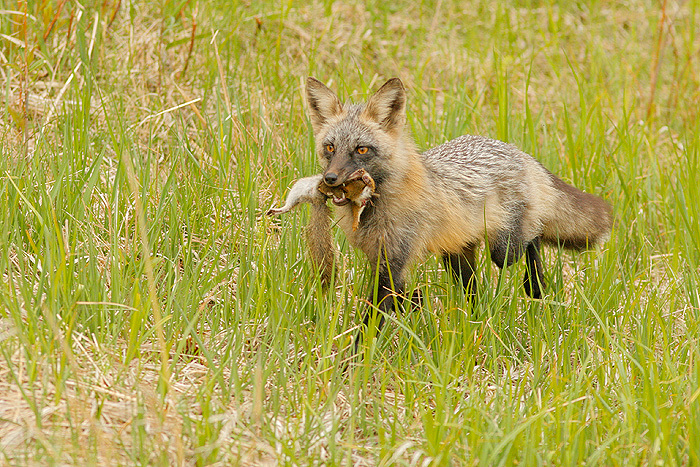}
	}\hfil
	\subfloat[GT: \inc{crate} \hspace*{4em} \\Pred: \inc{apiary}]{
		\includegraphics[width=0.23\linewidth]{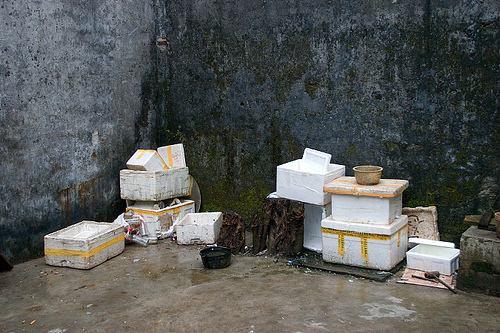}
	}\hfil
	\subfloat[GT: \inc{window screen}, \inc{cup} \hspace*{2em} \\Pred: \inc{eggnog}]{
		\includegraphics[width=0.23\linewidth]{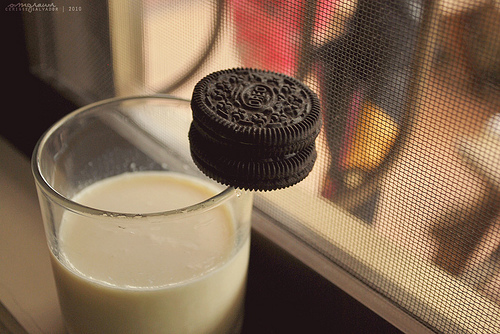}
	}\hfil
	\subfloat[GT: \inc{Norwich terrier} \hspace*{2em} \\Pred: \inc{cairn}]{
		\includegraphics[width=0.23\linewidth]{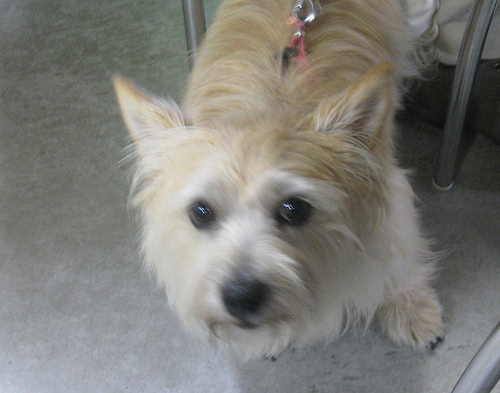}
	}
    \caption{
		\textbf{Fine-grained}: examples of common prediction errors (Pred)
		due to fine-grained errors.
		The correct (individual) multi-labels (GT) are separated by commas;
		the original \IN label is listed first.
	}
	\label{app:fig-fine-grained}
\end{figure}

\newpage

\begin{figure}[H]
    \captionsetup[subfloat]{labelformat=empty}
	\centering
	\subfloat[GT: \inc{sunglass}, \inc{dark glasses} \\Pred: \inc{miniature schnauzer} \\OOV: Other dog breed]{
		\includegraphics[width=0.23\linewidth]{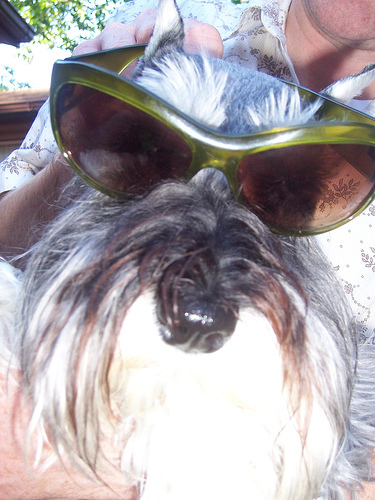}
	}\hfil
	\subfloat[GT: \inc{suit} \\Pred: \inc{bicycle-built-for-two}\\OOV: Single-person bicycle]{
		\includegraphics[width=0.23\linewidth]{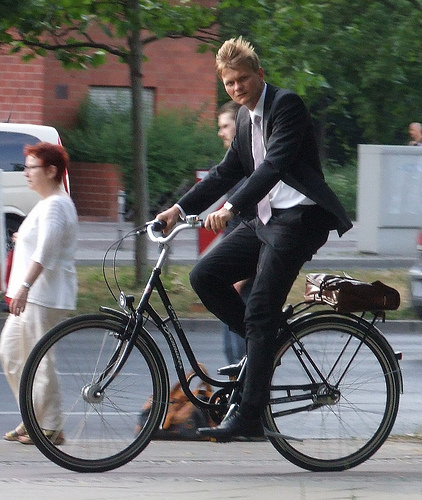}
	}\hfil
	\subfloat[GT: \inc{velvet} \\Pred: \inc{cardigan} \\ OOV: Garment]{
		\includegraphics[width=0.23\linewidth]{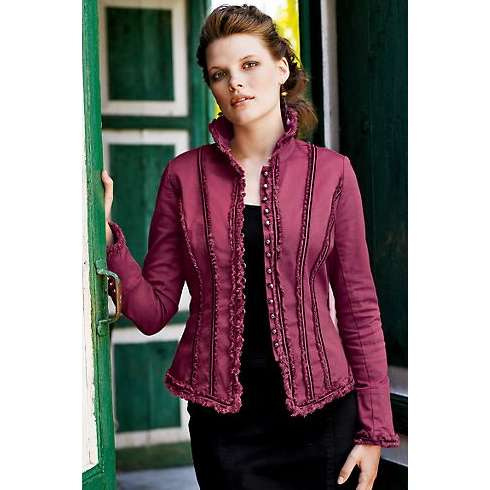}
	}\hfil
	\subfloat[GT: \inc{cup}, \inc{china cabinet} \\Pred: \inc{toyshop}\\OOV: Other shop or store]{
		\includegraphics[width=0.23\linewidth]{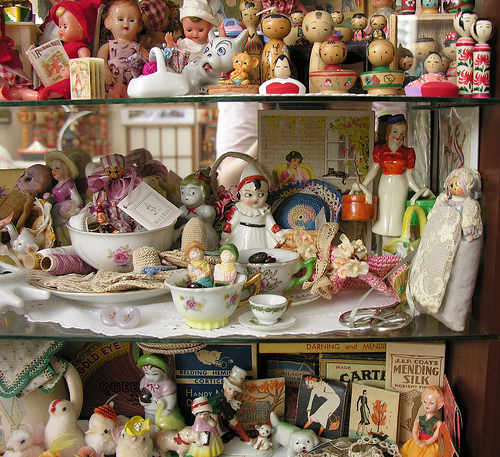}
	}\hfil
	\subfloat[GT: \inc{microwave} \\Pred: \inc{wardrobe}\\OOV: Cupboard or cabinet]{
		\includegraphics[width=0.23\linewidth]{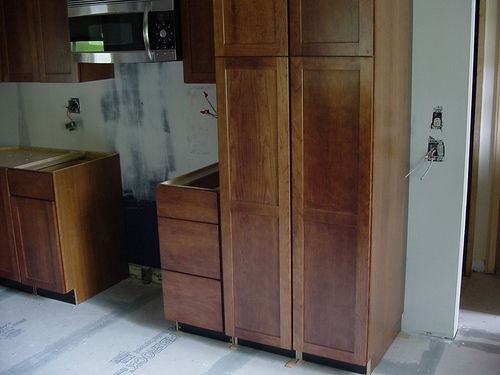}
	}\hfil
	\subfloat[GT: \inc{wardrobe} \\Pred: \inc{sliding door} \\ OOV: Wardrobe door]{
		\includegraphics[width=0.23\linewidth]{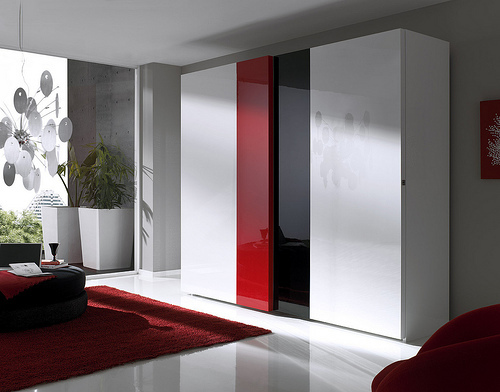}
	}\hfil
	\subfloat[GT: \inc{car wheel}, \inc{grille} \\Pred: \inc{beach wagon}\\OOV: Other car model]{
		\includegraphics[width=0.23\linewidth]{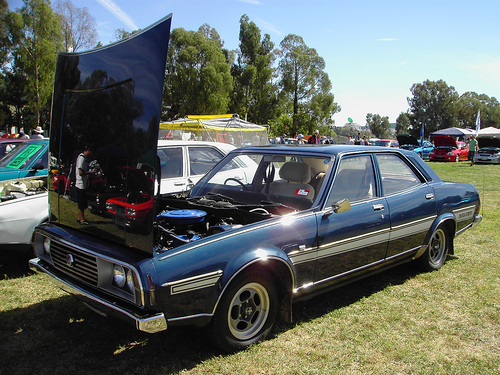}
	}\hfil
	\subfloat[GT: \inc{cannon} \\Pred: \inc{aircraft carrier}\\OOV: Military vessel]{
		\includegraphics[width=0.23\linewidth]{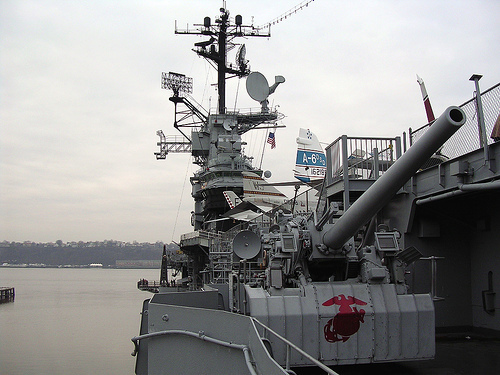}
	}\hfil
	\subfloat[GT: \inc{lumbermill} \\Pred: \inc{crane} \\OOV: Lifting device or facility]{
		\includegraphics[width=0.23\linewidth]{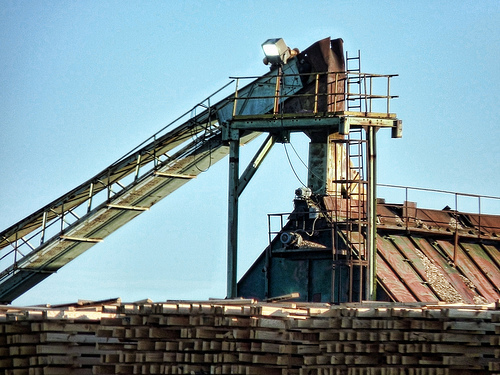}
	}\hfil
    \subfloat[GT: \inc{dock} \\Pred: \inc{aircraft carrier} \\ OOV: Guard ship or facility]{
		\includegraphics[width=0.23\linewidth]{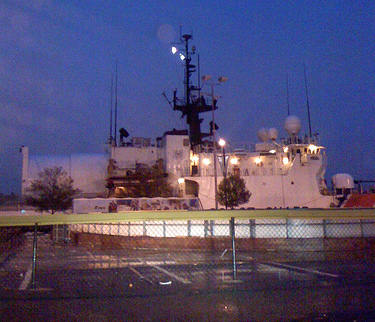}
	}\hfil
	\subfloat[GT: \inc{car wheel} \\Pred: \inc{sports car}\\ OOV: Coupe or other car model]{
		\includegraphics[width=0.23\linewidth]{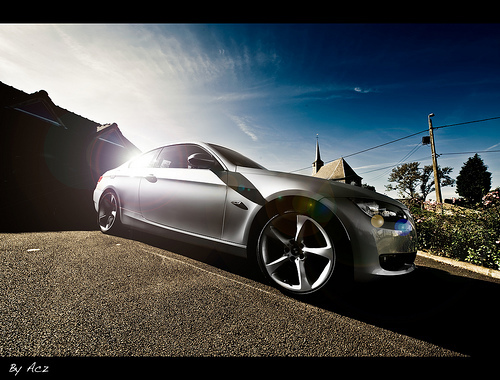}
	}\hfil
	\subfloat[GT: \inc{teddy bear} \\Pred: \inc{toyshop}\\OOV: Gift shop or other store]{
		\includegraphics[width=0.23\linewidth]{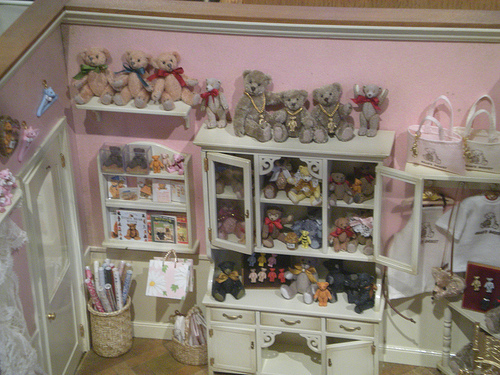}
	}\hfil
	\subfloat[GT: \inc{plate} \\Pred: \inc{ice cream} \\ OOV: Ice-cream cake or dessert]{
		\includegraphics[width=0.23\linewidth]{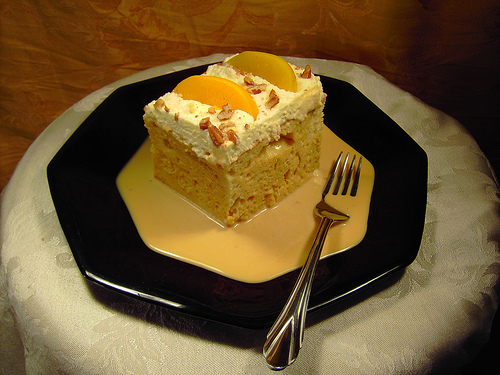}
	}\hfil
	\subfloat[GT: \inc{tiger cat} \\Pred: \inc{lynx} \\OOV: Baby ocelot]{
		\includegraphics[width=0.23\linewidth]{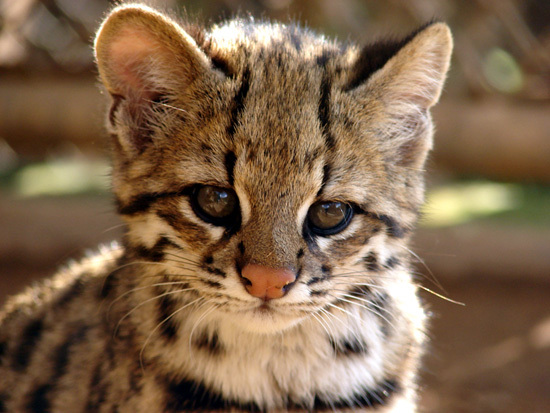}
	}\hfil
	\subfloat[GT: \inc{paddle} \\Pred: \inc{canoe}\\OOV: Rowboat]{
		\includegraphics[width=0.23\linewidth]{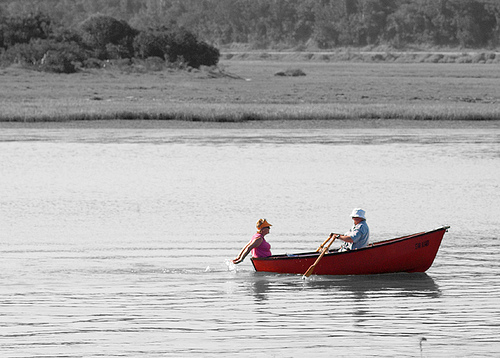}
	}\hfil
	\subfloat[GT: \inc{sea snake}, \inc{coral reef} \\Pred: \inc{pufferfish} \\OOV: Other fish]{
		\includegraphics[width=0.23\linewidth]{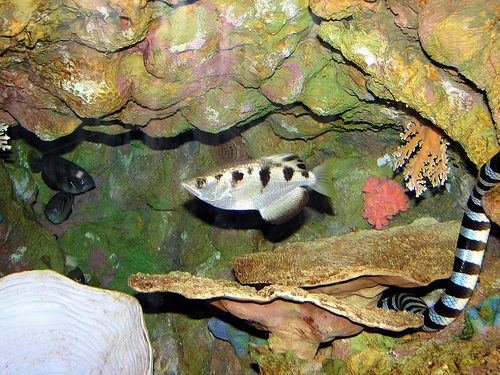}
	}\hfil
	\subfloat[GT: \inc{hammer}, \inc{hook} \\Pred: \inc{shovel} \\OOV: Other tool / shovel type]{
		\includegraphics[width=0.23\linewidth]{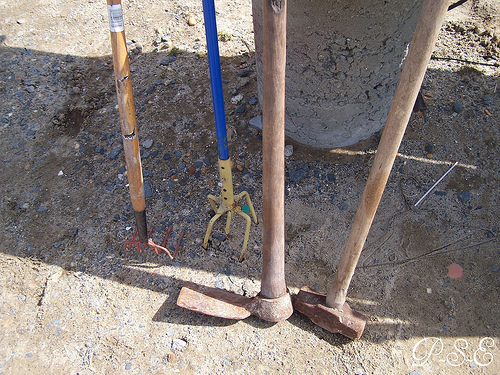}
	}\hfil
	\subfloat[GT: \inc{strainer} \\Pred: \inc{cucumber}\\OOV: Other vegetable]{
		\includegraphics[width=0.23\linewidth]{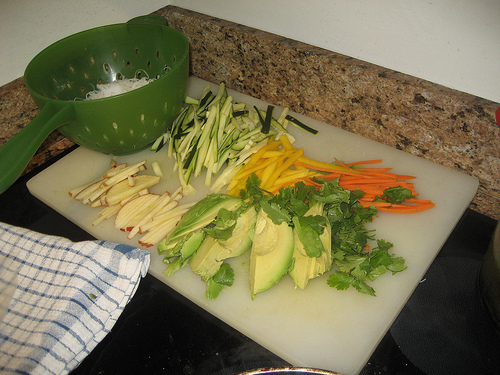}
	}\hfil
	\subfloat[GT: \inc{cradle}, \inc{barn}, \inc{hay} \\Pred: \inc{altar} \\ OOV: Altar crib]{
		\includegraphics[width=0.23\linewidth]{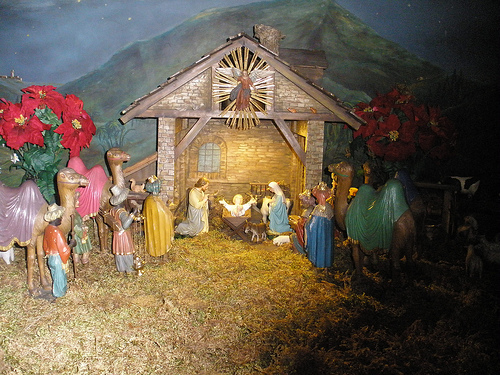}
	}\hfil
	\subfloat[GT: \inc{holster} \\Pred: \inc{revolver} \\ OOV: Semiautomatic pistol]{
		\includegraphics[width=0.23\linewidth]{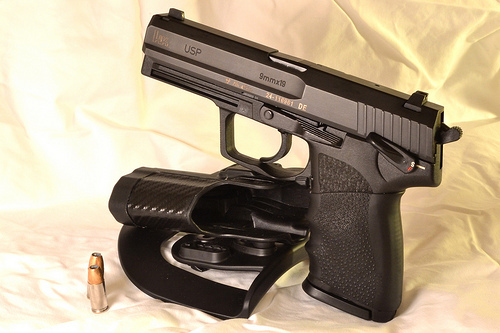}
	}
    \caption{
		\textbf{Fine-grained OOV}: examples of common prediction errors (Pred)
		due to fine-grained OOV errors.
		The correct (individual) multi-labels (GT) are separated by commas;
		the original \IN label is listed first.
	}
	\label{app:fig-fine-grained-OOV}
\end{figure}

\newpage

\begin{figure}[H]
    \captionsetup[subfloat]{labelformat=empty}
	\centering
	\subfloat[GT: \inc{pencil box}, \inc{purse} \\Pred: \inc{pillow}]{
		\includegraphics[width=0.23\linewidth]{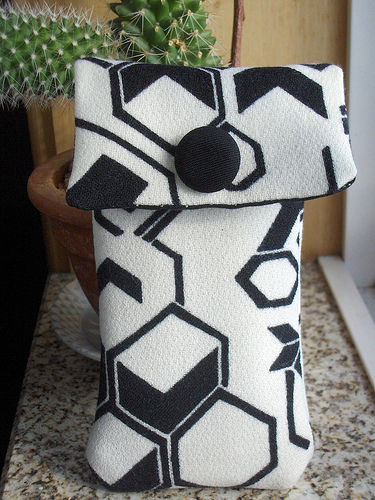}
	}\hfil
	\subfloat[GT: \inc{padlock} \hspace*{5em} \\Pred: \inc{shield}]{
		\includegraphics[width=0.23\linewidth]{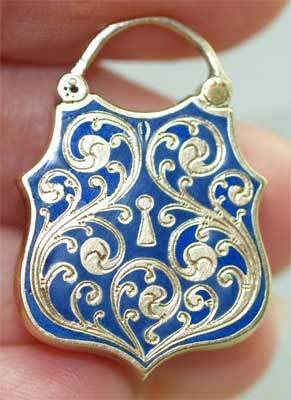}
	}\hfil
	\subfloat[GT: \inc{breastplate} \\Pred: \inc{water jug}]{
		\includegraphics[width=0.23\linewidth]{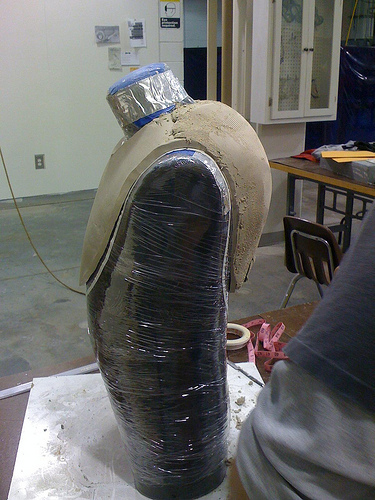}
	}\hfil
	\subfloat[GT: \inc{flute} \hspace*{5em}\\Pred: \inc{plunger}]{
		\includegraphics[width=0.23\linewidth]{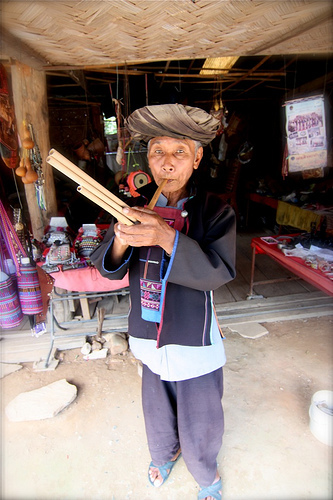}
	}\hfil
	\subfloat[GT: \inc{marmoset} \hspace*{5em} \\Pred: \inc{tarantula}]{
		\includegraphics[width=0.23\linewidth]{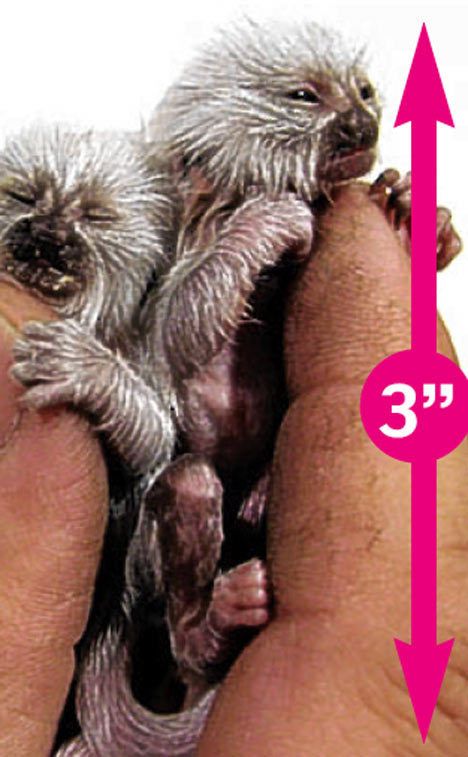}
	}\hfil
	\subfloat[GT: \inc{badger} \hspace*{5em} \\Pred: \inc{wombat}]{
		\includegraphics[width=0.23\linewidth]{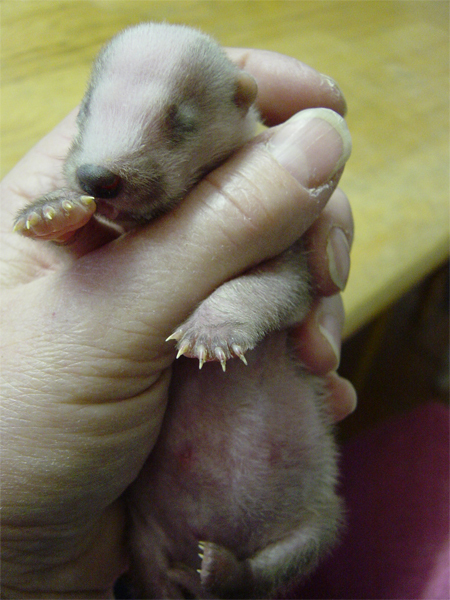}
	}\hfil
	\subfloat[GT: \inc{water jug} \\Pred: \inc{espresso maker}]{
		\includegraphics[width=0.23\linewidth]{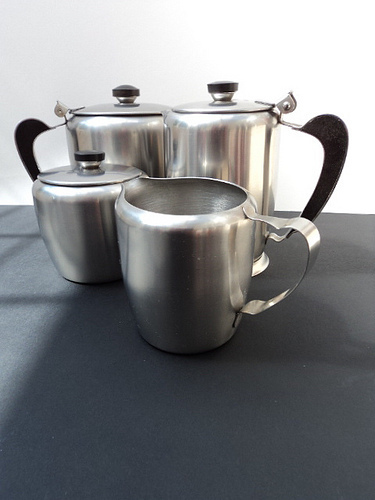}
	}\hfil
	\subfloat[GT: \inc{green lizard} \hspace*{3em} \\Pred: \inc{buckeye}]{
		\includegraphics[width=0.23\linewidth]{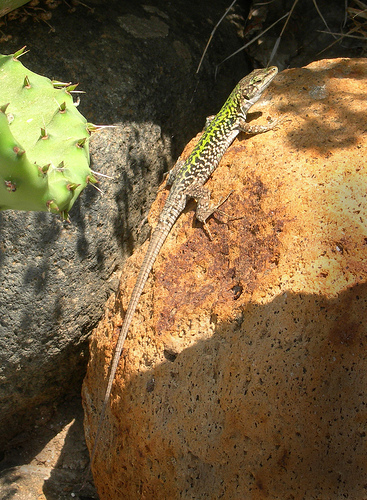}
	}\hfil
	\subfloat[GT: \inc{hair slide} \hspace*{5em}\\Pred: \inc{pinwheel}]{
		\includegraphics[width=0.23\linewidth]{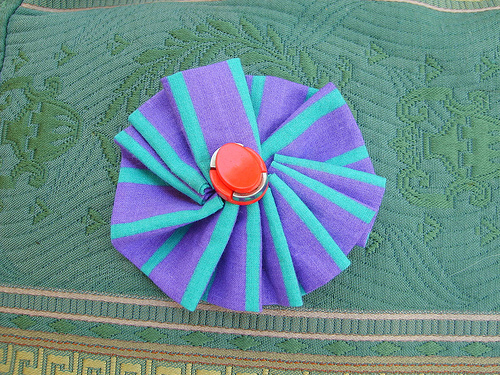}
	}\hfil
	\subfloat[GT: \inc{English foxhound} \\Pred: \inc{Staffordshire terrier}]{
		\includegraphics[width=0.23\linewidth]{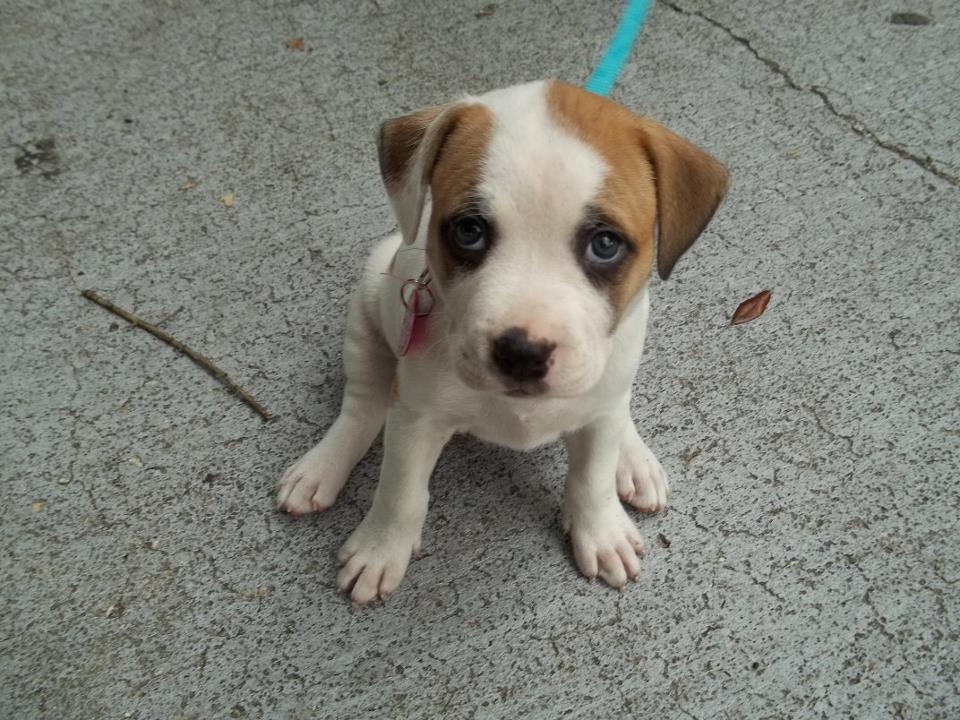}
	}\hfil
	\subfloat[GT: \inc{nipple} \hspace*{5em} \\Pred: \inc{candle}]{
		\includegraphics[width=0.23\linewidth]{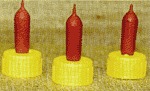}
	}\hfil
	\subfloat[GT: \inc{chest}, \inc{necklace} \hspace*{2em} \\Pred: \inc{radio}]{
		\includegraphics[width=0.23\linewidth]{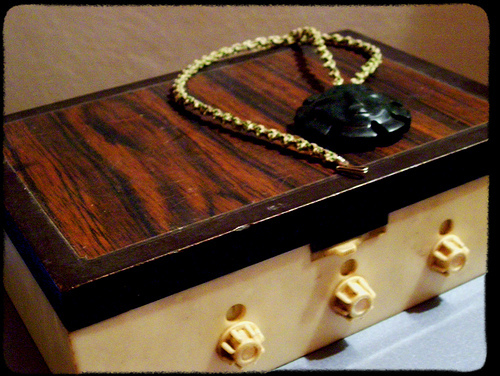}
	}\hfil
	\subfloat[GT: \inc{airliner} \hspace*{5em} \\Pred: \inc{wreck}]{
		\includegraphics[width=0.23\linewidth]{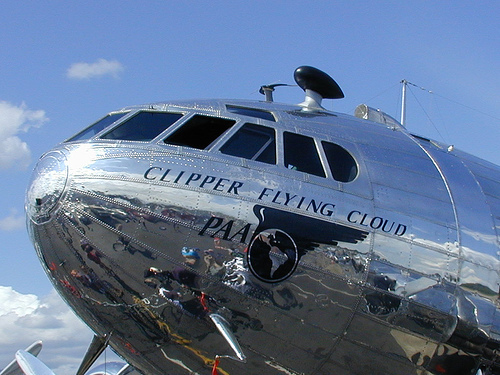}
	}\hfil
	\subfloat[GT: \inc{lipstick} \hspace*{5em}\\Pred: \inc{beaker}]{
		\includegraphics[width=0.23\linewidth]{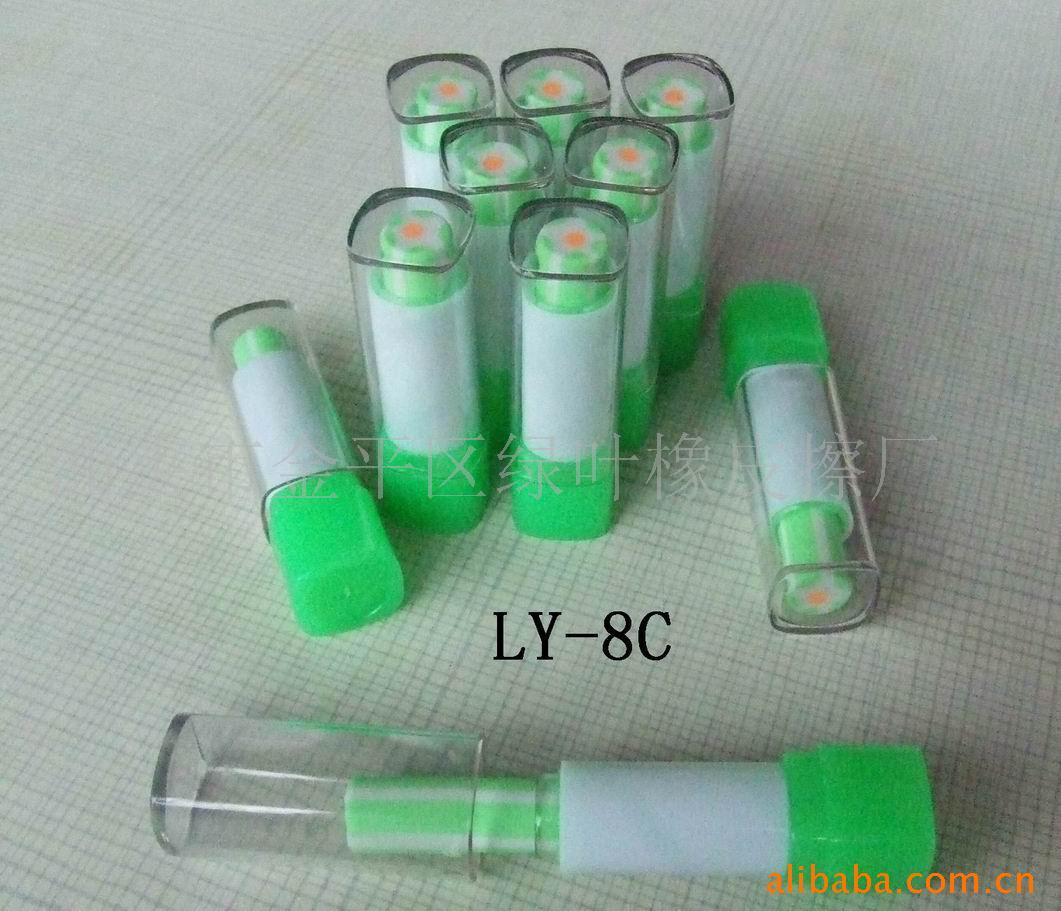}
	}\hfil
	\subfloat[GT: \inc{bonnet}, \inc{lipstick}, \\ \inc{saltshaker} \\Pred: \inc{bucket}]{
		\includegraphics[width=0.23\linewidth]{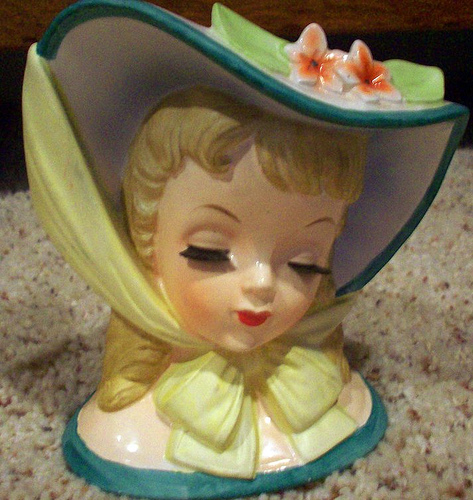}
	}\hfil
	\subfloat[GT: \inc{jack-o'-lantern} \\Pred: \inc{traffic light}]{
		\includegraphics[width=0.23\linewidth]{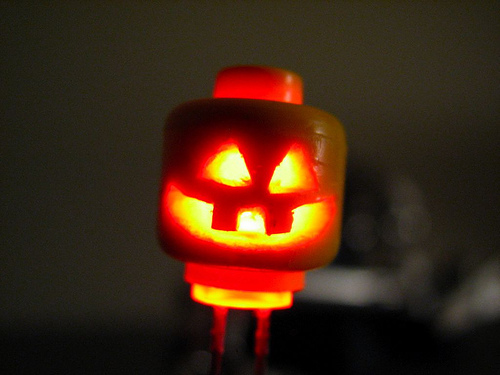}
	}\hfil
    \caption{
		\textbf{Non-prototypical}: examples of common prediction errors (Pred)
		due to non-prototypical instances.
		The correct (individual) multi-labels (GT) are separated by commas;
		the original \IN label is listed first.
	}
	\label{app:fig-non-proto}
\end{figure}

\newpage

\begin{figure}[H]
    \captionsetup[subfloat]{labelformat=empty}
	\centering
	\subfloat[GT: \inc{vestment} \hspace*{5em}\\Pred: \inc{altar}]{
		\includegraphics[width=0.23\linewidth]{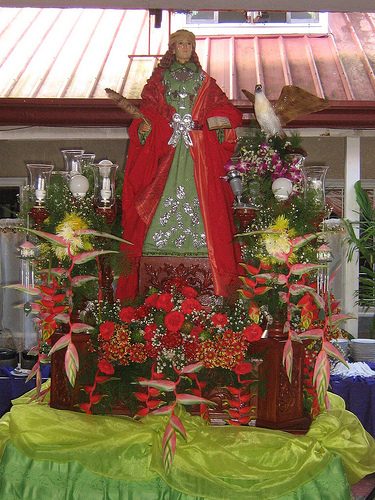}
	}\hfil
	\subfloat[GT: \inc{washbasin} \\Pred: \inc{medicine chest}]{
		\includegraphics[width=0.23\linewidth]{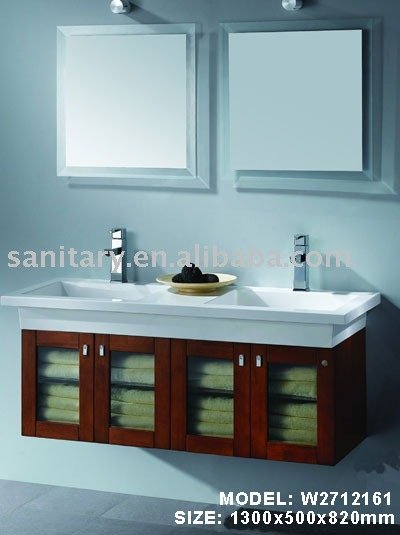}
	}\hfil
	\subfloat[GT: \inc{tank suit}, \inc{maillot} \\Pred: \inc{bathing cap}]{
		\includegraphics[width=0.23\linewidth]{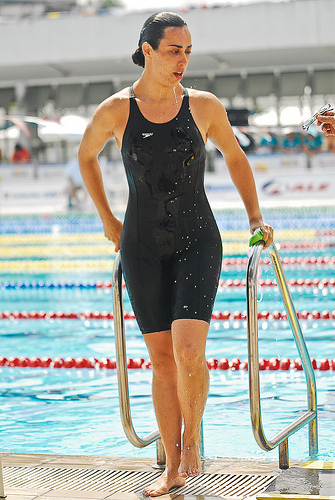}
	}\hfil
	\subfloat[GT: \inc{chain} \hspace*{5em} \\Pred: \inc{hook}]{
		\includegraphics[width=0.23\linewidth]{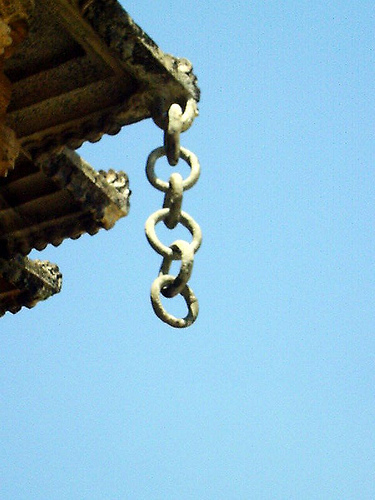}
	}\hfil
	\subfloat[GT: \inc{grille}\hspace*{5em} \\Pred: \inc{police van}]{
		\includegraphics[width=0.23\linewidth]{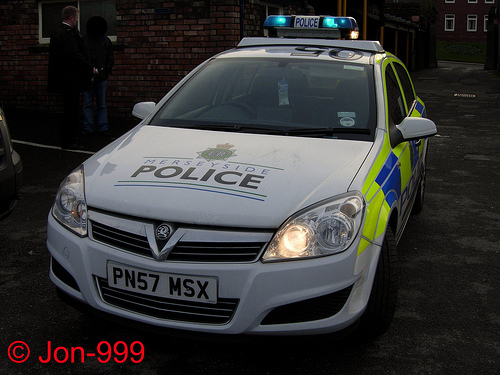}
	}\hfil
	\subfloat[GT: \inc{otterhound} \hspace*{5em} \\Pred: \inc{briard}]{
		\includegraphics[width=0.23\linewidth]{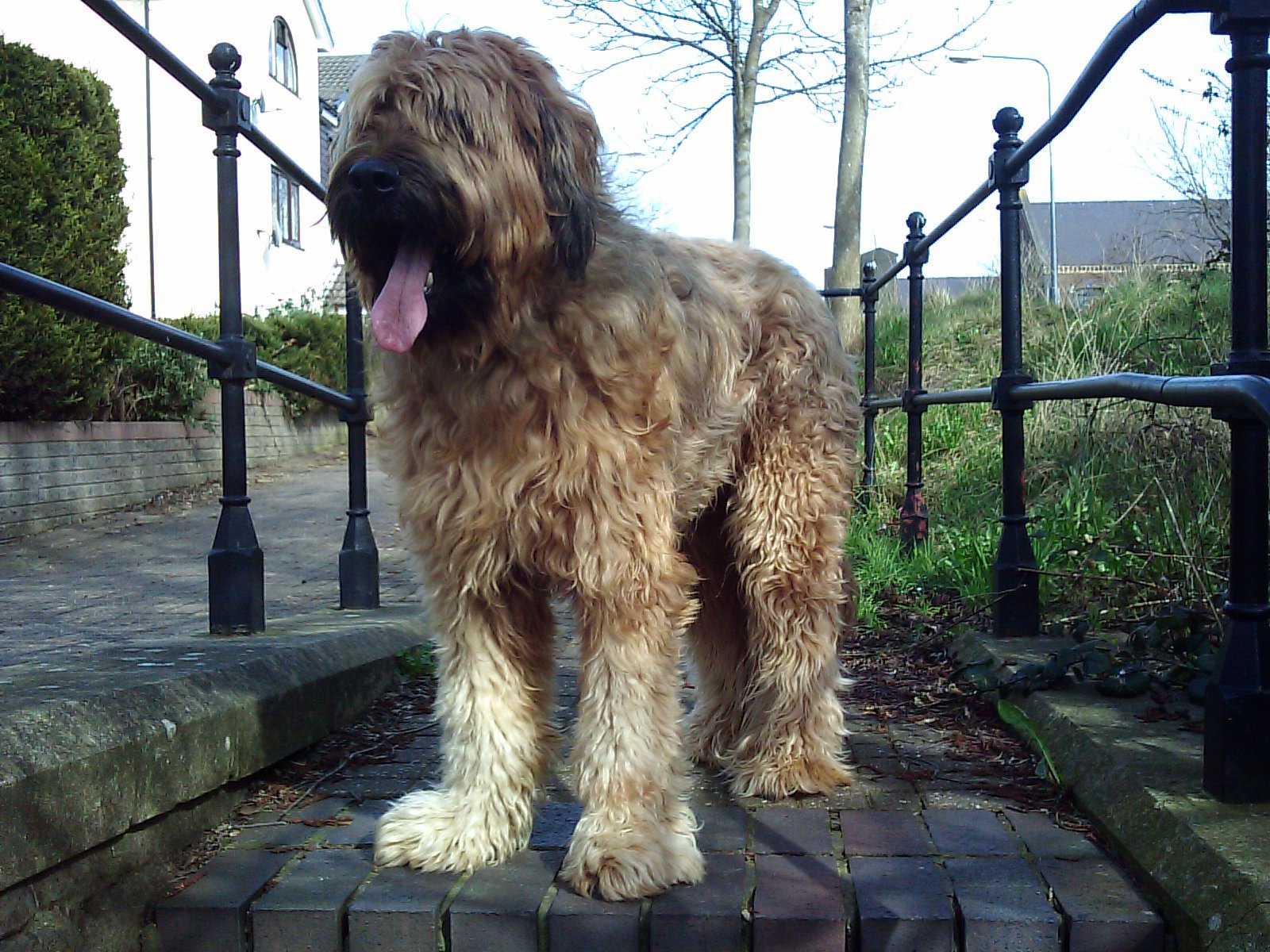}
	}\hfil
	\subfloat[GT: \inc{ski mask}, \inc{alp} \hspace*{3em} \\Pred: \inc{ski}]{
		\includegraphics[width=0.23\linewidth]{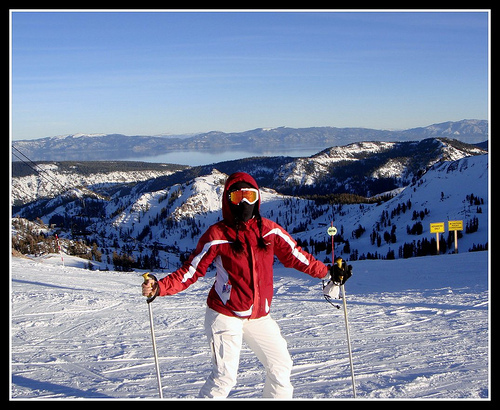}
	}\hfil
	\subfloat[GT: \inc{bookcase} \hspace*{5em} \\Pred: \inc{bookshop}]{
		\includegraphics[width=0.23\linewidth]{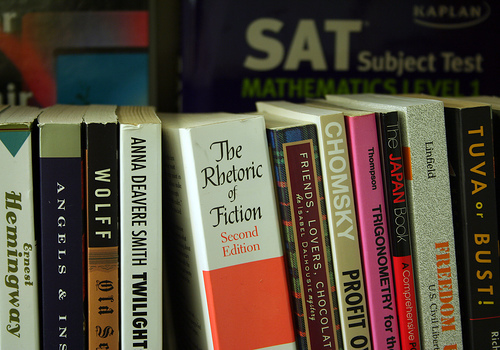}
	}\hfil
	\subfloat[GT: \inc{strawberry} \hspace*{5em} \\Pred: \inc{trifle}]{
		\includegraphics[width=0.23\linewidth]{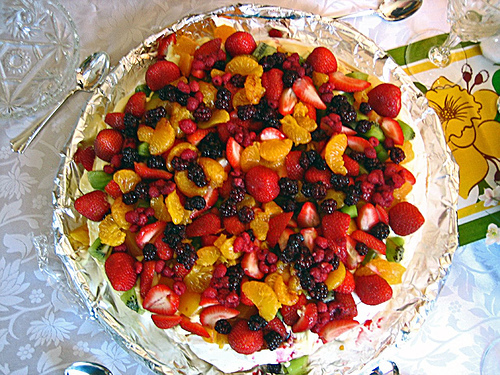}
	}\hfil
	\subfloat[GT: \inc{flowerpot} \hspace*{4em}\\Pred: \inc{greenhouse}]{
		\includegraphics[width=0.23\linewidth]{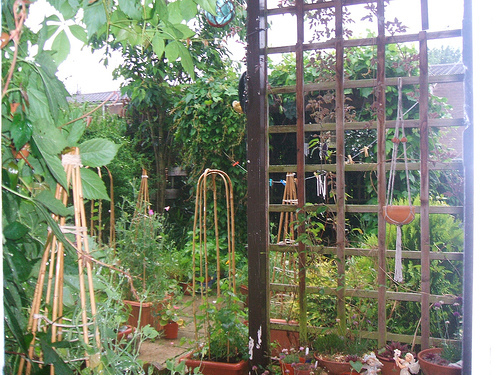}
	}\hfil
	\subfloat[GT: \inc{wool} \hspace*{5em} \\Pred: \inc{bonnet}]{
		\includegraphics[width=0.23\linewidth]{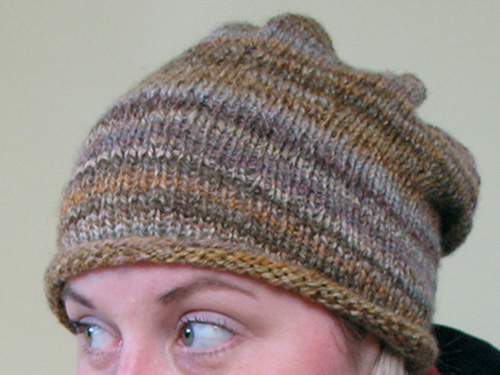}
	}\hfil
	\subfloat[GT: \inc{chain}, \inc{suspension bridge} \\Pred: \inc{swing}]{
		\includegraphics[width=0.23\linewidth]{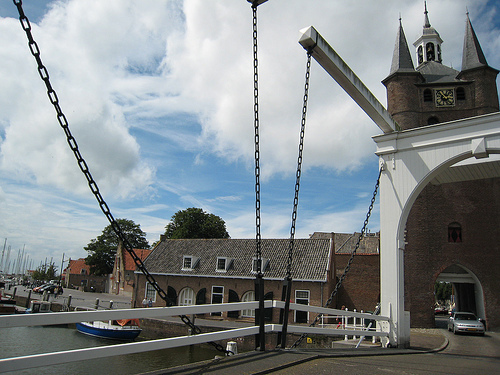}
	}\hfil
	\subfloat[GT: \inc{traffic light}, \\ \inc{crash helmet} \\Pred: \inc{bicycle-built-for-two}]{
		\includegraphics[width=0.23\linewidth]{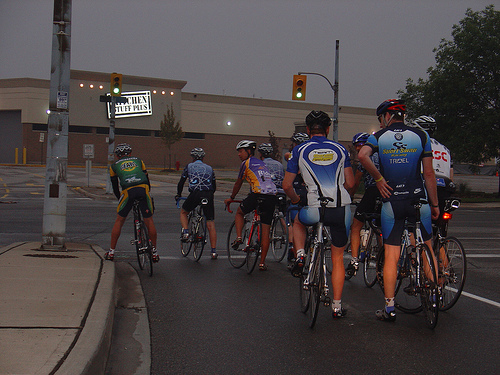}
	}\hfil
	\subfloat[GT: \inc{kuvasz} \hspace*{5em}\\Pred: \inc{Samoyed}]{
		\includegraphics[width=0.23\linewidth]{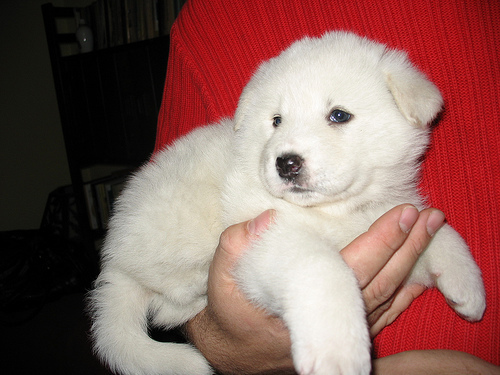}
	}\hfil
	\subfloat[GT: \inc{Windsor tie} \hspace*{5em} \\Pred: \inc{suit}]{
		\includegraphics[width=0.23\linewidth]{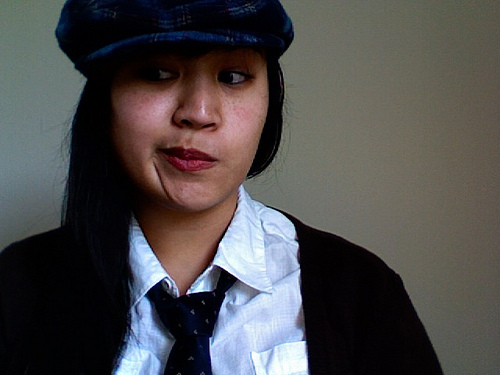}
	}\hfil
	\subfloat[GT: \inc{water jug} \hspace*{5em} \\Pred: \inc{water bottle}]{
		\includegraphics[width=0.23\linewidth]{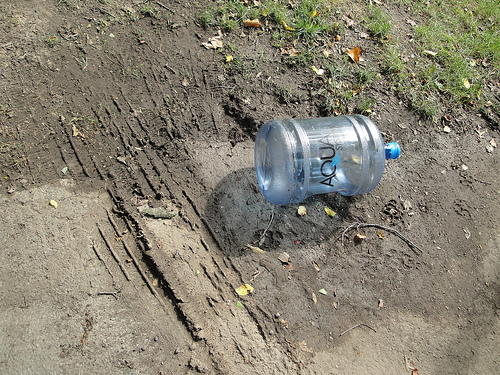}
	}\hfil
	\subfloat[GT: \inc{bulletproof vest}, \\ \inc{police van} \\Pred: \inc{military uniform}]{
		\includegraphics[width=0.23\linewidth]{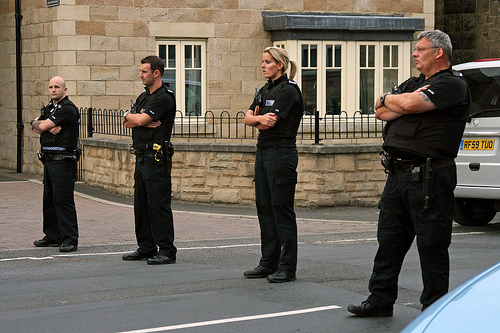}
	}\hfil
	\subfloat[GT: \inc{warplane}, \inc{pole}, \inc{chain} \\Pred: \inc{aircraft carrier}]{
		\includegraphics[width=0.23\linewidth]{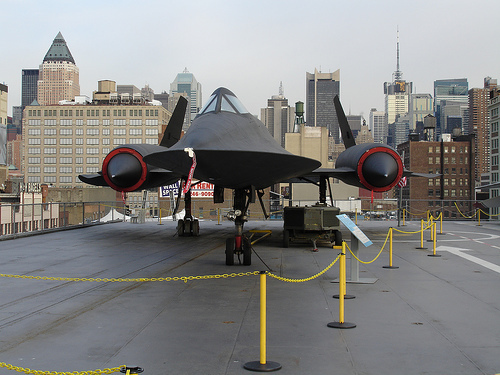}
	}\hfil
	\subfloat[GT: \inc{pole} \hspace*{5em} \\Pred: \inc{traffic light}]{
		\includegraphics[width=0.23\linewidth]{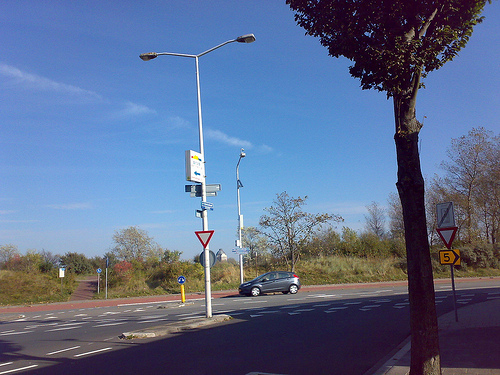}
	}\hfil
	\subfloat[GT: \inc{tobacco shop} \hspace*{2em} \\Pred: \inc{street sign}]{
		\includegraphics[width=0.23\linewidth]{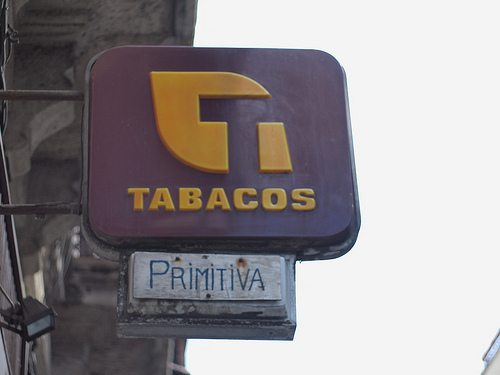}
	}
    \caption{
		\textbf{Spurious correlations}: examples of common prediction errors (Pred)
		due to spurious correlations.
		The correct (individual) multi-labels (GT) are separated by commas;
		the original \IN label is listed first.
	}
	\label{app:fig-spur-cor}
\end{figure}

\newpage

\begin{figure}[H]
    \captionsetup[subfloat]{labelformat=empty}
	\centering
	\subfloat[GT: \inc{wardrobe} \\Pred: \inc{refrigerator}]{
		\includegraphics[width=0.23\linewidth]{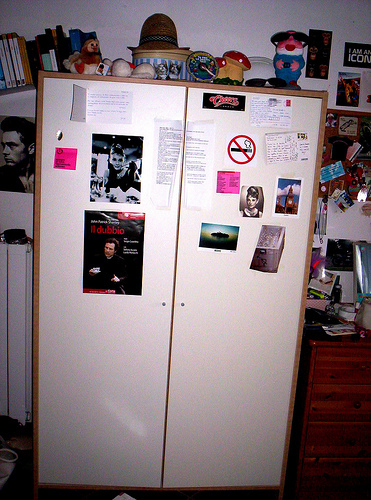}
	}\hfil
	\subfloat[GT: \inc{washer} \hspace*{5em} \\Pred: \inc{vending machine}]{
		\includegraphics[width=0.23\linewidth]{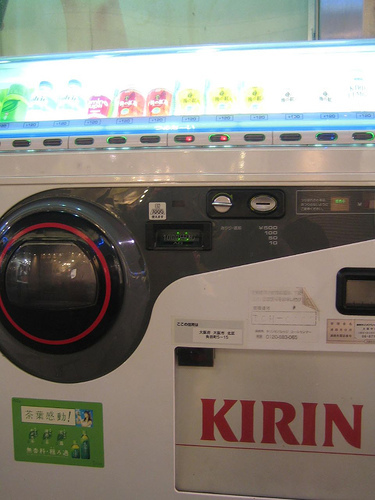}
	}\hfil
	\subfloat[GT: \inc{backpack} \hspace*{5em} \\Pred: \inc{unicycle}]{
		\includegraphics[width=0.23\linewidth]{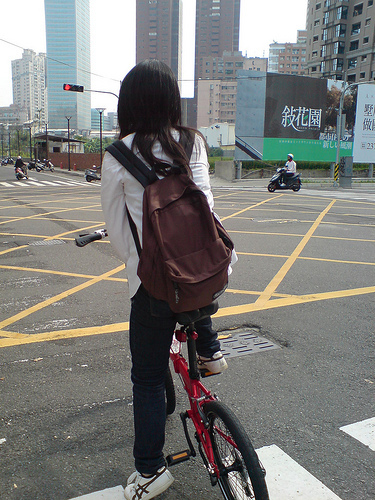}
	}\hfil
	\subfloat[GT: \inc{loggerhead} \hspace*{3em} \\Pred: \inc{scuba diver}]{
		\includegraphics[width=0.23\linewidth]{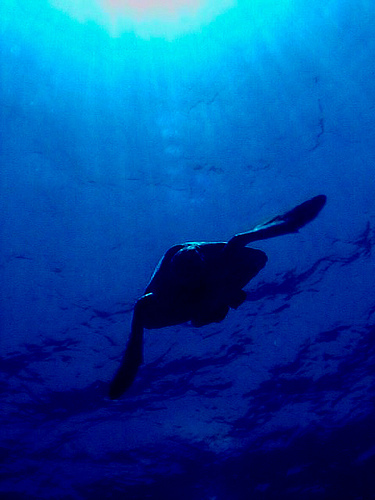}
	}\hfil
	\subfloat[GT: \inc{soccer ball} \hspace*{3em}\\Pred: \inc{potter's wheel}]{
		\includegraphics[width=0.23\linewidth]{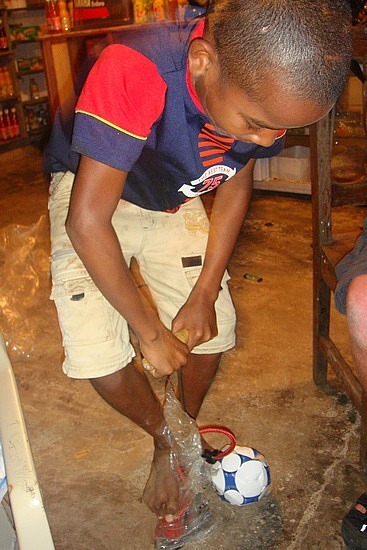}
	}\hfil
	\subfloat[GT: \inc{missile}, \inc{projectile} \\Pred: \inc{obelisk}]{
		\includegraphics[width=0.23\linewidth]{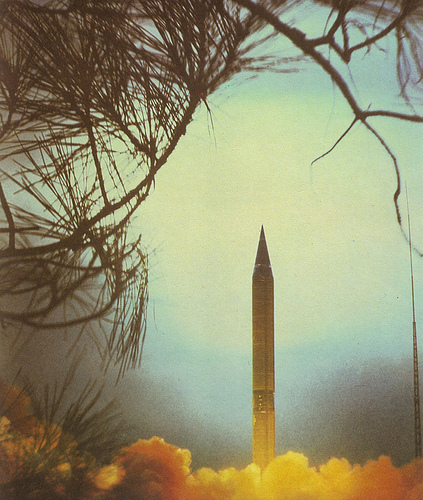}
	}\hfil
	\subfloat[GT: \inc{overskirt} \hspace*{4em} \\Pred: \inc{apron}]{
		\includegraphics[width=0.23\linewidth]{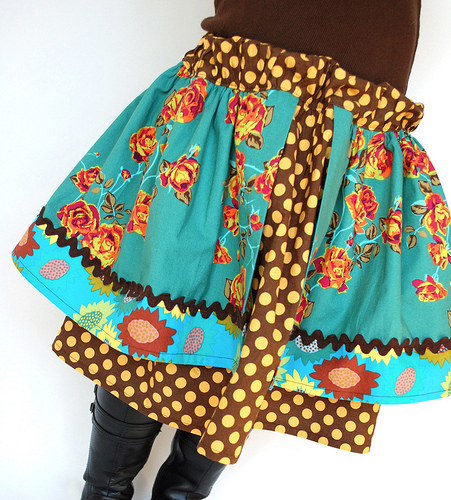}
	}\hfil
	\subfloat[GT: \inc{dumbbell} \hspace*{5em} \\Pred: \inc{abacus}]{
		\includegraphics[width=0.23\linewidth]{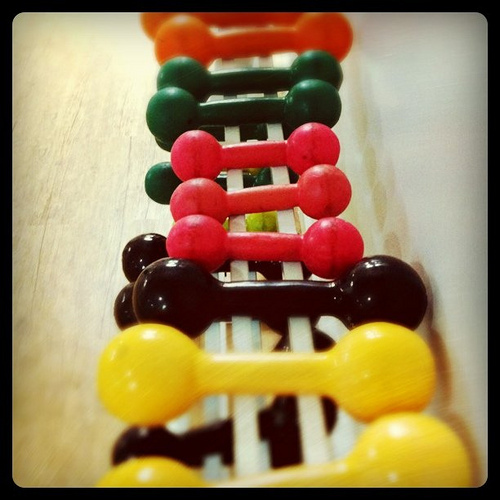}
	}\hfil
	\subfloat[GT: \inc{pay-phone} \hspace*{5em}\\Pred: \inc{cash machine}]{
		\includegraphics[width=0.23\linewidth]{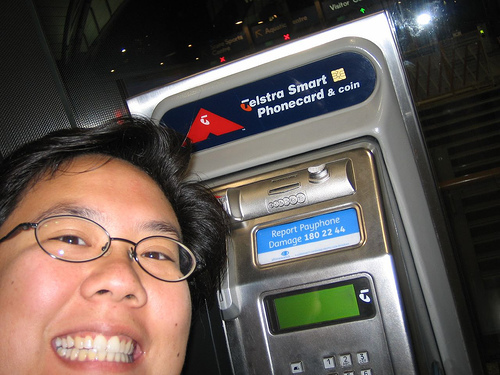}
	}\hfil
	\subfloat[GT: \inc{hatchet}, \inc{chain mail} \\Pred: \inc{pedestal}]{
		\includegraphics[width=0.23\linewidth]{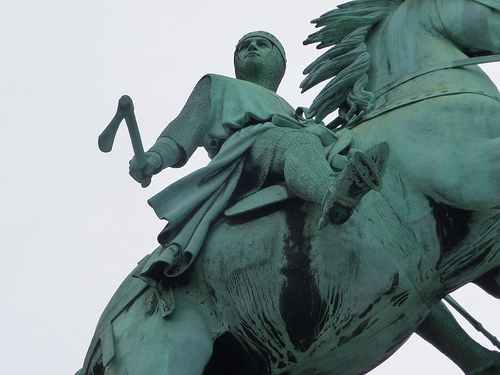}
	}\hfil
	\subfloat[GT: \inc{pop bottle}\hspace*{5em} \\Pred: \inc{confectionery}]{
		\includegraphics[width=0.23\linewidth]{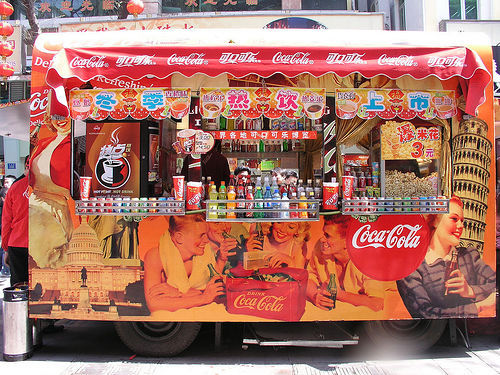}
	}\hfil
	\subfloat[GT: \inc{library} \hspace*{5em} \\Pred: \inc{lab coat}]{
		\includegraphics[width=0.23\linewidth]{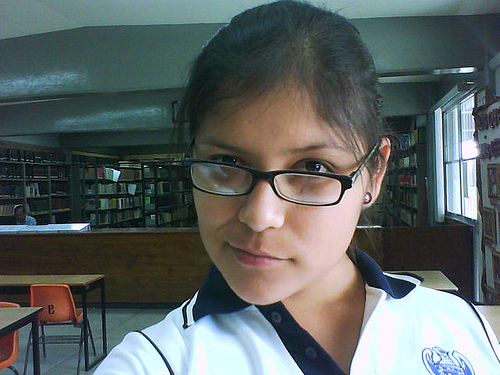}
	}\hfil
	\subfloat[GT: \inc{letter opener} \hspace*{2em} \\Pred: \inc{pencil sharpener}]{
		\includegraphics[width=0.23\linewidth]{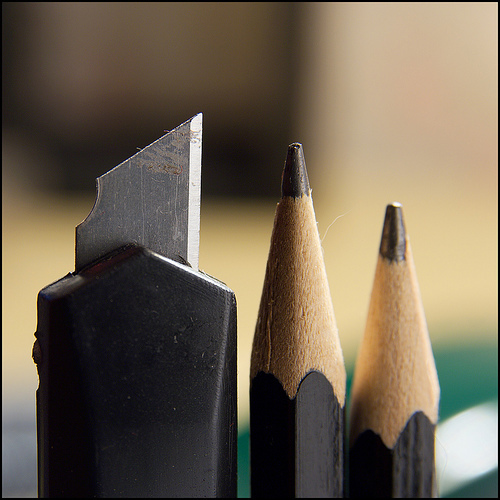}
	}\hfil
	\subfloat[GT: \inc{suit}, \inc{umbrella} \hspace*{3em} \\Pred: \inc{swing}]{
		\includegraphics[width=0.23\linewidth]{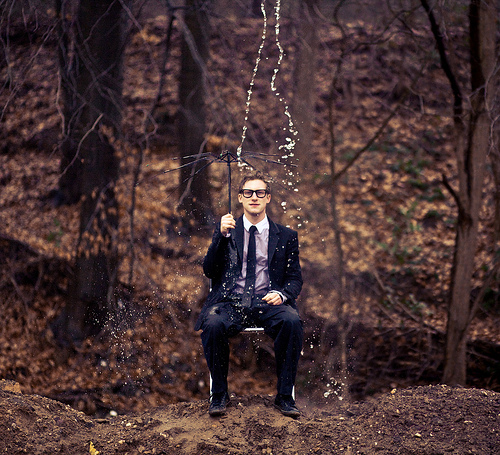}
	}\hfil
	\subfloat[GT: \inc{bannister} \hspace*{4em} \\Pred: \inc{crib}]{
		\includegraphics[width=0.23\linewidth]{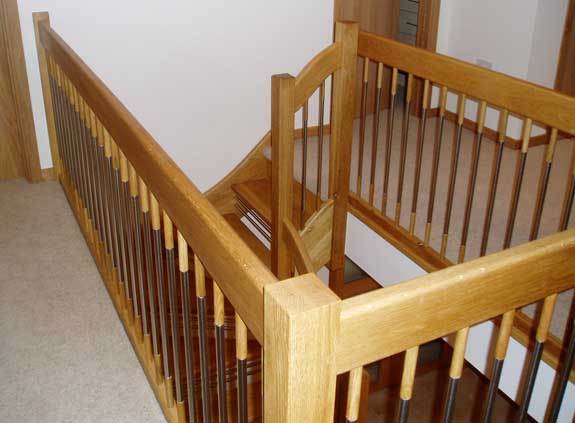}
	}\hfil
	\subfloat[GT: \inc{passenger car} \hspace*{5em} \\Pred: \inc{studio couch}]{
		\includegraphics[width=0.23\linewidth]{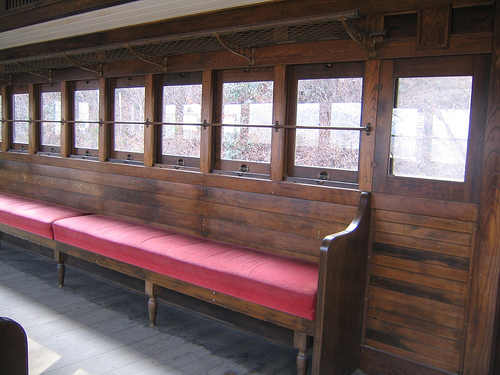}
	}\hfil
	\subfloat[GT: \inc{pig} \hspace*{5em} \\Pred: \inc{greenhouse}]{
		\includegraphics[width=0.23\linewidth]{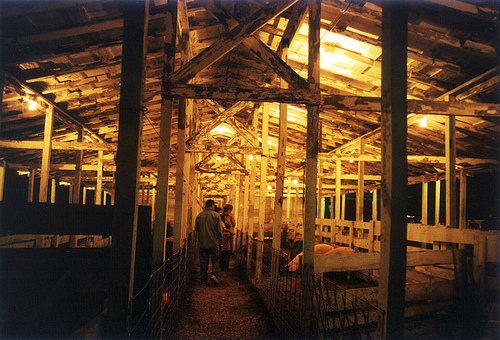}
	}\hfil
	\subfloat[GT: \inc{frying pan} \hspace*{4em}\\Pred: \inc{tray}]{
		\includegraphics[width=0.23\linewidth]{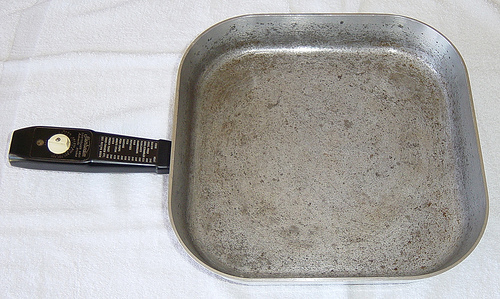}
	}\hfil
	\subfloat[GT: \inc{dark glasses}, \inc{sunglass} \\Pred: \inc{sunscreen}]{
		\includegraphics[width=0.23\linewidth]{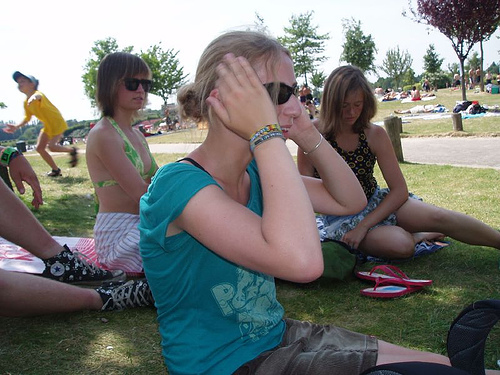}
	}\hfil
	\subfloat[GT: \inc{thimble} \hspace*{5em} \\Pred: \inc{microphone}]{
		\includegraphics[width=0.23\linewidth]{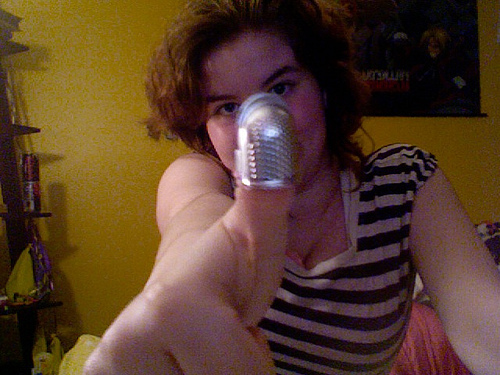}
	}
    \caption{
		\textbf{Model failures}: examples of common prediction errors (Pred)
		classified as model failures.
		The correct (individual) multi-labels (GT) are separated by commas;
		the original \IN label is listed first.
	}
	\label{app:fig-model-fail}
\end{figure}

\newpage
\section{Experimental Setup} \label{sec:setup}

In this section, we provide further details on our evaluation.

We publish our code and detailed instructions on how to reproduce our results at
\texttt{\href{https://github.com/eth-sri/automated-error-analysis}{https://github.com/eth-sri/automated-error-analysis}}.

\subsection{Computing the Multi-Label Accuracy}

When computing the models' multi-label accuracies, we closely follow the recommendations
by~\citet{ShankarRMFRS20}\footnote{\label{footnote:multilabel}\url{https://www.tensorflow.org/datasets/catalog/imagenet2012_multilabel}}.
In particular, we discard the set of problematic images, then compute per-class
accuracies and finally average them to obtain the multi-label accuracy. We consider
a model prediction to be correct if it is marked as correct or unclear in the
multi-label dataset.

\subsection{Datasets and Labels}
We consider the validation set of the ILSVRC-2012 subset of \IN \citep{DengDSLL009,RussakovskyDSKS15}, available under a non-commercial research license\footnote{Fore more details see \url{https://image-net.org/download}}. More concretely, we use the subset of this validation set labeled by \citet{ShankarRMFRS20} and then \citet{VasudevanCLFR22}, with the labels\textsuperscript{\ref{footnote:multilabel}} being available under Apache License 2.0. We further evaluate our pipeline on the ImageNet-A dataset~\citep{HendrycksZBSS21} available under MIT License\footnote{\url{https://github.com/hendrycks/natural-adv-examples}}.

\subsection{Summary of Evaluated Models}

\cref{tbl:models-summary} contains a list of all models we considered in this study
and a subset of their metadata. The models were obtained from multiple sources: Torchvision\footnote{\url{https://pytorch.org/vision/stable/models.html}}, \texttt{torch.hub}\footnote{\url{https://pytorch.org/hub/}}, HuggingFace\footnote{\url{https://huggingface.co/models}}, and \texttt{timm}\footnote{\url{https://github.com/huggingface/pytorch-image-models}}. They can all be automatically downloaded and evaluated using our open-sourced code. After collecting all model outputs (6 days for ImageNet and 1 day for ImageNet-A on a single GeForce RTX 2080 Ti GPU), running our error analysis pipeline on all models takes $12$ to $24$ hours using a single GeForce RTX 2080 Ti GPU for ImageNet and ImageNet-A respectively.

\begin{small}
\begin{longtable}{@{}rllll@{}}
	\caption{List of evaluated models and a subset of their metadata.}
	\endfirsthead
	\midrule
	\multicolumn{5}{@{}c@{}}%
	{Model summaries (continued from previous page)}\\
	\midrule
	No & Model ID &  Source & Arch & Dataset\\
	\midrule
	\endhead
	\midrule
	\multicolumn{5}{@{}c@{}}
	{Table continues onto next page}\\
	\midrule
	\endfoot
	\endlastfoot
    \toprule
    No & Model ID &  Source & Arch & Dataset \\
    \midrule
    1. & alexnet & torch & cnn & ImageNet-1k \\
    2. & bat\_resnext26ts.ch\_in1k & timm & cnn & ImageNet-1k \\
    3. & convformer\_b36.sail\_in1k & timm & cnn & ImageNet-1k \\
    4. & convformer\_b36.sail\_in1k\_384 & timm & cnn & ImageNet-1k \\
    5. & convformer\_b36.sail\_in22k\_ft\_in1k & timm & cnn & ImageNet-21k \\
    6. & convformer\_b36.sail\_in22k\_ft\_in1k\_384 & timm & cnn & ImageNet-21k \\
    7. & convformer\_m36.sail\_in1k & timm & cnn & ImageNet-1k \\
\end{longtable}
\end{small}

\subsection{Fine-grained Superclasses}

In this section, we detail the superclasses we use for detecting fine-grained and fine-grained OOV errors and their constituting \IN classes. When defining these superclasses, we focused on visual and semantic similarity rather than positions in the WordNet hierarchy. We believe that the latter is fundamentally unsuitable to obtain high-quality groupings, as visually similar or closely related classes often end up having a large WordNet distance, while very different classes are close.
Further, we ensured that superclass definitions are not overly generic, providing a much more fine-grained split than prior works. The superclasses are defined programmatically in our code and can be easily reused and adapted by others.

List of superclass definitions:
\begin{enumerate}[wide, labelwidth=!, labelindent=0pt]
	\item \textbf{\texttt{butterfly}} -- 6 classes: 
	
\begin{tabular}{p{0.5\linewidth}p{0.5\linewidth}}
		\inc{(n02276258, admiral)} & \inc{(n02277742, ringlet)}\\
		\inc{(n02279972, monarch)} & \inc{(n02280649, cabbage butterfly)}\\
		\inc{(n02281406, sulphur butterfly)} & \inc{(n02281787, lycaenid)}\\
	\end{tabular}
	\item \textbf{\texttt{beetle\_plus}} -- 13 classes: 
	
	\begin{tabular}{p{0.5\linewidth}p{0.5\linewidth}}
		\inc{(n02165105, tiger beetle)} & \inc{(n02165456, ladybug)}\\
		\inc{(n02167151, ground beetle)} & \inc{(n02168699, long-horned beetle)}\\
		\inc{(n02169497, leaf beetle)} & \inc{(n02172182, dung beetle)}\\
		\inc{(n02174001, rhinoceros beetle)} & \inc{(n02177972, weevil)}\\
		\inc{(n02226429, grasshopper)} & \inc{(n02229544, cricket)}\\
		\inc{(n02233338, cockroach)} & \inc{(n02256656, cicada)}\\
		\inc{(n02259212, leafhopper)} & \\
	\end{tabular}
	\item \textbf{\texttt{insect\_rest}} -- 13 classes: 
	
	\begin{tabular}{p{0.5\linewidth}p{0.5\linewidth}}
		\inc{(n02190166, fly)} & \inc{(n02206856, bee)}\\
		\inc{(n02219486, ant)} & \inc{(n02226429, grasshopper)}\\
		\inc{(n02229544, cricket)} & \inc{(n02231487, walking stick)}\\
		\inc{(n02233338, cockroach)} & \inc{(n02236044, mantis)}\\
		\inc{(n02256656, cicada)} & \inc{(n02259212, leafhopper)}\\
		\inc{(n02264363, lacewing)} & \inc{(n02268443, dragonfly)}\\
		\inc{(n02268853, damselfly)} & \\
	\end{tabular}
	\item \textbf{\texttt{isopod\_trilobite\_chiton}} -- 3 classes: 
	
	\begin{tabular}{p{0.5\linewidth}p{0.5\linewidth}}
		\inc{(n01768244, trilobite)} & \inc{(n01955084, chiton)}\\
		\inc{(n01990800, isopod)} & \\
	\end{tabular}
	\item \textbf{\texttt{crab\_scorpion}} -- 9 classes: 
	
	\begin{tabular}{p{0.5\linewidth}p{0.5\linewidth}}
		\inc{(n01770393, scorpion)} & \inc{(n01978287, Dungeness crab)}\\
		\inc{(n01978455, rock crab)} & \inc{(n01980166, fiddler crab)}\\
		\inc{(n01981276, king crab)} & \inc{(n01983481, American lobster)}\\
		\inc{(n01984695, spiny lobster)} & \inc{(n01985128, crayfish)}\\
		\inc{(n01986214, hermit crab)} & \\
	\end{tabular}
	\item \textbf{\texttt{spider}} -- 8 classes: 
	
	\begin{tabular}{p{0.5\linewidth}p{0.5\linewidth}}
		\inc{(n01770081, harvestman)} & \inc{(n01773157, black and gold garden spider)}\\
		\inc{(n01773549, barn spider)} & \inc{(n01773797, garden spider)}\\
		\inc{(n01774384, black widow)} & \inc{(n01774750, tarantula)}\\
		\inc{(n01775062, wolf spider)} & \inc{(n01776313, tick)}\\
	\end{tabular}
	\item \textbf{\texttt{marine\_invertebrate}} -- 7 classes: 
	
	\begin{tabular}{p{0.5\linewidth}p{0.5\linewidth}}
		\inc{(n01914609, sea anemone)} & \inc{(n01917289, brain coral)}\\
		\inc{(n01950731, sea slug)} & \inc{(n02317335, starfish)}\\
		\inc{(n02319095, sea urchin)} & \inc{(n09256479, coral reef)}\\
		\inc{(n12985857, coral fungus)} & \\
	\end{tabular}
	\item \textbf{\texttt{conch\_chambered\_nautilus}} -- 2 classes: 
	
	\begin{tabular}{p{0.5\linewidth}p{0.5\linewidth}}
		\inc{(n01943899, conch)} & \inc{(n01968897, chambered nautilus)}\\
	\end{tabular}
	\item \textbf{\texttt{invertebrate\_rest}} -- 7 classes: 
	
	\begin{tabular}{p{0.5\linewidth}p{0.5\linewidth}}
		\inc{(n01784675, centipede)} & \inc{(n01924916, flatworm)}\\
		\inc{(n01930112, nematode)} & \inc{(n01944390, snail)}\\
		\inc{(n01945685, slug)} & \inc{(n01950731, sea slug)}\\
		\inc{(n02321529, sea cucumber)} & \\
	\end{tabular}
	\item \textbf{\texttt{salamander\_lizard}} -- 16 classes: 
	
	\begin{tabular}{p{0.5\linewidth}p{0.5\linewidth}}
		\inc{(n01629819, European fire salamander)} & \inc{(n01630670, common newt)}\\
		\inc{(n01631663, eft)} & \inc{(n01632458, spotted salamander)}\\
		\inc{(n01632777, axolotl)} & \inc{(n01675722, banded gecko)}\\
		\inc{(n01677366, common iguana)} & \inc{(n01682714, American chameleon)}\\
		\inc{(n01685808, whiptail)} & \inc{(n01687978, agama)}\\
		\inc{(n01688243, frilled lizard)} & \inc{(n01689811, alligator lizard)}\\
		\inc{(n01692333, Gila monster)} & \inc{(n01693334, green lizard)}\\
		\inc{(n01694178, African chameleon)} & \inc{(n01695060, Komodo dragon)}\\
	\end{tabular}
	\item \textbf{\texttt{frog}} -- 3 classes: 
	
	\begin{tabular}{p{0.5\linewidth}p{0.5\linewidth}}
		\inc{(n01641577, bullfrog)} & \inc{(n01644373, tree frog)}\\
		\inc{(n01644900, tailed frog)} & \\
	\end{tabular}
	\item \textbf{\texttt{shark\_and\_aquatic\_mammal}} -- 7 classes: 
	
	\begin{tabular}{p{0.5\linewidth}p{0.5\linewidth}}
		\inc{(n01484850, great white shark)} & \inc{(n01491361, tiger shark)}\\
		\inc{(n01494475, hammerhead)} & \inc{(n02066245, grey whale)}\\
		\inc{(n02071294, killer whale)} & \inc{(n02074367, dugong)}\\
		\inc{(n02077923, sea lion)} & \\
	\end{tabular}
	\item \textbf{\texttt{ray}} -- 2 classes: 
	
	\begin{tabular}{p{0.5\linewidth}p{0.5\linewidth}}
		\inc{(n01496331, electric ray)} & \inc{(n01498041, stingray)}\\
	\end{tabular}
	\item \textbf{\texttt{fish\_rest}} -- 11 classes: 
	
	\begin{tabular}{p{0.5\linewidth}p{0.5\linewidth}}
		\inc{(n01440764, tench)} & \inc{(n01443537, goldfish)}\\
		\inc{(n02514041, barracouta)} & \inc{(n02526121, eel)}\\
		\inc{(n02536864, coho)} & \inc{(n02606052, rock beauty)}\\
		\inc{(n02607072, anemone fish)} & \inc{(n02640242, sturgeon)}\\
		\inc{(n02641379, gar)} & \inc{(n02643566, lionfish)}\\
		\inc{(n02655020, puffer)} & \\
	\end{tabular}
	\item \textbf{\texttt{turtle}} -- 5 classes: 
	
	\begin{tabular}{p{0.5\linewidth}p{0.5\linewidth}}
		\inc{(n01664065, loggerhead)} & \inc{(n01665541, leatherback turtle)}\\
		\inc{(n01667114, mud turtle)} & \inc{(n01667778, terrapin)}\\
		\inc{(n01669191, box turtle)} & \\
	\end{tabular}
	\item \textbf{\texttt{crocodilian\_reptile}} -- 2 classes: 
	
	\begin{tabular}{p{0.5\linewidth}p{0.5\linewidth}}
		\inc{(n01697457, African crocodile)} & \inc{(n01698640, American alligator)}\\
	\end{tabular}
	\item \textbf{\texttt{snake}} -- 17 classes: 
	
	\begin{tabular}{p{0.5\linewidth}p{0.5\linewidth}}
		\inc{(n01728572, thunder snake)} & \inc{(n01728920, ringneck snake)}\\
		\inc{(n01729322, hognose snake)} & \inc{(n01729977, green snake)}\\
		\inc{(n01734418, king snake)} & \inc{(n01735189, garter snake)}\\
		\inc{(n01737021, water snake)} & \inc{(n01739381, vine snake)}\\
		\inc{(n01740131, night snake)} & \inc{(n01742172, boa constrictor)}\\
		\inc{(n01744401, rock python)} & \inc{(n01748264, Indian cobra)}\\
		\inc{(n01749939, green mamba)} & \inc{(n01751748, sea snake)}\\
		\inc{(n01753488, horned viper)} & \inc{(n01755581, diamondback)}\\
		\inc{(n01756291, sidewinder)} & \\
	\end{tabular}
	\item \textbf{\texttt{green\_snake\_and\_lizards}} -- 5 classes: 
	
	\begin{tabular}{p{0.5\linewidth}p{0.5\linewidth}}
		\inc{(n01682714, American chameleon)} & \inc{(n01693334, green lizard)}\\
		\inc{(n01729977, green snake)} & \inc{(n01739381, vine snake)}\\
		\inc{(n01749939, green mamba)} & \\
	\end{tabular}
	\item \textbf{\texttt{non\_green\_snakes}} -- 15 classes: 
	
	\begin{tabular}{p{0.5\linewidth}p{0.5\linewidth}}
		\inc{(n01689811, alligator lizard)} & \inc{(n01728572, thunder snake)}\\
		\inc{(n01728920, ringneck snake)} & \inc{(n01729322, hognose snake)}\\
		\inc{(n01734418, king snake)} & \inc{(n01735189, garter snake)}\\
		\inc{(n01737021, water snake)} & \inc{(n01740131, night snake)}\\
		\inc{(n01742172, boa constrictor)} & \inc{(n01744401, rock python)}\\
		\inc{(n01748264, Indian cobra)} & \inc{(n01751748, sea snake)}\\
		\inc{(n01753488, horned viper)} & \inc{(n01755581, diamondback)}\\
		\inc{(n01756291, sidewinder)} & \\
	\end{tabular}
	\item \textbf{\texttt{cock\_hen\_gallinaceous}} -- 12 classes: 
	
	\begin{tabular}{p{0.5\linewidth}p{0.5\linewidth}}
		\inc{(n01514668, cock)} & \inc{(n01514859, hen)}\\
		\inc{(n01795545, black grouse)} & \inc{(n01796340, ptarmigan)}\\
		\inc{(n01797886, ruffed grouse)} & \inc{(n01798484, prairie chicken)}\\
		\inc{(n01806143, peacock)} & \inc{(n01806567, quail)}\\
		\inc{(n01807496, partridge)} & \inc{(n02017213, European gallinule)}\\
		\inc{(n02018207, American coot)} & \inc{(n02018795, bustard)}\\
	\end{tabular}
	\item \textbf{\texttt{small\_bird}} -- 15 classes: 
	
	\begin{tabular}{p{0.5\linewidth}p{0.5\linewidth}}
		\inc{(n01530575, brambling)} & \inc{(n01531178, goldfinch)}\\
		\inc{(n01532829, house finch)} & \inc{(n01534433, junco)}\\
		\inc{(n01537544, indigo bunting)} & \inc{(n01558993, robin)}\\
		\inc{(n01560419, bulbul)} & \inc{(n01580077, jay)}\\
		\inc{(n01582220, magpie)} & \inc{(n01592084, chickadee)}\\
		\inc{(n01601694, water ouzel)} & \inc{(n01824575, coucal)}\\
		\inc{(n01828970, bee eater)} & \inc{(n01833805, hummingbird)}\\
		\inc{(n01843065, jacamar)} & \\
	\end{tabular}
	\item \textbf{\texttt{bird\_of\_prey}} -- 3 classes: 
	
	\begin{tabular}{p{0.5\linewidth}p{0.5\linewidth}}
		\inc{(n01608432, kite)} & \inc{(n01614925, bald eagle)}\\
		\inc{(n01616318, vulture)} & \\
	\end{tabular}
	\item \textbf{\texttt{parrot}} -- 4 classes: 
	
	\begin{tabular}{p{0.5\linewidth}p{0.5\linewidth}}
		\inc{(n01817953, African grey)} & \inc{(n01818515, macaw)}\\
		\inc{(n01819313, sulphur-crested cockatoo)} & \inc{(n01820546, lorikeet)}\\
	\end{tabular}
	\item \textbf{\texttt{big\_beak\_bird}} -- 2 classes: 
	
	\begin{tabular}{p{0.5\linewidth}p{0.5\linewidth}}
		\inc{(n01829413, hornbill)} & \inc{(n01843383, toucan)}\\
	\end{tabular}
	\item \textbf{\texttt{aquatic\_bird\_and\_ostrich}} -- 25 classes: 
	
	\begin{tabular}{p{0.5\linewidth}p{0.5\linewidth}}
		\inc{(n01518878, ostrich)} & \inc{(n01847000, drake)}\\
		\inc{(n01855032, red-breasted merganser)} & \inc{(n01855672, goose)}\\
		\inc{(n01860187, black swan)} & \inc{(n02002556, white stork)}\\
		\inc{(n02002724, black stork)} & \inc{(n02006656, spoonbill)}\\
		\inc{(n02007558, flamingo)} & \inc{(n02009229, little blue heron)}\\
		\inc{(n02009912, American egret)} & \inc{(n02011460, bittern)}\\
		\inc{(n02012849, crane)} & \inc{(n02013706, limpkin)}\\
		\inc{(n02017213, European gallinule)} & \inc{(n02018207, American coot)}\\
		\inc{(n02018795, bustard)} & \inc{(n02025239, ruddy turnstone)}\\
		\inc{(n02027492, red-backed sandpiper)} & \inc{(n02028035, redshank)}\\
		\inc{(n02033041, dowitcher)} & \inc{(n02037110, oystercatcher)}\\
		\inc{(n02051845, pelican)} & \inc{(n02056570, king penguin)}\\
		\inc{(n02058221, albatross)} & \\
	\end{tabular}
	\item \textbf{\texttt{tusker\_elephant}} -- 3 classes: 
	
	\begin{tabular}{p{0.5\linewidth}p{0.5\linewidth}}
		\inc{(n01871265, tusker)} & \inc{(n02504013, Indian elephant)}\\
		\inc{(n02504458, African elephant)} & \\
	\end{tabular}
	\item \textbf{\texttt{primate}} -- 20 classes: 
	
	\begin{tabular}{p{0.5\linewidth}p{0.5\linewidth}}
		\inc{(n02480495, orangutan)} & \inc{(n02480855, gorilla)}\\
		\inc{(n02481823, chimpanzee)} & \inc{(n02483362, gibbon)}\\
		\inc{(n02483708, siamang)} & \inc{(n02484975, guenon)}\\
		\inc{(n02486261, patas)} & \inc{(n02486410, baboon)}\\
		\inc{(n02487347, macaque)} & \inc{(n02488291, langur)}\\
		\inc{(n02488702, colobus)} & \inc{(n02489166, proboscis monkey)}\\
		\inc{(n02490219, marmoset)} & \inc{(n02492035, capuchin)}\\
		\inc{(n02492660, howler monkey)} & \inc{(n02493509, titi)}\\
		\inc{(n02493793, spider monkey)} & \inc{(n02494079, squirrel monkey)}\\
		\inc{(n02497673, Madagascar cat)} & \inc{(n02500267, indri)}\\
	\end{tabular}
	\item \textbf{\texttt{bear\_panda\_marsupial}} -- 10 classes: 
	
	\begin{tabular}{p{0.5\linewidth}p{0.5\linewidth}}
		\inc{(n01877812, wallaby)} & \inc{(n01882714, koala)}\\
		\inc{(n01883070, wombat)} & \inc{(n02132136, brown bear)}\\
		\inc{(n02133161, American black bear)} & \inc{(n02134084, ice bear)}\\
		\inc{(n02134418, sloth bear)} & \inc{(n02457408, three-toed sloth)}\\
		\inc{(n02509815, lesser panda)} & \inc{(n02510455, giant panda)}\\
	\end{tabular}
	\item \textbf{\texttt{viverrine\_musteline\_rodent\_prototherian}} -- 19 classes: 
	
	\begin{tabular}{p{0.5\linewidth}p{0.5\linewidth}}
		\inc{(n01872401, echidna)} & \inc{(n01873310, platypus)}\\
		\inc{(n01877812, wallaby)} & \inc{(n02137549, mongoose)}\\
		\inc{(n02138441, meerkat)} & \inc{(n02342885, hamster)}\\
		\inc{(n02346627, porcupine)} & \inc{(n02356798, fox squirrel)}\\
		\inc{(n02361337, marmot)} & \inc{(n02363005, beaver)}\\
		\inc{(n02364673, guinea pig)} & \inc{(n02441942, weasel)}\\
		\inc{(n02442845, mink)} & \inc{(n02443114, polecat)}\\
		\inc{(n02443484, black-footed ferret)} & \inc{(n02444819, otter)}\\
		\inc{(n02445715, skunk)} & \inc{(n02447366, badger)}\\
		\inc{(n02454379, armadillo)} & \\
	\end{tabular}
	\item \textbf{\texttt{rabbit}} -- 3 classes: 
	
	\begin{tabular}{p{0.5\linewidth}p{0.5\linewidth}}
		\inc{(n02325366, wood rabbit)} & \inc{(n02326432, hare)}\\
		\inc{(n02328150, Angora)} & \\
	\end{tabular}
	\item \textbf{\texttt{ungulate}} -- 17 classes: 
	
	\begin{tabular}{p{0.5\linewidth}p{0.5\linewidth}}
		\inc{(n02389026, sorrel)} & \inc{(n02391049, zebra)}\\
		\inc{(n02395406, hog)} & \inc{(n02396427, wild boar)}\\
		\inc{(n02397096, warthog)} & \inc{(n02398521, hippopotamus)}\\
		\inc{(n02403003, ox)} & \inc{(n02408429, water buffalo)}\\
		\inc{(n02410509, bison)} & \inc{(n02412080, ram)}\\
		\inc{(n02415577, bighorn)} & \inc{(n02417914, ibex)}\\
		\inc{(n02422106, hartebeest)} & \inc{(n02422699, impala)}\\
		\inc{(n02423022, gazelle)} & \inc{(n02437312, Arabian camel)}\\
		\inc{(n02437616, llama)} & \\
	\end{tabular}
	\item \textbf{\texttt{domestic\_cat}} -- 5 classes: 
	
	\begin{tabular}{p{0.5\linewidth}p{0.5\linewidth}}
		\inc{(n02123045, tabby)} & \inc{(n02123159, tiger cat)}\\
		\inc{(n02123394, Persian cat)} & \inc{(n02123597, Siamese cat)}\\
		\inc{(n02124075, Egyptian cat)} & \\
	\end{tabular}
	\item \textbf{\texttt{wild\_cat\_big\_cat}} -- 8 classes: 
	
	\begin{tabular}{p{0.5\linewidth}p{0.5\linewidth}}
		\inc{(n02125311, cougar)} & \inc{(n02127052, lynx)}\\
		\inc{(n02128385, leopard)} & \inc{(n02128757, snow leopard)}\\
		\inc{(n02128925, jaguar)} & \inc{(n02129165, lion)}\\
		\inc{(n02129604, tiger)} & \inc{(n02130308, cheetah)}\\
	\end{tabular}
	\item \textbf{\texttt{wolf\_wild\_dog\_hyena\_fox}} -- 12 classes: 
	
	\begin{tabular}{p{0.5\linewidth}p{0.5\linewidth}}
		\inc{(n02114367, timber wolf)} & \inc{(n02114548, white wolf)}\\
		\inc{(n02114712, red wolf)} & \inc{(n02114855, coyote)}\\
		\inc{(n02115641, dingo)} & \inc{(n02115913, dhole)}\\
		\inc{(n02116738, African hunting dog)} & \inc{(n02117135, hyena)}\\
		\inc{(n02119022, red fox)} & \inc{(n02119789, kit fox)}\\
		\inc{(n02120079, Arctic fox)} & \inc{(n02120505, grey fox)}\\
	\end{tabular}
	\item \textbf{\texttt{pug\_boxer\_bulldog}} -- 9 classes: 
	
	\begin{tabular}{p{0.5\linewidth}p{0.5\linewidth}}
		\inc{(n02093256, Staffordshire bullterrier)} & \inc{(n02093428, American Staffordshire terrier)}\\
		\inc{(n02096585, Boston bull)} & \inc{(n02106550, Rottweiler)}\\
		\inc{(n02108089, boxer)} & \inc{(n02108422, bull mastiff)}\\
		\inc{(n02108915, French bulldog)} & \inc{(n02110958, pug)}\\
		\inc{(n02112706, Brabancon griffon)} & \\
	\end{tabular}
	\item \textbf{\texttt{toy\_dog}} -- 31 classes: 
	
	\begin{tabular}{p{0.5\linewidth}p{0.5\linewidth}}
		\inc{(n02085620, Chihuahua)} & \inc{(n02085782, Japanese spaniel)}\\
		\inc{(n02085936, Maltese dog)} & \inc{(n02086079, Pekinese)}\\
		\inc{(n02086240, Shih-Tzu)} & \inc{(n02086646, Blenheim spaniel)}\\
		\inc{(n02086910, papillon)} & \inc{(n02087046, toy terrier)}\\
		\inc{(n02093754, Border terrier)} & \inc{(n02094114, Norfolk terrier)}\\
		\inc{(n02094258, Norwich terrier)} & \inc{(n02094433, Yorkshire terrier)}\\
		\inc{(n02096177, cairn)} & \inc{(n02096294, Australian terrier)}\\
		\inc{(n02096437, Dandie Dinmont)} & \inc{(n02096585, Boston bull)}\\
		\inc{(n02097047, miniature schnauzer)} & \inc{(n02097474, Tibetan terrier)}\\
		\inc{(n02097658, silky terrier)} & \inc{(n02098105, soft-coated wheaten terrier)}\\
		\inc{(n02098286, West Highland white terrier)} & \inc{(n02098413, Lhasa)}\\
		\inc{(n02102318, cocker spaniel)} & \inc{(n02107312, miniature pinscher)}\\
		\inc{(n02108915, French bulldog)} & \inc{(n02110627, affenpinscher)}\\
		\inc{(n02112018, Pomeranian)} & \inc{(n02112706, Brabancon griffon)}\\
		\inc{(n02113624, toy poodle)} & \inc{(n02113712, miniature poodle)}\\
		\inc{(n02113799, standard poodle)} & \\
	\end{tabular}
	\item \textbf{\texttt{corgi}} -- 2 classes: 
	
	\begin{tabular}{p{0.5\linewidth}p{0.5\linewidth}}
		\inc{(n02113023, Pembroke)} & \inc{(n02113186, Cardigan)}\\
	\end{tabular}
	\item \textbf{\texttt{terrier}} -- 26 classes: 
	
	\begin{tabular}{p{0.5\linewidth}p{0.5\linewidth}}
		\inc{(n02090721, Irish wolfhound)} & \inc{(n02091635, otterhound)}\\
		\inc{(n02093647, Bedlington terrier)} & \inc{(n02093754, Border terrier)}\\
		\inc{(n02093859, Kerry blue terrier)} & \inc{(n02093991, Irish terrier)}\\
		\inc{(n02094114, Norfolk terrier)} & \inc{(n02094258, Norwich terrier)}\\
		\inc{(n02094433, Yorkshire terrier)} & \inc{(n02095314, wire-haired fox terrier)}\\
		\inc{(n02095570, Lakeland terrier)} & \inc{(n02095889, Sealyham terrier)}\\
		\inc{(n02096051, Airedale)} & \inc{(n02096177, cairn)}\\
		\inc{(n02096294, Australian terrier)} & \inc{(n02096437, Dandie Dinmont)}\\
		\inc{(n02097047, miniature schnauzer)} & \inc{(n02097130, giant schnauzer)}\\
		\inc{(n02097209, standard schnauzer)} & \inc{(n02097298, Scotch terrier)}\\
		\inc{(n02097474, Tibetan terrier)} & \inc{(n02097658, silky terrier)}\\
		\inc{(n02098105, soft-coated wheaten terrier)} & \inc{(n02098286, West Highland white terrier)}\\
		\inc{(n02098413, Lhasa)} & \inc{(n02110627, affenpinscher)}\\
	\end{tabular}
	\item \textbf{\texttt{poodle\_like}} -- 7 classes: 
	
	\begin{tabular}{p{0.5\linewidth}p{0.5\linewidth}}
		\inc{(n02093647, Bedlington terrier)} & \inc{(n02093859, Kerry blue terrier)}\\
		\inc{(n02099429, curly-coated retriever)} & \inc{(n02102973, Irish water spaniel)}\\
		\inc{(n02113624, toy poodle)} & \inc{(n02113712, miniature poodle)}\\
		\inc{(n02113799, standard poodle)} & \\
	\end{tabular}
	\item \textbf{\texttt{standard\_shepherd\_dog}} -- 11 classes: 
	
	\begin{tabular}{p{0.5\linewidth}p{0.5\linewidth}}
		\inc{(n02091467, Norwegian elkhound)} & \inc{(n02104365, schipperke)}\\
		\inc{(n02105056, groenendael)} & \inc{(n02105162, malinois)}\\
		\inc{(n02105412, kelpie)} & \inc{(n02105855, Shetland sheepdog)}\\
		\inc{(n02106030, collie)} & \inc{(n02106166, Border collie)}\\
		\inc{(n02106550, Rottweiler)} & \inc{(n02106662, German shepherd)}\\
		\inc{(n02112350, keeshond)} & \\
	\end{tabular}
	\item \textbf{\texttt{hairy\_shepherd\_dog}} -- 6 classes: 
	
	\begin{tabular}{p{0.5\linewidth}p{0.5\linewidth}}
		\inc{(n02097474, Tibetan terrier)} & \inc{(n02105251, briard)}\\
		\inc{(n02105505, komondor)} & \inc{(n02105641, Old English sheepdog)}\\
		\inc{(n02106382, Bouvier des Flandres)} & \inc{(n02110627, affenpinscher)}\\
	\end{tabular}
	\item \textbf{\texttt{beagle\_hound\_pointer\_spaniel\_setter}} -- 27 classes: 
	
	\begin{tabular}{p{0.5\linewidth}p{0.5\linewidth}}
		\inc{(n02087394, Rhodesian ridgeback)} & \inc{(n02088238, basset)}\\
		\inc{(n02088364, beagle)} & \inc{(n02088466, bloodhound)}\\
		\inc{(n02088632, bluetick)} & \inc{(n02089078, black-and-tan coonhound)}\\
		\inc{(n02089867, Walker hound)} & \inc{(n02089973, English foxhound)}\\
		\inc{(n02090379, redbone)} & \inc{(n02092339, Weimaraner)}\\
		\inc{(n02100236, German short-haired pointer)} & \inc{(n02100583, vizsla)}\\
		\inc{(n02100735, English setter)} & \inc{(n02100877, Irish setter)}\\
		\inc{(n02101006, Gordon setter)} & \inc{(n02101388, Brittany spaniel)}\\
		\inc{(n02101556, clumber)} & \inc{(n02102040, English springer)}\\
		\inc{(n02102177, Welsh springer spaniel)} & \inc{(n02102318, cocker spaniel)}\\
		\inc{(n02102480, Sussex spaniel)} & \inc{(n02102973, Irish water spaniel)}\\
		\inc{(n02107142, Doberman)} & \inc{(n02109047, Great Dane)}\\
		\inc{(n02110341, dalmatian)} & \inc{(n02110806, basenji)}\\
		\inc{(n02113978, Mexican hairless)} & \\
	\end{tabular}
	\item \textbf{\texttt{spaniels}} -- 5 classes: 
	
	\begin{tabular}{p{0.5\linewidth}p{0.5\linewidth}}
		\inc{(n02086646, Blenheim spaniel)} & \inc{(n02101388, Brittany spaniel)}\\
		\inc{(n02102040, English springer)} & \inc{(n02102177, Welsh springer spaniel)}\\
		\inc{(n02102318, cocker spaniel)} & \\
	\end{tabular}
	\item \textbf{\texttt{skinny\_hound\_rest}} -- 12 classes: 
	
	\begin{tabular}{p{0.5\linewidth}p{0.5\linewidth}}
		\inc{(n02088094, Afghan hound)} & \inc{(n02090622, borzoi)}\\
		\inc{(n02090721, Irish wolfhound)} & \inc{(n02091032, Italian greyhound)}\\
		\inc{(n02091134, whippet)} & \inc{(n02091244, Ibizan hound)}\\
		\inc{(n02091635, otterhound)} & \inc{(n02091831, Saluki)}\\
		\inc{(n02092002, Scottish deerhound)} & \inc{(n02107142, Doberman)}\\
		\inc{(n02110806, basenji)} & \inc{(n02113978, Mexican hairless)}\\
	\end{tabular}
	\item \textbf{\texttt{pinscher}} -- 3 classes: 
	
	\begin{tabular}{p{0.5\linewidth}p{0.5\linewidth}}
		\inc{(n02106550, Rottweiler)} & \inc{(n02107142, Doberman)}\\
		\inc{(n02107312, miniature pinscher)} & \\
	\end{tabular}
	\item \textbf{\texttt{sennenhunde\_and\_big\_dogs}} -- 15 classes: 
	
	\begin{tabular}{p{0.5\linewidth}p{0.5\linewidth}}
		\inc{(n02088364, beagle)} & \inc{(n02106166, Border collie)}\\
		\inc{(n02106550, Rottweiler)} & \inc{(n02107574, Greater Swiss Mountain dog)}\\
		\inc{(n02107683, Bernese mountain dog)} & \inc{(n02107908, Appenzeller)}\\
		\inc{(n02108000, EntleBucher)} & \inc{(n02108089, boxer)}\\
		\inc{(n02108422, bull mastiff)} & \inc{(n02108551, Tibetan mastiff)}\\
		\inc{(n02109047, Great Dane)} & \inc{(n02109525, Saint Bernard)}\\
		\inc{(n02111129, Leonberg)} & \inc{(n02111277, Newfoundland)}\\
		\inc{(n02112137, chow)} & \\
	\end{tabular}
	\item \textbf{\texttt{retriever\_like}} -- 10 classes: 
	
	\begin{tabular}{p{0.5\linewidth}p{0.5\linewidth}}
		\inc{(n02090379, redbone)} & \inc{(n02099267, flat-coated retriever)}\\
		\inc{(n02099429, curly-coated retriever)} & \inc{(n02099601, golden retriever)}\\
		\inc{(n02099712, Labrador retriever)} & \inc{(n02099849, Chesapeake Bay retriever)}\\
		\inc{(n02101006, Gordon setter)} & \inc{(n02104029, kuvasz)}\\
		\inc{(n02111277, Newfoundland)} & \inc{(n02111500, Great Pyrenees)}\\
	\end{tabular}
	\item \textbf{\texttt{husky}} -- 6 classes: 
	
	\begin{tabular}{p{0.5\linewidth}p{0.5\linewidth}}
		\inc{(n02091467, Norwegian elkhound)} & \inc{(n02109961, Eskimo dog)}\\
		\inc{(n02110063, malamute)} & \inc{(n02110185, Siberian husky)}\\
		\inc{(n02111889, Samoyed)} & \inc{(n02112350, keeshond)}\\
	\end{tabular}
	\item \textbf{\texttt{spitz}} -- 4 classes: 
	
	\begin{tabular}{p{0.5\linewidth}p{0.5\linewidth}}
		\inc{(n02111889, Samoyed)} & \inc{(n02112018, Pomeranian)}\\
		\inc{(n02112137, chow)} & \inc{(n02112350, keeshond)}\\
	\end{tabular}
	\item \textbf{\texttt{fungi}} -- 7 classes: 
	
	\begin{tabular}{p{0.5\linewidth}p{0.5\linewidth}}
		\inc{(n07734744, mushroom)} & \inc{(n12998815, agaric)}\\
		\inc{(n13037406, gyromitra)} & \inc{(n13040303, stinkhorn)}\\
		\inc{(n13044778, earthstar)} & \inc{(n13052670, hen-of-the-woods)}\\
		\inc{(n13054560, bolete)} & \\
	\end{tabular}
	\item \textbf{\texttt{geological\_formation}} -- 9 classes: 
	
	\begin{tabular}{p{0.5\linewidth}p{0.5\linewidth}}
		\inc{(n09193705, alp)} & \inc{(n09246464, cliff)}\\
		\inc{(n09288635, geyser)} & \inc{(n09332890, lakeside)}\\
		\inc{(n09399592, promontory)} & \inc{(n09421951, sandbar)}\\
		\inc{(n09428293, seashore)} & \inc{(n09468604, valley)}\\
		\inc{(n09472597, volcano)} & \\
	\end{tabular}
	\item \textbf{\texttt{bread}} -- 4 classes: 
	
	\begin{tabular}{p{0.5\linewidth}p{0.5\linewidth}}
		\inc{(n07684084, French loaf)} & \inc{(n07693725, bagel)}\\
		\inc{(n07695742, pretzel)} & \inc{(n07860988, dough)}\\
	\end{tabular}
	\item \textbf{\texttt{cruciferous\_vegetable}} -- 3 classes: 
	
	\begin{tabular}{p{0.5\linewidth}p{0.5\linewidth}}
		\inc{(n07714571, head cabbage)} & \inc{(n07714990, broccoli)}\\
		\inc{(n07715103, cauliflower)} & \\
	\end{tabular}
	\item \textbf{\texttt{cauliflower\_custard\_apple}} -- 2 classes: 
	
	\begin{tabular}{p{0.5\linewidth}p{0.5\linewidth}}
		\inc{(n07715103, cauliflower)} & \inc{(n07760859, custard apple)}\\
	\end{tabular}
	\item \textbf{\texttt{zucchini\_cucumber}} -- 2 classes: 
	
	\begin{tabular}{p{0.5\linewidth}p{0.5\linewidth}}
		\inc{(n07716358, zucchini)} & \inc{(n07718472, cucumber)}\\
	\end{tabular}
	\item \textbf{\texttt{squash}} -- 4 classes: 
	
	\begin{tabular}{p{0.5\linewidth}p{0.5\linewidth}}
		\inc{(n07716358, zucchini)} & \inc{(n07716906, spaghetti squash)}\\
		\inc{(n07717410, acorn squash)} & \inc{(n07717556, butternut squash)}\\
	\end{tabular}
	\item \textbf{\texttt{reproductive\_structure}} -- 12 classes: 
	
	\begin{tabular}{p{0.5\linewidth}p{0.5\linewidth}}
		\inc{(n07720875, bell pepper)} & \inc{(n07742313, Granny Smith)}\\
		\inc{(n07745940, strawberry)} & \inc{(n07747607, orange)}\\
		\inc{(n07749582, lemon)} & \inc{(n07753113, fig)}\\
		\inc{(n07754684, jackfruit)} & \inc{(n07760859, custard apple)}\\
		\inc{(n07768694, pomegranate)} & \inc{(n12267677, acorn)}\\
		\inc{(n12620546, hip)} & \inc{(n12768682, buckeye)}\\
	\end{tabular}
	\item \textbf{\texttt{corn}} -- 2 classes: 
	
	\begin{tabular}{p{0.5\linewidth}p{0.5\linewidth}}
		\inc{(n12144580, corn)} & \inc{(n13133613, ear)}\\
	\end{tabular}
	\item \textbf{\texttt{cardoon\_artichoke}} -- 3 classes: 
	
	\begin{tabular}{p{0.5\linewidth}p{0.5\linewidth}}
		\inc{(n07718747, artichoke)} & \inc{(n07730033, cardoon)}\\
		\inc{(n07753275, pineapple)} & \\
	\end{tabular}
	\item \textbf{\texttt{dessert}} -- 3 classes: 
	
	\begin{tabular}{p{0.5\linewidth}p{0.5\linewidth}}
		\inc{(n07613480, trifle)} & \inc{(n07614500, ice cream)}\\
		\inc{(n07836838, chocolate sauce)} & \\
	\end{tabular}
	\item \textbf{\texttt{baby\_bed}} -- 3 classes: 
	
	\begin{tabular}{p{0.5\linewidth}p{0.5\linewidth}}
		\inc{(n02804414, bassinet)} & \inc{(n03125729, cradle)}\\
		\inc{(n03131574, crib)} & \\
	\end{tabular}
	\item \textbf{\texttt{chair}} -- 4 classes: 
	
	\begin{tabular}{p{0.5\linewidth}p{0.5\linewidth}}
		\inc{(n02791124, barber chair)} & \inc{(n03376595, folding chair)}\\
		\inc{(n04099969, rocking chair)} & \inc{(n04429376, throne)}\\
	\end{tabular}
	\item \textbf{\texttt{entertainment\_center\_home\_theatre}} -- 2 classes: 
	
	\begin{tabular}{p{0.5\linewidth}p{0.5\linewidth}}
		\inc{(n03290653, entertainment center)} & \inc{(n03529860, home theater)}\\
	\end{tabular}
	\item \textbf{\texttt{cabinet}} -- 7 classes: 
	
	\begin{tabular}{p{0.5\linewidth}p{0.5\linewidth}}
		\inc{(n02870880, bookcase)} & \inc{(n03016953, chiffonier)}\\
		\inc{(n03018349, china cabinet)} & \inc{(n03290653, entertainment center)}\\
		\inc{(n03337140, file)} & \inc{(n03742115, medicine chest)}\\
		\inc{(n04550184, wardrobe)} & \\
	\end{tabular}
	\item \textbf{\texttt{table}} -- 2 classes: 
	
	\begin{tabular}{p{0.5\linewidth}p{0.5\linewidth}}
		\inc{(n03179701, desk)} & \inc{(n03201208, dining table)}\\
	\end{tabular}
	\item \textbf{\texttt{robe\_overgarment\_cloak}} -- 10 classes: 
	
	\begin{tabular}{p{0.5\linewidth}p{0.5\linewidth}}
		\inc{(n02667093, abaya)} & \inc{(n02669723, academic gown)}\\
		\inc{(n02730930, apron)} & \inc{(n03045698, cloak)}\\
		\inc{(n03404251, fur coat)} & \inc{(n03617480, kimono)}\\
		\inc{(n03630383, lab coat)} & \inc{(n03980874, poncho)}\\
		\inc{(n04479046, trench coat)} & \inc{(n04532106, vestment)}\\
	\end{tabular}
	\item \textbf{\texttt{sweater}} -- 3 classes: 
	
	\begin{tabular}{p{0.5\linewidth}p{0.5\linewidth}}
		\inc{(n02963159, cardigan)} & \inc{(n03980874, poncho)}\\
		\inc{(n04370456, sweatshirt)} & \\
	\end{tabular}
	\item \textbf{\texttt{suits}} -- 3 classes: 
	
	\begin{tabular}{p{0.5\linewidth}p{0.5\linewidth}}
		\inc{(n03763968, military uniform)} & \inc{(n03877472, pajama)}\\
		\inc{(n04350905, suit)} & \\
	\end{tabular}
	\item \textbf{\texttt{woolen\_pieces}} -- 7 classes: 
	
	\begin{tabular}{p{0.5\linewidth}p{0.5\linewidth}}
		\inc{(n02963159, cardigan)} & \inc{(n03207743, dishrag)}\\
		\inc{(n03775071, mitten)} & \inc{(n03980874, poncho)}\\
		\inc{(n04229816, ski mask)} & \inc{(n04325704, stole)}\\
		\inc{(n04599235, wool)} & \\
	\end{tabular}
	\item \textbf{\texttt{caps}} -- 3 classes: 
	
	\begin{tabular}{p{0.5\linewidth}p{0.5\linewidth}}
		\inc{(n02807133, bathing cap)} & \inc{(n02869837, bonnet)}\\
		\inc{(n04209133, shower cap)} & \\
	\end{tabular}
	\item \textbf{\texttt{swimsuit}} -- 4 classes: 
	
	\begin{tabular}{p{0.5\linewidth}p{0.5\linewidth}}
		\inc{(n02837789, bikini)} & \inc{(n02892767, brassiere)}\\
		\inc{(n03710637, maillot)} & \inc{(n03710721, maillot)}\\
	\end{tabular}
	\item \textbf{\texttt{woman\_dress}} -- 3 classes: 
	
	\begin{tabular}{p{0.5\linewidth}p{0.5\linewidth}}
		\inc{(n03450230, gown)} & \inc{(n03534580, hoopskirt)}\\
		\inc{(n03866082, overskirt)} & \\
	\end{tabular}
	\item \textbf{\texttt{neckwear}} -- 4 classes: 
	
	\begin{tabular}{p{0.5\linewidth}p{0.5\linewidth}}
		\inc{(n02865351, bolo tie)} & \inc{(n02883205, bow tie)}\\
		\inc{(n03814906, necklace)} & \inc{(n04591157, Windsor tie)}\\
	\end{tabular}
	\item \textbf{\texttt{sock\_and\_glove}} -- 3 classes: 
	
	\begin{tabular}{p{0.5\linewidth}p{0.5\linewidth}}
		\inc{(n03026506, Christmas stocking)} & \inc{(n03775071, mitten)}\\
		\inc{(n04254777, sock)} & \\
	\end{tabular}
	\item \textbf{\texttt{breathing\_device}} -- 3 classes: 
	
	\begin{tabular}{p{0.5\linewidth}p{0.5\linewidth}}
		\inc{(n03424325, gasmask)} & \inc{(n03868863, oxygen mask)}\\
		\inc{(n04251144, snorkel)} & \\
	\end{tabular}
	\item \textbf{\texttt{helmet}} -- 3 classes: 
	
	\begin{tabular}{p{0.5\linewidth}p{0.5\linewidth}}
		\inc{(n03127747, crash helmet)} & \inc{(n03379051, football helmet)}\\
		\inc{(n03929855, pickelhaube)} & \\
	\end{tabular}
	\item \textbf{\texttt{pickelhaube\_bearskin}} -- 2 classes: 
	
	\begin{tabular}{p{0.5\linewidth}p{0.5\linewidth}}
		\inc{(n02817516, bearskin)} & \inc{(n03929855, pickelhaube)}\\
	\end{tabular}
	\item \textbf{\texttt{hat}} -- 3 classes: 
	
	\begin{tabular}{p{0.5\linewidth}p{0.5\linewidth}}
		\inc{(n02869837, bonnet)} & \inc{(n03124170, cowboy hat)}\\
		\inc{(n04259630, sombrero)} & \\
	\end{tabular}
	\item \textbf{\texttt{armor}} -- 4 classes: 
	
	\begin{tabular}{p{0.5\linewidth}p{0.5\linewidth}}
		\inc{(n02895154, breastplate)} & \inc{(n03000247, chain mail)}\\
		\inc{(n03146219, cuirass)} & \inc{(n04192698, shield)}\\
	\end{tabular}
	\item \textbf{\texttt{cuirass\_bulletproof\_vest}} -- 2 classes: 
	
	\begin{tabular}{p{0.5\linewidth}p{0.5\linewidth}}
		\inc{(n02916936, bulletproof vest)} & \inc{(n03146219, cuirass)}\\
	\end{tabular}
	\item \textbf{\texttt{footwear\_rest}} -- 5 classes: 
	
	\begin{tabular}{p{0.5\linewidth}p{0.5\linewidth}}
		\inc{(n03047690, clog)} & \inc{(n03124043, cowboy boot)}\\
		\inc{(n03680355, Loafer)} & \inc{(n04120489, running shoe)}\\
		\inc{(n04133789, sandal)} & \\
	\end{tabular}
	\item \textbf{\texttt{top\_cover}} -- 4 classes: 
	
	\begin{tabular}{p{0.5\linewidth}p{0.5\linewidth}}
		\inc{(n02877765, bottlecap)} & \inc{(n03657121, lens cap)}\\
		\inc{(n03717622, manhole cover)} & \inc{(n04019541, puck)}\\
	\end{tabular}
	\item \textbf{\texttt{lamp\_light}} -- 5 classes: 
	
	\begin{tabular}{p{0.5\linewidth}p{0.5\linewidth}}
		\inc{(n02948072, candle)} & \inc{(n03590841, jack-o'-lantern)}\\
		\inc{(n03637318, lampshade)} & \inc{(n04286575, spotlight)}\\
		\inc{(n04380533, table lamp)} & \\
	\end{tabular}
	\item \textbf{\texttt{candle\_torch}} -- 2 classes: 
	
	\begin{tabular}{p{0.5\linewidth}p{0.5\linewidth}}
		\inc{(n02948072, candle)} & \inc{(n04456115, torch)}\\
	\end{tabular}
	\item \textbf{\texttt{ship\_boat}} -- 15 classes: 
	
	\begin{tabular}{p{0.5\linewidth}p{0.5\linewidth}}
		\inc{(n02687172, aircraft carrier)} & \inc{(n02951358, canoe)}\\
		\inc{(n02981792, catamaran)} & \inc{(n03095699, container ship)}\\
		\inc{(n03344393, fireboat)} & \inc{(n03447447, gondola)}\\
		\inc{(n03662601, lifeboat)} & \inc{(n03673027, liner)}\\
		\inc{(n03947888, pirate)} & \inc{(n04147183, schooner)}\\
		\inc{(n04273569, speedboat)} & \inc{(n04347754, submarine)}\\
		\inc{(n04483307, trimaran)} & \inc{(n04606251, wreck)}\\
		\inc{(n04612504, yawl)} & \\
	\end{tabular}
	\item \textbf{\texttt{tank}} -- 3 classes: 
	
	\begin{tabular}{p{0.5\linewidth}p{0.5\linewidth}}
		\inc{(n02704792, amphibian)} & \inc{(n03478589, half track)}\\
		\inc{(n04389033, tank)} & \\
	\end{tabular}
	\item \textbf{\texttt{snow\_sled}} -- 3 classes: 
	
	\begin{tabular}{p{0.5\linewidth}p{0.5\linewidth}}
		\inc{(n02860847, bobsled)} & \inc{(n03218198, dogsled)}\\
		\inc{(n04252077, snowmobile)} & \\
	\end{tabular}
	\item \textbf{\texttt{motor\_cycle}} -- 6 classes: 
	
	\begin{tabular}{p{0.5\linewidth}p{0.5\linewidth}}
		\inc{(n02835271, bicycle-built-for-two)} & \inc{(n03785016, moped)}\\
		\inc{(n03791053, motor scooter)} & \inc{(n03792782, mountain bike)}\\
		\inc{(n04482393, tricycle)} & \inc{(n04509417, unicycle)}\\
	\end{tabular}
	\item \textbf{\texttt{aircraft}} -- 3 classes: 
	
	\begin{tabular}{p{0.5\linewidth}p{0.5\linewidth}}
		\inc{(n02690373, airliner)} & \inc{(n02692877, airship)}\\
		\inc{(n04552348, warplane)} & \\
	\end{tabular}
	\item \textbf{\texttt{balloon}} -- 3 classes: 
	
	\begin{tabular}{p{0.5\linewidth}p{0.5\linewidth}}
		\inc{(n02692877, airship)} & \inc{(n02782093, balloon)}\\
		\inc{(n03888257, parachute)} & \\
	\end{tabular}
	\item \textbf{\texttt{train}} -- 6 classes: 
	
	\begin{tabular}{p{0.5\linewidth}p{0.5\linewidth}}
		\inc{(n02917067, bullet train)} & \inc{(n03272562, electric locomotive)}\\
		\inc{(n03393912, freight car)} & \inc{(n03895866, passenger car)}\\
		\inc{(n04310018, steam locomotive)} & \inc{(n04335435, streetcar)}\\
	\end{tabular}
	\item \textbf{\texttt{car\_bus\_truck}} -- 23 classes: 
	
	\begin{tabular}{p{0.5\linewidth}p{0.5\linewidth}}
		\inc{(n02701002, ambulance)} & \inc{(n02704792, amphibian)}\\
		\inc{(n02814533, beach wagon)} & \inc{(n02930766, cab)}\\
		\inc{(n03100240, convertible)} & \inc{(n03345487, fire engine)}\\
		\inc{(n03417042, garbage truck)} & \inc{(n03594945, jeep)}\\
		\inc{(n03670208, limousine)} & \inc{(n03769881, minibus)}\\
		\inc{(n03770679, minivan)} & \inc{(n03777568, Model T)}\\
		\inc{(n03796401, moving van)} & \inc{(n03930630, pickup)}\\
		\inc{(n03977966, police van)} & \inc{(n04037443, racer)}\\
		\inc{(n04065272, recreational vehicle)} & \inc{(n04146614, school bus)}\\
		\inc{(n04252225, snowplow)} & \inc{(n04285008, sports car)}\\
		\inc{(n04461696, tow truck)} & \inc{(n04467665, trailer truck)}\\
		\inc{(n04487081, trolleybus)} & \\
	\end{tabular}
	\item \textbf{\texttt{racer\_go\_kart}} -- 2 classes: 
	
	\begin{tabular}{p{0.5\linewidth}p{0.5\linewidth}}
		\inc{(n03444034, go-kart)} & \inc{(n04037443, racer)}\\
	\end{tabular}
	\item \textbf{\texttt{work\_cart}} -- 4 classes: 
	
	\begin{tabular}{p{0.5\linewidth}p{0.5\linewidth}}
		\inc{(n03384352, forklift)} & \inc{(n03444034, go-kart)}\\
		\inc{(n03445924, golfcart)} & \inc{(n03649909, lawn mower)}\\
	\end{tabular}
	\item \textbf{\texttt{open\_cart}} -- 4 classes: 
	
	\begin{tabular}{p{0.5\linewidth}p{0.5\linewidth}}
		\inc{(n02797295, barrow)} & \inc{(n03538406, horse cart)}\\
		\inc{(n03599486, jinrikisha)} & \inc{(n03868242, oxcart)}\\
	\end{tabular}
	\item \textbf{\texttt{rocket\_missile}} -- 3 classes: 
	
	\begin{tabular}{p{0.5\linewidth}p{0.5\linewidth}}
		\inc{(n03773504, missile)} & \inc{(n04008634, projectile)}\\
		\inc{(n04266014, space shuttle)} & \\
	\end{tabular}
	\item \textbf{\texttt{keyboard\_instrument}} -- 3 classes: 
	
	\begin{tabular}{p{0.5\linewidth}p{0.5\linewidth}}
		\inc{(n03452741, grand piano)} & \inc{(n03854065, organ)}\\
		\inc{(n04515003, upright)} & \\
	\end{tabular}
	\item \textbf{\texttt{stringed\_instrument}} -- 5 classes: 
	
	\begin{tabular}{p{0.5\linewidth}p{0.5\linewidth}}
		\inc{(n02676566, acoustic guitar)} & \inc{(n02787622, banjo)}\\
		\inc{(n02992211, cello)} & \inc{(n03272010, electric guitar)}\\
		\inc{(n04536866, violin)} & \\
	\end{tabular}
	\item \textbf{\texttt{wind\_instrument\_brass}} -- 4 classes: 
	
	\begin{tabular}{p{0.5\linewidth}p{0.5\linewidth}}
		\inc{(n03110669, cornet)} & \inc{(n03394916, French horn)}\\
		\inc{(n04141076, sax)} & \inc{(n04487394, trombone)}\\
	\end{tabular}
	\item \textbf{\texttt{wind\_instrument\_woodwind}} -- 4 classes: 
	
	\begin{tabular}{p{0.5\linewidth}p{0.5\linewidth}}
		\inc{(n02804610, bassoon)} & \inc{(n03372029, flute)}\\
		\inc{(n03838899, oboe)} & \inc{(n04141076, sax)}\\
	\end{tabular}
	\item \textbf{\texttt{wind\_instrument\_whistle}} -- 4 classes: 
	
	\begin{tabular}{p{0.5\linewidth}p{0.5\linewidth}}
		\inc{(n02804610, bassoon)} & \inc{(n03372029, flute)}\\
		\inc{(n03838899, oboe)} & \inc{(n04579432, whistle)}\\
	\end{tabular}
	\item \textbf{\texttt{percussion\_instrument}} -- 5 classes: 
	
	\begin{tabular}{p{0.5\linewidth}p{0.5\linewidth}}
		\inc{(n03017168, chime)} & \inc{(n03249569, drum)}\\
		\inc{(n03447721, gong)} & \inc{(n03721384, marimba)}\\
		\inc{(n04311174, steel drum)} & \\
	\end{tabular}
	\item \textbf{\texttt{public\_place}} -- 12 classes: 
	
	\begin{tabular}{p{0.5\linewidth}p{0.5\linewidth}}
		\inc{(n02776631, bakery)} & \inc{(n02791270, barbershop)}\\
		\inc{(n02871525, bookshop)} & \inc{(n02927161, butcher shop)}\\
		\inc{(n03032252, cinema)} & \inc{(n03089624, confectionery)}\\
		\inc{(n03461385, grocery store)} & \inc{(n03661043, library)}\\
		\inc{(n04081281, restaurant)} & \inc{(n04200800, shoe shop)}\\
		\inc{(n04443257, tobacco shop)} & \inc{(n04462240, toyshop)}\\
	\end{tabular}
	\item \textbf{\texttt{bridge\_pier\_dam}} -- 5 classes: 
	
	\begin{tabular}{p{0.5\linewidth}p{0.5\linewidth}}
		\inc{(n03160309, dam)} & \inc{(n03933933, pier)}\\
		\inc{(n04311004, steel arch bridge)} & \inc{(n04366367, suspension bridge)}\\
		\inc{(n04532670, viaduct)} & \\
	\end{tabular}
	\item \textbf{\texttt{game\_ball}} -- 9 classes: 
	
	\begin{tabular}{p{0.5\linewidth}p{0.5\linewidth}}
		\inc{(n02799071, baseball)} & \inc{(n02802426, basketball)}\\
		\inc{(n03134739, croquet ball)} & \inc{(n03445777, golf ball)}\\
		\inc{(n03942813, ping-pong ball)} & \inc{(n04118538, rugby ball)}\\
		\inc{(n04254680, soccer ball)} & \inc{(n04409515, tennis ball)}\\
		\inc{(n04540053, volleyball)} & \\
	\end{tabular}
	\item \textbf{\texttt{computer\_monitor}} -- 8 classes: 
	
	\begin{tabular}{p{0.5\linewidth}p{0.5\linewidth}}
		\inc{(n03180011, desktop computer)} & \inc{(n03485407, hand-held computer)}\\
		\inc{(n03642806, laptop)} & \inc{(n03782006, monitor)}\\
		\inc{(n03832673, notebook)} & \inc{(n03857828, oscilloscope)}\\
		\inc{(n04152593, screen)} & \inc{(n04404412, television)}\\
	\end{tabular}
	\item \textbf{\texttt{keyboard}} -- 6 classes: 
	
	\begin{tabular}{p{0.5\linewidth}p{0.5\linewidth}}
		\inc{(n03085013, computer keyboard)} & \inc{(n03485407, hand-held computer)}\\
		\inc{(n03642806, laptop)} & \inc{(n03832673, notebook)}\\
		\inc{(n04264628, space bar)} & \inc{(n04505470, typewriter keyboard)}\\
	\end{tabular}
	\item \textbf{\texttt{player}} -- 4 classes: 
	
	\begin{tabular}{p{0.5\linewidth}p{0.5\linewidth}}
		\inc{(n02979186, cassette player)} & \inc{(n02988304, CD player)}\\
		\inc{(n04041544, radio)} & \inc{(n04392985, tape player)}\\
	\end{tabular}
	\item \textbf{\texttt{small\_electronic\_equipment}} -- 9 classes: 
	
	\begin{tabular}{p{0.5\linewidth}p{0.5\linewidth}}
		\inc{(n02978881, cassette)} & \inc{(n02992529, cellular telephone)}\\
		\inc{(n03271574, electric fan)} & \inc{(n03485407, hand-held computer)}\\
		\inc{(n03492542, hard disc)} & \inc{(n03584254, iPod)}\\
		\inc{(n03602883, joystick)} & \inc{(n04074963, remote control)}\\
		\inc{(n04372370, switch)} & \\
	\end{tabular}
	\item \textbf{\texttt{phone}} -- 3 classes: 
	
	\begin{tabular}{p{0.5\linewidth}p{0.5\linewidth}}
		\inc{(n02992529, cellular telephone)} & \inc{(n03187595, dial telephone)}\\
		\inc{(n03902125, pay-phone)} & \\
	\end{tabular}
	\item \textbf{\texttt{radio\_modem}} -- 3 classes: 
	
	\begin{tabular}{p{0.5\linewidth}p{0.5\linewidth}}
		\inc{(n03492542, hard disc)} & \inc{(n03777754, modem)}\\
		\inc{(n04041544, radio)} & \\
	\end{tabular}
	\item \textbf{\texttt{big\_electronic\_equipment}} -- 6 classes: 
	
	\begin{tabular}{p{0.5\linewidth}p{0.5\linewidth}}
		\inc{(n03691459, loudspeaker)} & \inc{(n03777754, modem)}\\
		\inc{(n03857828, oscilloscope)} & \inc{(n03924679, photocopier)}\\
		\inc{(n04004767, printer)} & \inc{(n04009552, projector)}\\
	\end{tabular}
	\item \textbf{\texttt{clock}} -- 9 classes: 
	
	\begin{tabular}{p{0.5\linewidth}p{0.5\linewidth}}
		\inc{(n02708093, analog clock)} & \inc{(n02794156, barometer)}\\
		\inc{(n03706229, magnetic compass)} & \inc{(n03841143, odometer)}\\
		\inc{(n03891332, parking meter)} & \inc{(n04141975, scale)}\\
		\inc{(n04328186, stopwatch)} & \inc{(n04355338, sundial)}\\
		\inc{(n04548280, wall clock)} & \\
	\end{tabular}
	\item \textbf{\texttt{digital\_clock}} -- 5 classes: 
	
	\begin{tabular}{p{0.5\linewidth}p{0.5\linewidth}}
		\inc{(n03196217, digital clock)} & \inc{(n03197337, digital watch)}\\
		\inc{(n04141975, scale)} & \inc{(n04149813, scoreboard)}\\
		\inc{(n04328186, stopwatch)} & \\
	\end{tabular}
	\item \textbf{\texttt{ruler}} -- 3 classes: 
	
	\begin{tabular}{p{0.5\linewidth}p{0.5\linewidth}}
		\inc{(n04118776, rule)} & \inc{(n04238763, slide rule)}\\
		\inc{(n04376876, syringe)} & \\
	\end{tabular}
	\item \textbf{\texttt{firearm}} -- 3 classes: 
	
	\begin{tabular}{p{0.5\linewidth}p{0.5\linewidth}}
		\inc{(n02749479, assault rifle)} & \inc{(n04086273, revolver)}\\
		\inc{(n04090263, rifle)} & \\
	\end{tabular}
	\item \textbf{\texttt{gymnastic\_apparatus}} -- 3 classes: 
	
	\begin{tabular}{p{0.5\linewidth}p{0.5\linewidth}}
		\inc{(n02777292, balance beam)} & \inc{(n03535780, horizontal bar)}\\
		\inc{(n03888605, parallel bars)} & \\
	\end{tabular}
	\item \textbf{\texttt{barbell\_dumbbell}} -- 2 classes: 
	
	\begin{tabular}{p{0.5\linewidth}p{0.5\linewidth}}
		\inc{(n02790996, barbell)} & \inc{(n03255030, dumbbell)}\\
	\end{tabular}
	\item \textbf{\texttt{dish}} -- 13 classes: 
	
	\begin{tabular}{p{0.5\linewidth}p{0.5\linewidth}}
		\inc{(n04263257, soup bowl)} & \inc{(n07579787, plate)}\\
		\inc{(n07583066, guacamole)} & \inc{(n07584110, consomme)}\\
		\inc{(n07590611, hot pot)} & \inc{(n07697313, cheeseburger)}\\
		\inc{(n07697537, hotdog)} & \inc{(n07711569, mashed potato)}\\
		\inc{(n07831146, carbonara)} & \inc{(n07871810, meat loaf)}\\
		\inc{(n07873807, pizza)} & \inc{(n07875152, potpie)}\\
		\inc{(n07880968, burrito)} & \\
	\end{tabular}
	\item \textbf{\texttt{beverage\_pot\_jug}} -- 18 classes: 
	
	\begin{tabular}{p{0.5\linewidth}p{0.5\linewidth}}
		\inc{(n02815834, beaker)} & \inc{(n02823750, beer glass)}\\
		\inc{(n03062245, cocktail shaker)} & \inc{(n03063599, coffee mug)}\\
		\inc{(n03063689, coffeepot)} & \inc{(n03443371, goblet)}\\
		\inc{(n03733805, measuring cup)} & \inc{(n03950228, pitcher)}\\
		\inc{(n04131690, saltshaker)} & \inc{(n04254120, soap dispenser)}\\
		\inc{(n04398044, teapot)} & \inc{(n04522168, vase)}\\
		\inc{(n04560804, water jug)} & \inc{(n04579145, whiskey jug)}\\
		\inc{(n07892512, red wine)} & \inc{(n07920052, espresso)}\\
		\inc{(n07930864, cup)} & \inc{(n07932039, eggnog)}\\
	\end{tabular}
	\item \textbf{\texttt{spatula\_spoon\_ladle}} -- 3 classes: 
	
	\begin{tabular}{p{0.5\linewidth}p{0.5\linewidth}}
		\inc{(n03633091, ladle)} & \inc{(n04270147, spatula)}\\
		\inc{(n04597913, wooden spoon)} & \\
	\end{tabular}
	\item \textbf{\texttt{utensils}} -- 11 classes: 
	
	\begin{tabular}{p{0.5\linewidth}p{0.5\linewidth}}
		\inc{(n02909870, bucket)} & \inc{(n02939185, caldron)}\\
		\inc{(n03133878, Crock Pot)} & \inc{(n03259280, Dutch oven)}\\
		\inc{(n03400231, frying pan)} & \inc{(n03775546, mixing bowl)}\\
		\inc{(n03786901, mortar)} & \inc{(n03992509, potter's wheel)}\\
		\inc{(n04263257, soup bowl)} & \inc{(n04332243, strainer)}\\
		\inc{(n04596742, wok)} & \\
	\end{tabular}
	\item \textbf{\texttt{bucket\_pot}} -- 2 classes: 
	
	\begin{tabular}{p{0.5\linewidth}p{0.5\linewidth}}
		\inc{(n02909870, bucket)} & \inc{(n03991062, pot)}\\
	\end{tabular}
	\item \textbf{\texttt{bottle}} -- 4 classes: 
	
	\begin{tabular}{p{0.5\linewidth}p{0.5\linewidth}}
		\inc{(n02823428, beer bottle)} & \inc{(n03983396, pop bottle)}\\
		\inc{(n04557648, water bottle)} & \inc{(n04591713, wine bottle)}\\
	\end{tabular}
	\item \textbf{\texttt{tub\_basin}} -- 3 classes: 
	
	\begin{tabular}{p{0.5\linewidth}p{0.5\linewidth}}
		\inc{(n02808440, bathtub)} & \inc{(n04493381, tub)}\\
		\inc{(n04553703, washbasin)} & \\
	\end{tabular}
	\item \textbf{\texttt{barrel\_can}} -- 4 classes: 
	
	\begin{tabular}{p{0.5\linewidth}p{0.5\linewidth}}
		\inc{(n02747177, ashcan)} & \inc{(n02795169, barrel)}\\
		\inc{(n03764736, milk can)} & \inc{(n04049303, rain barrel)}\\
	\end{tabular}
	\item \textbf{\texttt{big\_appliance}} -- 7 classes: 
	
	\begin{tabular}{p{0.5\linewidth}p{0.5\linewidth}}
		\inc{(n03207941, dishwasher)} & \inc{(n03297495, espresso maker)}\\
		\inc{(n03761084, microwave)} & \inc{(n04070727, refrigerator)}\\
		\inc{(n04111531, rotisserie)} & \inc{(n04330267, stove)}\\
		\inc{(n04554684, washer)} & \\
	\end{tabular}
	\item \textbf{\texttt{kitchen\_appliance}} -- 4 classes: 
	
	\begin{tabular}{p{0.5\linewidth}p{0.5\linewidth}}
		\inc{(n03297495, espresso maker)} & \inc{(n03761084, microwave)}\\
		\inc{(n04442312, toaster)} & \inc{(n04542943, waffle iron)}\\
	\end{tabular}
	\item \textbf{\texttt{small\_appliance}} -- 4 classes: 
	
	\begin{tabular}{p{0.5\linewidth}p{0.5\linewidth}}
		\inc{(n03483316, hand blower)} & \inc{(n03584829, iron)}\\
		\inc{(n04179913, sewing machine)} & \inc{(n04517823, vacuum)}\\
	\end{tabular}
	\item \textbf{\texttt{oven\_stove}} -- 5 classes: 
	
	\begin{tabular}{p{0.5\linewidth}p{0.5\linewidth}}
		\inc{(n03761084, microwave)} & \inc{(n04111531, rotisserie)}\\
		\inc{(n04330267, stove)} & \inc{(n04442312, toaster)}\\
		\inc{(n04542943, waffle iron)} & \\
	\end{tabular}
	\item \textbf{\texttt{space\_heater\_radiator}} -- 2 classes: 
	
	\begin{tabular}{p{0.5\linewidth}p{0.5\linewidth}}
		\inc{(n04040759, radiator)} & \inc{(n04265275, space heater)}\\
	\end{tabular}
	\item \textbf{\texttt{paper\_tissue}} -- 2 classes: 
	
	\begin{tabular}{p{0.5\linewidth}p{0.5\linewidth}}
		\inc{(n03887697, paper towel)} & \inc{(n15075141, toilet tissue)}\\
	\end{tabular}
	\item \textbf{\texttt{cover\_presentation}} -- 10 classes: 
	
	\begin{tabular}{p{0.5\linewidth}p{0.5\linewidth}}
		\inc{(n02786058, Band Aid)} & \inc{(n02840245, binder)}\\
		\inc{(n03291819, envelope)} & \inc{(n03598930, jigsaw puzzle)}\\
		\inc{(n03871628, packet)} & \inc{(n06359193, web site)}\\
		\inc{(n06596364, comic book)} & \inc{(n06785654, crossword puzzle)}\\
		\inc{(n07248320, book jacket)} & \inc{(n07565083, menu)}\\
	\end{tabular}
	\item \textbf{\texttt{toiletry}} -- 5 classes: 
	
	\begin{tabular}{p{0.5\linewidth}p{0.5\linewidth}}
		\inc{(n03314780, face powder)} & \inc{(n03476991, hair spray)}\\
		\inc{(n03690938, lotion)} & \inc{(n03916031, perfume)}\\
		\inc{(n04357314, sunscreen)} & \\
	\end{tabular}
	\item \textbf{\texttt{bed}} -- 2 classes: 
	
	\begin{tabular}{p{0.5\linewidth}p{0.5\linewidth}}
		\inc{(n03388549, four-poster)} & \inc{(n04344873, studio couch)}\\
	\end{tabular}
	\item \textbf{\texttt{fabric}} -- 8 classes: 
	
	\begin{tabular}{p{0.5\linewidth}p{0.5\linewidth}}
		\inc{(n02808304, bath towel)} & \inc{(n02834397, bib)}\\
		\inc{(n03207743, dishrag)} & \inc{(n03223299, doormat)}\\
		\inc{(n03485794, handkerchief)} & \inc{(n03998194, prayer rug)}\\
		\inc{(n04209239, shower curtain)} & \inc{(n04525038, velvet)}\\
	\end{tabular}
	\item \textbf{\texttt{religious\_castle\_palace}} -- 12 classes: 
	
	\begin{tabular}{p{0.5\linewidth}p{0.5\linewidth}}
		\inc{(n02699494, altar)} & \inc{(n02825657, bell cote)}\\
		\inc{(n02980441, castle)} & \inc{(n03028079, church)}\\
		\inc{(n03220513, dome)} & \inc{(n03781244, monastery)}\\
		\inc{(n03788195, mosque)} & \inc{(n03877845, palace)}\\
		\inc{(n03956157, planetarium)} & \inc{(n04346328, stupa)}\\
		\inc{(n04486054, triumphal arch)} & \inc{(n04523525, vault)}\\
	\end{tabular}
	\item \textbf{\texttt{pole}} -- 3 classes: 
	
	\begin{tabular}{p{0.5\linewidth}p{0.5\linewidth}}
		\inc{(n03355925, flagpole)} & \inc{(n03733131, maypole)}\\
		\inc{(n03976657, pole)} & \\
	\end{tabular}
	\item \textbf{\texttt{column}} -- 4 classes: 
	
	\begin{tabular}{p{0.5\linewidth}p{0.5\linewidth}}
		\inc{(n03837869, obelisk)} & \inc{(n03903868, pedestal)}\\
		\inc{(n03976657, pole)} & \inc{(n04458633, totem pole)}\\
	\end{tabular}%

\end{enumerate}

}{}

\message{^^JLASTPAGE \thepage^^J}

\end{document}